\documentclass[krantz1,ChapterTOCs]{krantz}
\usepackage{amssymb}
\usepackage{amsmath}
\usepackage{graphicx}
\usepackage{subfigure}
\usepackage{makeidx}
\usepackage{multicol}
\usepackage[utf8]{inputenc}

\everymath{\displaystyle}

\usepackage[table]{xcolor}
\makeindex

\begin{document}

\frontmatter
%\newtheorem{theorem}{Theorem}
%\newtheorem{exercise}{Exercise}[chapter]
%\newtheorem{example}{Example}
%\newtheorem{definition}{Definition}
%\newtheorem{proof}{Proof}
 %place custom commands and macros here

\title{Octocopter Design: Modelling, Control and Motion Planning 
%in Socio-Environmental Systems, Public Health, and Insurance\\
%{\Large(Applied Environmental Statistics Series)}
}
\author{Nedim Osmic, Adnan Tahirovic and Bakir Lacevic}

\maketitle

%\include{frontmatter/dedication}
%\cleardoublepage
\setcounter{page}{7} %previous pages will be reserved for frontmatter to be added in later.
\tableofcontents
\include{placeins}
\listoffigures
\listoftables

\mainmatter

%\part{Extended abstract}
%\part{Modeling and control of octocopter}
%\include{chapters/Chapter0/ch0}
%\chapterauthor{Nedim Osmić}{dr. sc.}
%\chapterauthor{Adnan Tahirović}{dr. sc.}
%\chapterauthor{Bakir Lačević}{dr. sc.}
%\chapter{Introduction}

%\section*{Motion planning of aerial vehicle based on risk assessment of failure states}\label{intro}
%\thispagestyle{empty}

\chapter{Introduction}

\hspace*{\fill}

In recent years, unmanned aerial vehicles (UAV) have become one of the major fields of robotics research in academic and industrial communities due to the broad range of their potential applications, including search and rescue missions in urban \cite{Tomic_istr_platforme,NAdgledanje_2_doherty2007uav,Istorija_UAV_blom2010} and non-urban environments \cite{Morse}, aerial construction \cite{Kumar_konstrukcije}, precision agriculture \cite{Inspekcija_hrane_chang2013improvement}, \cite{Zhang_agrikultura}, disaster management \cite{Katastrofe_maza2011experimental,shen2014multi,lorincz2021novel}, remote sensing \cite{Fotografiranje_UAV_flener2013seamless,8}, power line and structural inspection \cite{Inspekcija_EE_Mreze_golightly2005visual,Inspekcija_EE_MREZA2_li2008knowledge,Termalna_inspekcija_dios2006automatic,Inspekcija_mostova_metni2007uav}, exploration and mapping of unknown environments \cite{Mapiranje_fraundorfer2012vision,tahirovic2013discussion,tahirovic2013convergent,tahirovic2016receding}, surveillance \cite{Nadgledanje_freed2004human},\cite{NAdgledanje_granica_girard2004border}, swarming \cite{Kumar_micro_UAV}, as well as monitoring and traffic analysis \cite{Nadzor_saobracaja_puri2005survey}. Multirotor aerial vehicles (MAV) are among the most popular UAV platforms due to characteristics such as small geometries, vertical takeoff and landing, low cost, simple construction, degrees of freedom, their inherent maneuverability, as well as ability to perform the tasks which are highly risk for humans. The miniaturization of actuators and sensors, power density improvements of batteries and the ubiquity of low cost computing platforms have additionally motivated research in the field of small UAVs \cite{bouabdallah2004design}. Nowadays, there exist different design solutions for MAVs, ranging from micro and mini to heavy MAVs with large endurance \cite{vrste_UAV}. Quadrotors have the simplest MAV configuration based on the four rotor actuation. Quadrotor platforms such as AR-Drone have become a standard aerial robotic research platform used in scientific community \cite{mahony2012multirotor,Edukacijske_platforme}. 

Regardless of the structural design type of a MAV, different faults may occur during the task execution. Faults can affect functional properties of actuators, sensors, and controllers and they can be of a structural nature. If a failure occurs, the mission execution may be terminated. The detailed Failure Mode and Effect Analysis (FMEA), presented in \cite{FTC_ETH_Thomas_Schneider}, identifies critical mechanical, electrical and software components of MAV systems. Reliability requirements of mechanical and electrical components are generally met via redundancy. The critical software components include the control system and the attitude estimation algorithm, which need to be fault-tolerant to satisfy reliability constraints. In \cite{mueller_I} and \cite{mueller_II}, the authors have investigated a control strategy for a quadcopter in the case of either a single or multiple rotor faults, including double-fault of two opposing propellers and all combinations of triple faults caused by failures of different propellers.

To increase a likelihood of mission accomplishment, a various types of redundancy can be imposed in the MAV design such as redundancy of the propulsion system. In \cite{Korejci_controllability_I,Korejci_controllability_II,shi_Analiza,Lunze_I,Lunze_II}, the authors investigated how the controllability property of a system with respect to different rotor faults could be preserved by increasing the number of rotors, or using a rotor with a possibility to tilt motors \cite{franchi_heksa}, \cite{mehmood_analiza}. As an example, in  \cite{FTC_ETH_Thomas_Schneider} it was demonstrated how to control degrees of freedom of an octocopter (a MAV designed with 8 rotors) for any potential double-fault scenarios.
To design a highly reliable MAV, it is inevitable to increase the total number of rotors in the initial system design. As an example, octocopters are inherently reliable with regards to double-faults scenarios, with full system controllability being preserved in 89\% of those scenarios \cite{FTC_ETH_Thomas_Schneider}. 

Regardless of whether the configuration of a MAV is redundant, the control algorithm has a significant role in improving its overall fault-tolerance. If a control algorithm is fault-ignorant, using redundant components in the initial design does not necessary increase reliability of a MAV system, that is the probability of completing the mission. A large number of fault-tolerant control algorithms that inherently posses a certain level of robustness with respect to possible failures include sliding mode control \cite{Sliding_mode_FTC_wang2012sliding}, adaptive fault-tolerant control \cite{Adaptivni_FTC_xu2013adaptive}, \cite{Adaptivno_robusni_FTC_jin2012robust}, control allocation method for MAVs  \cite{Control_Allocation_1_casavola2010fault}, \cite{Control_Allocation_2_oppenheimer2006control}, reconfigurable control \cite{Rekontigurabilni_FTC_drozeski2005fault}, backstepping method \cite{backstepping_1_harkegard2000backstepping}, \cite{Backsteping_2_glad2000flight}, model predictive control \cite{MPC_FTC_1_khan2011fault}, control based on linear quadratic regulator \cite{LQG_FTC_lemos2013actuator}, fuzzy predictive control \cite{Fuzzy_FTC_1681912}, to name a few.
 
Fault-tolerant controls and estimation techniques are based on Fault Detection and Isolation (FDI) algorithms, which identify and localize system faults. An extensive survey of these algorithms is presented in \cite{zhang2008bibliographical}, \cite{hwang2009survey}. In case of fault-tolerant MAV systems, the state-of-the-art methods heavily rely on observer based FDI techniques. Notable examples include Luenberger state estimaton \cite{Sharifi}, Moving Horizon Estimation (MHE) and Unscented Kalman Filtering (UKF) \cite{Izadi} for actuator fault diagnosis, as well as Thau observer \cite{Freddi} and multi-observer \cite{Berbra} for sensor fault diagnosis. 
A fault detection and
isolation, formulated as the least squares parameter identification
problem, has already been considered in [14] for linear time invariant state
space models. A more specific approach, tailored for a class
of nonlinear aerodynamic models can be found in [15].

Besides construction-based redundancy as well as control-induced fault tolerance, a further increase of mission reliability can be achieved at the motion planning stage. Namely, in case the system is aware of potential risks related to failure probabilities, it is then possible to carefully generate trajectories to make the system capable of dealing with different faults. Specifically, each faulty state may have some risky manoeuvres that the system may want to avoid, especially in case a faulty state is highly probable. A conservative approach to address this problem would be to generate the system trajectories based on safe maneuvers only. In that case, the system would significantly deteriorate its flying capabilities without achieving performances for which it is designed. However, a motion planner based on the decreased probability of moving with risky maneuvers for the given mission has been introduced in \cite{Osmic_automatika}. In this way, the planner compromises two extremes, a safe and reliable planner with bad performance (risk conservative planner) and a risk-ignorant planner with the best possible performance. 

The octocopter system is consisted of an 8-rotor system which makes it a representative example for all three reliability aspects. First, it is highly redundant by construction and capable of flying even in case of multiple motor faults simultaneously occur during the mission. Second, fault-tolerant control can come to the fore in case when there exist more fault-tolerant degrees of freedom. For instance, the octocopter system has more fault states in which it is still capable of completing the mission than the quadcopter and hexacopter systems. Third, the motion planner algorithms designed for the octocopter system are capable of taking into account a wide spectrum of potential faulty states into account. This may help to generate only those trajectories along which the octocopter would be likely ready for potential faults to occur preserving good performance of a risk-ignorant planner and safety of a risk conservative approach.

This book provides a solution to the control and motion planing design for an octocopter system. It includes a particular choice of control and motion planning algorithms which is based on the authors' previous research work, so it can be used as a reference design guidance for students, researchers as well as autonomous vehicles hobbyists. The control is constructed based on a fault tolerant approach aiming to increase the chances of the system to detect and isolate a potential failure in order to produce feasible control signals to the remaining active motors. The used motion planning algorithm is risk-aware by means that it takes into account the constraints related to the fault-dependant and mission-related maneuverability analysis of the octocopter system during the planning stage. Such a planner generates only those reference trajectories along which the octocopter system would be safe and capable of good tracking in case of a single motor fault and of majority of double motor fault scenarios. The control and motion planning algorithms presented in the book aim to increase overall reliability of the system for completing the mission. 

In Chapter 2, we derive the octocopter dynamics and provide its full state space model. Chapter 3 includes different controllers used in the ococopter control architecture. A variant of fault-tolerant controller is also described which can be used to increase overall system reliability. Chapter 4 explains how to conduct a maneuverability analysis aiming to understand octocopter controllability constraints related to different possible motor faults and to the given mission. This chapter also describes how it is possible to exploit such analysis in order to design a risk-aware motion planner.  Conclusion remarks are outlined in Chapter 5.

\chapter{Modelling}
\hspace*{\fill}
%\color{purple}

%\color{blue}

% In order to achieve satisfactory control of any MAV and to eventually improve the reliability of mission execution during the planning stage as well, it is necessary to understand all stability-influential elements of its design. 

Although the Lagrangian formulation of dynamics \cite{Samir_B.} \cite{Lagrange_raffo2010integral}, \cite{Lagrange_MARTINEZ_TORRES} can also be used to derive a general octocopter dynamical model, in this book the model is derived based on the Newton-Euler formulation from the direct interpretation of Newton's second law of motion \cite{Samir_B.}, \cite{Euler_garcia2006modelling}, \cite{Euler_madani2006control}, \cite{osmic_SMC}, \cite{osmic_modelica}. The dynamical model plays an important role in MAV motion analysis and provides understanding of system behaviour (position, orientation, velocity and acceleration) with respect to applied inputs (e.g., voltages applied to MAV rotors). The model is also used in silico for testing different control laws that can be potentially implemented aiming to force the system to behave in a desired manner. In order to present the underlying model equations, we first explain the working principle of an octocopter system that secures its desired movements (roll, pitch and yaw). Then, we present octocopter kinematics and dynamics of linear and angular motions together with motor dynamics. We also explain what types of forces and moments act on the octocopter and how they can be framed in order to be used in the final dynamical model.  \color{black}

\section{Mechanical design}
\begin{figure}[!tbph]
\begin{centering}
\includegraphics[width=8cm,height=8cm]{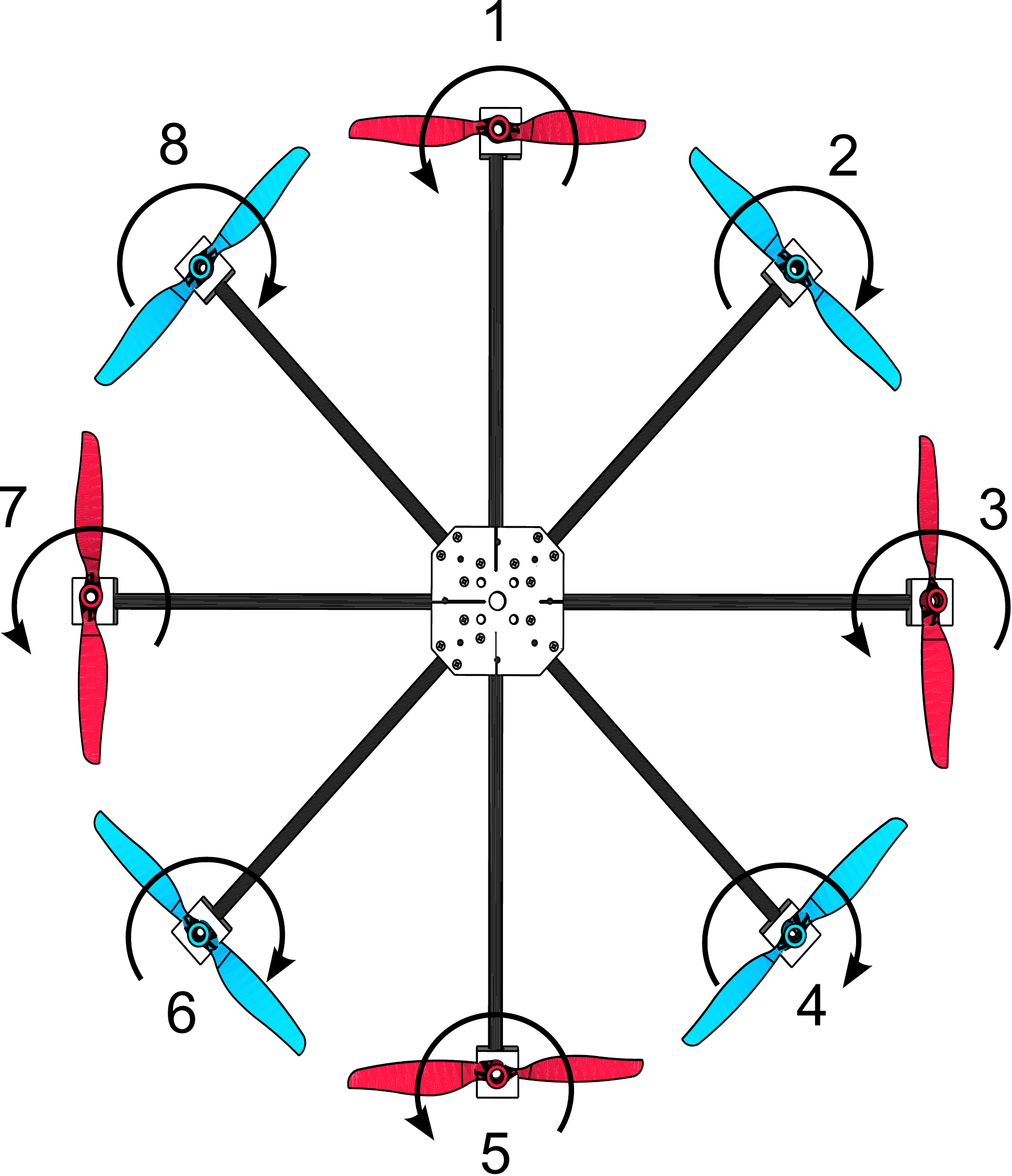}
\par\end{centering}
\caption{An octocopter design based on the PNPNPNPN configuration structure, where P and N indicate counterclockwise (motors $M_1$, $M_3$, $M_5$ and $M_7$) and clockwise (motors $M_2$, $M_4$, $M_6$ and $M_8$) directions.}
\label{prva slika}
\end{figure}

%\color{purple}
The octocopter system shown in Fig. \ref{prva slika} consists of eight arms of the same length $l$ that are fixed to a support plate, where each arm is equipped with a DC motor $M_i$ driving a fixed pitch rotor, where $i=1,..,8$. The angle between any two adjacent arms in an octocopter design is 45 degrees. The DC motors depicted with blue color ($M_2$, $M_4$, $M_6$ and $M_8$) rotate clockwise, while those depicted with red color ($M_1$, $M_3$, $M_5$ and $M_7$) rotate counterclockwise. Therefore, an octocopter system is inherently balanced with regard to the drag moment which can be generated by the motors included in the design. All other system components should be placed in a protective case which is mounted to the support plate. 

\begin{figure}[!tbph]
\begin{centering}
\includegraphics[width=10cm,height=6cm]{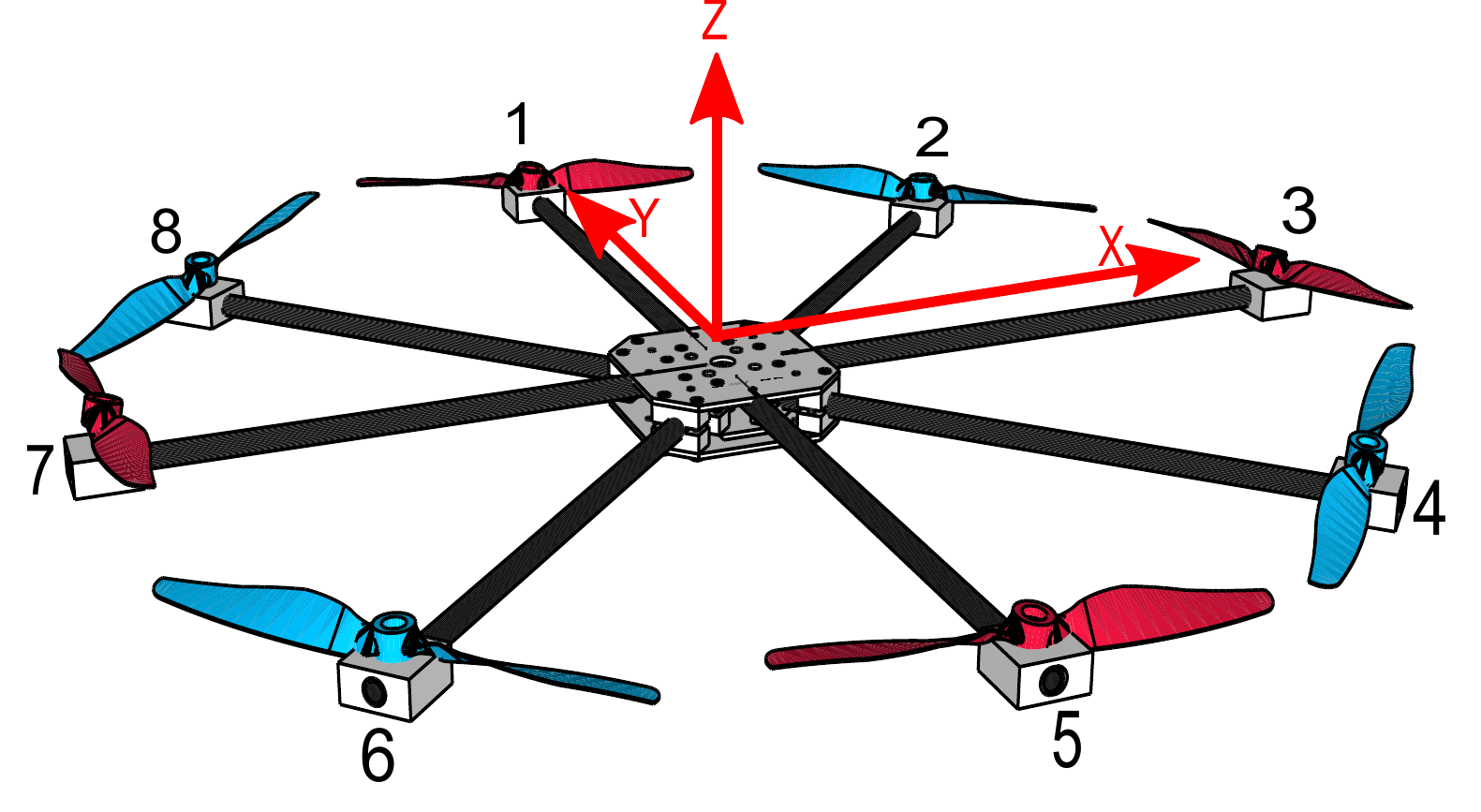}
\par\end{centering}
\caption{The local coordinate system attached to an octocopter.}
\label{druga slika}
\end{figure}

If the origin of the local coordinate system is fixed to the center of mass of the octocopter system (see Fig. \ref{druga slika}), where $X$-axis is taken along the DC motors $M_7$ and $M_3$ with the positive direction being from $M_7$ to $M_3$, $Y$-axis is taken along the rotors $M_5$ and $M_1$ with the positive direction being from $M_5$ to $M_1$, and $Z$-axis is directed upward, one gets the so-called $\textbf{'+'}$ with the PNPNPNPN configuration structure, where P and N indicate positive and negative motor rotations, respectively. Such a setup represents a classical octocopter configuration structure. However, the so-called $\textbf{'X'}$ formation based on the PPNNPPNN DC motor movement directions \cite{FTC_ETH_Thomas_Schneider} is also possible to use, as well as those with different motor distributions including the so-called $\textbf{'V'}$ and $\textbf{'X'}$ constructions based on the coaxial rotor setups \cite{Konfiguracije_rinaldi2014pid}.

Since the mechanical construction contains the motors attached to the propellers as the only moving parts of the octocopter, the input signals used to force the ocotcopter to change its position and orientation represent the control variables of interest. In the following subsections, we present different moving directions based on which it is possible to control an octocopter system. 

\medskip{}

%Višerotorska platforma koja se koristi u ovoj disertacije jest oktokopter.

%\color{magenta}

%The octorotor system shown in Fig. \ref{prva slika} consists of eight arms of the same length (lenght $l$) that are fixed to a support plate, where each arm is equipped with a DC motor driving a fixed pitch rotor. The spacing between two adjacent arms is 45 degrees. Rotors depicted with blue color (2, 4, 6 and 8) rotate clockwise, while the ones depicted with red color (1, 3, 5 and 7) rotate counterclockwise. Therefore, the octocopter is inherently balanced with regards to the drag moment generated by the rotors. All other system components (on a real system) should be placed in a protective case which is mounted to the support plate. 

\color{black}

%Da bi se objasnila osnovna gibanja oktokoptera (\emph{engl.} roll, pitch, yaw) neophodno
%je definirati i koordinatni sustav za opisivanje osnovnih gibanja.
%Ako se za ishodište lokalnog koordinatnog sustava izabere centar mase
%oktokoptera, gdje $X$ os leži na pravcu koji tvore motori broj 7
%i 3 (pozitivan smjer od motora 7 prema motoru 3), $Y$ os na pravcu
%kojeg tvore motori 5 i 1 (pozitivan smjer od motora 5 prema motoru
%1), dok je $Z$ os usmjerena prema gore, dobiva se + struktura oktokoptera
%(slika \ref{druga slika}) s PNPNPNPN smjerom vrtnje motora (P-pozitivan,
%N-negativan). Pored ove strukture postoje i druge strukture u formaciji
%$X$ %ili s PPNNPPNN smjerom vrtnje \cite{FTC_ETH_Thomas_Schneider},
%kao i različite inačice rasporeda motora poput oktokoptera V konstrukcije
%ili oktokoptera X konstrukcije s koaksijalno postavljenim motorima
%\cite{Konfiguracije_rinaldi2014pid}. Odabrana struktura predstavlja
%\textquotedbl{}klasičnu strukturu\textquotedbl{} s osam motora i naizmjeničnim
%smjerom vrtnje motora.

\medskip{}

%Budući da je mehanička konstrukcija oktokoptera jednostavna (od pokretnih
%dijelova sadrži samo motore na koje su pričvršćeni propeleri), očigledno
%je da su upravljački signali kojima se upravlja brzinom vrtnje svakog
%motora ponaosob jedine varijable na koje se može utjecati. Promjenom
%brzine vrtnje svakog motora (čime se mijenja i pojedinačna sila podizanja
%na svakom propeleru) možemo dovesti do promjene pozicije i orijentacije
%cjelokupnog sustava. U nastavku se ukratko opisuje svako od pojedinačnih
%gibanja koje se može ostvariti s pogodnim odabirom brzine vrtnje svakog
%od motora s propelerom.

\medskip{}
%\color{purple}
\section{Vertical motion - lift}
\color{black}
Fig. \ref{prva slika} illustrates a mechanical structure of an octocopter based on eight motors setup attached with eight propellers, where the motors ($M_1$, $M_3$, $M_5$, $M_7$) and ($M_2$, $M_4$, $M_6$, $M_8$) rotate in the opposite directions. In case when the equal input signals are applied to these motors, eight equal vertical forces ($F_{1}=F_{_{2}}=\cdots=F_{8}$) would be generated, all acting in the positive direction along $Z$-axis which is opposite to the direction of gravitational force. Whenever the sum of the generated forces, which are produced by each motor rotating with angular speed $\omega_{0}$ \textit{}{[}\textit{rads}$^{-1}${]}, is greater than the gravitational force, the octocopter starts moving upwards. To secure further upward moving, it is necessary to additionally increase angular speed of motors ($\omega_{0}+\Delta\omega$), where $\Delta\omega\in (0,\triangle\omega_{max})$, meaning that an octocopter can change its vertical position depending on the value of $\Delta\omega$. The upper limit $\triangle\omega_{max}$ is selected such that the system does not either face nonlinearities caused by the Coriolis and gyroscopic effects or the speed saturation.

In case when the octocopter is not in an ideal horizontal position then the resulting generated force can be decomposed along $X$ an $Y$ axes into two non-zero components that cause the horizontal movement. The inclination from an ideal horizontal position and rotation around $Z$-axis can be generated by applying asymmetrical motor speeds which allow the octocopter movement in any desired direction. In the following subsections, we explain how the forces and motor speeds are generated to force the octocopter rotation around $X$, $Y$ and $Z$ axes.

\begin{figure}[!tbph]
\begin{centering}
\includegraphics[scale=0.55]{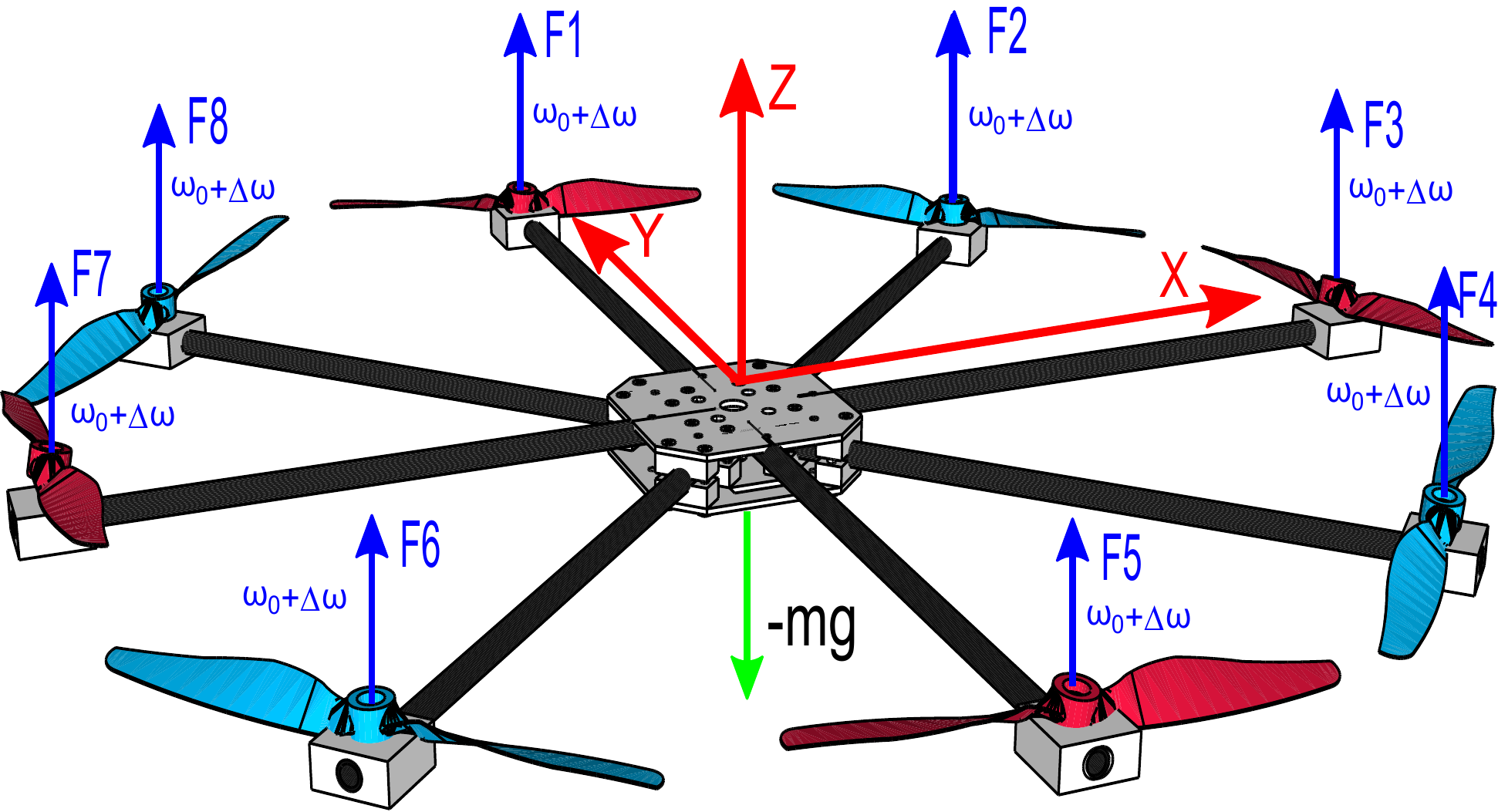}
\par\end{centering}
\caption{The forces that influence the movement of an octocopter system.}
\label{producirane_sile}
\end{figure}
\medskip{}

\section{Roll}

\medskip{}

%Ako se s $\phi$ označi rotacija oktokoptera oko $X$ osi (\emph{engl.}
%$roll$), tada je neophodno pronaći kombinaciju sila koje ostvaruju
%rotaciju oktokoptera samo oko $X$ osi, pri čemu se rotacija oko osi
%$Y$ i $Z$ ne ostvaruje. Raspored brzine vrtnje motora s propelerima,
%odnosno generirane sile, predočen je slikom \ref{Rotacija_X}.

\begin{figure}[!tbph]
\begin{centering}
\includegraphics[scale=0.5]{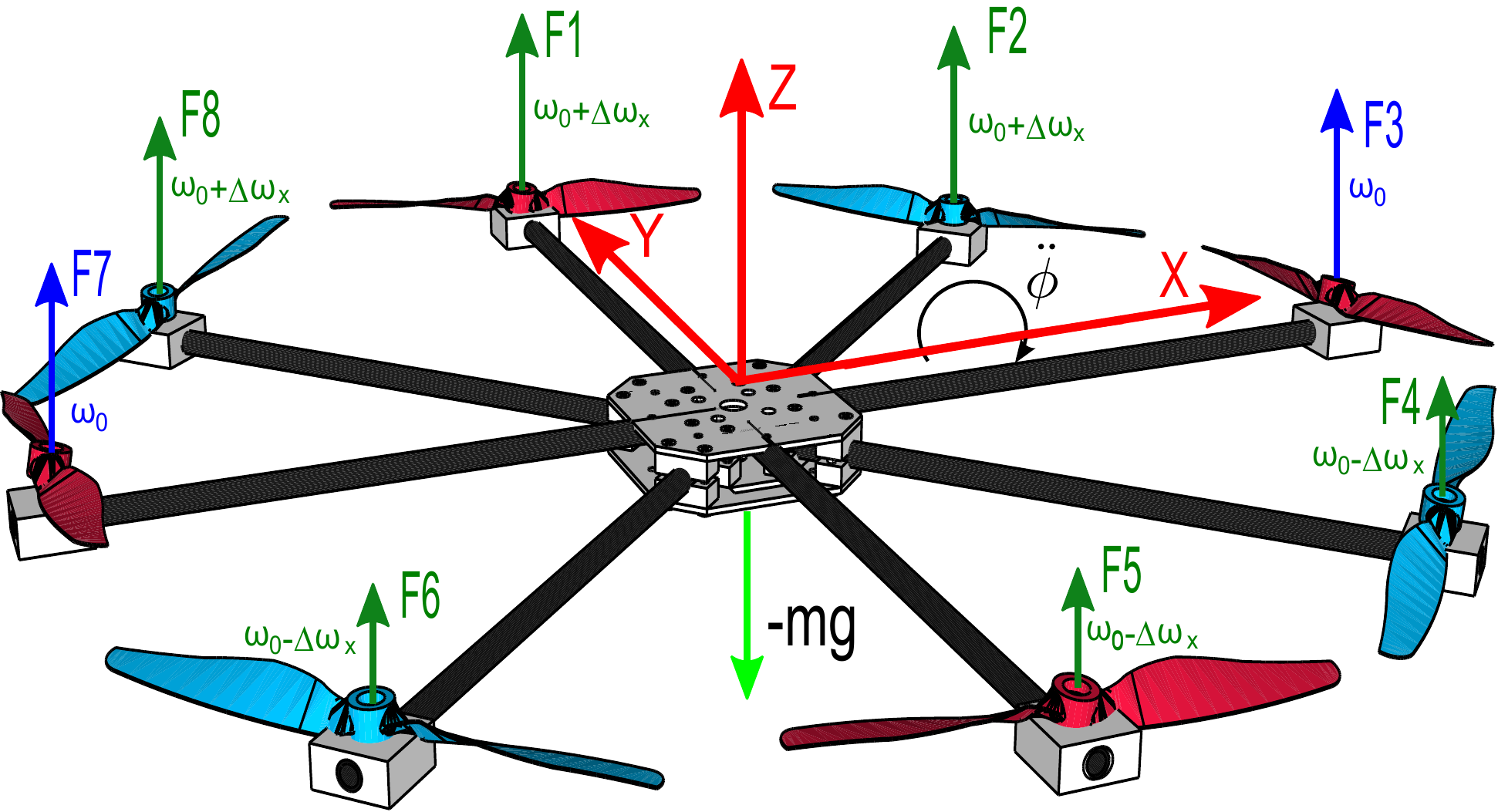}
\par\end{centering}
\caption{Distribution of the forces to ensure rotation of an octocopter system around $X$-axis of the local coordinate system.}
\label{Rotacija_X}
\end{figure}

\medskip{}

% \color{red}

Rolling around $X$-axis (see Fig. \ref{Rotacija_X}) with an angle $\phi$ can be achieved by increasing the angular velocity of the DC motors $M_1$, $M_2$ and $M_8$ for a value $\triangle\omega_{x}$, while decreasing the angular velocity of the DC motors $M_4$, $M_5$ and $M_6$ for a value $\triangle\omega_{x}$ and maintaining the same angular velocity $\omega_{0}$ of the remaining motors $M_3$ and $M_7$. In this way, the intensity of generated forces $F_{1}=F_{2}=F_{8}$ are greater than the intensity of forces $F_{3}=F_{7}$ and $F_{4}=F_{5}=F_{6}$. Having now $F_{1}+F_{2}+F_{8}>F_{4}+F_{5}+F_{6}$ implies $\phi>0$, i.e. a positive rotation of the octocopter around $X$-axis of the local coordinate system. It can be seen from Fig. \ref{Rotacija_X} that the total sum of generated forces is still the same which preserves the same vertical octocopter position. It can also be noticed that there is a full balance between $(F_{2}, F_{3}, F_{4})$ and $(F_{6}, F_{7}, F_{8})$, and between $(F_{2}, F_{4}, F_{6}, F_{8})$ and $(F_{1}, F_{3}, F_{5}, F_{7})$, providing no additional rotations around $Y$-axis and $Z$-axis, respectively. However, in order to rotate an octocopter system only around $X$-axis but in the opposite direction, one should secure $F_{1}+F_{2}+F_{8}<F_{4}+F_{5}+F_{6}$ while preserving $F_{3}=F_{7}$.

%On the other hand forces $F_{3}$ and $F_{7}$ they do not generate any moment around $X$-body axis. 
%There is no rotation about the $Z$-body axis for this propeller motor rotation because the total momentum generated around the $ Z $ axis remained the same and equal is zero (provided that the signal level $\triangle\omega_{x}$ is not too large because then the effects of nonlinearity or saturation of the DC motors come to the fore).
%Also, due to the shown way of generating forces, there is no change in the total force of vertical lifting so that the octocopter remains at the same height, while the UAV rolling around the $X$-body axis.
%To rotate the octocopter system in the opposite direction, force generated on motors 1, 2 and 8 are less than nominal while they are generated on motors 4, 5, and 6 greater than the nominal, i.e. $F_{1}+F_{2}+F_{8}<F_{4}+F_{5}+F_{6}$.

\section{Pitch}

%Na sličan način kako je to objašnjeno u prethodnom potpoglavlju za
%rotaciju oko $X$ osi generiraju se sile za rotaciju oko $Y$ osi.
%Ako se s $\theta$ označi rotacija oktokoptera oko $Y$ osi (\emph{engl.}
%$pitch)$, tada je neophodno pronaći kombinaciju sila koje osiguravaju
%rotaciju oktokoptera samo oko $Y$ osi, pri čemu se ne ostvaruje rotacija
%oko osi $X$ i $Z$. Raspored brzina vrtnje motora s propelerima,
%odnosno generiranih sila, predočen je slikom \ref{Rotacija_Y} 

\begin{figure}[!tbph]
\begin{centering}
\includegraphics[scale=0.5]{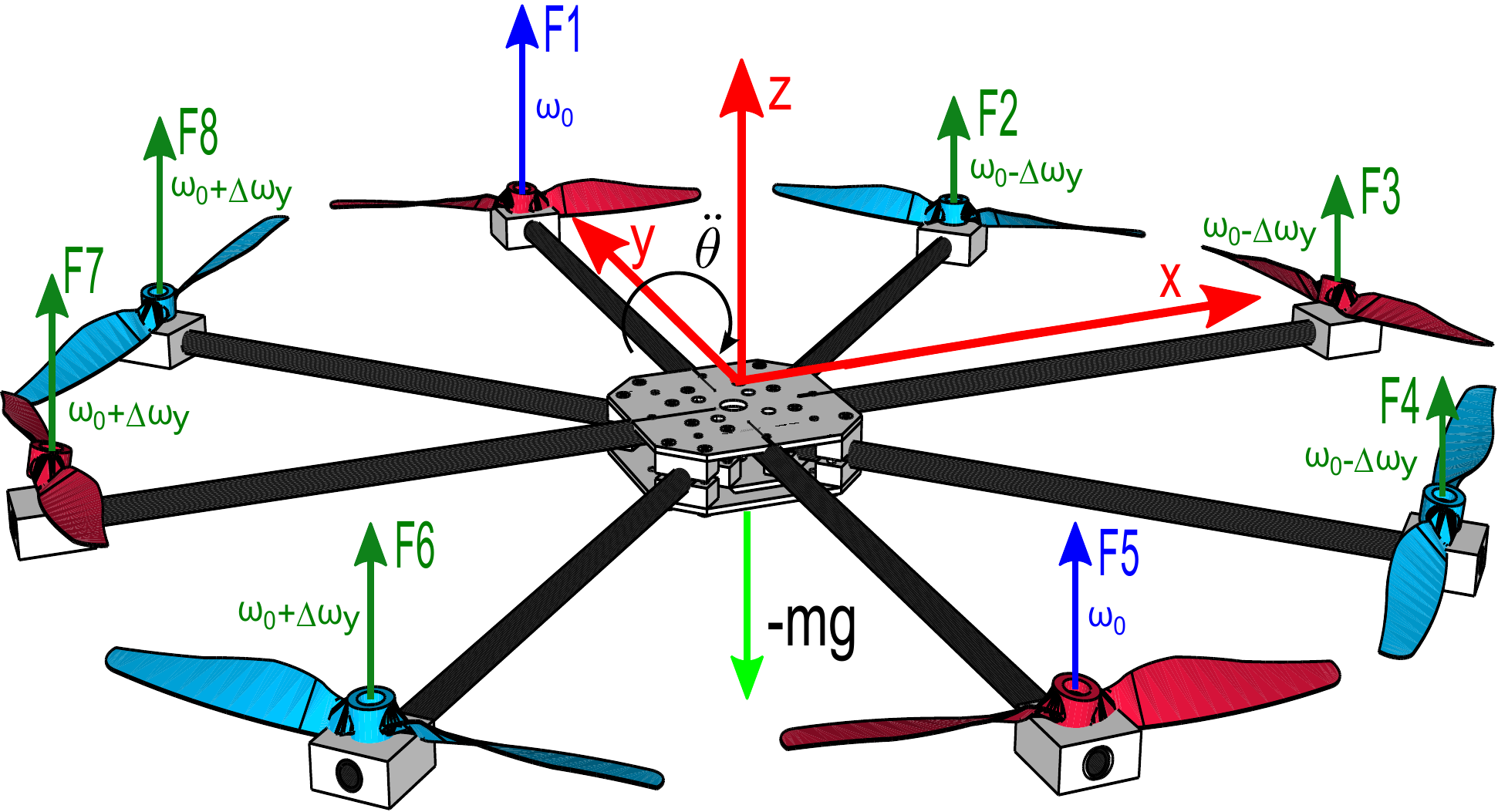}
\par\end{centering}
\caption{Distribution of the forces to ensure rotation of an octocopter system around $Y$-axis of the local coordinate system.}
\label{Rotacija_Y}
\end{figure}

%\color{red}

Pitching around $Y$-axis (see Fig. \ref{Rotacija_Y}) with an angle $\theta$ can be achieved by increasing the angular velocity of the DC motors $M_6$, $M_7$ and $M_8$ for a value $\triangle\omega_{x}$, while decreasing the angular velocity of the DC motors $M_2$, $M_3$ and $M_4$ for a value $\triangle\omega_{x}$ and maintaining the same angular velocity of the remaining DC motors $M_1$ and $M_5$. In this way, the intensity of generated forces $F_{6}=F_{7}=F_{8}$ are greater than the intensity of forces $F_{1}=F_{5}$ and $F_{2}=F_{3}=F_{4}$. Having now $F_{6}+F_{7}+F_{8}>F_{2}+F_{3}+F_{4}$ implies $\theta>0$, i.e. a positive rotation of the octocopter around $Y$-axis. It can be seen from Fig. \ref{Rotacija_Y} that the total sum of generated forces is still the same which preserves the same vertical octocopter position. It can also be noticed that there is a full balance between $(F_{1}, F_{2}, F_{8})$ and $(F_{4}, F_{5}, F_{6})$, and between $(F_{2}, F_{4}, F_{6}, F_{8})$ and $(F_{1}, F_{3}, F_{5}, F_{7})$, providing no additional rotations around $X$-axis and $Z$-axis, respectively. However, in order to rotate an octocopter system only around $Y$-axis but in the opposite direction, one should secure $F_{6}+F_{7}+F_{8}<F_{2}+F_{3}+F_{4}$ while preserving $F_{1}=F_{5}$.

%Pitch is achieved when the UAV performs a rotation around the $Y$-body axis. This is achieved by increasing the angular velocity of the rotors 6., 7. and 8. for value $\triangle\omega_{x}$ while decreasing the angular velocity of the rotors 2., 3. and 4. for value $\triangle\omega_{x}$,  maintaining the same angular velocity remaining rotors (Fig. \ref{Rotacija_X} ). In this way, the generated forces $F_{6}=F_{7}=F_{8}$ are greater, in intensity, than the forces $F_{1}=F_{5}$, as well as forces $F_{2}=F_{3}=F_{4}$ (these forces are less than the nominal forces $F_{1}=F_{5}$).
%Knowing, $F_{6}+F_{7}+F_{8}>F_{2}+F_{3}+F_{4}$, this results in a positive pitch or rotation around the $Y$-body axis. On the other hand forces $F_{1}$ and $F_{5}$ they do not generate any moment around $Y$-body axis.
%There is no rotation about the $Z$-body axis for this propeller motor rotation because the total momentum generated around the $Z$ axis remained the same and equal is zero (provided that the signal level $\triangle\omega_{x}$ is not too large because then the effects of nonlinearity or saturation of the DC motors come to the fore).
%Also, due to the shown way of generating forces, there is no change in the total force of vertical lifting so that the octocopter remains at the same height, while the UAV rolling around the $X$-body axis.
%To rotate the octocopter system in the opposite direction, force generated on motors 1, 2 and 8 are less than nominal while they are generated on motors 4, 5, and 6 greater than the nominal, i.e. $F_{6}+F_{7}+F_{8}<F_{2}+F_{3}+F_{4}$.

\section{Yaw}

\begin{figure}[!tbph]
\begin{centering}
\includegraphics[scale=0.5]{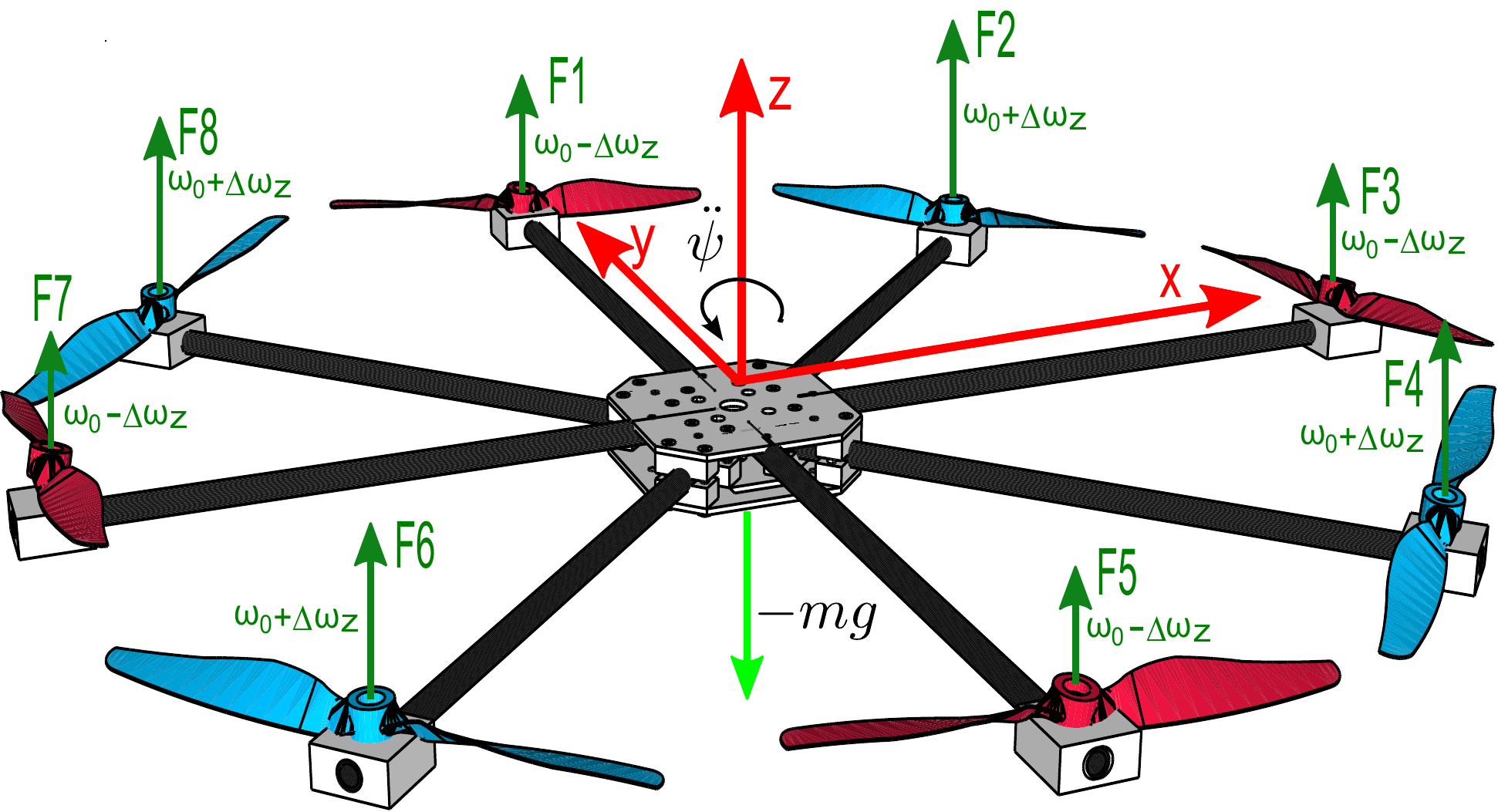}
\par\end{centering}
\caption{Distribution of the forces to ensure rotation of an octocopter system around $Z$-axis of the local coordinate system.}
\label{Rotacija_Z}
\end{figure}

Yawing around $Z$-axis (see Fig. \ref{Rotacija_Z}) with an angle $\psi$ can be achieved by increasing the angular velocity of the DC motors $M_2$, $M_4$, $M_6$ and $M_8$ for a value $\triangle\omega_{x}$, while decreasing the angular velocity of the DC motors $M_1$, $M_3$, $M_5$ and $M_7$ for a value $\triangle\omega_{x}$. In this way, the intensity of generated forces $F_{2}=F_{4}=F_{6}=F_{8}$ over the P type of rotors are greater than the intensity of forces $F_{1}=F_{3}=F_{5}=F_{7}$ over the N type of rotors. This imbalance implies $\psi>0$, i.e. a positive rotation of the octocopter around $Z$-axis. It can be seen from Fig. \ref{Rotacija_Z} that the total sum of generated forces is still the same which preserves the same vertical octocopter position. It can also be noticed that there is a full balance between $(F_{1}, F_{2}, F_{8})$ and $(F_{4}, F_{5}, F_{6})$, and between $(F_{2}, F_{3}, F_{4})$ and $(F_{6}, F_{7}, F_{8})$, providing no additional rotations around $X$-axis and $Y$-axis, respectively. However, in order to rotate an octocopter system only around $Z$-axis but in the opposite direction, one should apply the opposite logic with respect to P and N types of rotors. 

%Označimo s $\psi$ rotaciju oko $Z$ osi (\emph{engl.} $yaw$). \color{red} It is necessary to find a combination of generated forces that will not lead to an overall lifting of the UAV or to a rotation about the $X$-body and $Y$-body axes. In order to rotate the octocopter system about the $Z$-body axis, it is possible increase (or decrease) the DC motor speed 2, 4, 6, and 8 with simultaneous decrease (increase) of DC motor speed 1., 3., 5.
%and 7 (Fig. \ref{Rotacija_Z})..

%In this way, rotation about the $Z$ axis for the desired value is achieved angle, where there is no change in the total force of vertical lifting so  that  the  octocopter  remains  at  the same height, while the UAV rotate around the $Z$-body axis. Rotations around the $X$-body and $Y$-body axes are not generated because the generated forces are completely balanced.

Now, it is possible to describe the octocopter system motion towards a given reference point in space, as illustrated in the following subsections.
\color{black}

\section{Frames of reference}

\begin{figure}[!b]
\centering
\includegraphics[width=0.9\textwidth, height=0.4\textheight]{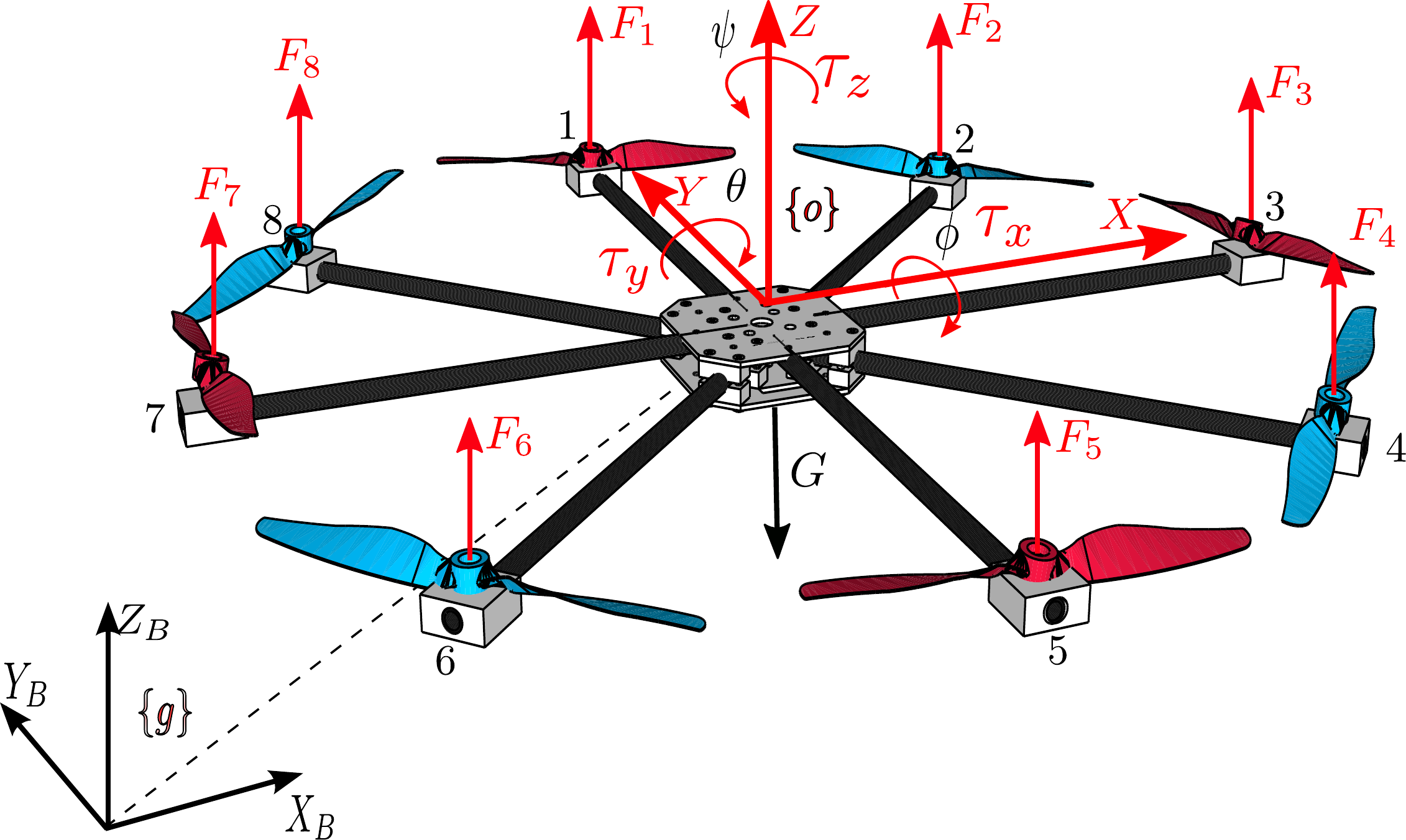}
\caption{Body and ground fixed frames for a PNPNPNPN octocopter system, where P and N indicate counter and clockwise directions depicted with blue and red colors respectively
 \cite{osmic_SMC}.} 
\label{fig:Frame}
\end{figure}

The octocopter system based on the PNPNPNPN configuration design, its body (local) and ground (global) fixed frames are shown in Fig. \ref{fig:Frame}. As earlier stated, each motor and its associated propeller included in the octocopter design is mounted on an arm of length $l$. The adjacent arms are equally distant from each other by $45^{\circ}$, that is ($360^{\circ}/n$), where $n=8$ is the total number of motors. 

Two reference frames are used to derive the model, one for a local coordinate system $\{o\}$ attached to the octocopter system and one representing a global coordinate system $\{g\}$ fixed to the ground. For $\{g\}$, the ENU convention is used to represent the axes, meaning that the axes $X_B$, $Y_B$ and $Z_B$ are pointing to the north, east and up, where $\textbf{\textit{x}} =[x\enspace y \enspace z]^{T}$ and $\boldsymbol{\Psi} = [\phi \enspace \theta \enspace \psi]^T$ indicate the position and the Euler-based orientation. The linear velocity $\boldsymbol{v} = [u \enspace v \enspace w]^T$ and the angular velocity $\boldsymbol{P} = [P \enspace Q \enspace R]^T$ are represented in $\{o\}$. The positive directions of $\phi$, $\theta$ and $\psi$ are chosen to coincide with the positive directions of $P$, $Q$ and $R$, respectively.

Rotations around the axes of the local coordinate system in terms of minimal representation can be described by the Euler angles $\phi$, $\theta$ and $\psi$ which are also known as roll, pitch and yaw, respectively. These are elementary rotations and can be expressed with the rotation matrices \cite{Sicilijano_Robotika}:
\begin{align}
\boldsymbol{R}(X, \phi) &= \begin{bmatrix}
1 & 0 & 0 \\
0 & c_{\phi} & -s_{\phi} \\
0 & s_{\phi} & c_{\phi} 
\end{bmatrix}, 
\label{Rx}\\
\boldsymbol{R}(Y, \theta) &= \begin{bmatrix}
c _{\theta} & 0 & s _{\theta}\ \\
0 & 1 & 0 \\
-s _{\theta} & 0 & c _{\theta} 
\end{bmatrix},
\label{Ry}\\
\boldsymbol{R}(Z, \psi) &= \begin{bmatrix}
c _{\psi} & -s _{\psi} & 0 \\
s _{\psi} & c _{\psi} & 0 \\
0 & 0 & 1 
\end{bmatrix},
\label{Rz}
\end{align}
where the positive directions of the roll, pitch and yaw are defined by the right hand rule.

If we assume that the local and global coordinate systems are perfectly aligned for $(\phi, \theta, \psi) = (0, 0, 0)$, then any vector $\boldsymbol{p}^o = \begin{bmatrix} p^o_x & p^o_y & p^o_z\end{bmatrix}^T$ given in the reference frame $\{o\}$ can be expressed in terms of the reference frame $\{g\}$ as a vector $\boldsymbol{p}^g = \begin{bmatrix}p^g_x & p^g_y & p^g_z\end{bmatrix}^T$, that is
\begin{equation}
\boldsymbol{p}^g = \boldsymbol{R}^g_o \boldsymbol{p}^o,
\label{euler_rotacije}
\end{equation}
where $\boldsymbol{R}^g_o$ is a rotation matrix describing the total rotation between the local and global coordinate systems. Finally, if we assume the $ZYX$ convention then the rotation matrix is given by
\begin{align}
\boldsymbol{R}^{g}_o &= \boldsymbol{R}(\phi, \theta, \psi) = \boldsymbol{R}(Z, \psi) \boldsymbol{R}(Y, \theta) \boldsymbol{R}(X, \phi) \nonumber \\
&= \begin{bmatrix}
c_{\psi} \, c_{\theta} & c_{\psi} \, s_{\theta} \, s_{\phi} - s_{\psi} \, c_{\phi} & c_{\psi} \, s_{\theta} \, c_{\phi} + s_{\psi} \, s_{\phi} \\
s_{\psi} \, c_{\theta} & s_{\psi} \, s_{\theta} \, s_{\phi} + c_{\psi} \, c_{\phi} & s_{\psi} \, s_{\theta} \, c_{\phi} - c_{\psi} \, s_{\phi} \\
-s_{\theta} & c_{\theta} \, s_{\phi} & c_{\theta} \, c_{\phi}
\end{bmatrix}.
\end{align}

\color{black}
\section{Forces, moments and control inputs}

The forces and torques acting on the system are shown in Fig. \ref{fig:Frame}. There are only two types of forces acting on the system, i.e. thrust $T$ and gravitational force $G$. Based on a static approximation, the motor thrust can be computed as follows \cite{mahony2012multirotor}:
\begin{equation}
T = \sum_{i = 1}^8 F_i = b \sum_{i = 1}^8 \Omega_i^2,
\end{equation} 
where $b \, [Ns^2/rad^2]$ is the motor thrust constant, while $\Omega_i \, [rad/s]$ and $F_i \, [N]$ are the angular velocity and the thrust force associated with the $i^{th}$ motor, respectively. If we denote the octocopter mass with $m_o$, then the gravitational force can be expressed as $G = m_o g$, where $g \approx 9.81 \, [m/s^2]$ is the gravitional acceleration of Earth. The thrust $T$ acts along the $Z$ direction of the local coordinate system and therefore can be written as the vector $\boldsymbol{T}^o = \begin{bmatrix}0 & 0 & T\end{bmatrix}^T$, while the gravitational force $G$ acts along the $Z_B$ direction of the global coordinate system and can be written as the vector $\boldsymbol{G}^g = \begin{bmatrix}0 & 0 & G\end{bmatrix}^T$. 

Let the torques acting around the $X$, $Y$ and $Z$ axes of the local coordinate system be denoted as $\tau_x$, $\tau_y$ and $\tau_z$, respectively (see Fig. \ref{fig:Frame}). Then, we assume that the distances from the octocopter center of mass from the center of mass of each single motor are equal and denoted by $l$. Now, the torques around the $X$ and $Y$ axes can be computed as:

\begin{align}
\tau_{x} &= l \left(F_1 + \frac{\sqrt{2}}{2} F_2 + \frac{\sqrt{2}}{2} F_8 - F_5 - \frac{\sqrt{2}}{2} F_4 - \frac{\sqrt{2}}{2} F_6 \right) \nonumber \\
&= b l \left(\Omega_1^2 + \frac{\sqrt{2}}{2} \Omega_2^2 + \frac{\sqrt{2}}{2} \Omega_8^2 - \Omega_5^2 - \frac{\sqrt{2}}{2} \Omega_4^2 - \frac{\sqrt{2}}{2} \Omega_6^2 \right), \nonumber \\
\tau_{y} &= l \left(F_7 + \frac{\sqrt{2}}{2} F_6 + \frac{\sqrt{2}}{2} F_8 - F_3 - \frac{\sqrt{2}}{2} F_2 - \frac{\sqrt{2}}{2} F_4 \right) \nonumber \\
&= b l \left(\Omega_7^2 + \frac{\sqrt{2}}{2} \Omega_6^2 + \frac{\sqrt{2}}{2} \Omega_8^2 - \Omega_3^2 - \frac{\sqrt{2}}{2} \Omega_2^2 - \frac{\sqrt{2}}{2} \Omega_4^2 \right). \nonumber
\end{align}

The angular motion of any motor included in the design causes a drag moment which is opposite to the direction of the motion according to Newton's third law. Therefore, if we also assume a static approximation of the drag \cite{mahony2012multirotor}, we can model the torque around the $Z$ axis as
\begin{align}
\tau_{z} &= -M_1 + M_2 - M_3 + M_4 - M_5 + M_6 - M_7 + M_8 \nonumber \\
		   &= d \left(-\Omega_1^2 + \Omega_2^2 - \Omega_3^2 + \Omega_4^2 - \Omega_5^2 + \Omega_6^2 - \Omega_7^2 + \Omega_8^2\right),
\end{align}
where $d \, [N m s^2/rad^2]$ is the rotor drag constant and $M_i \, [Nm]$ ($i = \overline{1..8}$) is the drag moment of the $i^{th}$ motor.

In an octocopter system, the algebraic sum
\begin{equation}
W_{G}=-\Omega_{1}+\Omega_{2}-\Omega_{3}+\Omega_{4}-\Omega_{5}+\Omega_{6}-\Omega_{7}+\Omega_{8} \label{eq:ziroskopska_smjetnja}
\end{equation}
of the angular velocities of all eight rotors is usually kept equal to zero. However, if $W_{G}\neq 0$ an additional moment $M_g$ appears caused by gyroscopic effect. 

Gyroscopic effect appears when the octocopter rotates around an axis (the spin axis in Fig. 2.8) during which there exist a force which is perpendicular to the plane of rotation causing the related moment along the input axis (Fig. 2.8). This results in a precession movement which rotates the octocopter around the output axis (Fig. 2.8).  
\medskip{}
\begin{figure}[!tbph]
\begin{centering}
\includegraphics[width=0.8\textwidth, height=0.3\textheight]{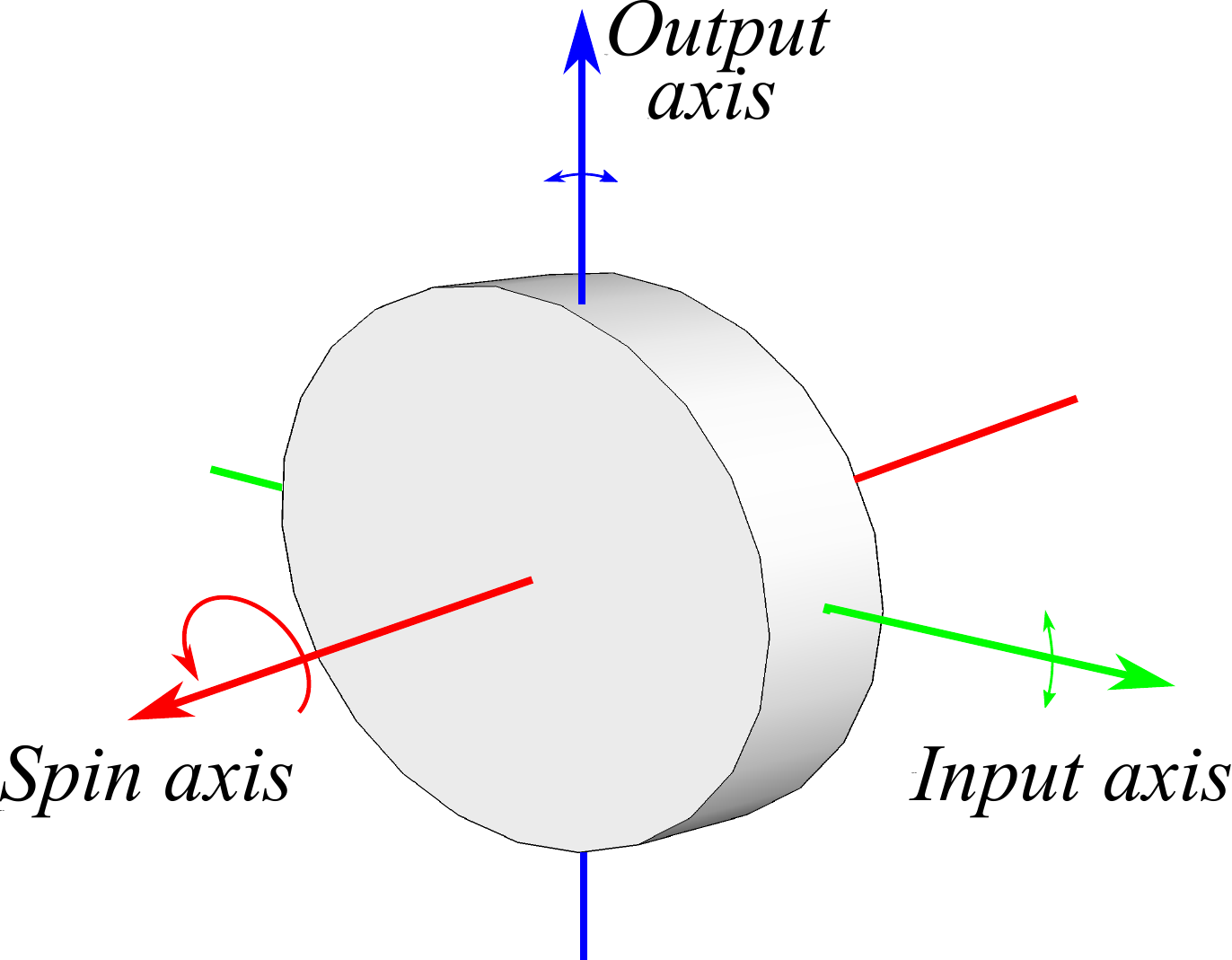}
\par\end{centering}
\caption{Gyroscopic effect.}
\label{Ziroskopski}
\end{figure}
\medskip{}

\color{black}
The gyroscopic moment is proportional to the sum $W_g$ \cite{Paralelni_thorem_morin2008introduction} and can be written as
\color{black}
\begin{equation}
\scriptstyle
M_{G}=\omega\times\left(J_{G}\cdot\left[\begin{array}{c}
0\\
0\\
\scriptstyle \sum (-1)^{i}\cdot\Omega_{i}
\end{array}\right]\right)=\omega\times\left[\begin{array}{ccc}
\scriptstyle I_{XXm} & 0 & 0\\
0 & \scriptstyle I_{YYm} & 0\\
0 & 0 & \scriptstyle I_{ZZm}
\end{array}\right]\cdot\left[\begin{array}{c}
0\\
0\\
\scriptstyle \sum(-1)^{i}\cdot\Omega_{i}
\end{array}\right],\label{eq:Ziroskopsi_efekt_I-1}
\normalsize
\end{equation}
where $M_{G}$, $\omega$, $J_{G}$ and $\Omega_i$ indicate the resulting moment,  the octocopter angular velocity, the motor inertia and the rotational velocity of the motor $M_i$, respectively. Since $\omega=\left[\begin{array}{ccc}
P & Q & R\end{array}\right]^{T}$, (\ref{eq:Ziroskopsi_efekt_I-1}) can be written in a more compact form as 
\begin{equation}
M_{G}=\left[\begin{array}{c}
P\\
Q\\
R
\end{array}\right]\times\left[\begin{array}{c}
0\\
0\\
I_{ZZm}\cdot W_{G}
\end{array}\right].\label{eq:Ziroskopsk_konacna}
\end{equation}
If we now use the skew-symmetric vector operator $S(\cdot)$ (see, e.g. \cite{Cross_produkt_Gilbert_Strang})
 which transforms a vector $a=[\begin{array}{ccc}
a_{1} &  a_{2} &  a_{3}\end{array}]^{T}$ into a skew-symmetric matrix 
\begin{equation}
S(a)=-S^{T}(a)=\left[\begin{array}{ccc}
0 & -a_{3} & a_{2}\\
a_{3} & 0 & -a_{1}\\
-a_{2} & a_{1} & 0
\end{array}\right],\label{eq:Skew_simetric-matrix-1}
\end{equation}
then the outer product (\ref{eq:Ziroskopsk_konacna}) can be written as a matrix product
\begin{equation}
M_{G}=\left[\begin{array}{ccc}
0 & R & -Q\\
-R & 0 & P\\
Q & -P & 0
\end{array}\right]\cdot\left[\begin{array}{c}
0\\
0\\
\scriptstyle I_{ZZm}\cdot W_{G}
\end{array}\right]=\left[\begin{array}{c}
Q\cdot \scriptstyle I_{ZZm}\cdot W_{G}\\
-P\cdot \scriptstyle I_{ZZm}\cdot W_{G}\\
0
\normalsize
\end{array}\right].
\label{eq:Ziroskopsk_konacna-1}
\end{equation}

\begin{figure}[!tbph]
\begin{centering}
\includegraphics{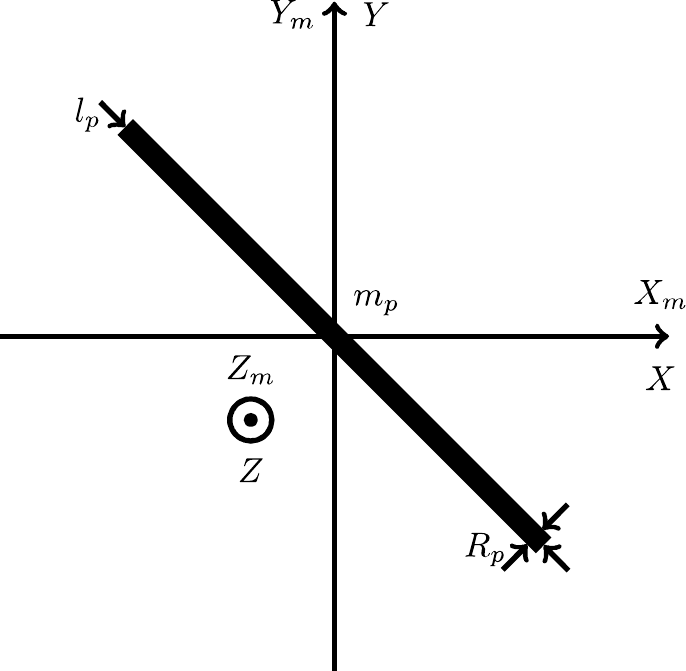}
\par\end{centering}
\caption{Simplified rotor geometry.\label{fig:Propeler}}

\end{figure}

It should be noted that $M_g \neq 0$ only when the yawing and rolling occur simultaneously. However, the moment caused by gyroscopic effect can be often neglected in comparison to other moments and forces generated in the octocopter system. We also use a simplified rotor geometry as shown in Fig. \ref{fig:Propeler} to derive the moment of inertia $I_{zzm}$. We assume that the rotor can be modelled as an infinitely thin rod ($R_p \to 0$) of length $l_p$ and mass $m_p$. We also assume a unit transmission between the motor shaft and the rotor and neglect the moment of inertia of the motor shaft due to its relative small size with regard to the rotor geometry. Therefore, the inertia moment $I_{zzm}$ is described as \cite{Paralelni_thorem_morin2008introduction}:
\begin{equation}
I_{zzm} = \frac{m_p l_p^2}{12}.
\end{equation}

In order to control the octocopter system, we generate different thrust and torque values by changing the angular velocity of the rotors. Therefore, if we define the virtual control vector $\boldsymbol{u} = \begin{bmatrix} T & \boldsymbol{\tau} \end{bmatrix}^T = \begin{bmatrix} T & \tau_x & \tau_y & \tau_z \end{bmatrix}^T$ and the actuation matrix $\boldsymbol{A}$ as:

\begin{equation}
\boldsymbol{A}=\begin{bmatrix}
b & b & b & b & b & b & b & b \\
bl & \frac{\sqrt{2}}{2} bl & 0 & -\frac{\sqrt{2}}{2} bl & -bl & -\frac{\sqrt{2}}{2} bl & 0 & \frac{\sqrt{2}}{2} bl\\
0 & -\frac{\sqrt{2}}{2} bl & -bl & -\frac{\sqrt{2}}{2} bl & 0 & \frac{\sqrt{2}}{2} bl & bl & \frac{\sqrt{2}}{2} bl\\
-d & d & -d & d & -d & d & -d & d \\
\end{bmatrix}, \nonumber
\label{eq:matrica_A}
\end{equation}

the system actuation  can be finally described as $\boldsymbol{u} = \boldsymbol{A} \boldsymbol{\Omega_s},$ where $\boldsymbol{\Omega_s}$ is the squared rotor velocity vector defined as:
\begin{equation}
\boldsymbol{\Omega_s} = \begin{bmatrix}\Omega_1^2 & \Omega_2^2 & \Omega_3^2 & \Omega_4^2 & \Omega_5^2 & \Omega_6^2 & \Omega_7^2 & \Omega_8^2\end{bmatrix}^T. 
\label{aktuacijska_m}
\end{equation}

Consequently, the control inputs of the system are represented with the motor speeds $\Omega_1$, $\Omega_2$, $\Omega_3$, $\Omega_4$, $\Omega_5$, $\Omega_6$, $\Omega_7$ and $\Omega_8$. Changing the rotor velocity of the motors in range $0\leq\varOmega_{i}\leq\Omega_{max}$, $i = \overline{1..8}$, the different thrust force ($T$) and the torque ($\tau_{x}$, $\tau_{y}$ and $\tau_{z}$) about the $x$, $y$ and $z$ axes can be achieved.

\medskip{}

\color{black}
In order to control the octocopter using rotor angular velocities as input variables, it is necessary to understand all forces that act on the system. Since each motor can have different angular velocity, different forces $F_{i}$ ($i=1,...,8,$) directed along positive direction of $Z$-axis and moments $\tau_{x},\tau_{y},\tau_{z}$ with respect to all three axes, $X$, $Y$, and $Z$ can be generated. Every generated force is proportional to the square of angular velocity, that is
\begin{equation}
F_{i}=b\cdot\Omega_{i}^{2},\label{eq:producirane_sile}
\end{equation}
\label{eq:Sila_potiska}\medskip{}
where the coefficient $b$ can be experimentally found for each type of propellers. 

Alongside with the generated forces $F_{i}$, the gravitational force also acts on the system which is balanced with the thrust force generated with the rotor movements, where the thrust force can be computed as
\begin{equation}
T=\sum\limits _{i=1}^{8}F_{i}=\sum\limits _{i=1}^{8}b\cdot\Omega_{i}^{2}=b\cdot(\Omega_{1}^{2}+\Omega_{2}^{2}+\Omega_{3}^{2}+\Omega_{4}^{2}+\Omega_{5}^{2}+\Omega_{6}^{2}+\Omega_{7}^{2}+\Omega_{8}^{2}).\label{eq:Sila_potiska-1}
\end{equation}

\color{black}

\medskip{}

\medskip{}
 
\section{Octocopter kinematics}

Let the linear velocities along the $X$, $Y$ and $Z$ axes of the local coordinate system be denoted as $u$, $v$ and $w$, respectively. Let also the linear velocities along the $X_B$, $Y_B$ and $Z_B$ axes of the global coordinate system be denoted as $\dot{x}$, $\dot{y}$ and $\dot{z}$, respectively. By rewriting these velocities in a vector form as $\boldsymbol{v}^o = \boldsymbol{v} = \begin{bmatrix}u & v & w\end{bmatrix}^T$ and $\dot{\boldsymbol{x}}^g = \dot{\boldsymbol{x}} = \begin{bmatrix} \dot{x} & \dot{y} & \dot{z} \end{bmatrix}^T$, we can directly apply the rotation matrix $R^g_o$ to derive the kinematic model of the linear motion as
\begin{equation}
\dot{\boldsymbol{x}} = \boldsymbol{R}(\phi, \theta, \psi) \boldsymbol{v}.
\label{lin_kinematics}
\end{equation}

The derivation of the kinematic model of the angular motion is also straightforward but it has some subtleties that need to be discussed. First, let the angular velocities around the $X$, $Y$ and $Z$ axes of the local coordinate system be denoted as $P$, $Q$ and $R$, respectively. Recall that the total rotation of the local coordinate system with respect to the global coordinate system can be expressed in terms of the Euler angles $\phi$, $\theta$ and $\psi$. By calculating the derivative of the Euler angles, we get the rotational velocities $\dot{\phi}$, $\dot{\theta}$ and $\dot{\psi}$, which in general do not have a physical interpretation. If we rewrite the angular velocities in the vector form as $\boldsymbol{P}^o = \boldsymbol{P} = \begin{bmatrix} P & Q & R \end{bmatrix}^T$ and the rotational velocities in the vector form as $\boldsymbol{\Psi} = \begin{bmatrix}\phi & \theta & \psi\end{bmatrix}^T$, then we can calculate the contributions of each rotational velocity component to the angular velocity components as \cite{Sicilijano_Robotika}:
\begin{align}
\boldsymbol{P} = 
&\left(\boldsymbol{R}(Z, \psi) \boldsymbol{R}(Y, \theta) \boldsymbol{R}(X, \phi)\right)^{-1}
\begin{bmatrix}
0 \\
0 \\
\dot{\psi}
\end{bmatrix} + \nonumber \\
&\left(\boldsymbol{R}(Y, \theta) \boldsymbol{R}(X, \phi)\right)^{-1}
\begin{bmatrix}
0 \\
\dot{\theta} \\
0
\end{bmatrix} +  
\boldsymbol{R}^{-1}(X, \phi)
\begin{bmatrix}
\dot{\phi} \\
0 \\
0
\end{bmatrix} \nonumber \\
= &\boldsymbol{R}_A(\phi, \theta, \psi)
\begin{bmatrix}
\dot{\phi} \\
\dot{\theta} \\
\dot{\psi}
\end{bmatrix} = \boldsymbol{R}_A(\phi, \theta, \psi) \dot{\boldsymbol{\Psi}},
\end{align}
where $\boldsymbol{R}_A(\phi, \theta, \psi)$ is defined as 
\begin{equation}
\boldsymbol{R}_A(\phi, \theta, \psi) = \begin{bmatrix}
1 & 0 & -s_{\theta} \\ 
0 & c_{\phi} & s_{\phi}c_{\theta}	\\
0 & -s_{\phi} & c_{\phi}c_{\theta}
\end{bmatrix}.
\label{lin_kinematics_rot}
\end{equation}
By calculating the determinant of $\boldsymbol{R}_A(\phi, \theta, \psi)$, it can be shown that the inverse mapping is singular for $c_{\theta} = 0$, which yields $\theta = \pm \frac{\pi}{2}$ to be singular configurations. In real world applications, we measure the angular velocities $P$, $Q$ and $R$ with inertial measurement units (IMU) and try to estimate the Euler angles, therefore the inverse mapping  
\begin{equation}
\dot{\boldsymbol{\Psi}} = \boldsymbol{R}^{-1}_A(\phi, \theta, \psi) \boldsymbol{P},
\label{orien_kinematics}
\end{equation}
is more useful, where the matrix $\boldsymbol{R}^{-1}_A(\phi, \theta, \psi)$ is defined as
\begin{equation}
\boldsymbol{R}^{-1}_A(\phi, \theta, \psi) = \begin{bmatrix}
1 & s_{\phi}t_{\theta} & c_{\phi}t_{\theta} \\
0 & c_{\phi} & -s_{\phi}	\\
0 & \frac{s_{\phi}}{c_{\theta}} & \frac{c_{\phi}}{c_{\theta}}
\end{bmatrix}.
\label{orient_rot_matrix}
\end{equation}

\section{Octocopter dynamics}

In order to derive the dynamical model of the linear motion, we apply Newton's third law in the form
\begin{equation}
\boldsymbol{F}^o_{\text{net}} = m_o \boldsymbol{\dot{v}}^o_{\text{net}},
\label{newton_lin_01}
\end{equation}
where $\boldsymbol{F}^o_{\text{net}}$ is the net force acting on the system and $\boldsymbol{\dot{v}}^o_{\text{net}}$ is the net linear acceleration of the system with respect to the local reference frame $\{o\}$. The net force is the sum of the thrust and the gravitational force expressed in the local coordinate system, therefore we can write
\begin{align}
\boldsymbol{F}^o_{\text{net}} &= \boldsymbol{T}^o + \boldsymbol{G}^o = \boldsymbol{T}^o + \boldsymbol({R}^o_g)^{-1} \boldsymbol{G}^g = \\
&= \boldsymbol{T}^o + \left(\boldsymbol{R}(Z, \psi) \boldsymbol{R}(Y, \theta) \boldsymbol{R}(X, \phi)\right)^{-1} \boldsymbol{G}^g \nonumber \\
&= \begin{bmatrix}
0 \\
0 \\
T
\end{bmatrix} + m_o g 
\begin{bmatrix}
s_{\theta} \\
-s_{\phi}\;c_{\theta} \\
-c_{\phi}\;c_{\theta}
\end{bmatrix}.
\label{newton_lin_02}
\end{align}
Given that $\{o\}$ is a non-inertial reference frame, the net acceleration is calculated as \cite{Klasicna_mehanika}:
\begin{align}
\boldsymbol{\dot{v}}^o_{\text{net}} &= \boldsymbol{\dot{v}}^o + \boldsymbol{P}^o \times \boldsymbol{v}^o = \boldsymbol{\dot{v}} + \boldsymbol{P} \times \boldsymbol{v} =\boldsymbol{\dot{v}} + \boldsymbol{S} \boldsymbol{v},
\label{newton_lin_03}
\end{align}
where $\boldsymbol{S}$ is a skew-symmetric matrix defined as \cite{Sicilijano_Robotika}:
\begin{equation}
\boldsymbol{S} = 
\begin{bmatrix}
0 & -R & Q \\
R & 0 & -P \\
-Q & P & 0
\end{bmatrix}.
\label{skew}
\end{equation}
The resulting linear motion dynamics, obtained by combining the equations \eqref{newton_lin_01}, \eqref{newton_lin_02} and \eqref{newton_lin_03}, can by written in the form
\begin{align}
\boldsymbol{\dot{v}} &= \frac{\boldsymbol{F}^o}{m_o} - \boldsymbol{S} \boldsymbol{v} 
= \begin{bmatrix}
0 \\
0 \\
\frac{T}{m_o}
\end{bmatrix} + g 
\begin{bmatrix}
s_{\theta} \\
-s_{\phi}\;c_{\theta} \\
-c_{\phi}\;c_{\theta}
\end{bmatrix}
- \boldsymbol{S} \boldsymbol{v}. 
\label{dinamika_01}
\end{align}

In the same manner, to derive the dynamic model of the angular motion we apply Newton's third law in the form
\begin{equation}
\boldsymbol{\tau}^o_{\text{net}} = \boldsymbol{J} \boldsymbol{\dot{\omega}}^o_{\text{net}},
\label{newton_ang_01}
\end{equation}
where $\boldsymbol{\tau}^o_{\text{net}}$ is the net torque acting on the system and $\boldsymbol{\dot{\omega}}^o_{\text{net}}$ is the net angular acceleration of the system with respect to the reference frame $\{o\}$. $\boldsymbol{J}$ is the inertia tensor of the octocopter and, based on the fact that $X$, $Y$ and $Z$ are principle axes of intertia, can be written in the form:
\begin{equation}
\boldsymbol{J} =
\begin{bmatrix}
I_{xx} & 0 & 0 \\
0 & I_{yy} & 0 \\
0 & 0 & I_{zz}
\end{bmatrix},
\end{equation}
where $I_{xx}$, $I_{yy}$ and $I_{zz}$ are the moments of inertia around the $X$, $Y$ and $Z$ axes, respectively. The net torque acting on the system is described by $\boldsymbol{\tau}^o_{\text{net}} = \boldsymbol{\tau}$ and the net angular acceleration is given as \cite{Klasicna_mehanika}:
\begin{align}
\boldsymbol{\dot{\omega}}^o_{\text{net}} &= \boldsymbol{\dot{P}}^o + \boldsymbol{J}^{-1} \boldsymbol{P}^o \times \boldsymbol{J} \boldsymbol{P}^o = \boldsymbol{\dot{P}} + \boldsymbol{J}^{-1} \boldsymbol{P} \times \boldsymbol{J} \boldsymbol{P} \nonumber \\
&= \boldsymbol{\dot{P}} + \boldsymbol{J}^{-1} \boldsymbol{S} \boldsymbol{J} \boldsymbol{P}.
\label{newton_ang_02}
\end{align}
The resulting angular motion dynamics can be obtained by combining the equations \eqref{newton_ang_01} and \eqref{newton_ang_02} and can by written as
\begin{equation}
\boldsymbol{\dot{P}} = \boldsymbol{J}^{-1} \left(\boldsymbol{\tau} - \boldsymbol{S} \boldsymbol{J} \boldsymbol{P}\right).
\label{angular_motion}
\end{equation}

If we also want to include the Gyroscopic effect into the dynamic model of the angular motion, we have to add the gyroscopic term \cite{johansen2013control} 
\begin{equation}
-\boldsymbol{P} \times \boldsymbol{J}_m \begin{bmatrix}
0 \\
0 \\
W_g
\end{bmatrix},
\end{equation}
where $W_g$ is the difference of the rotor velocities given as \eqref{eq:ziroskopska_smjetnja} and $\boldsymbol{J}_m$ is the inertia tensor of the rotors given as
\begin{equation}
\boldsymbol{J}_m =
\begin{bmatrix}
I_{xxm} & 0 & 0 \\
0 & I_{yym} & 0 \\
0 & 0 & I_{zzm}
\end{bmatrix}.
\end{equation}
The $I_{xxm}$, $I_{yym}$ and $I_{zzm}$ are the moments of inertia around the axes of the rotor reference frame. The gyroscopic term can be simplified to 
\begin{equation}
-\boldsymbol{P} \times \boldsymbol{J}_m \begin{bmatrix}
0 \\
0 \\
W_g
\end{bmatrix} = -\boldsymbol{S} \begin{bmatrix}
0 \\
0 \\
I_{zzm} W_g
\end{bmatrix},
\end{equation}
and finally, in order to include Gyroscopic effect, the angular motion dynamics can be rewritten as
\begin{equation}
\boldsymbol{\dot{P}} = \boldsymbol{J}^{-1} \left(\boldsymbol{\tau} - \boldsymbol{S} \boldsymbol{J} \boldsymbol{P} - \boldsymbol{S} \begin{bmatrix}
0 \\
0 \\
I_{zzm} W_g
\end{bmatrix} \right).
\label{Dinamika_rotacija}
\end{equation}

\begin{figure}[!tbph]
\begin{centering}
\includegraphics[width=8cm,height=8cm]{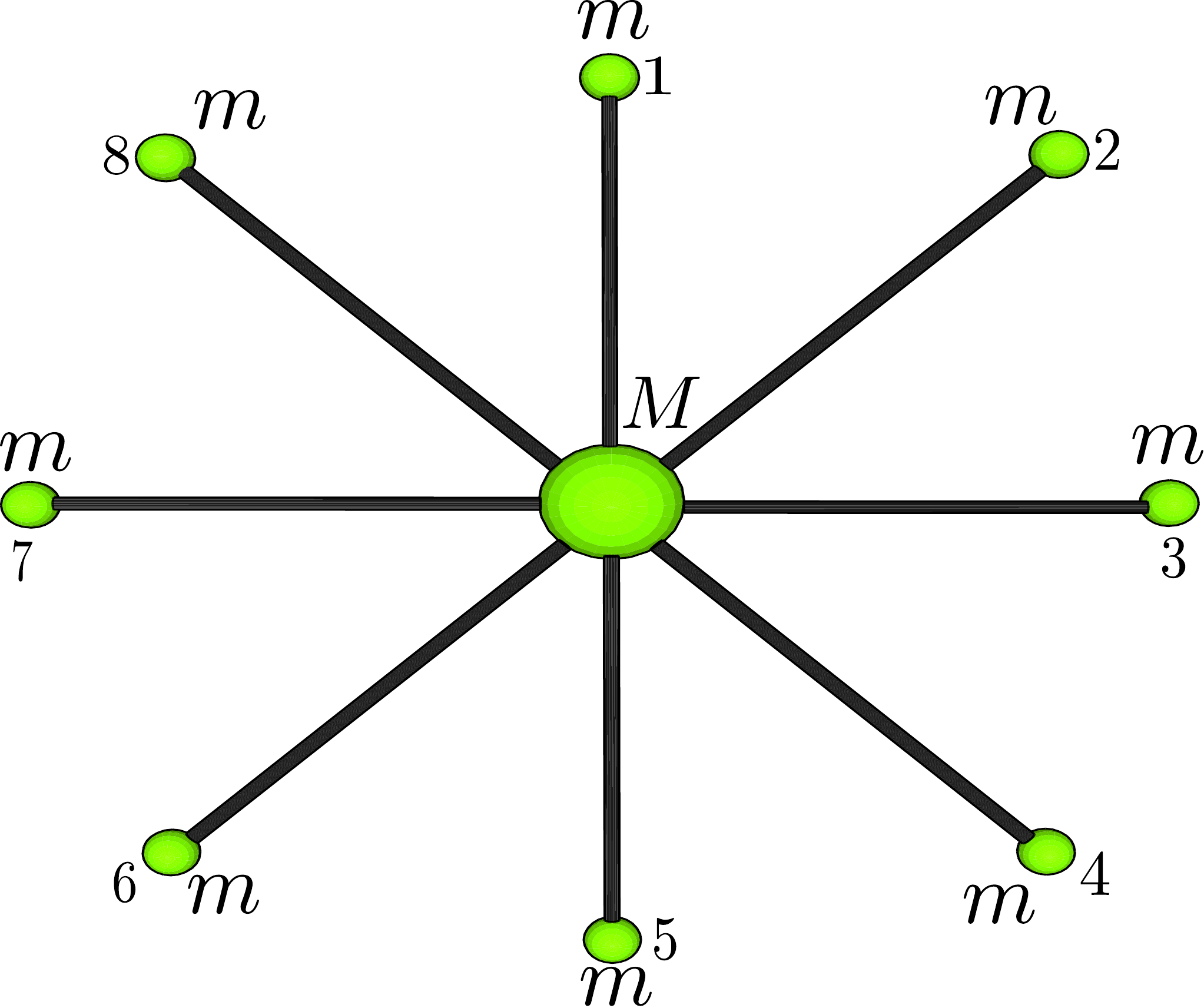}
\par\end{centering}
\caption{A simplified octocopter configuration structure.}
\label{mase_sustava}
\end{figure}

In order to model the inertia moments $I_{xx}$, $I_{yy}$ and $I_{zz}$ we use the parallel axis theorem (also called Huygens–Steiner theorem) \cite{Paralelni_thorem_morin2008introduction}. We assume a simplified octocopter confiuration structure as shown in Fig. \ref{mase_sustava}, where the motors are particles with mass $m$, the support plate with the mounted hardware is a solid sphere of radius $r$ and mass $M$, while the mass of the rotors is considered negligible.

Therefore, the inertia moments $I_{xx}$ and $I_{yy}$ are given as
\begin{align}
I_{xx} = I_{yy} &= \frac{2 M r^2}{5} + 2 m l^2 + 4 \left(\frac{\sqrt{2}}{2} l\right)^2 \nonumber \\
&= \frac{2 M r^2}{5} + 4 m l^2,
\end{align}
while the inertia moment $I_{zz}$ is given by
\begin{equation}
I_{zz} = \frac{2 M r^2}{5} + 8 m l^2.
\end{equation}

\section{Motor dynamics}

\begin{figure}[!tbph]
\centering{}\includegraphics{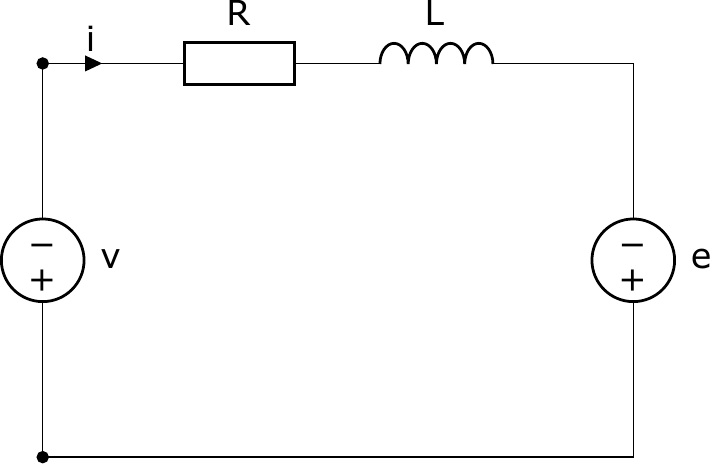}
\caption{Equivalent electric circuit of the armature.
\label{fig:DC_motor}}
\end{figure}

The actuator used to drive propellers mounted on an cotocopter system (or any other MAV) is usually a DC motor which can be modelled with (\ref{eq:Jed_DC_motor}) and represented in Fig.  \ref{fig:DC_motor}
\begin{equation}
v=Ri+L\frac{di}{dt}+e,\label{eq:Jed_DC_motor}
\end{equation}
where $R$ is the resistance and $L$ is the inductance of the motor \color{black} windings, while $v$ is the voltage applied to the motor, $i$ is the motor current and $e$ is the counter-electromotive force induced into the motor windings.
This can be simplified to the algebraic model $v = R i + e$ if we neglect inductive losses of the motor armature. This is a reasonable assumption based on the fact that DC motors used in robotics are in general constructed to minimize the inductive losses. We can rewrite the counter-electromotive force as $e = K_e \Omega$, where $K_e \, [Vs/rad]$ is the so-called electrical motor constant and $\Omega$ is the angular velocity of the motor (which is in our case also the angular velocity of the rotor). Therefore, the electrical model of the motor can by written in the form:
\begin{equation}
v=Ri+K_{e}\Omega.\label{eq:napon rotorskog kruga}
\end{equation}

The dynamic model of the rotor angular motion can be obtained by applying Newton's second law \cite{Klasicna_mehanika} in the form:
\begin{equation}
I_{zzm}\dot{\Omega}=\tau_{m}-\tau_{l},\label{eq:rot_kretanje}
\end{equation}
where $I_{zzm}$ is moment of inertia of the motor, $\tau_m$ and $\tau_l$ are the motor and load torque, respectively. We can rewrite the motor torque as $\tau_m = K_m i$,
where $K_m \, [N m/A]$ is the so-called mechanical motor constant. The resulting motor dynamics can be obtained by combining (\ref{eq:napon rotorskog kruga})
i (\ref{eq:rot_kretanje}) with the previously defined motor torque term $\tau_m$ in the form
\begin{equation}
I_{zzm}\dot{\Omega} + \frac{K_m K_e}{R} \Omega = \frac{K_m}{R} v - \tau_l.
\label{motor_1}
\end{equation}
By treating the DC motor as an isolated system and writing out the power balance equations, it is possible to show that the electrical and mechanical motor constants have the same numeric values given in different units. Another subtlety we need to include in the model is the motor voltage saturation given as $0 \leq v \leq v_{\text{max}}$.

Finally, assuming that all motors have the same parameter values $J_{m},$ $K_{m},$ $K_{e}$ i $R$, we can model the octocopter motor dynamics as 
\begin{equation}
I_{zzm}\dot{\Omega}_i + \frac{K_m K_e}{R} \Omega_i = \frac{K_m}{R} v_i - \tau_{li}, \; \; i = \overline{1..8},
\label{motor_2}
\end{equation}
where the load torque is the drag moment given as
\begin{equation}
\tau_{li} = d \Omega_i^2, \; \; i = \overline{1..8},
\label{aerodinamicki}
\end{equation}
and $d$ being the rotor drag constant.

\color{black}
\section{State space model}
In this section we provide the complete state space model of an octocopter which includes linear and angular velocities with respect to the local coordinate system, $(u, v, w)$ and $(P,Q,R)$, Euler angles $\phi,\theta,\psi$ and the octocopter position $(x, y, z)$ with respect to the global reference system. 

The resulting force ${F^o}$ which acts on the octocopter consists of the gravitational force $F_G^B$ presented in the global coordinate system and the thrust which is generated as a sum of all forces caused by rotation of motors, that is
\[
{F}^o=F_{G}^{B}+T.
\]
From (\ref{dinamika_01}), one can obtain
\begin{equation}
\dot{v}=-Sv+\frac{F_{G}^{B}}{m}+\frac{T}{m},\label{eq:vremesnki_clan_III}
\end{equation}

which can be written in the form
\begin{equation}
\left[\begin{array}{c}
\dot{u}\\
\dot{v}\\
\dot{w}
\end{array}\right]=-
\begin{bmatrix}
\scriptstyle 0 &  \scriptstyle -R &  \scriptstyle Q \\
 \scriptstyle R & \scriptstyle 0 &  \scriptstyle -P \\
 \scriptstyle -Q & \scriptstyle  P & \scriptstyle 0
\end{bmatrix}.
\left[\begin{array}{c}
u\\
v\\
w
\end{array}\right]+\left[\begin{array}{c}
g\cdot s_{\theta}\\
-g\cdot c_{\theta}\cdot s_{\phi}\\
-g\cdot c_{\theta}\cdot c_{\phi}
\end{array}\right]+\dfrac{1}{m}\left[\begin{array}{c}
\scriptstyle F_{X}\\
\scriptstyle F_{Y}\\
\scriptstyle F_{Z}
\end{array}\right].\label{eq:Vremsnki_clan_IV}
\end{equation}
If we assume that the thrust force acts only along the direction of $Z$-axis, it yields

\begin{equation}
\left[\begin{array}{c}
\dot{u}\\
\dot{v}\\
\dot{w}
\end{array}\right]=-\left[\begin{array}{ccc}
\scriptstyle 0 & \scriptstyle R & \scriptstyle -Q\\
\scriptstyle -R & \scriptstyle 0 & \scriptstyle P\\
\scriptstyle Q & \scriptstyle -P & \scriptstyle 0
\end{array}\right]\cdot\left[\begin{array}{c}
u\\
v\\
w
\end{array}\right]+\left[\begin{array}{c}
g\cdot s_{\theta}\\
-g\cdot c_{\theta}\cdot s_{\phi}\\
-g\cdot c_{\theta}\cdot c_{\phi}
\end{array}\right]+\left[\begin{array}{c}
0\\
0\\
 \dfrac{\scriptstyle F_{Z}}{\scriptstyle m}
\end{array}\right],\label{eq:vremenski clan V}
\end{equation}
that is
\label{eq:rotacijski_clan_II}
\begin{equation}
\left[\begin{array}{c}
\dot{u}\\
\dot{v}\\
\dot{w}
\end{array}\right]=\left[\begin{array}{c}
R\cdot v-Q\cdot w\\
P\cdot w-R\cdot u\\
Q\cdot u-P\cdot v
\end{array}\right]+\left[\begin{array}{c}
g\cdot s_{\theta}\\
-g\cdot c_{\theta}\cdot s_{\phi}\\
-g\cdot c_{\theta}\cdot c_{\phi}
\end{array}\right]+\left[\begin{array}{c}
0\\
0\\
\dfrac{T}{m}
\end{array}\right].\label{eq:Vremenski_clan_konacno}
\end{equation}

From (\ref{Dinamika_rotacija})\color{black}, one can obtain
\begin{equation}
%\boldsymbol{\dot{P}}=J^{-1}\cdot\left\{S\cdot\left\ J\cdot \boldsymbol{P} +\right\tau\right\}, 
\boldsymbol{\dot{P}} = \boldsymbol{J}^{-1} \left(\boldsymbol{\tau} - \boldsymbol{S} \boldsymbol{J} \boldsymbol{P}\right).
\label{eq:rotacijski clan II}
\end{equation}

that is

\begin{equation}
\scriptstyle
\left[\begin{array}{c}
\dot{\scriptstyle P}\\
\dot{\scriptstyle Q}\\
\dot{\scriptstyle R}
\end{array}\right]=J^{-1}\cdot\left\{ \left[\begin{array}{ccc}
\scriptstyle 0 & \scriptstyle R & \scriptstyle -Q\\
\scriptstyle -R & \scriptstyle 0 & \scriptstyle P\\
\scriptstyle Q & \scriptstyle -P & \scriptstyle 0
\end{array}\right]\cdot\left[\begin{array}{ccc}
\scriptstyle I_{XX} & 0 & 0\\
0 & \scriptstyle I_{YY} & 0\\
0 & 0 & \scriptstyle I_{ZZ}
\end{array}\right]\cdot\left[\begin{array}{c}
P\\
Q\\
R
\end{array}\right]+\left[\begin{array}{c}
\tau_{x}\\
\tau_{y}\\
\tau_{z}
\end{array}\right]\right\}, \label{eq:rotacijski_clan_III}
\end{equation}
which gives the form
\begin{equation}
\begin{array}{ccl}
\left[\begin{array}{c}
\dot{P}\\
\dot{Q}\\
\dot{R}
\end{array}\right] & = & \left[\begin{array}{ccc}
\scriptstyle I_{XX}^{-1} & 0 & 0\\
0 & \scriptstyle I_{YY}^{-1} & 0\\
0 & 0 & \scriptstyle I_{ZZ}^{-1}
\end{array}\right]\cdot\left\{ \left[\begin{array}{c}
(\scriptstyle I_{YY}-\scriptstyle I_{ZZ})\cdot QR\\
(\scriptstyle I_{ZZ}-\scriptstyle I_{XX})\cdot PR\\
(\scriptstyle I_{XX}- \scriptstyle I_{YY})\cdot PQ
\end{array}\right]+\left[\begin{array}{c}
\tau_{x}\\
\tau_{y}\\
\tau_{z}
\end{array}\right]\right\} \\
\\
 & = & \left[\begin{array}{c}
\dfrac{\scriptstyle I_{YY}- \scriptstyle I_{ZZ}}{\scriptstyle I_{XX}}\cdot \scriptstyle QR\\
\dfrac{\scriptstyle I_{ZZ}-\scriptstyle I_{XX}}{\scriptstyle I_{YY}}\cdot \scriptstyle PR\\
\dfrac{\scriptstyle I_{XX}-\scriptstyle I_{YY}}{\scriptstyle I_{ZZ}}\cdot \scriptstyle PQ
\end{array}\right]+\left[\begin{array}{c}
\dfrac{\tau_{x}}{\scriptstyle I_{XX}}\\
\dfrac{\tau_{y}}{\scriptstyle I_{YY}}\\
\dfrac{\tau_{z}}{\scriptstyle I_{ZZ}}
\end{array}\right].
\end{array}\label{eq:Rotacijski_clan_konacno}
\end{equation}
Furthermore, from (\ref{euler_rotacije}) which indicates the relation between Euler angles and angular velocities, one obtains 
\begin{equation}
\begin{array}{ccl}
\dot{\phi} & = & P+s_{\phi}\cdot t_{\theta}\cdot Q+c_{\phi}\cdot t_{\theta}\cdot R\\
\dot{\theta} & = & c\cdot Q-s_{\phi}\cdot R\\
\dot{\psi} & = & \dfrac{s_{\phi}}{c_{\theta}}\cdot Q+\dfrac{c_{\phi}}{c_{\theta}}\cdot R,
\end{array}.\label{eq:izvodi eulera i ugaone}
\end{equation}
providing the relations between the time derivatives of Euler angles from which one can compute the system orientation $\phi,\theta,\psi$ at each time instant.

Finally, by taking into account the previously derived relations (\ref{lin_kinematics}) and (\ref{lin_kinematics_rot}) one obtains the full state space model of the octocopter system as follows
\medskip{}

\begin{equation}
\begin{array}{ccl}
\dot{x} & = & c_{\theta}\cdot c_{\psi}\cdot u+\left(s_{\phi}\cdot s_{\theta}\cdot c_{\psi}-c_{\phi}\cdot s_{\psi}\right)\cdot v+\left(c_{\phi}\cdot s_{\phi}\cdot c_{\psi}+s_{\phi}\cdot s_{\psi}\right)\cdot w\\
\dot{y} & = & c_{\theta}\cdot s_{\psi}\cdot u+\left(s_{\phi}\cdot s_{\theta}\cdot s_{\psi}+c_{\phi}\cdot c_{\psi}\right)\cdot v+\left(c_{\phi}\cdot s_{\phi}\cdot s_{\psi}-s_{\phi}\cdot c_{\psi}\right)\cdot w\\
\dot{z} & = & -s_{\theta}\cdot u+c_{\theta}\cdot s_{\phi}\cdot v+c_{\theta}\cdot c_{\phi}\cdot w\\
\dot{u} & = & R\cdot v-Q\cdot w+g\cdot s_{\theta}\\
\dot{v} & = & P\cdot w-R\cdot u-g\cdot c_{\theta}\cdot s_{\phi}\\
\dot{w} & = & Q\cdot u-P\cdot v-g\cdot c_{\theta}\cdot c_{\phi}+\frac{T}{m}\\
\dot{P} & = & \dfrac{\scriptstyle I_{YY}- \scriptstyle I_{ZZ}}{\scriptstyle I_{XX}}\cdot Q\cdot R+\dfrac{\tau_{x}}{\scriptstyle I_{XX}}-\dfrac{\scriptstyle I_{ZZM}}{\scriptstyle I_{XX}}\cdot Q\cdot W_{G}\\
\dot{Q} & = & \dfrac{\scriptstyle I_{ZZ}-\scriptstyle I_{XX}}{\scriptstyle I_{YY}}\cdot P\cdot R+\dfrac{\tau_{y}}{I_{YY}}+\dfrac{\scriptstyle I_{ZZM}}{\scriptstyle I_{YY}}\cdot P\cdot W_{G}\\
\dot{R} & = & \dfrac{\scriptstyle I_{XX}- \scriptstyle I_{YY}}{\scriptstyle I_{ZZ}}\cdot P\cdot Q+\dfrac{\tau_{z}}{\scriptstyle I_{ZZ}}\\
\dot{\phi} & = & P+Q\cdot s_{\phi}\cdot t_{\theta}+R\cdot c_{\phi}\cdot t_{\theta}\\
\dot{\theta} & = & Q\cdot c_{\phi}-R\cdot s_{\phi}\\
\dot{\psi} & = & Q\cdot\dfrac{s_{\phi}}{c_{\theta}}+R\cdot\dfrac{c_{\phi}}{c_{\theta}}
\end{array}\label{eq:Konacne_Jednacine}
\end{equation}

%\section{Conclusion}

%In this chapter, a general mathematical model of multi-rotor craft has been developed. A symmetrical geometry with an even number of rotors is assumed. It is shown how a specific realization of the craft (number of rotors, arms’ length, and direction of rotors’ rotation) affects the resulting actuation matrix, which is a central figure for realizing the control of the overall UAV. To control the multi-rotor UAV, it is necessary to develop basic controllers for tracking individual relevant quantities, in particular the Cartesian coordinates x,y, and z, and orientation angles $\phi,\theta$ and $\psi$. Moreover, it is necessary to establish how each individual motor/propeller, under the assumption of specific geometry, affects the control and stability of the overall UAV system. These aspects are addressed in more detail in the following chapters.

\chapter{Control}

This chapter describes the control architecture for tracking reference trajectories. The architecture comprises position and orientation controllers, control allocation algorithm, and motor speed controllers. To facilitate the control design, a nonlinear mathematical model of the multi-rotor craft is linearized with respect to equilibrium point. The resulting controllers enable satisfactory tracking of reference trajectories. In the remainder of the chapter, a tracking controller based on the architecture from \cite{mahony2012multirotor} and a simple PD control law is designed. Despite its simplicity, the control system is able to track considerably complex reference trajectories. Moreover, the considered UAV system may be equipped with additional actuators and hence the capacity to continue the mission. However, a mechanism for fault detection is also necessary, together with the suitable control system capable of utilizing the information about the fault state occurrence, in terms of its location and severity. This information should be used, if possible, to mitigate the consequences caused by the fault state. To this end, this chapter also presents some relevant aspects of fault-tolerant control.

\color{black}

\section{Relevant background work} 
First controllers designed for the purpose of multi-rotor UAV are proposed in the aforementioned PhD dissertation of Samir Bouabdallah \cite{Samir_B.}, where the first fully autonomous quad-rotor craft has been developed. A simple stabilizing PID controller is designed, and it was shown that it was possible to perform the orientation stabilization. Based on results from \cite{bouabdallah2004design}, the author has compared the PID-based control with adaptive LQR-based controller, and shown that the former performs better. Continuing on his work on UAV control, he designed a backstepping-based controller, and a sliding-mode-based controller \cite{bouabdallah2005backstepping}. Remarkably, the backstepping control ensured better performance. It was argued that this was the consequence of switching nature of the sliding mode control that caused unwanted oscillations within UAV control.
Beside the seminal work of Bouabdallah, the collaboration of Robert Mahony, Vijay Kumar and Peter Corke yielded several papers that attracted substantial attention  and paved the path for further research in the field of UAV control. The methods proposed by the abovementioned authors have been validated both in simulated environment and on real UAVs. In \cite{michael2010grasp}, the results show the tracking of reference trajectories in laboratory-controlled conditions, whilst \cite{shen2011autonomous} and \cite{shen2014multi} demonstrate the validation in real-world indoor and outdoor environments. Soon after the first real-world validations, the controllers were developed capable of navigating UAVs in obstacle-filled environments and performing complex maneuvers \cite{mellinger2012trajectory}, \cite{mahony2012multirotor}. 
Further enhancements in multi-rotor UAV control can be found in the works of Mark W. Mueller and Raffaell D'Andrea, which enabled “aggressive” maneuvers \cite{flaying_arena_Muler}. For instance, in \cite{muller2011_tenis}, the authors developed a quad-copter equipped with a tennis racket, along with a perception system with eight cameras. The system was able to predict the motion of the tennis ball, enabling the real-time trajectory planning for the UAV to timely place the racket and reflect the approaching ball. 
The same authors successfully showed that multiple UAVs can be coordinated for performing complex tasks \cite{Muler_cooperative}, while in \cite{mueller2014stability} it was shown that the craft can be controlled even in case when one or more rotors are in fault state.

\color{black}
\section{Motor Controller}
The motor dynamic model can be expressed with eq. (\ref{motor_1})-(\ref{aerodinamicki}). 
A robust multirotor motor control approach with respect to battery voltage changes can be found in \cite{mahony2012multirotor} and is given by
\begin{equation}
v^{\text{des}}=K_{\Omega}\left(\Omega_{\text{ref}}-\Omega\right)+v_{\text{ff}}(\Omega_{\text{ref}}),\label{eq:napon napajanja motra}
\end{equation}
where $\Omega_{\text{ref}}$ is a constant referent velocity, $K_{\Omega} > 0$ is the proportional controller gain and $v_{\text{ff}}(\Omega_{\text{ref}})$ is the voltage feedforward term ensuring drag moment compensation at $\Omega_{\text{ref}}$ velocity. The feedforward term can be calculated from the static rotor drag characteristics either from a parametric approximation or a lookup table. 

\begin{figure}[p]
\begin{centering}
\includegraphics[scale=0.7]{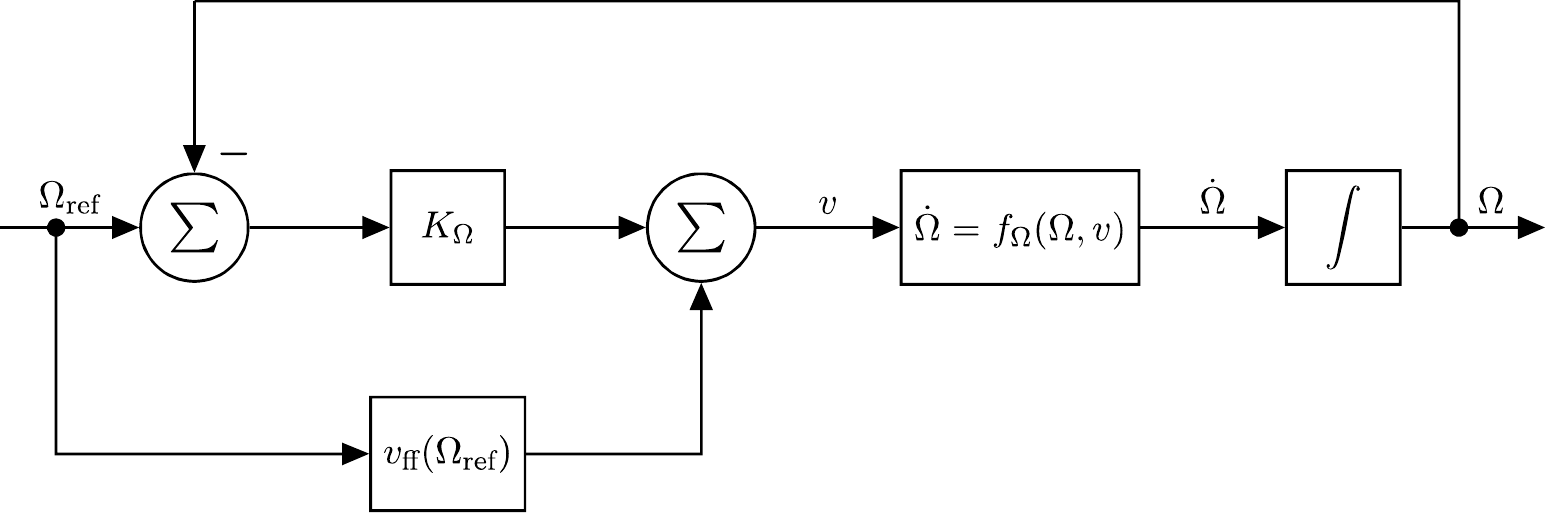}
\par\end{centering}
\caption{Motor speed control diagram .\label{fig:Blok_shema_regulatora}}
\end{figure}

Assuming that the angular rotor velocity $\Omega$ achieved $\Omega_{\text{ref}}$, eq. (\ref{motor_1}) can be expressed as
\begin{equation}
\frac{K_{m}K_{e}}{R}\Omega=\frac{K_{m}}{R}v-\tau_{l}.\label{eq:diff_jednacina_motora}
\end{equation}
If we now plug (\ref{aerodinamicki}) into (\ref{eq:diff_jednacina_motora}) and solve for the input voltage $v$, the feedforward term can be obtained in the form
\begin{equation}
v_{\text{ff}}(\Omega_{\text{ref}})=K_{e}\Omega_{\text{ref}}+\frac{Rd}{K_{m}}\Omega_{\text{ref}}^{2}.\label{eq:feed_forvard_clan}
\end{equation}
To derive the final model of the motor dynamics, eq. (\ref{motor_1}) can be written in the form
\begin{equation}
\dot{\Omega}=-\frac{K_{m}K_{e}}{I_{zzm}R}\Omega+\frac{K_{m}}{I_{zzm}R}v-\frac{\tau_{l}}{I_{zzm}}=f_{\Omega}(\Omega,v),\label{eq:diff_jednacina_motora_I}
\end{equation}
from which one can illustrate the use of the angular velocity controller $K_{\Omega}$ as in Fig. \ref{fig:Blok_shema_regulatora}.

\begin{figure}[p]
\begin{centering}
\includegraphics[scale=0.6]{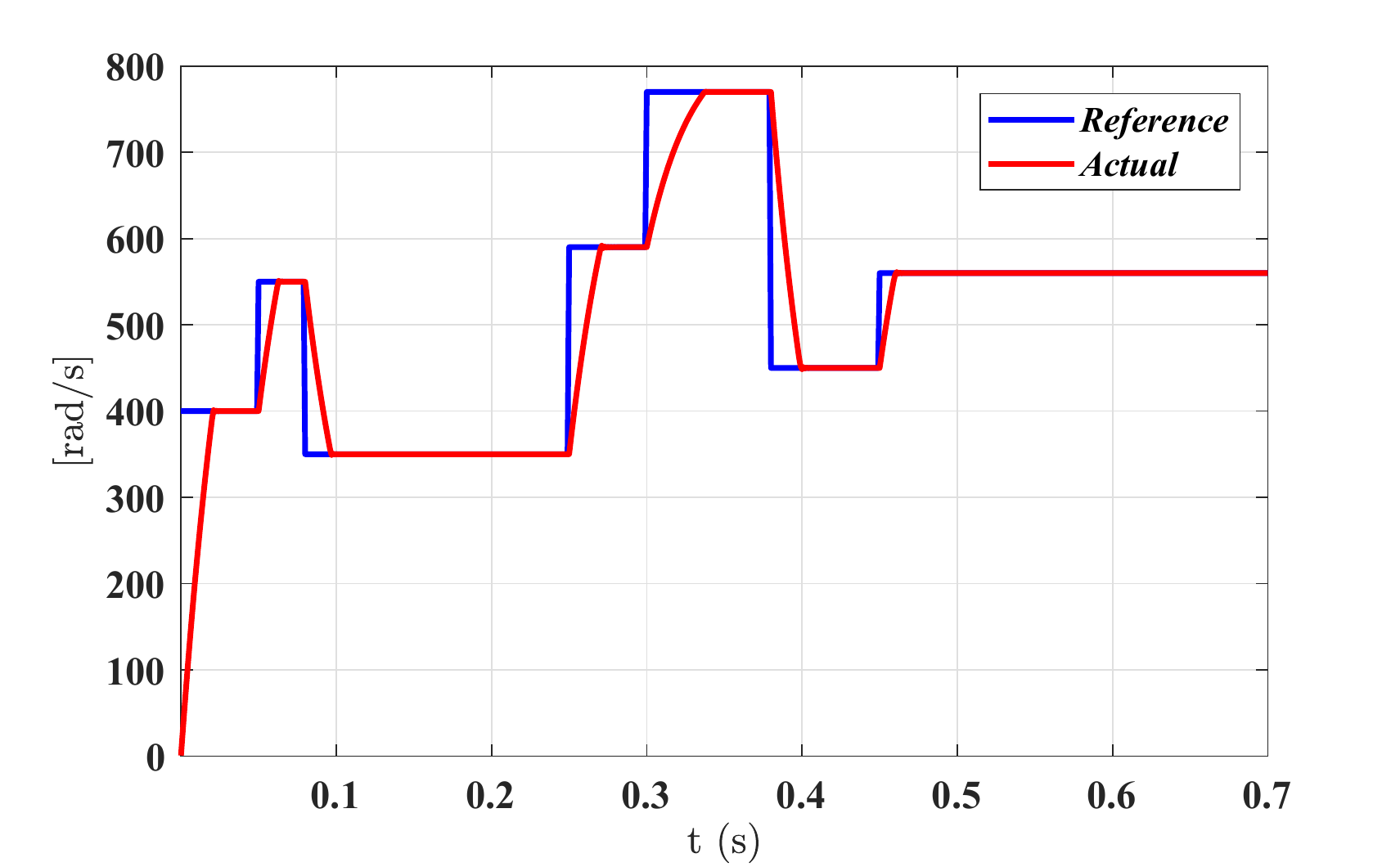}
\par\end{centering}
\caption{Motor speed tracking.\label{fig:Ref-brzina_motora}}

\end{figure}

\begin{figure}
\begin{centering}
\includegraphics[scale=0.6]{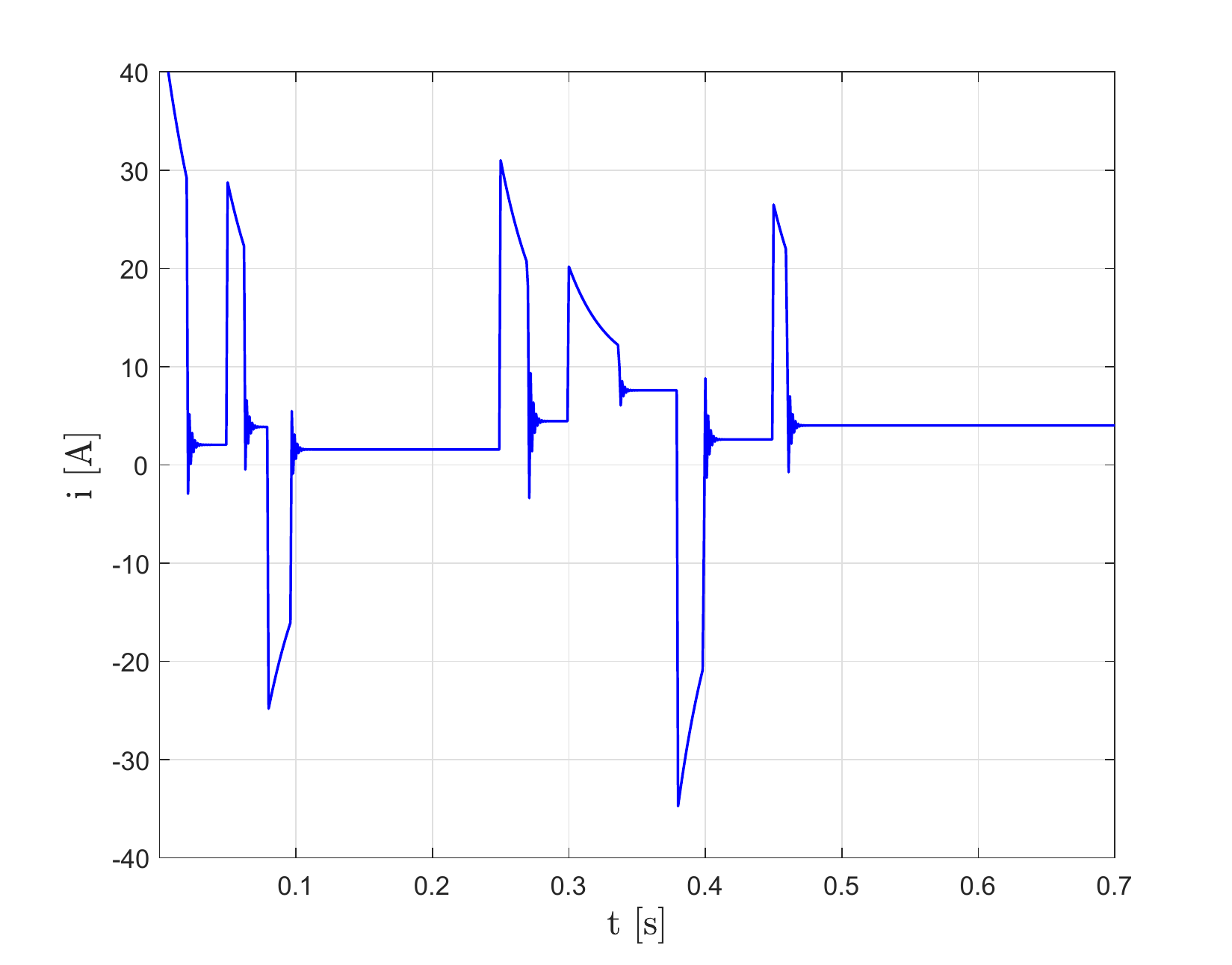}
\par\end{centering}
\caption{Time response of the armature current.\label{fig:Struja-na-namotima}}

\end{figure}

\begin{figure}[p]
\begin{centering}
\includegraphics[scale=0.6]{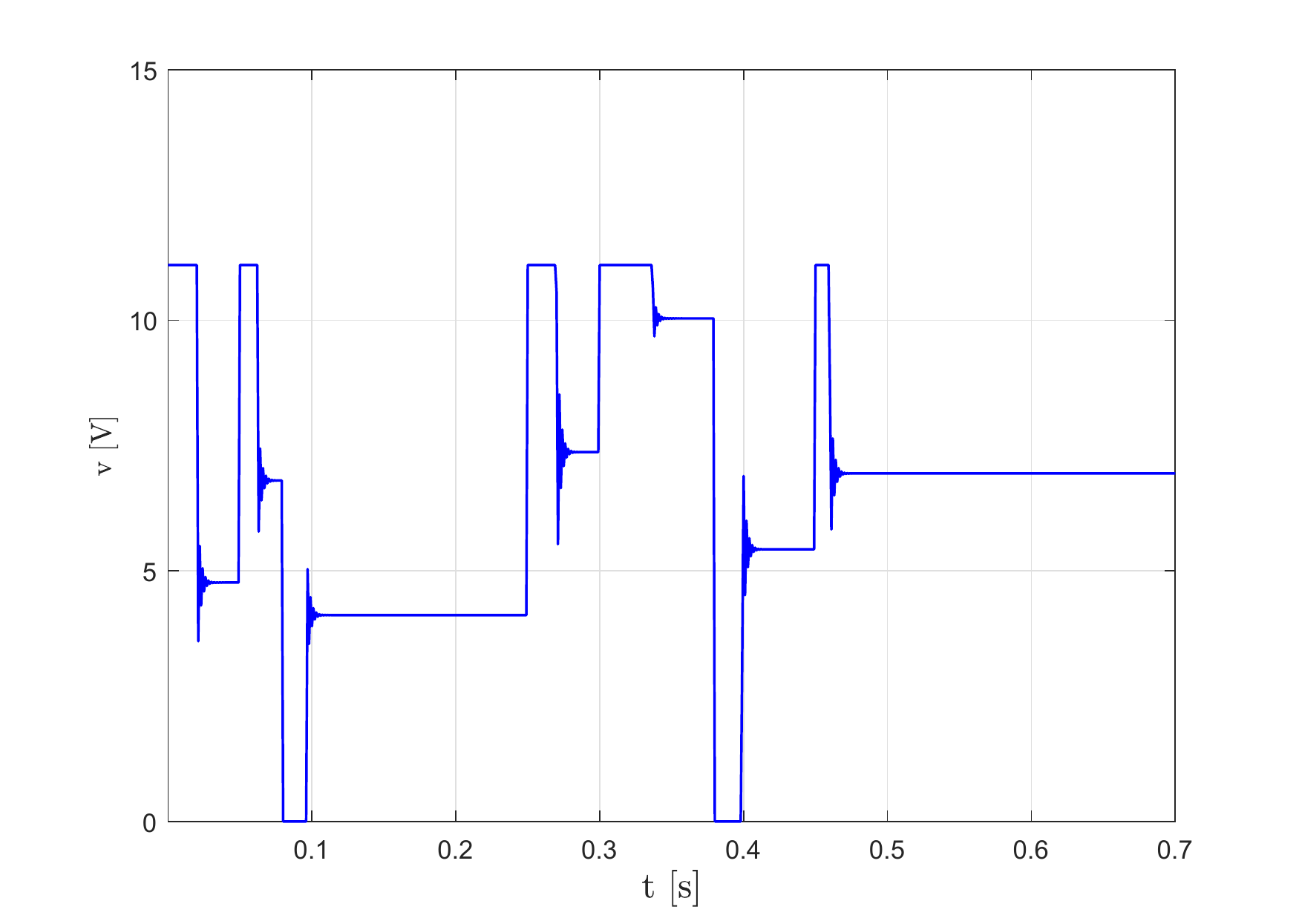}
\par\end{centering}
\caption{Time response of the armature voltage.\label{fig:Napon-na-namotima}}

\end{figure}

Fig. \ref{fig:Ref-brzina_motora} indicates how well angular velocity is controlled using only proportional component for $K_{\Omega}$. It can be seen that the output velocity $\Omega$ (red) tracks quickly the reference input $\Omega_{ref}$ (blue) and the transient phase is a bit longer when the reference input is larger. This occurs when the reference value is close to the velocity saturation of the motor. The controller gain ($K_{\Omega}=2$) is selected to decrease the oscillations in the motor current and voltage, as shown respectively in Figs. \ref{fig:Struja-na-namotima}
i \ref{fig:Napon-na-namotima}, in order to prolong its life cycle. 

\begin{figure}[p]
\begin{centering}
\includegraphics[scale=0.7]{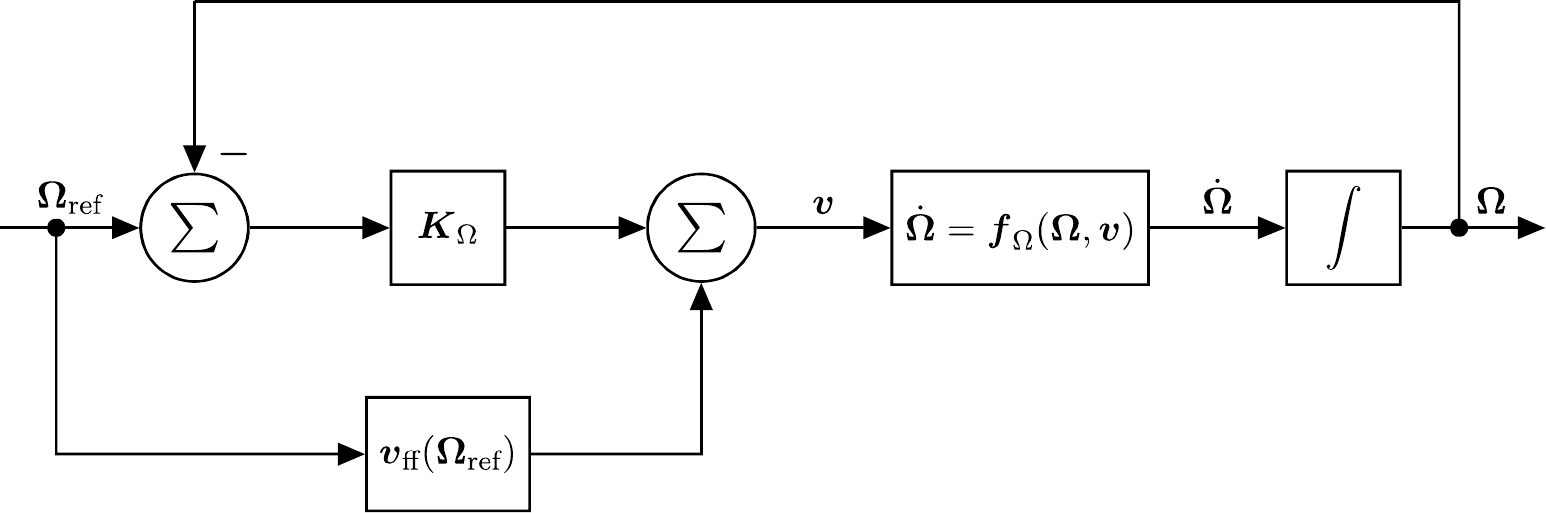}
\caption{Vectorized control of motor speed.
\label{fig:Blok_shema_vektoriziranog_reg}}
\par\end{centering}
\end{figure}

If we now aim to apply a controller $K_{\omega}$ to an octocopter, it is necessary to define the vector consisting of all motor angular velocities
\begin{equation}
\boldsymbol{\Omega}=\begin{bmatrix}\Omega_{1} & \Omega_{2} & \Omega_{3} & \Omega_{4} & \Omega_{5} & \Omega_{6} & \Omega_{7} & \Omega_{8}\end{bmatrix}^{T},\label{eq:vektor brzine}
\end{equation}
the reference vector for each single motor
\begin{equation}
\boldsymbol{\Omega}_{\text{ref}}=\begin{bmatrix}\Omega_{\text{ref},1} & \Omega_{\text{ref},2} & \Omega_{\text{ref},3} & \Omega_{\text{ref},4} & \Omega_{\text{ref},5} & \Omega_{\text{ref},6} & \Omega_{\text{ref},7} & \Omega_{\text{ref},8}\end{bmatrix}^{T},\label{eq:vektor referentnih brzina}
\end{equation}
the vector of control inputs
\begin{equation}
\boldsymbol{v}=\begin{bmatrix}v_{1} & v_{2} & v_{3} & v_{4} & v_{5} & v_{6} & v_{7} & v_{8}\end{bmatrix}^{T},\label{eq:vektor upravljackih ulaza}
\end{equation}
and the vector of feedforward terms as
\begin{equation}
\boldsymbol{v}_{\text{ff}}(\boldsymbol{\Omega}_{\text{ref}})=\begin{bmatrix}v_{\text{ff}}(\Omega_{\text{ref},1})\\
v_{\text{ff}}(\Omega_{\text{ref},2})\\
v_{\text{ff}}(\Omega_{\text{ref},3})\\
v_{\text{ff}}(\Omega_{\text{ref},4})\\
v_{\text{ff}}(\Omega_{\text{ref},5})\\
v_{\text{ff}}(\Omega_{\text{ref},6})\\
v_{\text{ff}}(\Omega_{\text{ref},7})\\
v_{\text{ff}}(\Omega_{\text{ref},8})
\end{bmatrix}.\label{eq:vektorski feedforvard clan}
\end{equation}
In this case, the controller can be generalized using the form

\begin{equation}
\boldsymbol{v}^{\text{des}}=\boldsymbol{K}_{\Omega}\left(\boldsymbol{\Omega}_{\text{ref}}-\boldsymbol{\Omega}\right)+\boldsymbol{v}_{\text{ff}}(\boldsymbol{\Omega}_{\text{ref}}),\label{eq:vektor_generalizirani}
\end{equation}
where $\boldsymbol{K}_{\Omega}$ is a diagonal matrix given as $\boldsymbol{K}_{\Omega} = K_\Omega \boldsymbol{I}_{8 \times 8}$. If we now decribe the motor dynamic model using the following compact form
\begin{equation}
\boldsymbol{f}_{\Omega}(\boldsymbol{\Omega},\boldsymbol{v})=\begin{bmatrix}\dot{\Omega}_{1}\\
\dot{\Omega}_{2}\\
\dot{\Omega}_{3}\\
\dot{\Omega}_{4}\\
\dot{\Omega}_{5}\\
\dot{\Omega}_{6}\\
\dot{\Omega}_{7}\\
\dot{\Omega}_{8}
\end{bmatrix}=\begin{bmatrix}f_{\Omega}(\Omega_{1},v_{1})\\
f_{\Omega}(\Omega_{2},v_{2})\\
f_{\Omega}(\Omega_{3},v_{3})\\
f_{\Omega}(\Omega_{4},v_{4})\\
f_{\Omega}(\Omega_{5},v_{5})\\
f_{\Omega}(\Omega_{6},v_{6})\\
f_{\Omega}(\Omega_{7},v_{7})\\
f_{\Omega}(\Omega_{8},v_{8})
\end{bmatrix},\label{eq:dinamicki clan}
\end{equation}
then the vector of the angular velocity controllers can be illustrated as \color{black} in Fig. \ref{fig:Blok_shema_vektoriziranog_reg}.

\section{Control allocation algorithm}

\color{black}

 The relation between the control inputs $u$ (the reference thrust  force $T$ and torques $\boldsymbol{\tau}$) and the rotation velocity $\boldsymbol{\Omega}_s$ of DC motors (see chapter 2) is given with:
\begin{equation}
\boldsymbol{u_{ref}} = \boldsymbol{A} \boldsymbol{\Omega}_s \label{uref_linear}
\end{equation}

\noindent where

\begin{equation}
\boldsymbol{u_{ref}}=\begin{bmatrix}T & \boldsymbol{\tau}\end{bmatrix}^{T}=\begin{bmatrix}T & \tau_{x} & \tau_{y} & \tau_{z}\end{bmatrix}^{T},\label{eq:virtualno_upravljanje}
\end{equation}

\noindent $\boldsymbol{A}$ being the system actuation matrix defined as:

\begin{equation}
\boldsymbol{A}=\begin{bmatrix}b & b & b & b & b & b & b & b\\
bl & \frac{\sqrt{2}}{2}bl & 0 & -\frac{\sqrt{2}}{2}bl & -bl & -\frac{\sqrt{2}}{2}bl & 0 & \frac{\sqrt{2}}{2}bl\\
0 & -\frac{\sqrt{2}}{2}bl & -bl & -\frac{\sqrt{2}}{2}bl & 0 & \frac{\sqrt{2}}{2}bl & bl & \frac{\sqrt{2}}{2}bl\\
-d & d & -d & d & -d & d & -d & d
\end{bmatrix},\label{eq:matrica_aktuacije}
\end{equation}
while $\boldsymbol{\Omega}_s \in D_{\Omega_s} \subset \mathbb{R}^8$ represents the squared rotor velocity vector given as

\begin{equation}
\boldsymbol{\Omega}_{s}=\begin{bmatrix}\Omega_{1}^{2} & \Omega_{2}^{2} & \Omega_{3}^{2} & \Omega_{4}^{2} & \Omega_{5}^{2} & \Omega_{6}^{2} & \Omega_{7}^{2} & \Omega_{8}^{2}\end{bmatrix}^{T}.\label{eq:Vektor_Ugaonih_brzina}
\end{equation}
$\boldsymbol{u} \in D_{u} \subset \mathbb{R}^4$ 
is a surjective (onto) mapping $D_{\Omega_s} \mapsto D_{u}$ implying that multiple $\boldsymbol{\Omega}_s$ values map to the same $\boldsymbol{u}$ value. Consequently, the so-called control allocation problem given by the inverse mapping $D_{u} \mapsto D_{\Omega_s}$ does not have a unique solution. Due to the motor voltage constraints, the set $D_{\Omega_s}$ is defined based on the squared rotor velocity constraints:
\begin{equation}
\label{omega_ogranicenjee}
0 \leq \Omega^2_i \leq \Omega^2_{\text{max}}, \; \; i = \overline{1..8}.
\end{equation}

Assuming that the DC motor velocity is limited between 0 and  $\omega_{max}$ \eqref{omega_ogranicenjee} and the mapping is defined by the linear relation (\ref{uref_linear}), it means that the set $D_{u}$ represents a polytope in space $\mathbb{R}^4$  due to the linear actuation in conjunction with the box constrained inputs. Another interesting subtlety that arises due to the decentralized control approach is that the altitude and attitude controllers can generate desired control input values $\boldsymbol{u}^{\text{des}} \not\in D_{u}$ that may cause actuator saturation. 

An extensive survey on algorithms that are used to solve the control allocation problem is considered in \cite{johansen2013control}. For simplicity, we introduce the equality constraints
\begin{align}
\Omega_1 = \Omega_3, \quad
\Omega_2 = \Omega_4, \quad 
\Omega_5 = \Omega_7, \quad
\Omega_6 = \Omega_8,
\end{align}
and therefore reduce the actuation matrix $\boldsymbol{A}$ to its square (invertible) form
\begin{equation}
\boldsymbol{A}_{f}=\begin{bmatrix}2b & 2b & 2b & 2b\\
bl & 0 & -bl & 0\\
-bl & -\sqrt{2}bl & bl & \sqrt{2}bl\\
-2d & 2d & -2d & 2d
\end{bmatrix}.\label{eq:jedinicna_matrica_aktuacije}
\end{equation}

\begin{figure}[p]
\begin{centering}
\includegraphics[scale=0.7]{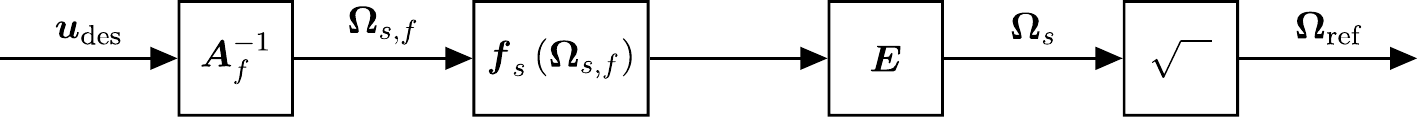}
\par\end{centering}
\caption{Actuation generation diagram.
\label{fig:Blok_shema_aktuacije}}
\end{figure}

\noindent If we also introduce an auxiliary vector in the form

\begin{equation}
\boldsymbol{\Omega}_{s,f}=\begin{bmatrix}\Omega_{s,f,1}\\
\Omega_{s,f,2}\\
\Omega_{s,f,3}\\
\Omega_{s,f,4}
\end{bmatrix}=\begin{bmatrix}\Omega_{1}^{2}\\
\Omega_{2}^{2}\\
\Omega_{5}^{2}\\
\Omega_{6}^{2}
\end{bmatrix},\label{eq:vektor_rotacije}
\end{equation}

\noindent it is then possible to solve the control allocation problem as:

\[
\boldsymbol{\Omega}_{s,f}=\boldsymbol{A}_{f}^{-1}\boldsymbol{u}^{\text{des}},
\]
and by introducing the selection operator

\begin{equation}
\boldsymbol{E}=\left[\begin{array}{cccccccc}
1 & 0 & 1 & 0 & 0 & 0 & 0 & 0\\
0 & 1 & 0 & 1 & 0 & 0 & 0 & 0\\
0 & 0 & 0 & 0 & 1 & 0 & 1 & 0\\
0 & 0 & 0 & 0 & 0 & 1 & 0 & 1
\end{array}\right]^{T},\label{eq:operator_inverzije}
\end{equation}
it is also possible to reconstruct the squared rotor velocity vector as:
\begin{equation}
\boldsymbol{\Omega}_{s}=\boldsymbol{E}\,\boldsymbol{\Omega}_{s,f}.\label{eq:upravljacki_vektor}
\end{equation}
In order to include the angular velocity saturations as well, one can define the function
\begin{equation}
f_{s}\left(\Omega_{s,f}\right)=\begin{cases}
0,\hspace{14.1mm}\Omega_{s,f}<0\\
\Omega_{s,f},\;\;0\leq\Omega_{s,f}\leq\Omega_{\text{max}}^{2}\\
\Omega_{\text{max}}^{2},\hspace{8mm}\Omega_{s,f}>\Omega_{\text{max}}^{2}
\end{cases}\label{eq:funcija_zasicenja}
\end{equation} 
with its full vector form 
\begin{equation}
\boldsymbol{f}_{s}\left(\boldsymbol{\Omega}_{s,f}\right)=\left[\begin{array}{cccc}
f_{s}\left(\Omega_{s,f,1}\right) & f_{s}\left(\Omega_{s,f,2}\right) & f_{s}\left(\Omega_{s,f,3}\right) & f_{s}\left(\Omega_{s,f,4}\right)\end{array}\right]^{T},\label{eq:Vektorizacija}
\end{equation}
which is used in control allocation as shown in  Fig. \ref{fig:Blok_shema_aktuacije} \color{black}.
If we further define the full vector of squares of the angular velocities of each single motor
\begin{equation}
\boldsymbol{f}_{m}(\boldsymbol{\Omega})=\begin{bmatrix}\Omega_{1}^{2} & \Omega_{2}^{2} & \Omega_{3}^{2} & \Omega_{4}^{2} & \Omega_{5}^{2} & \Omega_{6}^{2} & \Omega_{7}^{2} & \Omega_{8}^{2}\end{bmatrix}^{T}=\boldsymbol{\Omega}_{s},\label{eq:Vek_fja_aktuacije}
\end{equation}
than the whole actuation can be shown in a compact form as in Fig. \ref{fig:Blok_shema_aktuacije}. 

\begin{figure}[p]
\begin{centering}
\includegraphics[scale=0.9]{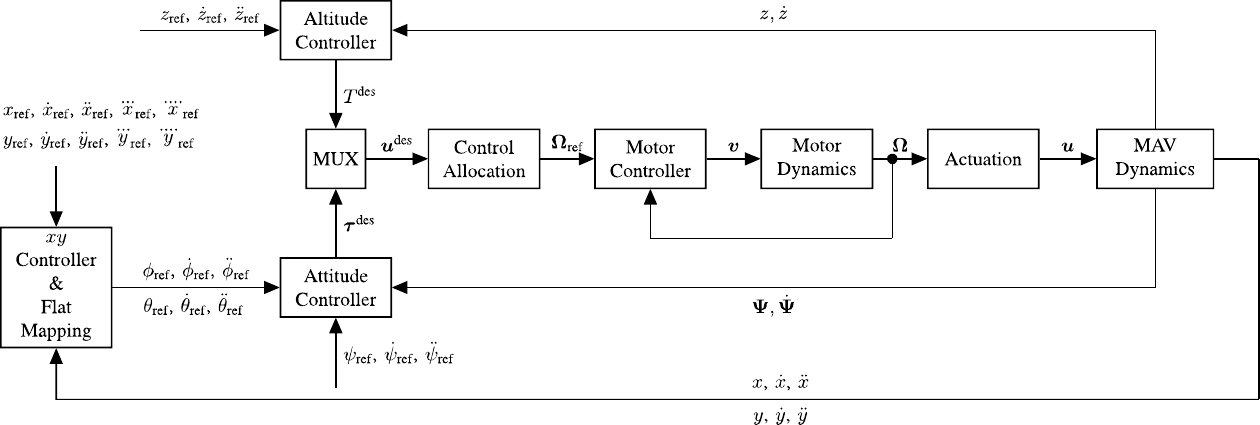}
\par\end{centering}
\caption{Control architecture.
\label{fig:Regulator_sve}}

\end{figure}

\section{Linearized model}

\color{black}
The controlled variables are represented with the vehicle position $x$, $y$, $z$ and its attitude $\boldsymbol{\Psi}$, while the overall architecture includes $xy$, altitude, attitude and motor controllers as well as the control allocation and system dynamics consisted of the motor, actuation and the octocopter dynamics (see Fig. \ref{fig:Regulator_sve}). 

The desired linear motion reference values along a mission trajectory are applied to the $xy$ controller, while the corresponding desired attitude and altitude reference values are handled by the related controllers. The outputs of these controllers form the desired values of the total force and torques to the control allocation algorithm to deal with the over-actuated system. This algorithm then distributes these desired values onto the desired velocity vector $\boldsymbol{\Omega_{ref}}$ to provide speeds for each motor. For the purpose of this work, we use a pseudo-inverse control allocation \cite{FTC_ETH_Thomas_Schneider},\cite{basson_pseudo}. The motor controller is used as a low-level controller to force the motor velocity vector $\boldsymbol{\Omega}$ to follow the reference values from $\boldsymbol{\Omega_{ref}}$. 

To design a PD tracking controller, it is common practice to linearize the octocopter dynamics around the hover configuration 
\begin{align}
(x_e, y_e, z_e) &= (x, y, z), \nonumber \\
(u_e, v_e, w_e) &= (0, 0, 0), \nonumber \\
(P_e, Q_e, R_e) &= (0, 0, 0), \nonumber \\
(\phi_e, \theta_e, \psi_e) &= (0, 0, \psi).
\end{align}
The linearized kinematic model of the linear motion can be described with:
\begin{equation}
\begin{array}{c}
\dot{x}=c_{\psi_{e}}u-s_{\psi_{e}}v\\
\dot{y}=s_{\psi_{e}}u+c_{\psi_{e}}v\\
\dot{z}=w,
\end{array}\label{eq:lin_koordinate}
\end{equation}
while the linearized kinematic model of angular motion is represented by:
\begin{equation}
\begin{array}{c}
\dot{\phi}=P\\
\dot{\theta}=Q\\
\dot{\psi}=R.
\end{array}\label{eq:kin-model_kruznog}
\end{equation}
Similarly, the linearized dynamic model of linear motion become
\begin{equation}
\begin{array}{c}
\dot{u}=g\theta\\
\dot{v}=-g\phi\\
\dot{w}=\frac{T}{m_{o}}-g,
\end{array}\label{eq:din_model_pravolinijskog}
\end{equation}
while the linearized dynamic model of angular motion is simplified to
\begin{equation}
\begin{array}{c}
\dot{P}=\frac{\tau_{x}}{I_{xx}}\\
\dot{Q}=\frac{\tau_{y}}{I_{yy}}\\
\dot{R}=\frac{\tau_{z}}{I_{zz}}.
\end{array}\label{din_model_kruznog}
\end{equation}

The compact form of these models can be written as
\begin{align}
\label{sim_lin_kin}
\boldsymbol{\dot{x}} &= \boldsymbol{R}(Z, \psi_e) \, \boldsymbol{v} \\
\label{sim_ang_kin}
\boldsymbol{\dot{\Psi}} &= \boldsymbol{P} \\
\label{sim_lin_dyn}
\boldsymbol{\dot{v}} &= \begin{bmatrix}
0 \\
0 \\
\frac{T}{m_o}
\end{bmatrix} + g 
\begin{bmatrix}
\theta \\
-\phi \\
-1
\end{bmatrix} \\
\label{sim_ang_dyn}
\boldsymbol{\dot{P}} &= \boldsymbol{J}^{-1} \boldsymbol{\tau},
\end{align}
where $\boldsymbol{R}(Z, \psi_e)$ represents the rotation matrix around the $z$-axis.

\section{Altitude control near equilibrium}

\color{black}
In order to design the tracking controller for maintaining the desired height of the octocopter, we start from the linearized dynamic model of linear motion in the form
\begin{equation}
\ddot{z}=\frac{T}{m_{o}}-g,\label{eq:lin_visine}
\end{equation}
and form the control error as
\begin{equation}
e_{z}=z_{\text{ref}}-z,\label{eq:greska_visine}
\end{equation}
where $z_{\text{ref}}$ is a reference height. The PD control law can now be constructed in the form
\begin{equation}
T^{\text{des}}=m_{o}\left(g+\ddot{z}_{\text{ref}}+K_{dz}\dot{e}_{z}+K_{pz}e_{z}\right),\label{eq:upravljacki zakon}
\end{equation}
where $g$ is the gravitational acceleration, while $K_{pz}$ and $K_{dz}$ proportional and derivative gains. Inserting (\ref{eq:upravljacki zakon}) into (\ref{eq:lin_visine}), one can obtain the error dynamics as
\begin{equation}
\ddot{e}_{z}+K_{dz}\dot{e}_{z}+K_{pz}e_{z}=0.\label{eq:dinamika_greske_visine}
\end{equation}
Since (\ref{eq:dinamika_greske_visine}) is the second-order linear differential equation, the error $e_z$ exponentially vanishes for positive values of $K_{pz}$ and $K_{dz}$.

\begin{figure}[p]
\begin{centering}
\includegraphics[scale=0.7]{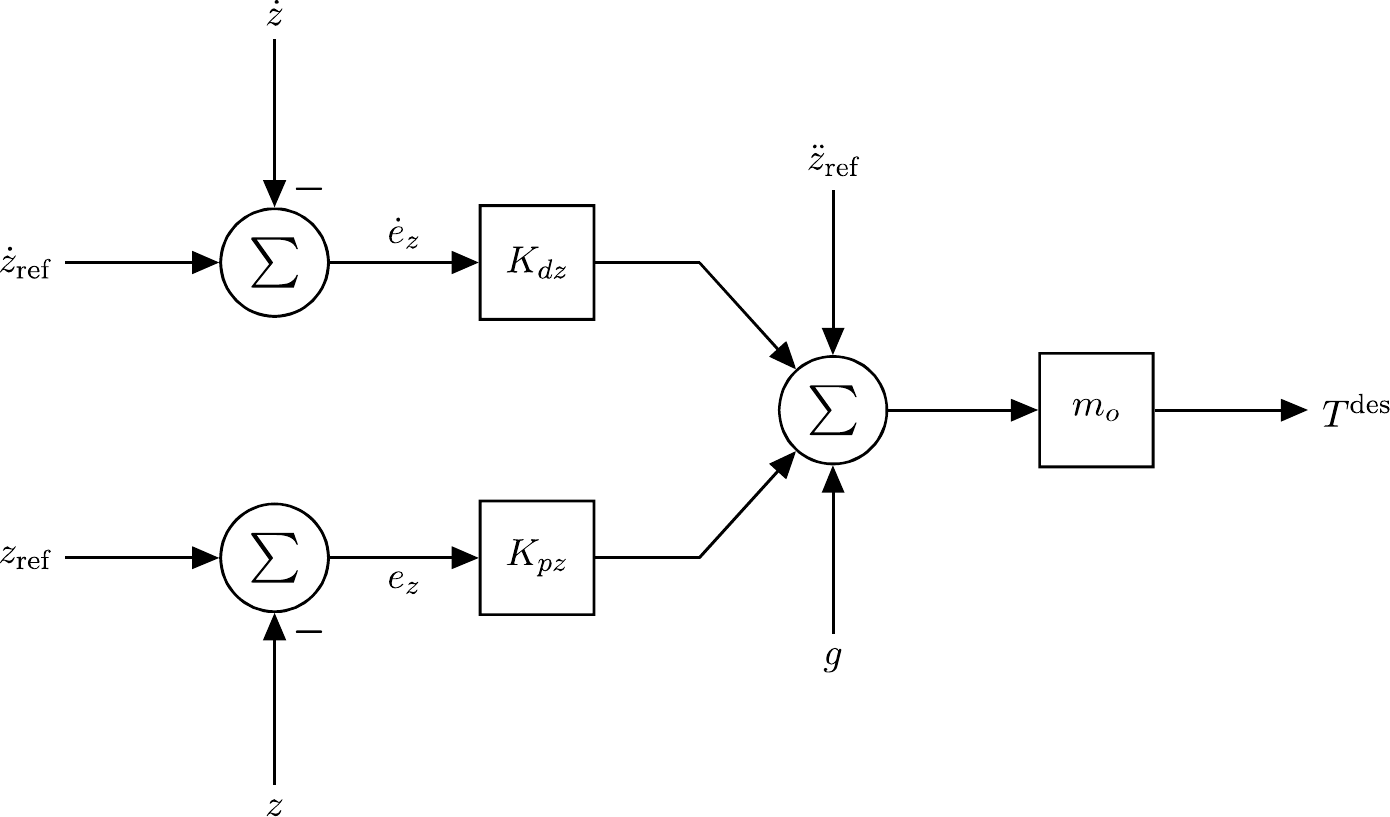}
\par\end{centering}
\caption{Altitude tracking control diagram.
\label{fig:regulator_visine}}
\end{figure}

\begin{figure}[p]
\begin{centering}
\includegraphics[scale=0.6]{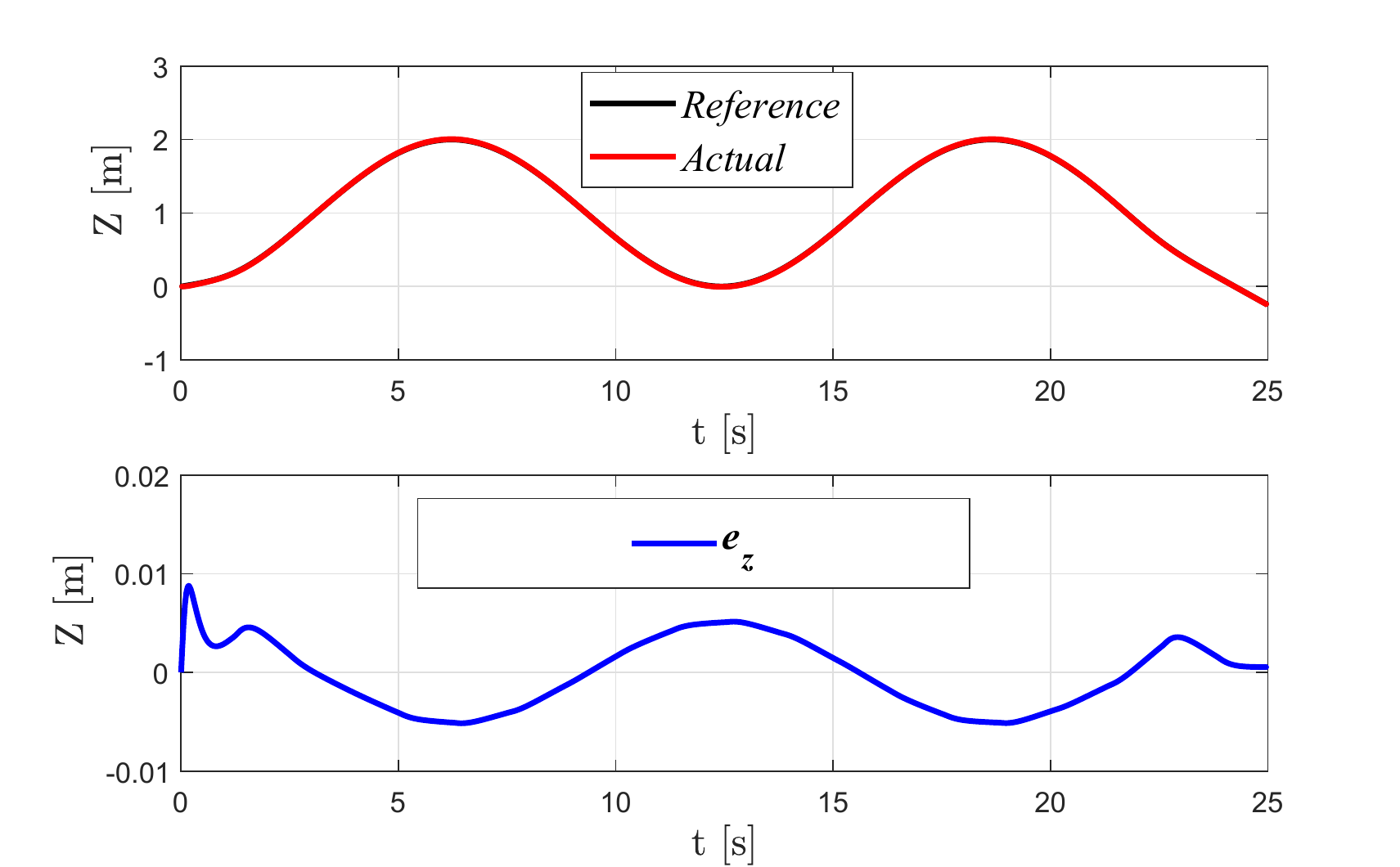}
\par\end{centering}
\caption{Altitude tracking results.\label{fig:rezultati_pracenja_visine}}
\end{figure}

Fig. \ref{fig:regulator_visine} demonstrates the block structure of the selected height tracking controller, while Fig. \ref{fig:rezultati_pracenja_visine} shows the simulation results obtained for $K_{dz}=5$ i
$K_{pz}=6.25$. One can notice from Fig. \ref{fig:rezultati_pracenja_visine} that the octocopter satisfactory tracks the sinusoidal shape of the reference values.

\section{Orientation control near equilibrium}

\color{black}
In order to design the tracking controller for maintaining the desired orientation of the octocopter, we start from the linearized dynamic model of angular motion in the form
\begin{equation}
\ddot{\boldsymbol{\Psi}}=\boldsymbol{J}^{-1}\boldsymbol{\tau},\label{eq:din_-model_orijentacije}
\end{equation}
where $\boldsymbol{J}$ is the inertia tensor of the octocopter given as
\begin{equation}
\boldsymbol{J}=\begin{bmatrix}I_{xx} & 0 & 0\\
0 & I_{yy} & 0\\
0 & 0 & I_{zz}
\end{bmatrix},\label{eq:tenzor_inercije_okto}
\end{equation}
and $\dot{\boldsymbol{\Psi}}=\boldsymbol{P}$, where $\boldsymbol{\Psi}=[\phi\:\: \theta\:\: \psi]^T$ represent the angles with respect to the axes $X$, $Y$ and $Z$, and $\boldsymbol{P}=[P\:\: Q\:\: R]^T$ being the vector of angular velocities in the local coordinate system. 

\noindent \medskip{}
The vector of control errors can be formed as
\begin{equation}
\boldsymbol{e}_{\Psi}=\begin{bmatrix}\phi_{\text{ref}}\\
\theta_{\text{ref}}\\
\psi_{\text{ref}}
\end{bmatrix}-\begin{bmatrix}\phi\\
\theta\\
\psi
\end{bmatrix}=\boldsymbol{\Psi}_{\text{ref}}-\boldsymbol{\Psi},
\label{eq:greska_odstupanja_kutova}
\end{equation} and the PD control law can be constructed in the form

\begin{equation}
\boldsymbol{\tau}^{\text{des}}=\boldsymbol{J}\left(\ddot{\boldsymbol{\Psi}}_{\text{ref}}+\boldsymbol{K}_{d}\dot{\boldsymbol{e}}_{\Psi}+\boldsymbol{K}_{p}\boldsymbol{e}_{\Psi}\right),
\label{eq:Upravljacki_zakon_visina}
\end{equation}
yielding the error dynamics given in the vector form
\begin{equation}
\ddot{\boldsymbol{e}}_{\Psi}+\boldsymbol{K}_{d}\dot{\boldsymbol{e}}_{\Psi}+\boldsymbol{K}_{p}\boldsymbol{e}_{\Psi}=\boldsymbol{0}_{3\times1}.
\label{eq:din_model_greske_orjen}
\end{equation}
Since (\ref{eq:din_model_greske_orjen}) can be separated into three second-order differential equations, all three errors exponentially vanish for positive values of proportial and derivative gains $K_d$ and $K_p$, where $\boldsymbol{K}_{d}=K_{d}\boldsymbol{I}_{3\times3}$ and $\boldsymbol{K}_{p}=K_{p}\boldsymbol{I}_{3\times3}$.

\begin{figure}
\begin{centering}
\includegraphics[scale=0.7]{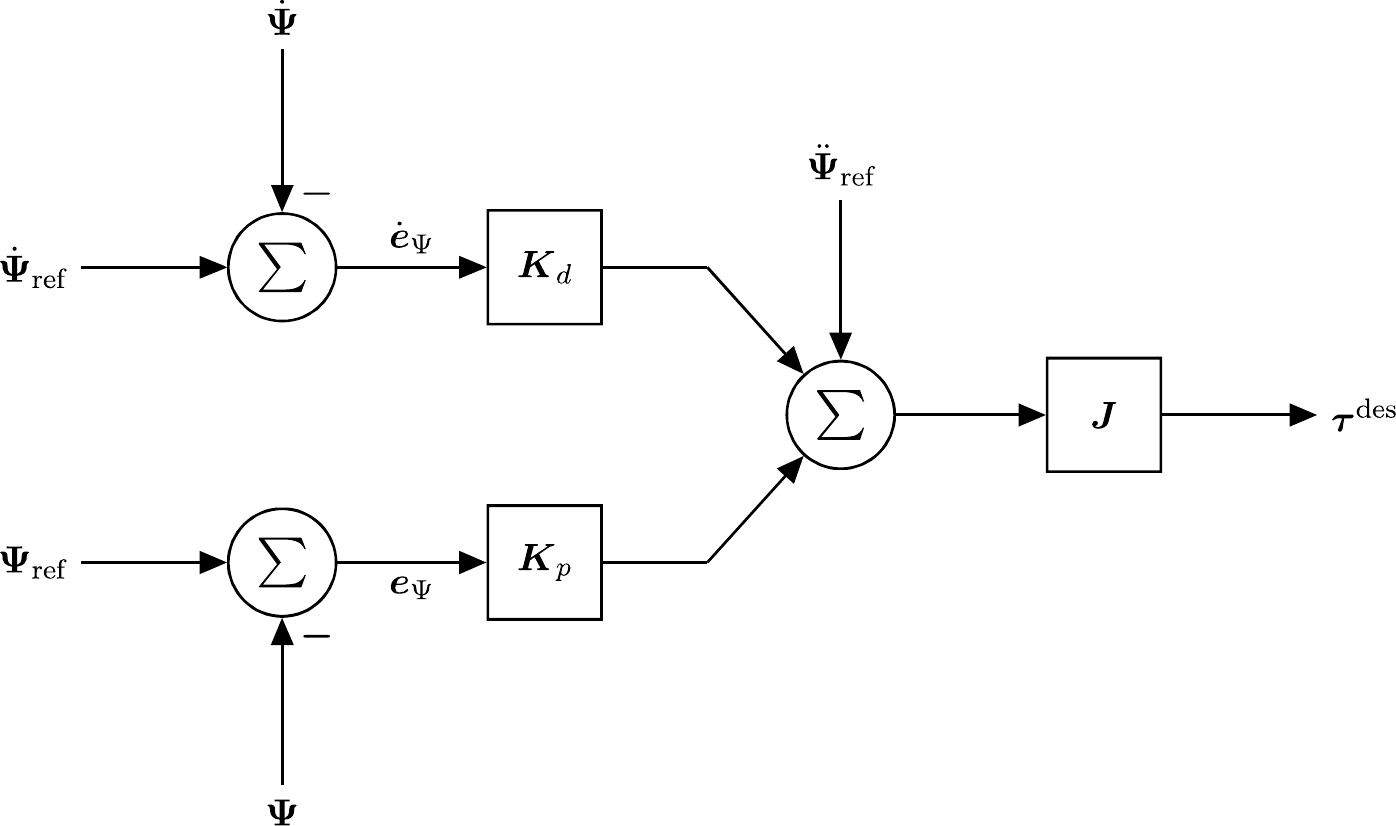}
\par\end{centering}
\caption{Orientation tracking control diagram.\label{fig:Regulator_orijentacije}}
\end{figure}

\begin{figure}
\begin{centering}
\includegraphics[scale=0.42]{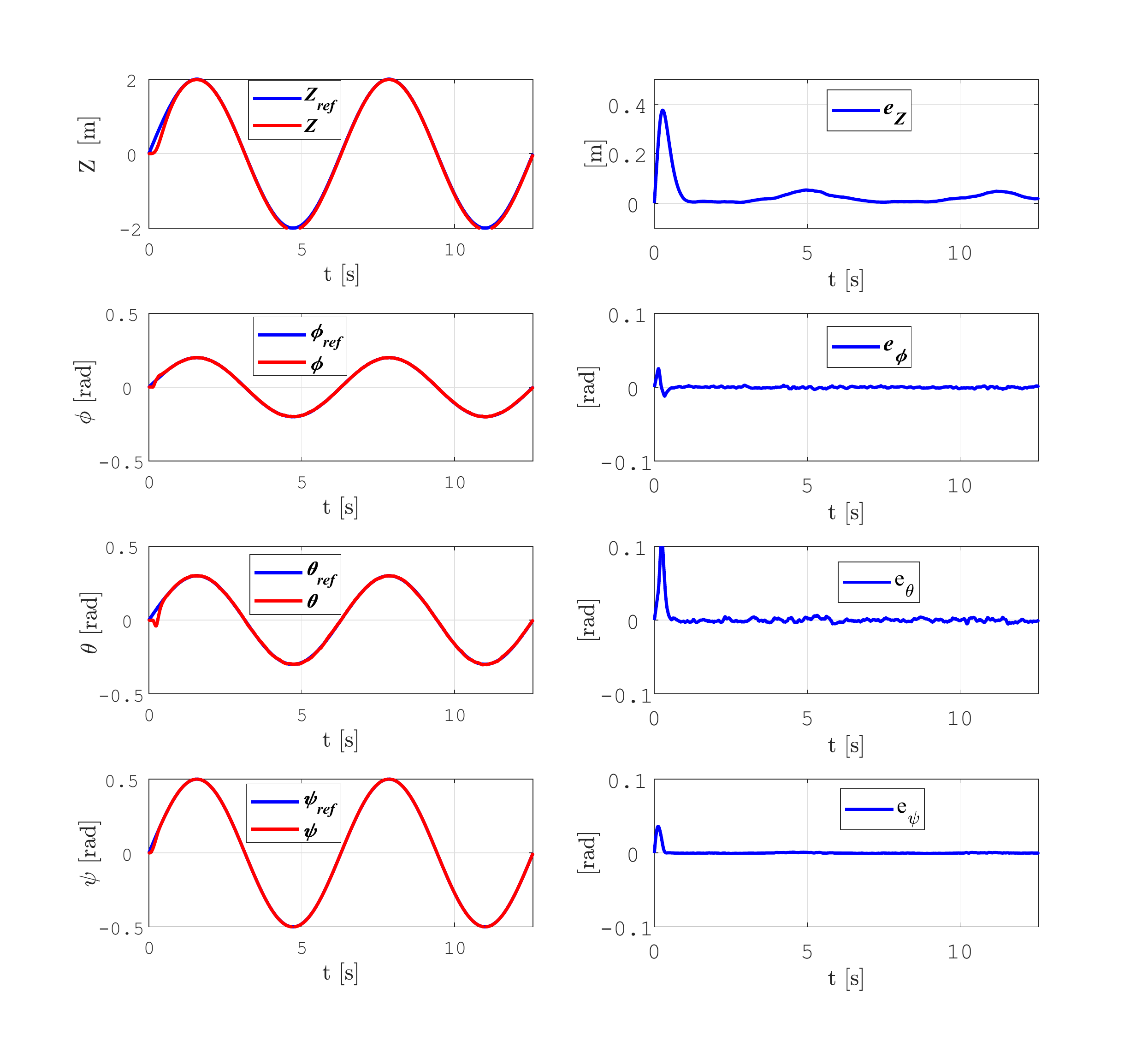}\end{centering}
\caption{Altitude and orientation tracking results.\label{fig:Pracenje_orijentacije}}
\end{figure}

Fig. \ref{fig:Regulator_orijentacije} demonstrates the block structure of the selected orientation tracking controller, while Fig. \ref{fig:Pracenje_orijentacije} shows the simulation results obtained of the simultaneous tracking of the height and orientations for ${K_{dz}}=30$, ${K_{pz}}=225$, ${K_{d}}=20$ and ${K_{p}}=100$. It can be seen from Fig. \ref{fig:Pracenje_orijentacije} that all tracking errors vanish within one second interval and that the tracking performance can be considered satisfactory despite the fact that the tracking is conducted simultaneously with respect to all control variables.


\section{Position control near equilibrium}

\color{black}
In order to design the tracking controller for maintaining the desired octocopter position $(x,y)$ in the global reference frame, we start from the linearized motion dynamics in the form 
\begin{equation}
\begin{bmatrix}\ddot{x}\\
\ddot{y}
\end{bmatrix}=g\begin{bmatrix}c_{\psi_{e}} & -s_{\psi_{e}}\\
s_{\psi_{e}} & c_{\psi_{e}}
\end{bmatrix}\begin{bmatrix}\theta\\-\phi

\end{bmatrix},
\label{eq:flat_maping1}
\end{equation}
that is
\begin{equation}
\begin{bmatrix}\phi\\
\theta
\end{bmatrix}=\frac{1}{g}\begin{bmatrix}s_{\psi_{e}} & -c_{\psi_{e}}\\
c_{\psi_{e}} & s_{\psi_{e}}
\end{bmatrix}\begin{bmatrix}\ddot{x}\\
\ddot{y}
\end{bmatrix}.\label{eq:flat_maping}
\end{equation}
Therefore, in order to achieve tracking of the $x$ and $y$ position coordinates, we need a flat mapping between those positions and the altitude and attitude coordinates. For that reason, let the tracking errors $e_x = x_{\text{ref}} - x$ and $e_x = y_{\text{ref}} - y$ be introduced, where $x_{\text{ref}}$ and $y_{\text{ref}}$ are the reference values. If one wants the tracking errors to decay exponentially, it is sufficient that the following holds
\begin{equation}
\label{greska_flat}
\begin{bmatrix}
\ddot{e}_x \\
\ddot{e}_y
\end{bmatrix} +
K_{d}
\begin{bmatrix}
\dot{e}_x \\
\dot{e}_y
\end{bmatrix} +
K_{p}
\begin{bmatrix}
{e}_x \\
{e}_y
\end{bmatrix} = 
\boldsymbol{0}_{2 \times 1},
\end{equation}
which can be rewritten as \eqref{xy_rewrite}
\begin{equation}
\begin{bmatrix}
\ddot{x} \\
\ddot{y}
\end{bmatrix}
= 
\begin{bmatrix}
\ddot{x}_{\text{ref}} \\
\ddot{y}_{\text{ref}}
\end{bmatrix} +
K_d 
\begin{bmatrix}
\dot{e}_x \\
\dot{e}_y
\end{bmatrix} +
K_p
\begin{bmatrix}
{e}_x \\
{e}_y
\end{bmatrix}.
\label{xy_rewrite}
\end{equation}
From \eqref{eq:flat_maping1} and assuming that $\Psi_e = \Psi_{\text{ref}} = const$, the reference values of the roll $\phi_{\text{ref}}$ and pitch $\theta_{\text{ref}}$ become
\begin{equation}
\label{ugao_rewrite}
\begin{bmatrix}
\phi_{\text{ref}} \\
\theta_{\text{ref}}
\end{bmatrix}
= 
\frac{1}{g}
\begin{bmatrix}
s \psi_{\text{ref}} & -c \psi_{\text{ref}} \\
c \psi_{\text{ref}} & s \psi_{\text{ref}}
\end{bmatrix}
\begin{bmatrix}
\ddot{x} \\
\ddot{y}  
\end{bmatrix}.
\end{equation}

The attitude controller \eqref{eq:Upravljacki_zakon_visina} requires the first and the second derivation of the roll and pitch reference values. By differentiating \eqref{ugao_rewrite}, we get
\begin{align}
\label{izvodi_uglova}
\begin{bmatrix}
\dot{\phi}_{\text{ref}} \\
\dot{\theta}_{\text{ref}}
\end{bmatrix}
&= 
\frac{1}{g}
\begin{bmatrix}
s \psi_{\text{ref}} & -c \psi_{\text{ref}} \\
c \psi_{\text{ref}} & s \psi_{\text{ref}}
\end{bmatrix}
\begin{bmatrix}
\dddot{x} \\
\dddot{y}  
\end{bmatrix}, \nonumber \\
\begin{bmatrix}
\ddot{\phi}_{\text{ref}} \\
\ddot{\theta}_{\text{ref}}
\end{bmatrix}
&= 
\frac{1}{g}
\begin{bmatrix}
s \psi_{\text{ref}} & -c \psi_{\text{ref}} \\
c \psi_{\text{ref}} & s \psi_{\text{ref}}
\end{bmatrix}
\begin{bmatrix}
\ddddot{x} \\
\ddddot{y}  
\end{bmatrix}.
\end{align}
If $\dddot{x}$, $\dddot{y}$, $\ddddot{x}$, $\ddddot{y}$ are known, then one can obtain $\dot{\phi}_{\text{ref}}$, $\dot{\theta}_{\text{ref}}$, $\ddot{\phi}_{\text{ref}}$ and $\ddot{\theta}_{\text{ref}}$ from \eqref{izvodi_uglova}. By conducting additional differentiation of \eqref{xy_rewrite}, we get $(\dddot{x},\dddot{y})$ and $(\ddddot{x},\ddddot{y})$ as follows
\begin{align}
\begin{bmatrix}
\dddot{x} \\
\dddot{y}
\end{bmatrix}
&= 
\begin{bmatrix}
\dddot{x}_{\text{ref}} \\
\dddot{y}_{\text{ref}}
\end{bmatrix} +
K_{d} 
\begin{bmatrix}
\ddot{e}_x \\
\ddot{e}_y
\end{bmatrix} +
K_{p}
\begin{bmatrix}
\dot{e}_x \\
\dot{e}_y
\end{bmatrix},
\nonumber \\
\begin{bmatrix}
\ddddot{x} \\
\ddddot{y}
\end{bmatrix}
&= 
\begin{bmatrix}
\ddddot{x}_{\text{ref}} \\
\ddddot{y}_{\text{ref}}
\end{bmatrix} +
K_{d} 
\begin{bmatrix}
\dddot{e}_x \\
\dddot{e}_y
\end{bmatrix} +
K_{p}
\begin{bmatrix}
\ddot{e}_x \\
\ddot{e}_y
\end{bmatrix}.
\label{4th_der_pos1}
\end{align}
Eqs. \eqref{xy_rewrite} to \eqref{4th_der_pos1} represent the $xy$ position controller and the flat mapping between $(x, y)$ and $(\phi, \theta)$. The proposed controller is able to track the reference $x_{ref}$, $y_{ref}$, and $z_{ref}$ as well as $\psi_{ref}$. The proposed architecture has been exploited in \cite{Osmic_AIM} to control the position and orientation of the octocopter. 

It is necessary to emphasize that the reference trajectories of the $x$ and $y$ position coordinates must be at least four times differentiable, while trajectories for the altitude  $z$ and orientation $\psi$ must have the first and the second derivation. These trajectories are provided by an adequate motion planning algorithm, while references for $\phi$ and $\theta$ orientation coordinates are provided as the output of the \textit{xy} controller. 

\begin{figure}\begin{center}
\includegraphics[scale=0.9]{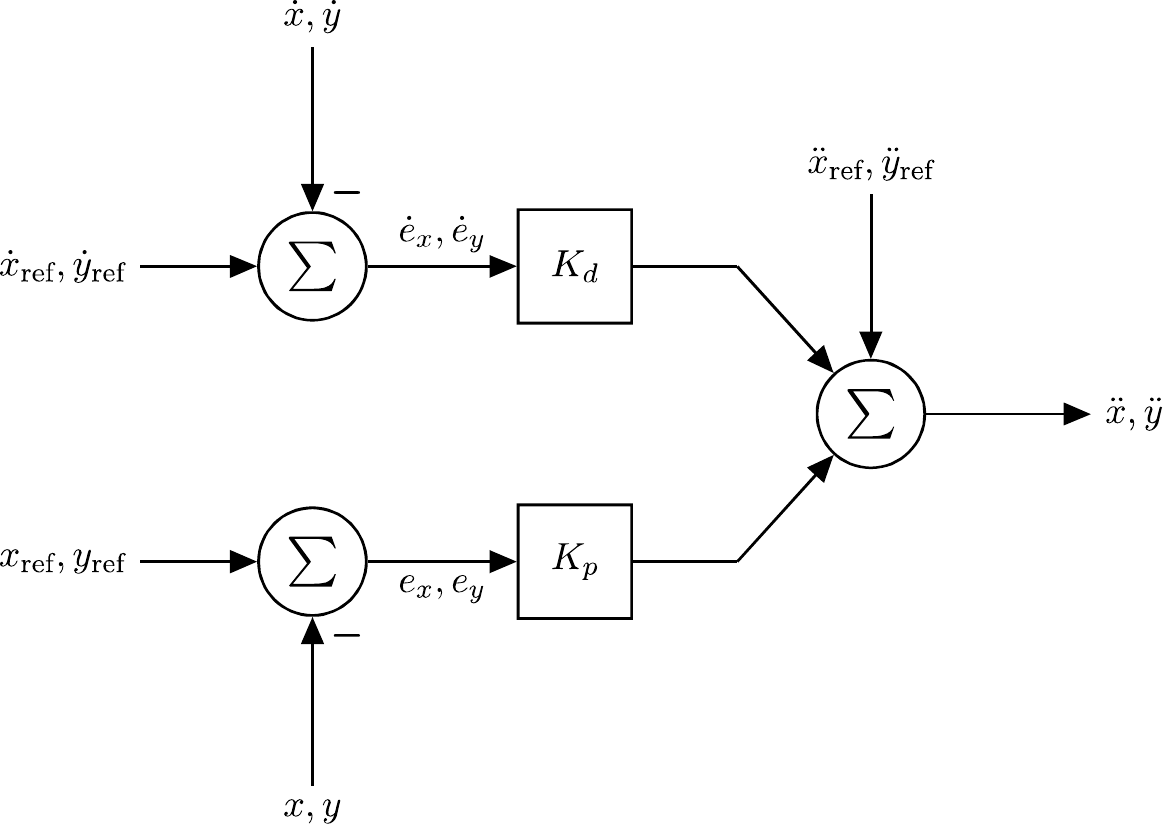}\end{center}
\caption{Control diagram for tracking $x$ and $y$ Cartesian coordinates.\label{fig:lPracenje_pozicije}}
\end{figure}

\begin{figure}
\begin{centering}
\includegraphics[scale=0.7]{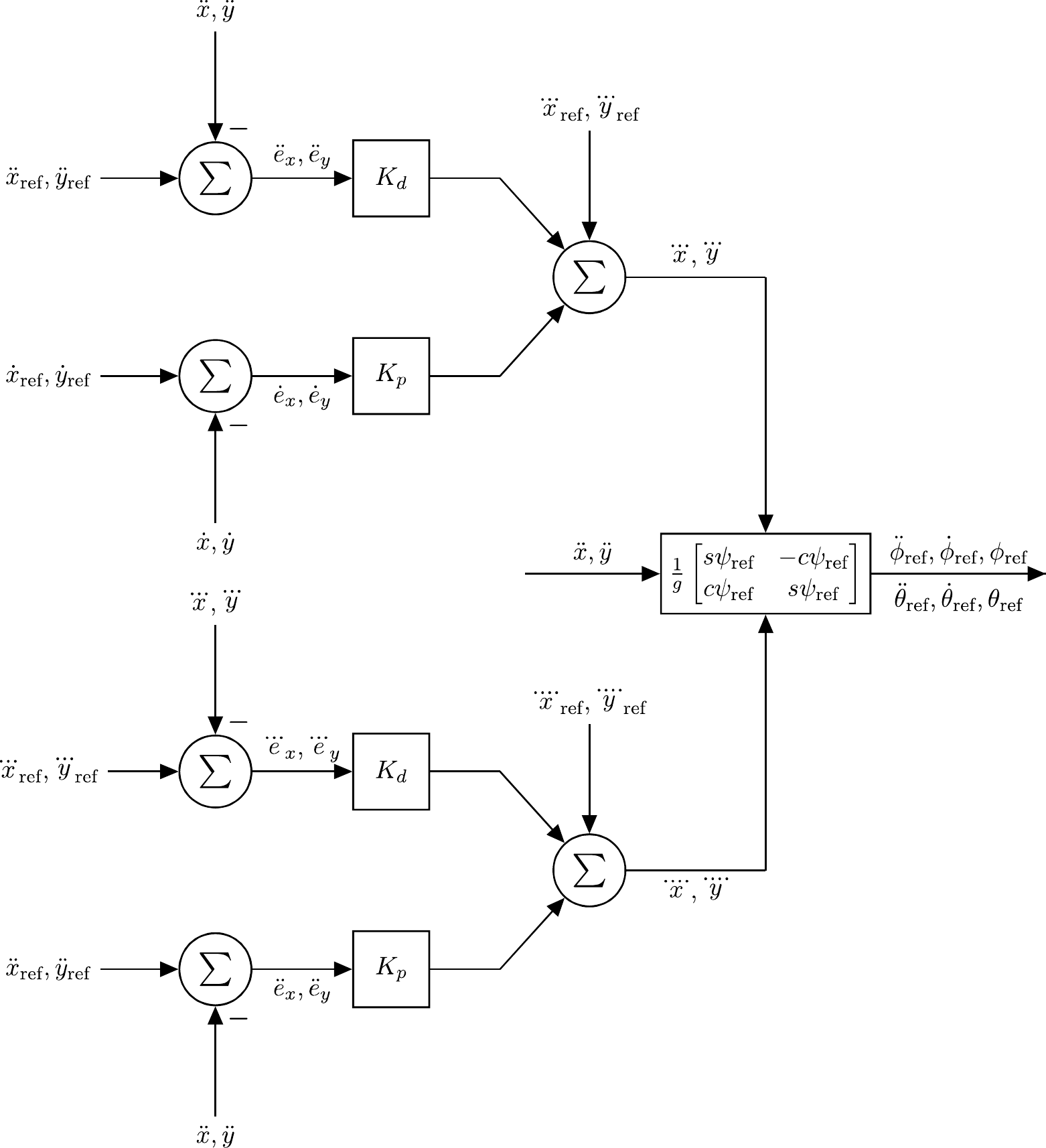}
\par\end{centering}
\caption{Flatness-based control diagram.\label{fig:preslikavanje-flatness}}

\end{figure}

Fig. \ref{fig:lPracenje_pozicije} shows the block structure of the position tracking controller, while Fig. \ref{fig:preslikavanje-flatness} illustrates how to transform $x$ i $y$ positions based on the flat mapping (\eqref{xy_rewrite} to \eqref{4th_der_pos1}). Figs. \ref{fig:lPracenje_pozicije} and \ref{fig:preslikavanje-flatness}, together with the altitude controller shown in Fig. (\ref{fig:regulator_visine}) construct the position controller of the octocopter. 

\begin{figure}[p]
\begin{centering}
\includegraphics[scale=0.42]{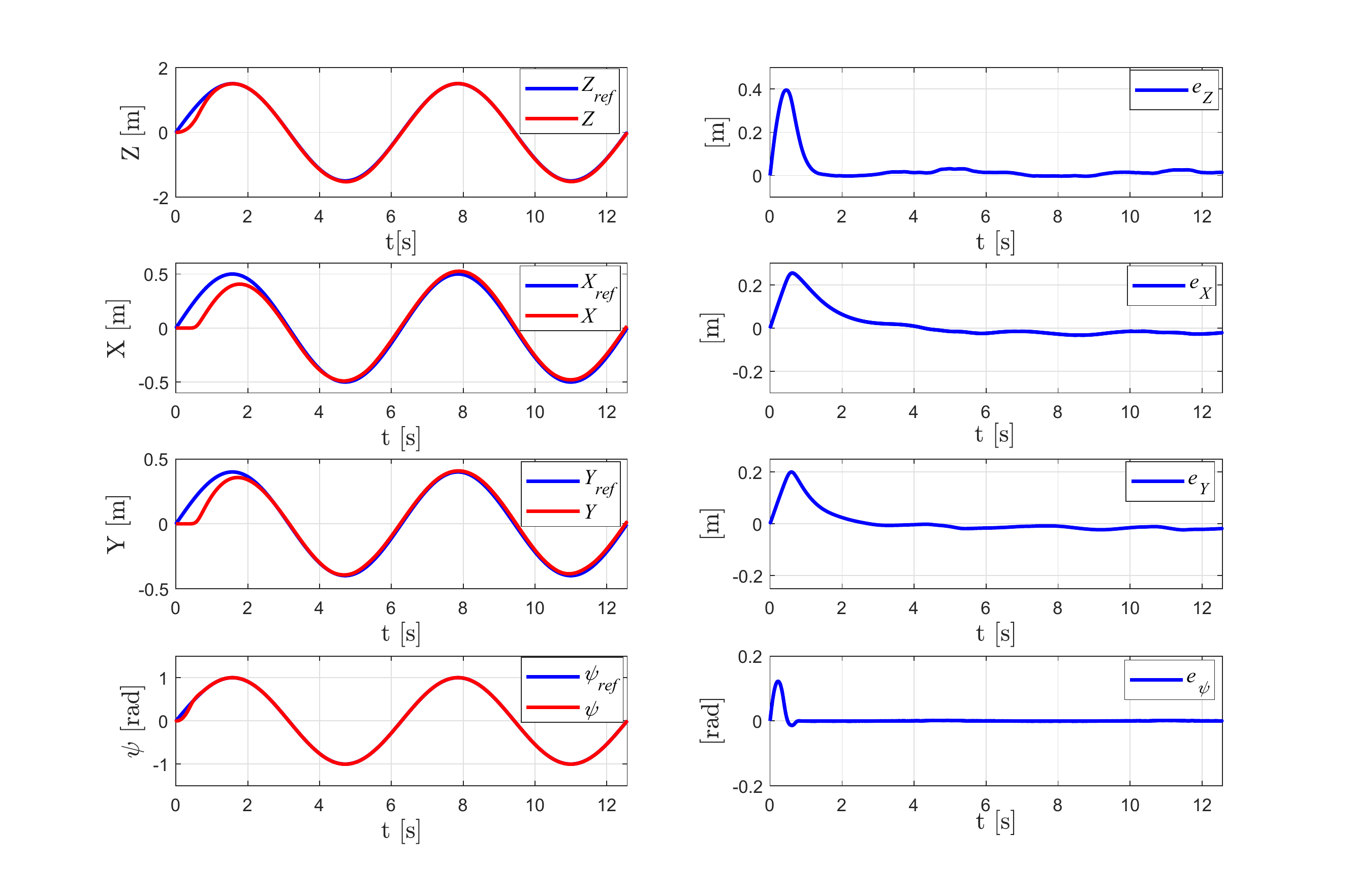}
\par\end{centering}
\caption{Position and orientation tracking results.\label{fig:pracenja_pozicije_sve}}

\end{figure}

Fig. \ref{fig:pracenja_pozicije_sve} shows the results obtained for position and orientation tracking for $K_{dz}=14.5$,
$K_{pz}=52.56$, $K_{d}=45$ and $K_{p}=506.5$. Although the reference trajectories are quite demanding, it can be seen that the tracking errors vanish within a 4-second interval and the overall tracking performance can be considered satisfactory.   

\section{Overall UAV control architecture}
\color{black}
The final control architecture for trajectory tracking which includes $x$, $y$, $z$ coordinates, as well as orientation  $\psi$, is shown in Fig. \ref{fig:Regulator_sve}. The architecture contains the $xy$ positional controller which includes flat mapping, altitude controller and attitude tracking controllers, control allocation and low-level control of motor angular velocity for each single motor. 

As previously indicated, the reference trajectories for $x$ and $y$ coordinates should be at least four times differentiable, while the reference trajectories for the altitude $z$ and the orientation $\psi$ should be at least two times differentable. In addition, one can notice that the change of $\phi$ and $\theta$ angles will be the result of the $xy$ position controller. 

\begin{figure}[p]
\begin{centering}
\includegraphics[scale=0.8]{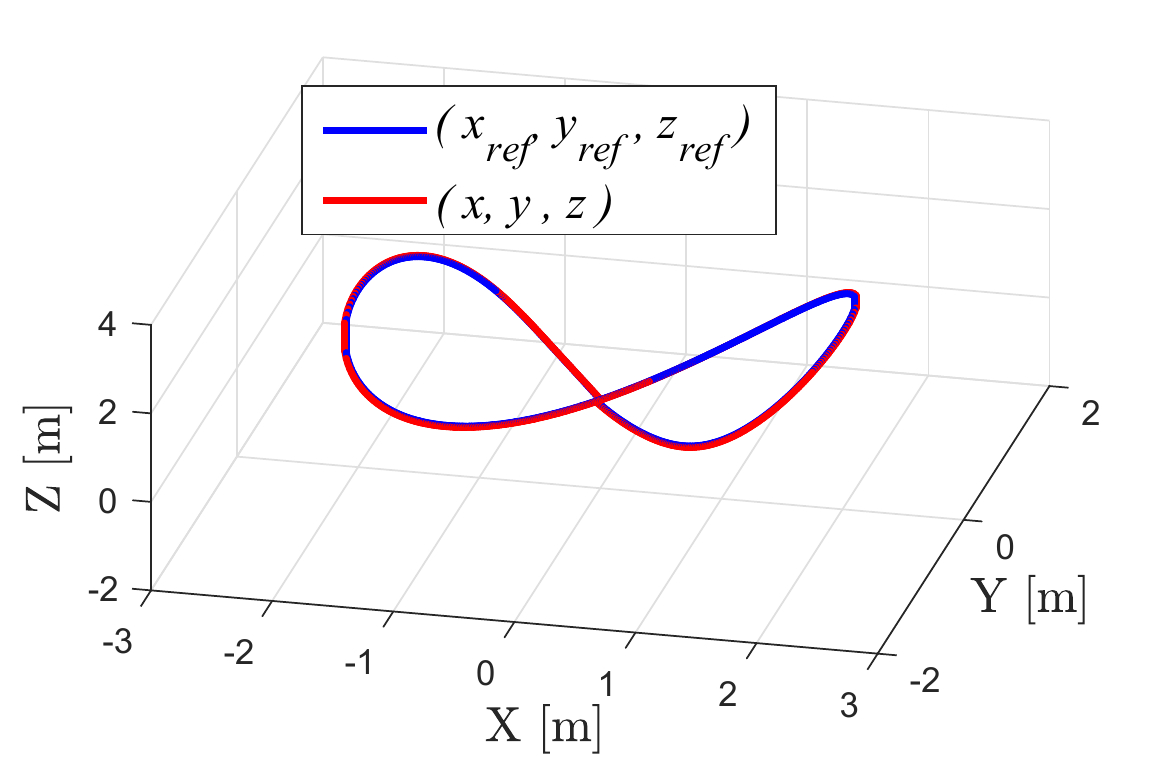}
\par\end{centering}
\caption{Trajectory tracking of Viviani curve achieved with an PNPNPNPN octocopter configuration structure without any failure states.
\label{fig:Vivijani_3D_ispravno}}

\end{figure}

\begin{figure}[p]
\includegraphics[scale=0.42]{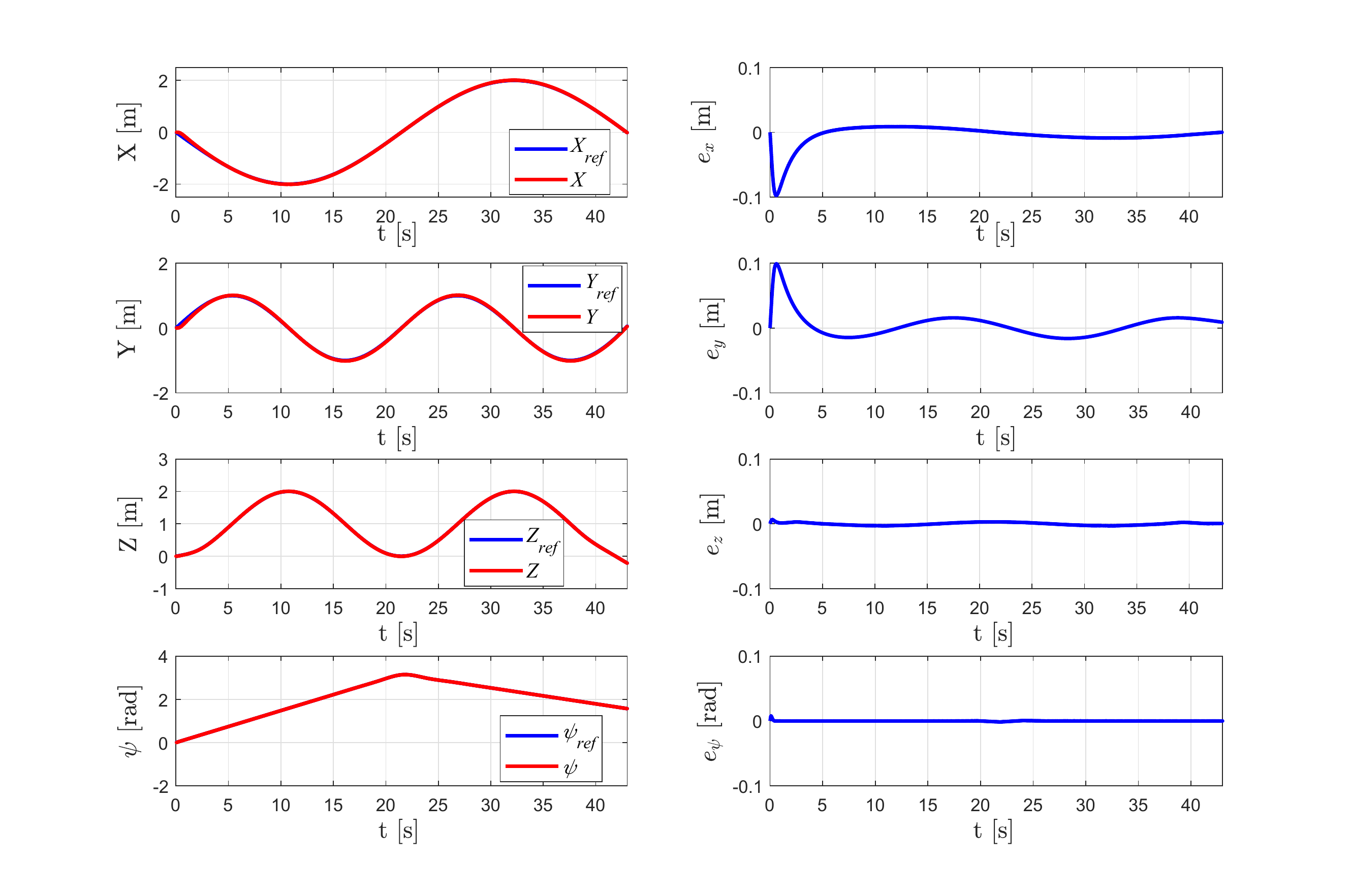}
\caption{Trajectory tracking for each position and orientation achieved with an PNPNPNPN octocopter configuration structure without any failure states. Left: reference and achieved values. Right: tracking errors.
\label{fig:vivijani_pracenje_bez_greske} }
\end{figure}

To illustrate the proposed control architecture for the position and orientation tracking, we consider the Vivian curve (ococopter with PPNNPPNN configuration) in the three-dimensional space, which is represented in Fig. \ref{fig:Vivijani_3D_ispravno}. The tracking errors with respect to the coordinates $x$, $y$, $z$ and the orientation $\psi$ are give in Fig. \ref{fig:vivijani_pracenje_bez_greske} from which one can observe a small tracking errors with respect to the chosen reference trajectory. 

\medskip{}

\section{RLS-based fault tolerant control}

\begin{figure}[p]
\begin{centering}
\includegraphics[scale=0.5]{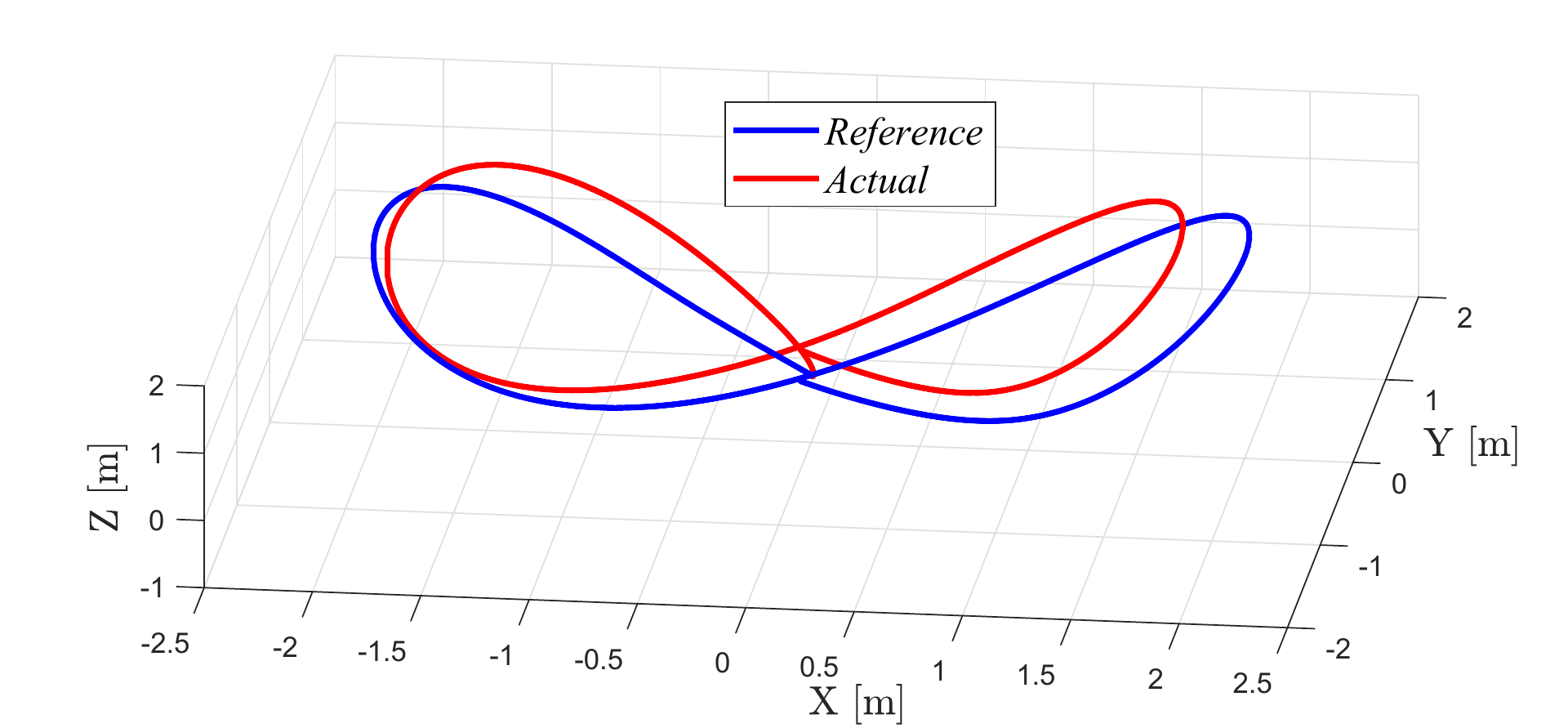}
\par\end{centering}
\caption{Trajectory tracking achieved with an PNPNPNPN octocopter configuration structure with the motor $M_3$ being in a fault state.\label{fig:Vivijani_3d_neispravan}}
\end{figure}

\color{black}
As a motivational example for the next subsection, we present here the tracking performance in case one motor (e.g., $M_3$) is in fault state from the start of the flying mission. As it can be seen from Fig. \ref{fig:Vivijani_3d_neispravan}, there exist a permanent tracking error, where, for instance, the tracking error for $x$ coordinate becomes 0.5 meters at some time instants. From this simple example, one can conclude that a single motor failure may cause a permanent error with respect to some coordinates. Depending on which motor is in a fault state, the tracking error may vary along different coordinates. It is also worth noting that the octocopcter is a system with a redundant structure (there exist the remaining active 7 motors) and the controller used in this example does not exploit any information about the motor failure. For this reason, the control allocation generates the control inputs to that motor as well, producing undesired system behaviour. 

In order to fully exploit the redundant structure of an octocopter system, it is worth designing a controller that will take into account the information about motor failure and allocate the control inputs only to those active motors in order to achieve a satisfactory tracking performance. 

\medskip{}

\medskip{}

\begin{figure}[p]
\begin{centering}
\includegraphics[scale=0.42]{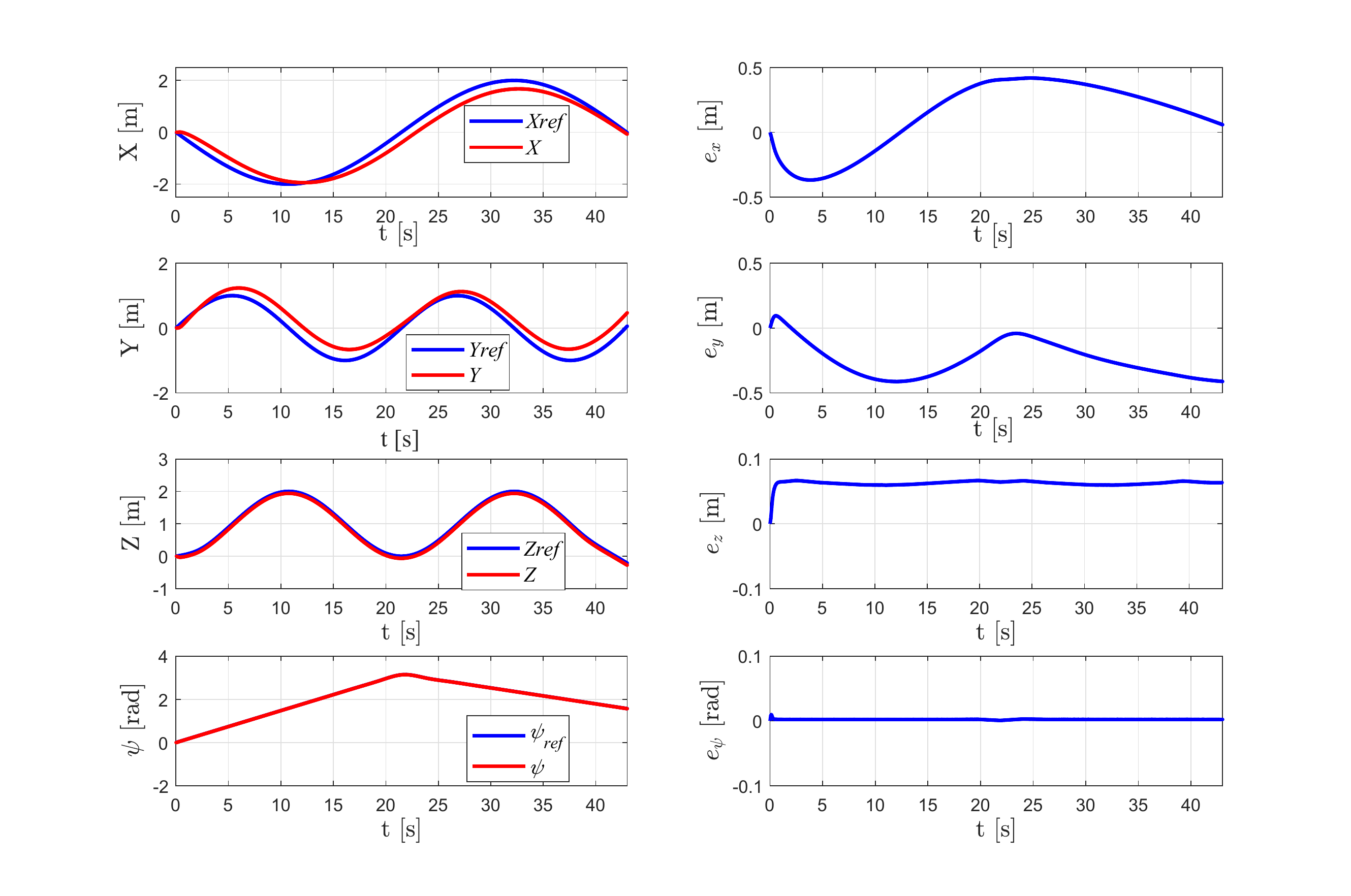}
\par\end{centering}
\caption{Trajectory tracking achieved with an PNPNPNPN octocopter configuration structure and $M_3$ being in a fault state. Left:
reference and achieved values, Right: tracking errors. \label{fig:Vivijani_pogreska_neispravan}}
\end{figure}

\medskip{}

\subsection{Fault-tolerant PD tracking control}
\color{black}
To design a fault-tolerant control, it is necessary to include information about fault states of DC motors into the actuation matrix. Accordingly, we can rewrite the actuation matrix as:
\begin{equation}
\boldsymbol{u} = \boldsymbol{A} \text{diag}\left(\boldsymbol{\Omega}_s\right)\boldsymbol{\theta},
\label{lin_model}
\end{equation}
where $\boldsymbol{\theta} = \begin{bmatrix}
\theta_1 &\theta_2 &\theta_3 &\theta_4 &\theta_5 &\theta_6 &\theta_7 &\theta_8 \end{bmatrix}^T$ represents the fault state vector of DC motors. The coefficients $0 \leq \theta_i \leq 1 \; (i = \overline{1..8})$ represent the failure level of related DC motor, where $\theta_i=1$ represents a fully available $M_i$ motor, $\theta_i=0$ a failed $M_i$ motor, while all other values in between represent a partial loss of the related DC motor functionality that is their working capacity level. If we rewrite the actuation matrix as $\boldsymbol{A} = \begin{bmatrix} \boldsymbol{A}_T & \boldsymbol{A}_1 & \boldsymbol{A}_2 & \boldsymbol{A}_3 \end{bmatrix}^T$, the components of the $\boldsymbol{\tau_{x}}$, $\boldsymbol{\tau_{y}}$ and  $\boldsymbol{\tau_{z}}$ vectors of the controllable control signals $\boldsymbol{u}$ can be represented as weighted scalar products in the form 
\begin{align}
\label{aktuacijapa_paramterizirana}
T &= \boldsymbol{A}_T \text{diag}\left(\boldsymbol{\Omega}_s\right) \boldsymbol{\theta} \nonumber \\
\tau_{x} &= \boldsymbol{A}_1 \text{diag}\left(\boldsymbol{\Omega}_s\right) \boldsymbol{\theta} \nonumber \\
\tau_{y} &= \boldsymbol{A}_2 \text{diag}\left(\boldsymbol{\Omega}_s\right) \boldsymbol{\theta} \nonumber \\
\tau_{z} &= \boldsymbol{A}_3 \text{diag}\left(\boldsymbol{\Omega}_s\right) \boldsymbol{\theta}.
\end{align}

As we can see from \eqref{aktuacijapa_paramterizirana}, the control output can be represented by its four linearly dependant components. Based on these values, it is possible to estimate the parameter vector $\boldsymbol{\theta}$ in a least-squares manner. For the estimation, it is necessary to know all other parameters in \eqref{aktuacijapa_paramterizirana}. The basic requirement is that the values of the actuation matrix $\boldsymbol {A}$ are a priori known. We only use gyroscopic data to detect and isolate failures, so that the first equality from \eqref{aktuacijapa_paramterizirana} can be omitted, so the final model which will be used for prediction is
\begin{align}
\label{predikcija}
\tau_{x} &= \boldsymbol{A}_1 \text{diag}\left(\boldsymbol{\Omega}_s\right) \boldsymbol{\theta} \nonumber \\
\tau_{y} &= \boldsymbol{A}_2 \text{diag}\left(\boldsymbol{\Omega}_s\right) \boldsymbol{\theta} \nonumber \\
\tau_{z} &= \boldsymbol{A}_3 \text{diag}\left(\boldsymbol{\Omega}_s\right) \boldsymbol{\theta}.
\end{align} 
If we have $N$ measurements at time instances  $1$ to $N$, then for each sample $i=\overline{1..N}$, based on \eqref{predikcija}, we have the following model to predict the output 
\begin{align}
\label{predikcija_01}
\hat{\tau}_x(i) &= \boldsymbol{\varPsi}_{1}^{r}(i) \boldsymbol{\theta}, \nonumber \\
\hat{\tau}_y(i) &= \boldsymbol{\varPsi}_{2}^{r}(i) \boldsymbol{\theta}, \nonumber \\
\hat{\tau}_z(i) &= \boldsymbol{\varPsi}_{3}^{r}(i) \boldsymbol{\theta},
\end{align}
where each regressor of output models is defined as 
\begin{align}
\label{regresor}
\boldsymbol{\varPsi}_{1}^{r}(i) &= \boldsymbol{A}_1 \text{diag}\left(\boldsymbol{\Omega}_s(i)\right),\nonumber \\
\boldsymbol{\varPsi}_{2}^{r}(i) &= \boldsymbol{A}_2 \text{diag}\left(\boldsymbol{\Omega}_s(i)\right),\nonumber \\
\boldsymbol{\varPsi}_{3}^{r}(i) &= \boldsymbol{A}_3 \text{diag}\left(\boldsymbol{\Omega}_s(i)\right).
\end{align}
Using the property that each tensor can be represented as a matrix using the skew-symmetric matrix $\boldsymbol{S}$ defined in \eqref{skew}, we can represent the inertia tensor $\boldsymbol{J} = \text{diag}\left(\begin{bmatrix}I_{xx} & I_{yy} & I_{zz}\end{bmatrix}\right)$ as a symmetric matrix. Now, we can rewrite \eqref{angular_motion} as
\begin{align}
{\tau}_x(i) &= I_{xx} \dot{P}(i) - (I_{yy} - I_{zz}) Q(i) R(i),\nonumber \\
{\tau}_y(i) &= I_{yy} \dot{Q}(i) - (I_{zz} - I_{xx}) P(i) R(i),\nonumber \\
{\tau}_z(i) &= I_{zz} \dot{R}(i)- (I_{xx} - I_{yy}) P(i) Q(i). 
\end{align}

We can now formulate the FDI technique for the propulsion system as a recursive least square (RLS) estimation problem of the rotor capacity vector $\boldsymbol{\theta}$ in the following way:
\begin{equation}
\label{N_uzoraka}
    \resizebox{0.89\hsize}{!}{%
        $\boldsymbol{\tau} =
\begin{bmatrix}
{\tau}_x(1) &
{\tau}_y(1) &
{\tau}_z(1) &
... &
{\tau}_x(N) &
{\tau}_y(N) &
{\tau}_z(N) 
\end{bmatrix}^T,$      
        }
\end{equation}
where the data matrix $\boldsymbol{\varPsi}^{r}$ has the form
\begin{equation}
    \resizebox{0.89\hsize}{!}{%
        $\boldsymbol{\varPsi}^{r} = 
\begin{bmatrix}
\boldsymbol{\varPsi}_{1}^{rT}(1) &
\boldsymbol{\varPsi}_{2}^{rT}(1) &
\boldsymbol{\varPsi}_{3}^{rT}(1) &
... &
\boldsymbol{\varPsi}_{1}^{rT}(N) &
\boldsymbol{\varPsi}_{2}^{rT}(N) &
\boldsymbol{\varPsi}_{3}^{rT}(N) 
\end{bmatrix}^T.$      
        }
\end{equation}
By using a classical non-recursive least-square method \cite{isermann_ident}, we can express the coefﬁcient $\boldsymbol{\hat{\theta}}$ as
\begin{equation}
\label{Thete_proracun}
\boldsymbol{\hat{\theta}}=\left(\boldsymbol{\varPsi}^{rT}\boldsymbol{\varPsi}^{r}\right)^{-1}\boldsymbol{\varPsi}^{rT}\boldsymbol{\tau}.
\end{equation}

It is necessary to emphasize that this RLS technique has a linear configuration, while obtaining values for the vector rotor capacity comes from the nonlinear octocopter model \eqref{angular_motion}. The proposed method can be applied to any type of MAV with $2n$ rotors mounted in a planar plane. 

To apply the aforementioned RLS technique to an ococopter system, the RLS with a forgetting factor is used. The values for $\theta_i$ obtained by the RLS algorithm are used as a feedback to update the actuation matrix. Note that $\boldsymbol{\theta}$ and $\boldsymbol{\Omega_s}$ represent vectors of the same size, so  \eqref{lin_model} can be rewritten as 
\begin{equation}
\boldsymbol{u} = \boldsymbol{A} \text{diag}\left(\boldsymbol{\theta}\right) \boldsymbol{\Omega}_s.
\label{nova_jedl}
\end{equation}
Furthermore, by introducing a new matrix $\boldsymbol{B}$ as $\boldsymbol{B} = \boldsymbol{A} \text{diag}\left(\boldsymbol{\theta}\right)$, we can calculate the velocity for all DC motors that can achieve the reference thrust and torques as
\begin{equation}
\label{pseudo_konacana}
\boldsymbol{\Omega_s}= \boldsymbol{B}^{+}\boldsymbol{u},
\end{equation}
where $\boldsymbol{B}^{+}=\boldsymbol{B}^{T}(\boldsymbol{B\boldsymbol{B}^{T}})^{-1}$.

\subsection{Simulation results for Fault-tolerant PD tracking control}

In this subsection, we present the simulation results for the RLS-based technique for detection and isolation  of DC motor failures on an octocopter system. We used a fault-tolerant PD tracking control system around the hovering configuration. To illustrate that the designed controller is capable of handling a fault state on an octocopter system, we consider the PPNNPPNN conﬁguration. 

\begin{figure}[!t]
\centering
\includegraphics[scale=0.7]{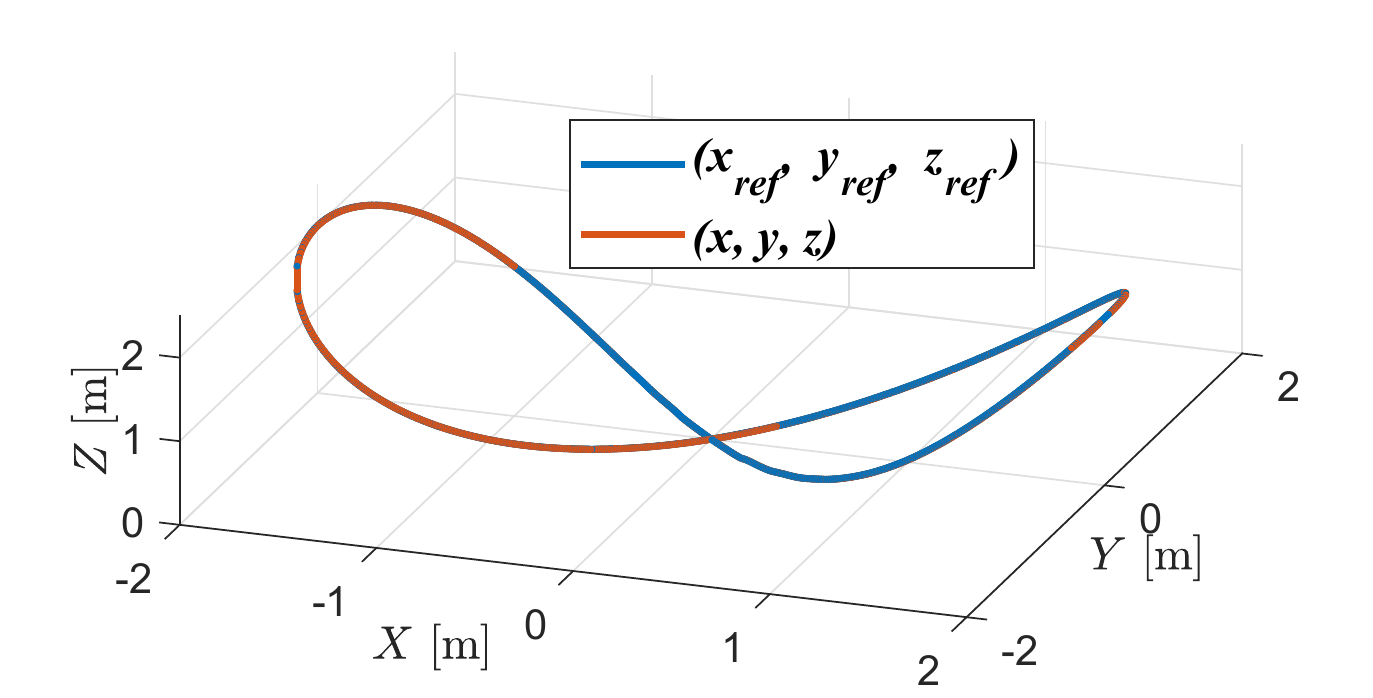}
\caption{Tracking performance achieved in 3D space with an PNPNPNPN octocopter conﬁguration structure based on a PD controller and control allocation algorithm with a fault state in motor $M_3$ starting from $t=5$ $[s]$.} 
\label{fig:3D_fault}
\end{figure}
\begin{figure}[!t]
\centering
\includegraphics[scale=0.7]{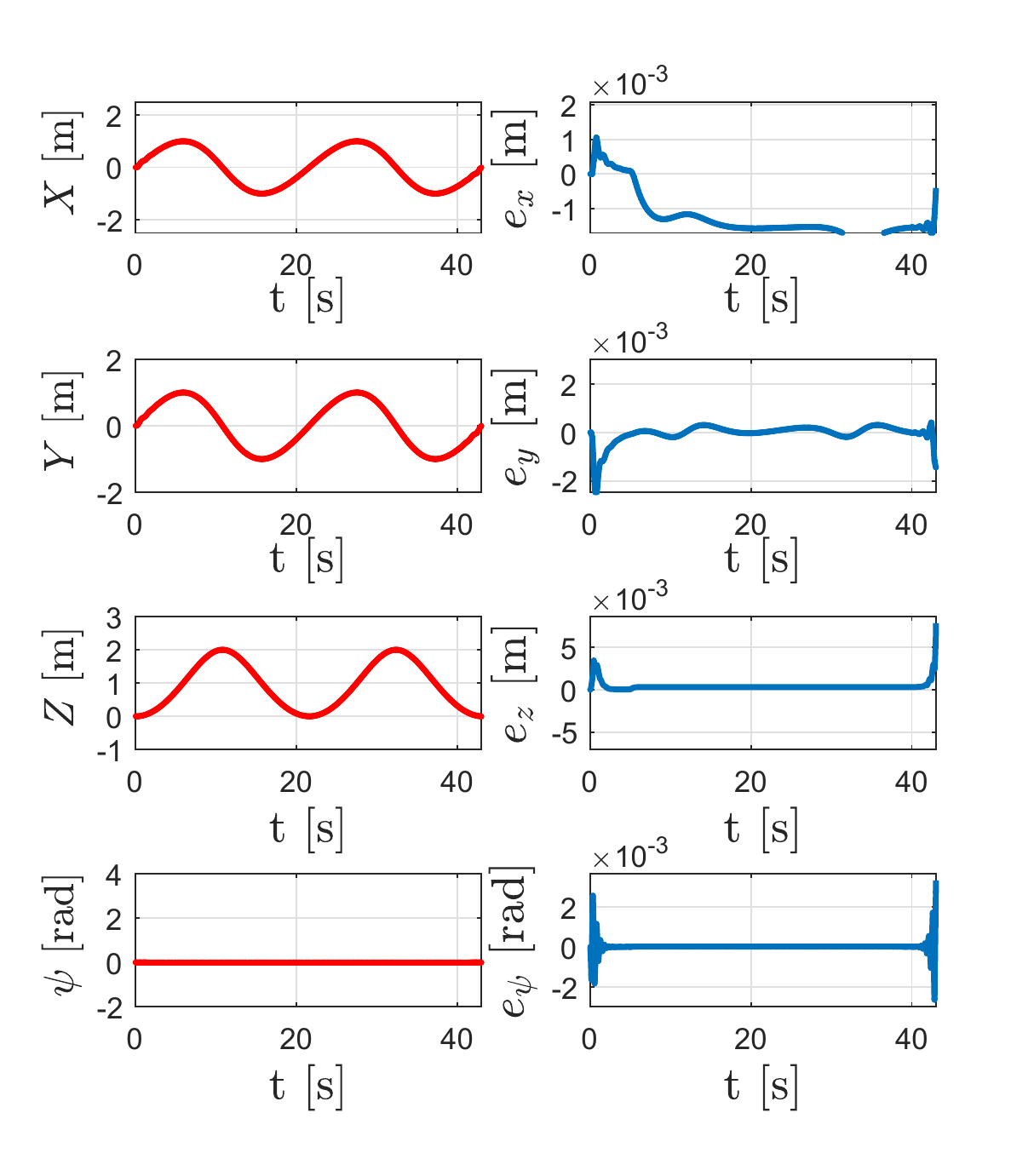}
\caption{Tracking performance achieved with an PNPNPNPN octocopter configuration based on a PD control with a fault state in motor $M_3$ starting from $t=5$ $[s]$. Left: reference values. Right: tracking error.} 
\label{fig:Pracenje_fault}
\end{figure}
\begin{figure}[!t]
\centering
\includegraphics[scale=0.7]{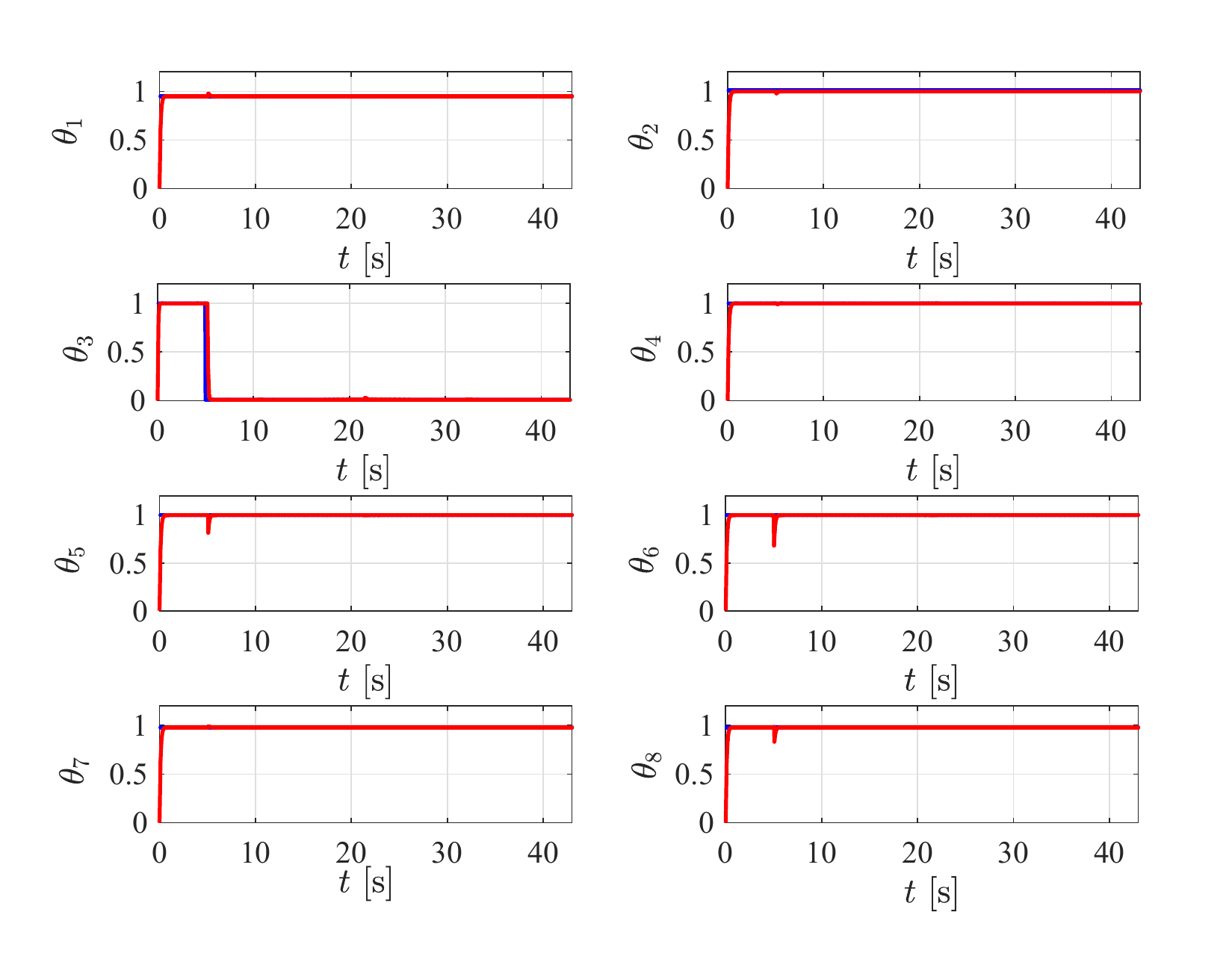}
\caption{Estimation of rotor capacities  during the tracking task.} 
\label{fig:Parametri_fault}
\end{figure}

It should be emphasized that the presented approach was tested for different types of possible faults (different numbers of failed motors and different values of rotor capacity). For the purpose of the RLS algorithm \eqref{N_uzoraka}, we need to determine eight unknowns parameters $\theta_{i}$, so we need at least eight equations. It follows that the number of samples $N$ must be $N\geq8$.  However, to eliminate the impact of noise, it is advisable to use a larger number of equations, that is $N\gg8$. So, we set the forgetting factor to 0.8 in order to take into account the measurements from the previous 0.8 seconds to provide a sufficient number of samples for the RLS algorithm. Let now a failure related to the motor $M_3$ occurs at time $t=5$ $[s]$, which can be expressed by the rotor capacity vector $\boldsymbol{\theta} = \begin{bmatrix} 1 & 1 & 0 & 1 & 1 & 1 & 1 & 1\end{bmatrix}^T$. 
Figs. \ref{fig:3D_fault} and \ref{fig:Pracenje_fault} depict the performance of the RLS-based PD tracking controller merged with the control allocation for the given failure. Fig. \ref{fig:Parametri_fault} also shows that the relative capacity of each motor is properly estimated. 

[scale=0.4]

From Fig. \ref{fig:Pracenje_fault}, we can conclude that the presented RLS controller has an acceptable tracking performance which is similar to the tracking performance of a healthy octocopter system (an octocopter system without any faulty states). Such a result is expected and it can be attributed to fault-tolerability that an octocopter system inherently possesses by design.

%\section{Conclusion}

%In this chapter, we presented a novel RLS-based technique for detection and isolation of propulsion system faults in MAVs. Based on this technique, we developed a fault-tolerant PD tracking control system around hover configuration. Simulation results, on a severe triple rotor fault scenario, indicated proper fault detection and isolation, which resulted in notable tracking performance regardless of the multirotor fault present in the system. In the next chapter, we provide a thorough analysis of the effects which different motor failures may cause in order to further exploit this information and to increase the reliability of the octocopter system. As we have seen from this chapter, this can be done using a fault-tolerant control. However, we illustrate how this goal can be achieved during the planning stage as well, aiming to consider the so-called risk-aware trajectories which would secure the octocopter system to be properly prepared (positioned) in case the motor related failures occur during the mission execution.

\chapter{Maneuverability }

\color{black}

In order to develop a motion planner and estimate a possibility of completing a pre-planned mission, it is necessary to determine whether the system is capable of generating necessary thrust and torques, to be able to reach the waypoints generated by the motion planner with available DC motors. Regardless of whether the system contains a redundant actuation or not, it is possible to have a case when a control algorithm is not able to track the referent trajectory (e.g, a fault state). This problem has been addressed in \cite{FTC_ETH_Thomas_Schneider,Korejci_controllability_I,Korejci_controllability_II,shi_Analiza,Lunze_I,Lunze_II} for a MAV designed with fixed and classical rotor configurations (quadcopter, hexacopter and octocopter). In addition, in \cite{franchi_heksa} and \cite{mehmood_analiza}, the controllability analysis has been considered for a MAV designed with a tilted rotor, while for non-classical (coaxial) octocopters the relevant analysis has been given in \cite{coaxial_octo}. In this chapter, an empirical method is presented which can be used for any MAV configuration designed with different number of rotors and their rotational directions.

\section{Fault-dependent admissible set of thrust force
and torques}

To find an admissible set of thrust force and torques in control space, it is necessary to check whether the system can reach and stay in a hovering state without any rotation. For illustration purposes, consider again an octocopter with the PNPNPNPN configuration. The relation between the control inputs $u$ (the reference thrust  force $T$ and torques $\boldsymbol{\tau}$) and the rotation velocity $\boldsymbol{\Omega}_s$ of DC motors is given with $\boldsymbol{u_{ref}} = \boldsymbol{A} \boldsymbol{\Omega}_s$ (see Chapter 3), where the control vector $\boldsymbol{u_{ref}}$ is represented by
\begin{equation}
u_{ref}=[\begin{array}{cc}
T & \tau\end{array}]^{T}=[\begin{array}{cccc}
T & \boldsymbol{\tau_{x}} & \boldsymbol{\tau_{y}} & \boldsymbol{\tau_{z}}\end{array}]^{T},\label{zeljene_sile_i_momenti}
\end{equation}
$\boldsymbol{\Omega}_s \in D_{\Omega_s} \subset \mathbb{R}^8$ and $\boldsymbol{u} \in D_{u} \subset \mathbb{R}^4$. The set $D_{\Omega_s}$ is defined based on velocity constraints of DC motors
\begin{equation}
\label{omega_ogranicenje}
0 \leq \Omega^2_i \leq \Omega^2_{\text{max}}, \; \; i = \overline{1..8}.
\end{equation}

\begin{figure}[p]
\begin{centering}
\includegraphics[width=7cm,height=7cm]{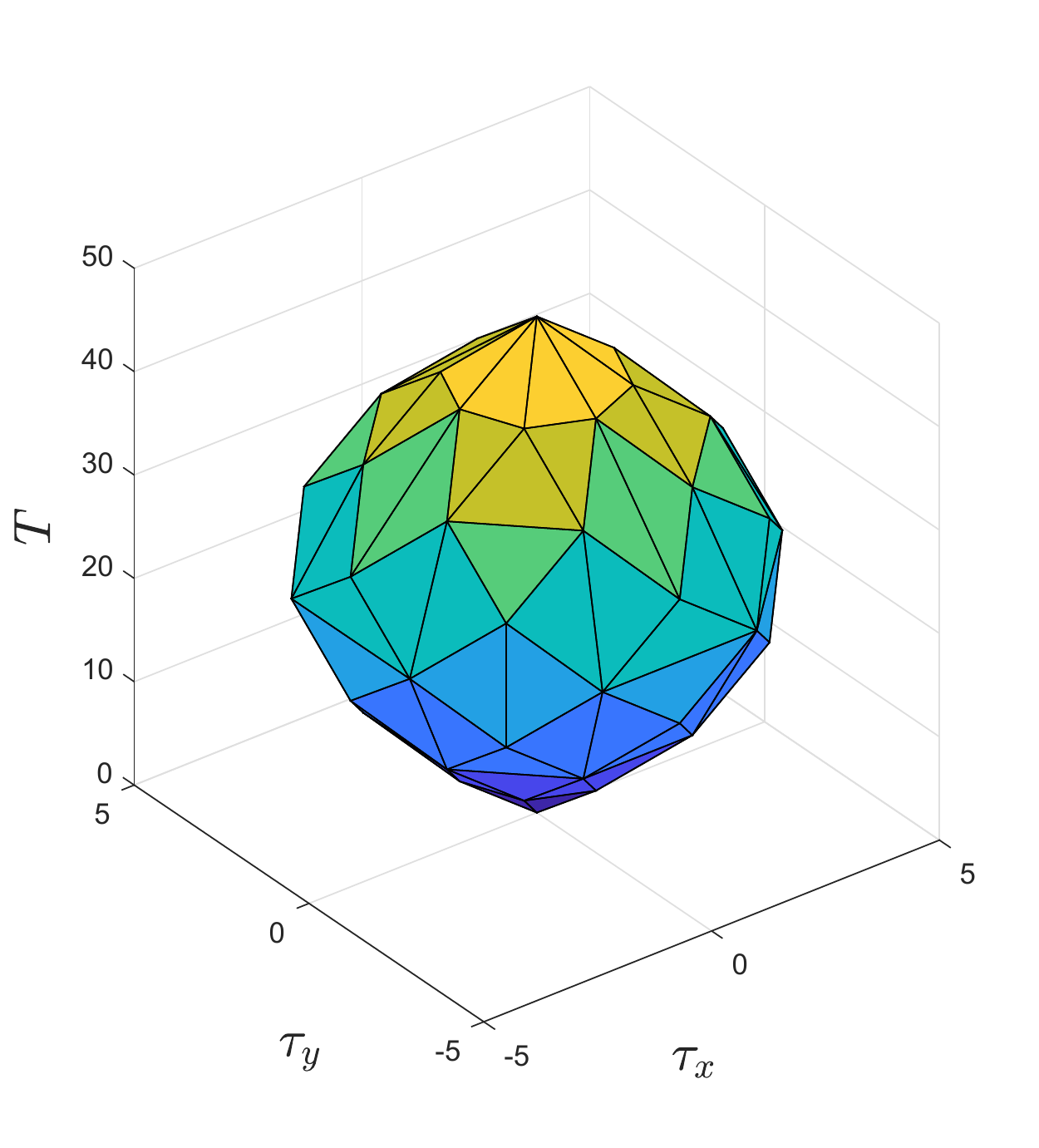}
\caption{{Representation of the four-dimensional admissible region ($T$, $\tau_{x}$, $\tau_{y}$ $\tau_{z}$) in three-dimensional space ($T$, $\tau_{x}$, $\tau_{y}$) when $\tau_{z}=0$.
\label{fig:Oblast_X_Y}}}
\par\end{centering}
\end{figure}

\begin{figure}[p]
\begin{centering}
\includegraphics[width=7cm,height=7cm]{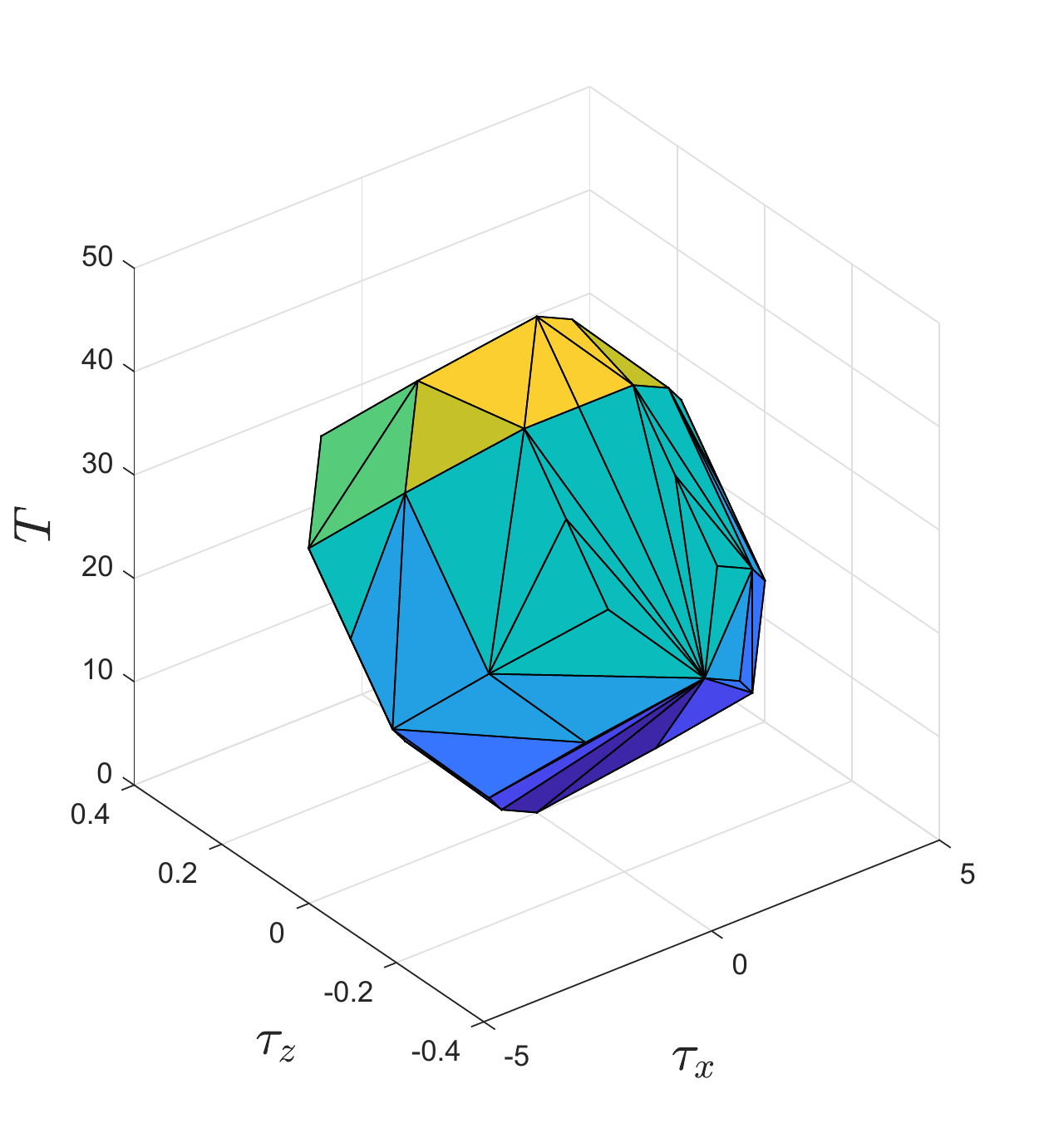}
\par\end{centering}
\caption{{Representation of the four-dimensional admissible region ($T$, $\tau_{x}$, $\tau_{y}$ $\tau_{z}$) in three-dimensional space ($T$, $\tau_{x}$, $\tau_{z}$) when $\tau_{y}=0$.
 \label{fig:Oblast_X_Z}}}

\end{figure}

\begin{figure}[p]
\begin{centering}
\includegraphics[width=7cm,height=7cm]{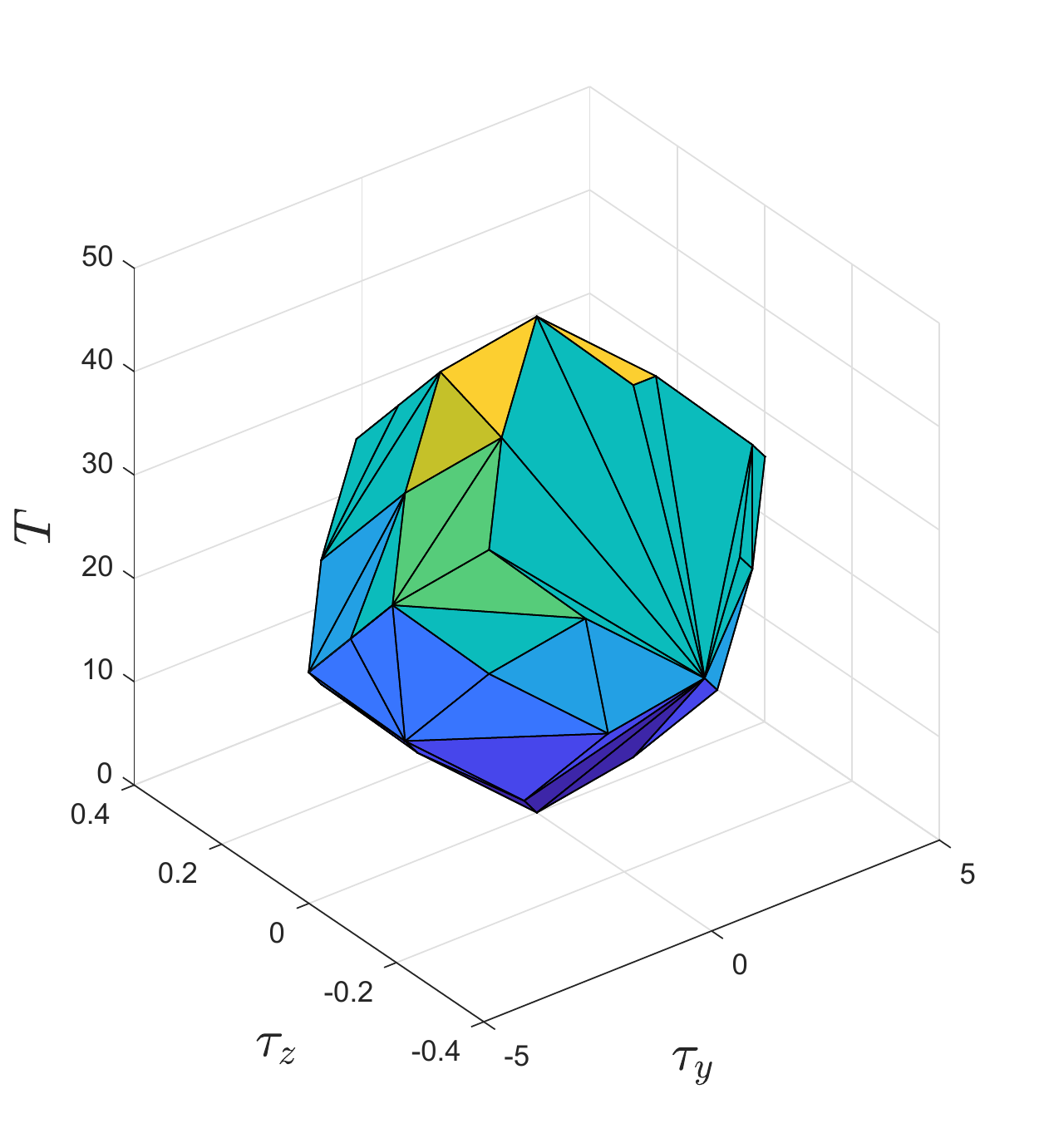}
\par\end{centering}
\caption{{Representation of the four-dimensional admissible region ($T$, $\tau_{x}$, $\tau_{y}$ $\tau_{z}$) in three-dimensional space ($T$, $\tau_{y}$, $\tau_{z}$) when $\tau_{x}=0$. 
\label{fig:Oblast_Y_Z}}}
\end{figure}

Assuming that the DC motor velocity is limited between 0 and  $\omega_{max}$ \eqref{omega_ogranicenje} and the mapping is defined by the linear relation $\boldsymbol{u_{ref}} = \boldsymbol{A} \boldsymbol{\Omega}_s$, it means that the set $D_{u}$ represents a polytope in space $\mathbb{R}^4$. If velocities of all DC motors are equal to zero when all components of the control input are zero-valued, we get the first point in control space determined with 4 coordinates ($T$, $\tau_{x}$, $\tau_{y}$ and $\tau_{z}$). If we now set the angular velocity of the first DC motor to its maximum value, we get the second point in control space. The total number of these combinations is $2^{2n}$, where $n$ is the number of pairs DC motors. For the octocopter example, it is possible to construct one hyper-plane for each tuple (4 control components)  of 256 points in total. However, only those hyper-planes that form an outer region are relevant to define the admissible set. In this way, one can construct a convex polytope-like admissible region in four-dimensional space. These admissible sets can be used to impose additional constraints during the planning stage in order to generate only those referent trajectories which the octocopter will be capable of tracking. Since the obtained region is constructed in four-dimensional control space, we only illustrate three simplified cases for which $\tau_{x}=0$, $\tau_{y}=0$ and $\tau_{z}=0$. An orthogonal projection of the polytope of the set $D_{u}$ (with coordinates ($T$,
$\tau_{x}$, $\tau_{y}$), ($T$, $\tau_{x}$, $\tau_{z}$) i ($T$,
$\tau_{y}$, $\tau_{z}$)) is shown in Figs. \ref{fig:Oblast_X_Y}, \ref{fig:Oblast_X_Z} and \ref{fig:Oblast_Y_Z}, respectively.  

\begin{figure}[p]
\begin{centering}
\includegraphics[width=7cm,height=7cm]{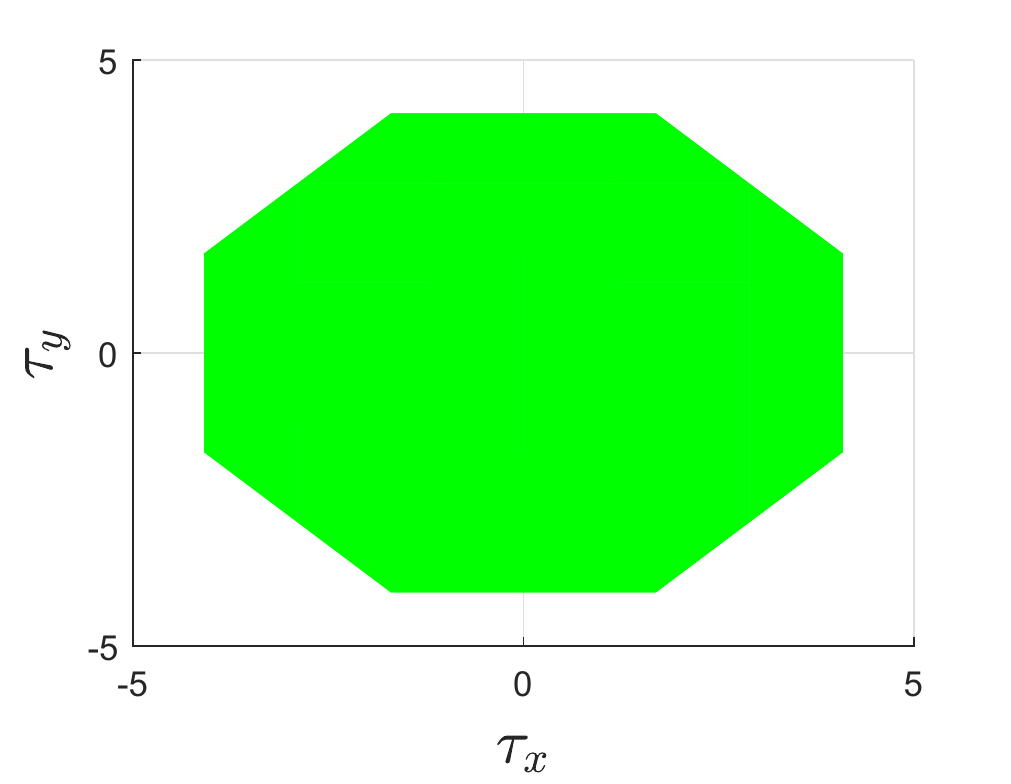} 
\par\end{centering}
\caption{{The projection of torques $\tau_{x}$, $\tau_{y}$ onto the plane $T=mg$ along its orthogonal direction for a PNPNPNPN octocopter configuration structure without any fault states.\label{fig:projekcija_ispravno}}}
\end{figure}

\begin{figure}[p]
\begin{centering}
\includegraphics[width=7cm,height=7cm]{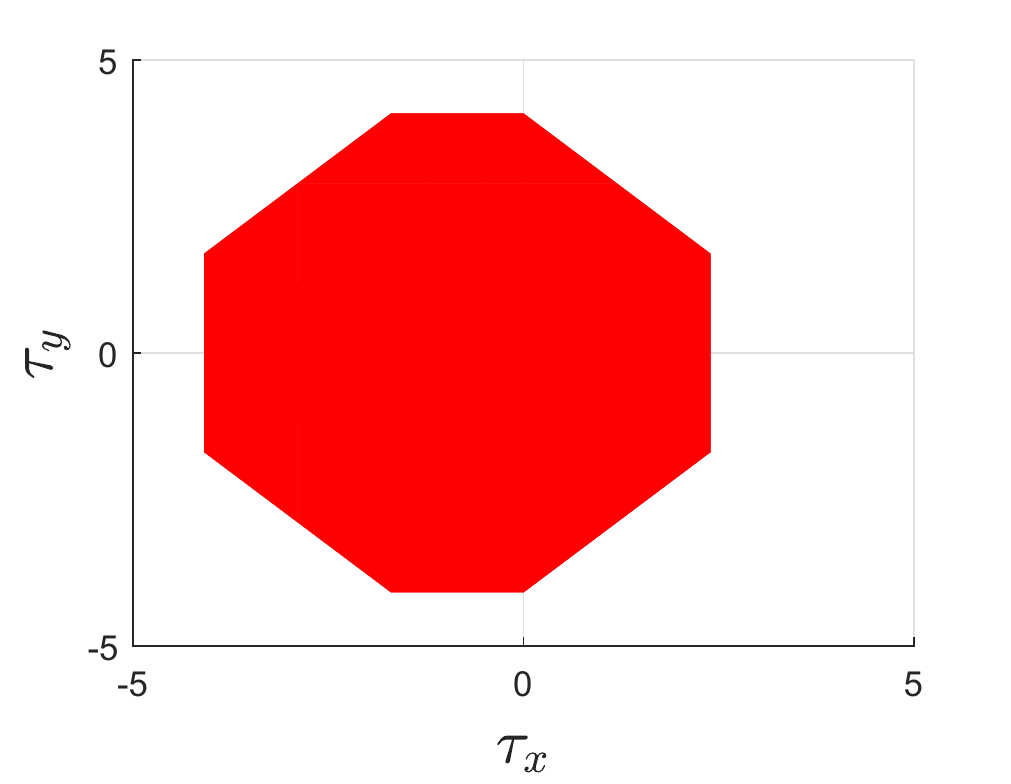} 
\par\end{centering}
\caption{{The projection of torques $\tau_{x}$ and $\tau_{y}$ onto the plane $T=mg$ along its orthogonal direction for a PNPNPNPN octocopter configuration structure with a fault state related to the DC motor $M_1$.\label{fig:projekcija_jednostruki}}}

\end{figure}

\begin{figure}[p]
\begin{centering}
\includegraphics[width=7cm,height=7cm]{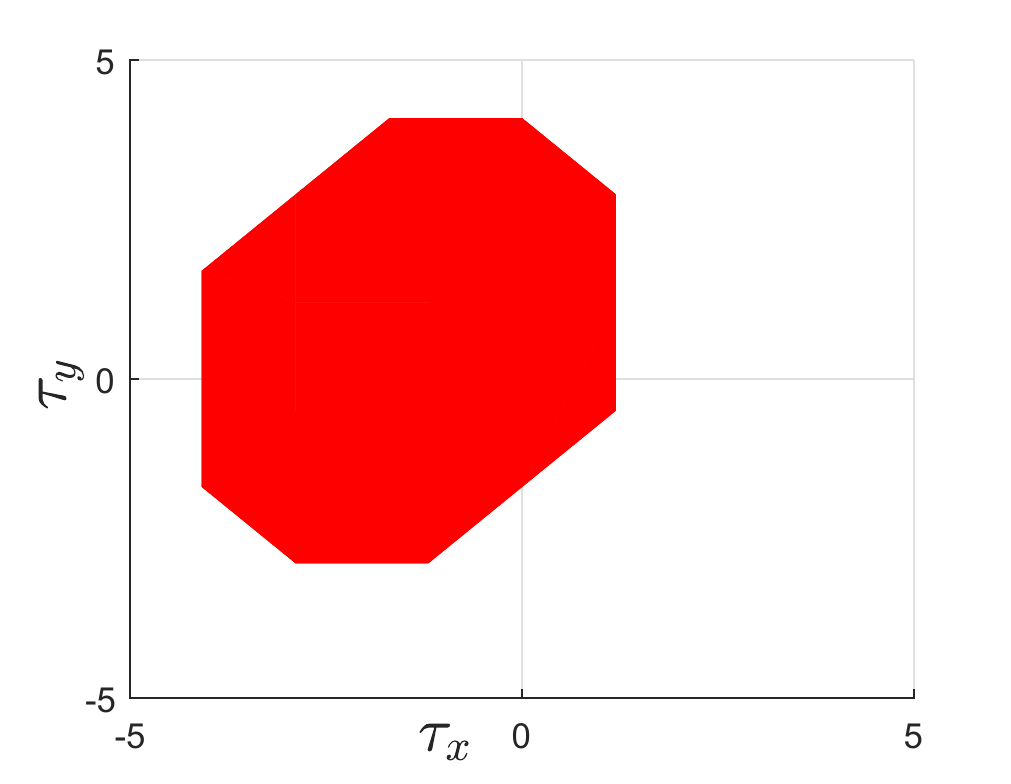}\caption{{The projection of torques $\tau_{x}$ and $\tau_{y}$ onto the plane $T=mg$ along its orthogonal direction, for a PNPNPNPN octocopter configuration structure with a double fault state related to the DC motors $M_1$ and $M_2$.\label{fig:projekcija_dvostruki}}}
\par\end{centering}
\end{figure}

In order to make an octocopter system stable at an arbitrarily hovering point, it is obvious that the thrust force should compensate the gravitational force, that is $T=mg$. If the represented three-dimensional set from Fig.  \ref{fig:Oblast_X_Y} is projected onto the plane $T=mg$, then the projection is shown in Fig. \ref{fig:projekcija_ispravno}. It can be observed that the torques $\tau_{x}$ and $\tau_{y}$ have symmetric values and that they are mutually constrained, meaning that is  not possible to simultaneously reach maximal values of the torques $\tau_{x}$ and $\tau_{y}$. Consider now the DC motor $M_1$ is in a fault state. The projection of the torques $\tau_ {x}$ and $\tau_ {y}$ onto the plane $T=mg$ is shown in Fig. \ref{fig:projekcija_jednostruki}. In case of a double fault (fault states of DC motors $M_1$ and $M_2$), the projection of the torques onto the plane $T=mg$ is shown in Fig. \ref{fig:projekcija_dvostruki}.

It can be seen from Fig. \ref{fig:projekcija_jednostruki} and \ref{fig:projekcija_dvostruki} that the admissible set for $\tau_ {x}$ and $\tau_ {y}$ is reduced with respect to a healthy octocopter system shown in Fig. \ref{fig:projekcija_ispravno}. Depending on the type and combination of faults occurred, some of the planned maneuvers for stabilizing a hovering state will not be possible. The final control admissible set depends on the failure mode occurred and ultimatelly influence the maneavarebility of the octocopter system. For different combinations of failure modes, the system will not be capable to achieve and stay at different hovering states. The obtained admissible sets have been illustrated only to understand that each DC motor has a different effect on the generation of thrust $T$ and torques $\tau_{x}$, $\tau_{y}$ and $\tau_{z}$. 

\color{black}
\section{Fault-dependent controllability test procedure}
To understand whether an octocopter system is controllable (or at least stabilizabile) in case of a single fault (or any multiple-faults combination), we introduce a testing procedure to check whether a hovering state is reachable or not. We say that an octocopter system is controllable with respect to a certain state in case there is a control input that moves the octocopter to that state. In case the controller is not capable to influence yaw-torque $\tau_z$ only, we say the system is stabilizable in that state. The latter means that the vehicle is capable to remain at the given position only by rotating around $z-axes$. 

The task of the control allocation algorithm is to distribute DC motor velocities $\boldsymbol{\Omega}_s$ to each motor in order to achieve the referent thrust force and torques for reaching a waypoint generated by a motion planner. For all MAVs for which $n>2$, there is an infinite number of realization to achieve the same result in case a feasible solution exists. To check whether a feasible solution exists, we define the optimization problem to generate the optimal solution $\boldsymbol{\Omega}^{*}_s$ that minimizes the square-error between the reference $u_{ref}=\boldsymbol{A}\boldsymbol{\Omega}_{s}$ and the achieved control over the feasible control region ($0\leq\boldsymbol{\Omega}_s\leq\omega^{2}_{max}$)
\begin{equation}
\boldsymbol{\Omega}_{s}^{*}=\underset{0\leq\boldsymbol{\Omega}_{s}\leq\omega_{max}^{2}}{argmin}(\Vert\boldsymbol{e_{p}}\Vert^{2})=\underset{0\leq\boldsymbol{\Omega}_{s}\leq\omega_{max}^{2}}{argmin}(\Vert{\boldsymbol{u}_{ref}-\boldsymbol{A}\boldsymbol{\Omega}_{s}}\Vert^{2}),\label{eq:norma}
\end{equation}
where the hovering reference control is defined as  $u_{ref}=[\begin{array}{cccc}
mg & 0 & 0 & 0\end{array}]^{T}$. 

For the optimization problem (\ref{eq:norma}), one can obtain three different cases. 
\begin{enumerate}
\item In case there is no feasible solution, i.e. the octocopter is not capable of achieving the desired hovering state, the system is not controllable with respect to the control admissible set. 
\hspace*{\fill}
\item In case there is a feasible solution yilding zero-valued $e_{p}=[\begin{array}{cccc}
0 & 0 & 0 & 0\end{array}]^{T}$, then the hovering state is achievable and the octocopter is controllable with respect to the control admissible set. 
\hspace*{\fill}
\item In case there is a feasible solution for which there is at least one non-zero error component, we have two additional cases. 
\begin{enumerate}
\item If we allow free movements around $z$-axis, we exclude the torque $\theta_z$, that is  $u_{ref}^-=[\begin{array}{ccc}
T & \boldsymbol{\tau_{x}} & \boldsymbol{\tau_{y}}\end{array}]^{T}=[\begin{array}{ccc}
mg & 0 & 0\end{array}]^{T}$  and repeat the optimization. In case we obtain zero-valued error $e_{p}^-=[\begin{array}{ccc}
0 & 0 & 0\end{array}]^{T}$, then the octocopter is capable to achieve and stay at the hovering state, although it will be rotating around $z$-axis. In this case, we say the octocopter is stabilizabile with respect to the control admissible set. 
\item In case the error $e_{p}^-$ has at least one non-zero component, the octocopter is not controllable with respect to the control admissble set. However, in this case the non-zero error indicates the closest state to the desired hovering state from which one can understand the resulted behaviour of the octocopter system. For instance, we can conclude whether the system will increase ($T>mg$) or decrease ($T<mg$) its height with respect to the desired hovering state or the octocopter system will rotate around an axis in one or the other direction.

\end{enumerate}
\end{enumerate}

By checking whether the hovering point is reachable or not, we can understand if the octocopter system is capable for given mission regardless of the faults. In the following subsection, we show how to thoroughly analyze controllability of different types of MAVs by examining different single and multiple faults including th octocopter system. For clarity, we include the related procedures for the quadcopter and hexacopter systems as well. 

\section{Fault-dependent controllability analysis for a quadcopter system}

Consider a quadcopter without any fault states (a healthy quadcopter system). The quadcopter is designed based on the following parameters \cite{osmic_modelica}:  $m_o=1.32$ $[kg]$, $l=0.211$ $[m]$, $I_{xx}=I_{yy}=0.0128$ $[kg m^2]$, $I_{zz}=0.0239$ $[kg m^2]$, $I_{zzm}=4.3 \cdot 10^{-5}$ $[kg m^2]$, $b=9.9865 \cdot 10^{-6}$   $[\frac{N s^2}{rad^2}]$, $d=1.5978 \cdot 10^{-7}$ $[\frac{N m s^2}{rad^2}]$, $\omega_{max}=840$ $[rad/s]$. 
 
 Solving the optimization problem \eqref{eq:norma}, in which the referent thrust force and torques are given as $\boldsymbol{u_{ref}}=\begin{bmatrix} T & \boldsymbol{\tau}\end{bmatrix}^T=\begin{bmatrix} mg & 0 & 0 & 0 \end{bmatrix}^T$, we obtain $\omega_{1}=\omega_{2}=\omega_{3}=\omega_{4}=569.35$ $[rad/s]$ and $e_{p}=[\begin{array}{cccc} 0 & 0 & 0 & 0\end{array}]^{T}$. This means that the hovering point is reached and it is possible to stabilize the MAV at this point.  Let now the same optimization problem be considered for the quadcopter case for each possible single fault with the same reference $\boldsymbol{u_{ref}}=\begin{bmatrix} T & \boldsymbol{\tau}\end{bmatrix}^T=\begin{bmatrix} mg & 0 & 0 & 0 \end{bmatrix}^T$. The results of these optimizations are shown in Table \ref{tab:Quad}. As it can be seen, the quadcopter cannot be fully controlled at the hovering point for every single failure. Such cases are indicated by red color in Table \ref{tab:Quad}. For example, in case of a failure occurred in the DC motor $M_1$, we obtain $e_{p}=\text{[\ensuremath{\begin{array}{cccc} 0  &  -0.03  &  0  &  0.21\end{array}]^{T}}}$. For this case the quadcopter is stabilizable, but there is a constant rotation in the negative direction about the $x$-axis as well as intense rotation about the $z$-axis. 

\begin{table}[p]
\caption{Quadcopter: Analysis of single-fault cases.\label{tab:Quad} }

\begin{center} \begin{tabular}{|c|c|} \hline Fault                &                                                                                                              $\boldsymbol{e_p} ={\boldsymbol{u_{ref}} - \boldsymbol{A} \boldsymbol{\Omega}_s}$ \\ \hline                               &                                                                                                                               \\ \hline Motor $M_1$                       &  \cellcolor[HTML]{FF8080} $\begin{bmatrix}\ 0.0 & -0.03 & 0.0 & 0.21\ \end{bmatrix}^T$ \\ \hline Motor $M_2$                       &  \cellcolor[HTML]{FF8080}$\begin{bmatrix}\ 0.0 & 0.0 & 0.03 & -0.21\ \end{bmatrix}^T$ \\ \hline Motor $M_3$                       &  \cellcolor[HTML]{FF8080}$\begin{bmatrix}\ 0.0 & 0.03 &  0.0 & & 0.21\ \end{bmatrix}^T$ \\ \hline Motor $M_4$                       &  \cellcolor[HTML]{FF8080}$\begin{bmatrix}\ 0.0 & 0.0 & -0.03 & -0.21\ \end{bmatrix}^T$ \\ \hline \end{tabular} \end{center} 
\end{table}

%Predočimo dopušteni skup upravljačkih varijabli kvadkoptera
%za nastanak kvarnog stanja na motoru 1. Kako se može uočiti s %prikazanih
%slika \ref{fig:Quad_jednostruki_X_Y} i %\ref{fig:Quad_jednostruki_X_Z},
%moment oko $X$-osi može imati vrijednosti u opsegu (-1.5, 0), dok
%moment oko $Z$-osi može imati vrijednosti u opsegu (-0.12, 0.22).
%Oko $X$-osi uvijek je prisutan moment u negativnom smjeru vrtnje
%te se za minimalnu vrijednost momenta oko $X$-osi proračunava %optimalna
%vrijednost za moment oko $Z$-osi. Dobiveni su rezultati očekivani
%i u skladu su sa stanjem u području.
The obtained results are expected since they are in line with the state-of-the-art work. In \cite{Lunze_I}, the authors have shown that the quadcopter does not have a redundant configuration and its controllability will be lost in case any of the DC motors fails.
These results indicate that a quadcopter structure designed with only four motors cannot be reliable system in cases when there is a high probability of any single motor failure. 
\begin{figure}[p]
\centering{}\includegraphics[width=7cm,height=7cm]{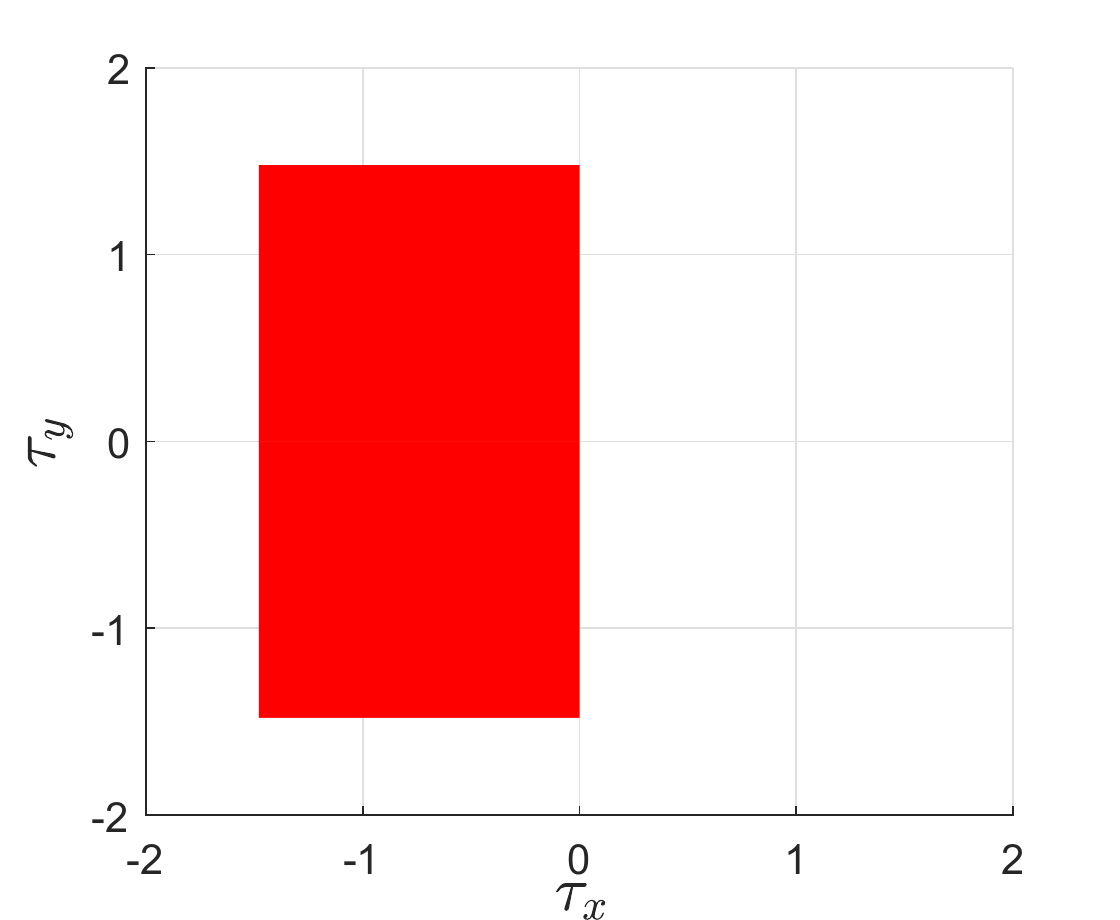}
\caption{{The projection of torques $\tau_{x}$ and $\tau_{y}$ onto the plane $T=mg$ along its orthogonal direction for a PNPNPNPN octocopter configuration structure with a fault state related to the DC motor $M_1$. 
\label{fig:Quad_jednostruki_X_Y}}}
\end{figure}

\begin{figure}[p]
\centering{}\includegraphics[width=7cm,height=7cm]{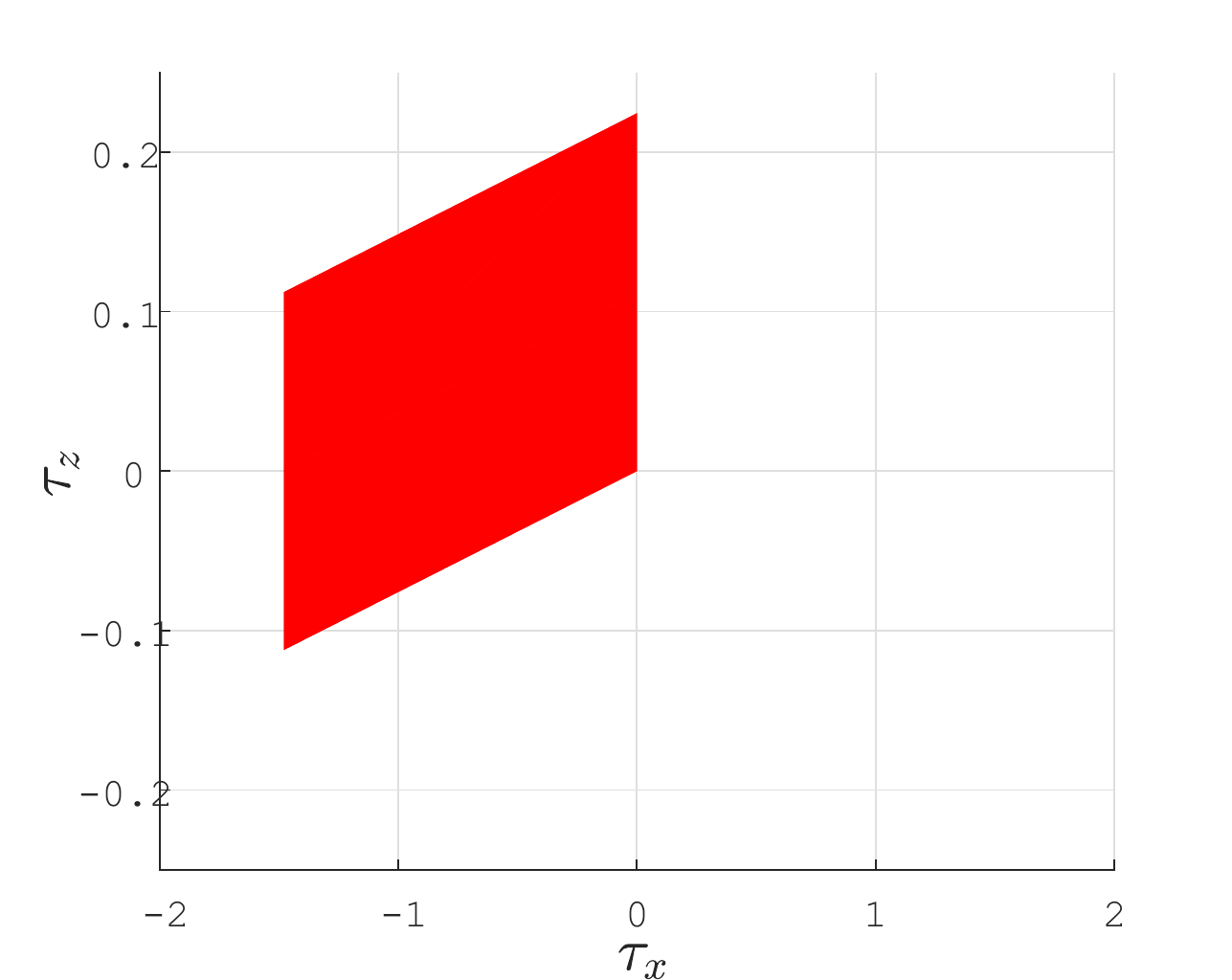}
\caption{{The projection of torques $\tau_{x}$ and $\tau_{z}$ onto the plane $T=mg$ along its orthogonal direction a PNPNPNPN octocopter configuration structure with a fault state related to the DC motor $M_1$.
\label{fig:Quad_jednostruki_X_Z}}}
\end{figure}

\subsection{Fault-dependent controllability analysis for a hexacopter system}
In this subsection, we analyze two types of hexacopter design, including the PNPNPN and the PPNNPN rotation configuration. The hexacopter is designed based on the following parameters: $m_o=1.54$ $[kg]$, $l=0.211$ $[m]$, $I_{xx}=I_{yy}=0.0168$ $[kg m^2]$, $I_{zz}=0.0308$ $[kg m^2]$, $I_{zzm}=2 \cdot 10^{-5}$ $[kg m^2]$, $b=8.5485 \cdot 10^{-6}$   $[\frac{N s^2}{rad^2}]$, $d=1.3678 \cdot 10^{-7}$ $[\frac{N m s^2}{rad^2}]$, $\omega_{max}=874$ $[rad/s]$.

First, we consider controllability of both configurations without fault states by using the same control reference $\boldsymbol{u_{ref}}=\begin{bmatrix} T & \boldsymbol{\tau}\end{bmatrix}^T=\begin{bmatrix} mg & 0 & 0 & 0 \end{bmatrix}^T$. As it can be seen from Table \ref{tab:Heksa_jednostruki}, both versions of the hexacopter are inherently fault-tolerant with respect to a single DC motor failure. For this reason, we consider different double-fault cases (left columns in Tables \ref{tab:Heksa_dvostruki_PNPNPN}
and \ref{tab:Heksa_dvostruki_PPNNPN}) and the cases where the hovering point is stabilizable but not controllable when the reference is used in the form $\boldsymbol{u_{ref}}=\begin{bmatrix} T & \boldsymbol{\tau_{x}} & \boldsymbol{\tau_{y}} \end{bmatrix}^T= \begin{bmatrix} mg & 0 & 0 \end{bmatrix}^T$ (right columns in Tables \ref{tab:Heksa_dvostruki_PNPNPN}
and \ref{tab:Heksa_dvostruki_PPNNPN}). Although the hexacopter may loose controllability, the latter case is important to be examined since a safety landing can be performed which can protect the vehicle and its equipment from potential damage.

\begin{table}[p]
\caption{Hexacopter: Analysis of single-fault cases for the PNPNPN and PPNNPN configurations.\label{tab:Heksa_jednostruki} }

\begin{center} \begin{tabular}{|c|c|c|} \hline Fault   & $e_p$ for the PNPNPN                                                        & $e_p$ for the PPNNPN   \\ \hline         &                                                              &                                                                                      \\ \hline $M_1$ & \cellcolor[HTML]{67FD9A}$\begin{bmatrix}\ 0.0 & 0.0 & 0.0 & 0.0\ \end{bmatrix}^T$ & \cellcolor[HTML]{67FD9A}$\begin{bmatrix}\ 0.0 & 0.0 & 0.0 & 0.0\ \end{bmatrix}^T$ \\ \hline $M_2$ & \cellcolor[HTML]{67FD9A}$\begin{bmatrix}\ 0.0 & 0.0 & 0.0 & 0.0\ \end{bmatrix}^T$ & \cellcolor[HTML]{67FD9A}$\begin{bmatrix}\ 0.0 & 0.0 & 0.0 & 0.0\ \end{bmatrix}^T$                         \\ \hline $M_3$ & \cellcolor[HTML]{67FD9A}$\begin{bmatrix}\ 0.0 & 0.0 & 0.0 & 0.0\ \end{bmatrix}^T$ & \cellcolor[HTML]{67FD9A}$\begin{bmatrix}\ 0.0 & 0.0 & 0.0 & 0.0\ \end{bmatrix}^T$                         \\ \hline $M_4$ & \cellcolor[HTML]{67FD9A}$\begin{bmatrix}\ 0.0 & 0.0 & 0.0 & 0.0\ \end{bmatrix}^T$ & \cellcolor[HTML]{67FD9A}$\begin{bmatrix}\ 0.0 & 0.0 & 0.0 & 0.0\ \end{bmatrix}^T$                         \\ \hline $M_5$ & \cellcolor[HTML]{67FD9A}$\begin{bmatrix}\ 0.0 & 0.0 & 0.0 & 0.0\ \end{bmatrix}^T$ & \cellcolor[HTML]{67FD9A}$\begin{bmatrix}\ 0.0 & 0.0 & 0.0 & 0.0\ \end{bmatrix}^T$                         \\ \hline $M_6$ & \cellcolor[HTML]{67FD9A}$\begin{bmatrix}\ 0.0 & 0.0 & 0.0 & 0.0\ \end{bmatrix}^T$ & \cellcolor[HTML]{67FD9A}$\begin{bmatrix}\ 0.0 & 0.0 & 0.0 & 0.0\ \end{bmatrix}^T$                         \\ \hline \end{tabular} \end{center} 
\end{table}

From the results presented in Tables  \ref{tab:Heksa_dvostruki_PNPNPN}
and \ref{tab:Heksa_dvostruki_PPNNPN} related to the two different hexacopter orientation configurations, $80\%$ of total double-fault cases lead to the loss of controllability (red color in the left columns), while $40\%$ cases are unstable without possibility for a safe landing (red colors in the right columns). This further means that the hexacopter will have a potential to continue the mission only in three cases (green color in the left columns) and to be safe in $60\%$ (green color in right columns). One can also conclude that the hexacopter is single-fault-tolerant, while it is quite sensitive to double faults in terms of mission execution. However, it possesses a certain level of safety robustness. Since the obtained results are similar, there is no advantage of using any particular hexacopter configuration over the other one.

\begin{table}[p]
\caption{Hexacopter: Analysis of double-fault cases for the
PNPNPN configuration.\label{tab:Heksa_dvostruki_PNPNPN} }

\begin{center}  \begin{tabular}{|c|c|c|} \hline Fault    & $\boldsymbol{e_p}=\begin{bmatrix} 0 & 0 & 0 & 0 \end{bmatrix}^T$                               & $\boldsymbol{e_p}=\begin{bmatrix} 0 & 0 & 0  \end{bmatrix}^T$ \\ \hline         &                                                              &                                                                                      \\ \hline $M_{12}$ & \cellcolor[HTML]{FF8080}$\begin{bmatrix}\ 0.0 & -0.44 & 0.25 & 0.0\ \end{bmatrix}^T$ & \cellcolor[HTML]{FF8080}$\begin{bmatrix}\ 0.0 & -0.44  & 0.25\ \end{bmatrix}^T$ \\ \hline $M_{14}$ & \cellcolor[HTML]{FF8080}$\begin{bmatrix}\ 0.0 & 0.0 & 0.03 & 0.13\ \end{bmatrix}^T$ & \cellcolor[HTML]{67FD9A}$\begin{bmatrix}\ 0.0 & 0.0  & 0.0\ \end{bmatrix}^T$                         \\ \hline $M_{14}$ & \cellcolor[HTML]{67FD9A}$\begin{bmatrix}\ 0.0 & 0.0 & 0.0 & 0.0\ \end{bmatrix}^T$ & \cellcolor[HTML]{67FD9A}$\begin{bmatrix}\ 0.0 & 0.0  & 0.0\ \end{bmatrix}^T$                         \\ \hline $M_{15}$ & \cellcolor[HTML]{FF8080}$\begin{bmatrix}\ 0.0 & -0.2 & 0.0 & 0.12\ \end{bmatrix}^T$ & \cellcolor[HTML]{67FD9A}$\begin{bmatrix}\ 0.0 & 0.0  & 0.0\ \end{bmatrix}^T$                         \\ \hline $M_{16}$ & \cellcolor[HTML]{FF8080}$\begin{bmatrix}\ 0.0 & -0.6 & -0.3 & 0.0\ \end{bmatrix}^T$ & \cellcolor[HTML]{FF8080}$\begin{bmatrix}\ 0.0 & -0.44  & -0.25\ \end{bmatrix}^T$                         \\ \hline $M_{23}$ & \cellcolor[HTML]{FF8080}$\begin{bmatrix}\ 0.0 & 0.0 & 0.51 & 0.0\ \end{bmatrix}^T$ & \cellcolor[HTML]{FF8080}$\begin{bmatrix}\ 0.0 & 0.5  & 0.0\ \end{bmatrix}^T$                         \\ \hline $M_{24}$ & \cellcolor[HTML]{FF8080}$\begin{bmatrix}\ 0.0 & 0.24 & 0.42 & -0.13\ \end{bmatrix}^T$ & \cellcolor[HTML]{67FD9A}$\begin{bmatrix}\ 0.0 & 0.0  & 0.0\ \end{bmatrix}^T$                         \\ \hline $M_{25}$ & \cellcolor[HTML]{67FD9A}$\begin{bmatrix}\ 0.0 & 0.0 & 0.0 & 0.0\ \end{bmatrix}^T$ & \cellcolor[HTML]{67FD9A}$\begin{bmatrix}\ 0.0 & 0.0  & 0.0\ \end{bmatrix}^T$                         \\ \hline $M_{26}$ & \cellcolor[HTML]{FF8080}$\begin{bmatrix}\ 0.0 & -0.04 & 0.0 & -0.12\ \end{bmatrix}^T$ & \cellcolor[HTML]{67FD9A}$\begin{bmatrix}\ 0.0 & 0.0  & 0.0\ \end{bmatrix}^T$                         \\ \hline $M_{34}$ & \cellcolor[HTML]{FF8080}$\begin{bmatrix}\ 0.0 & 0.64 & 0.36 & 0.0\ \end{bmatrix}^T$ & \cellcolor[HTML]{FF8080}$\begin{bmatrix}\ 0.0 & 0.44  & 0.25\ \end{bmatrix}^T$                         \\ \hline $M_{35}$ & \cellcolor[HTML]{FF8080}$\begin{bmatrix}\ 0.0 & 0.04 & 0.0 & 0.12\ \end{bmatrix}^T$ & \cellcolor[HTML]{67FD9A}$\begin{bmatrix}\ 0.0 & 0.0  & 0.0\ \end{bmatrix}^T$                         \\ \hline $M_{36}$ & \cellcolor[HTML]{67FD9A}$\begin{bmatrix}\ 0.0 & 0.0 & 0.0 & 0.0\ \end{bmatrix}^T$ & \cellcolor[HTML]{67FD9A}$\begin{bmatrix}\ 0.0 & 0.0  & 0.0\ \end{bmatrix}^T$                         \\ \hline $M_{45}$ & \cellcolor[HTML]{FF8080}$\begin{bmatrix}\ 0.0 & 0.04 & -0.25 & 0.12\ \end{bmatrix}^T$ & \cellcolor[HTML]{FF8080}$\begin{bmatrix}\ 0.0 & 0.61  & -0.35\ \end{bmatrix}^T$                         \\ \hline $M_{46}$ & \cellcolor[HTML]{FF8080}$\begin{bmatrix}\ 0.0 & 0.02 & -0.03 & -0.12\ \end{bmatrix}^T$ & \cellcolor[HTML]{67FD9A}$\begin{bmatrix}\ 0.0 & 0.0  & 0.0\ \end{bmatrix}^T$                         \\ \hline $M_{56}$ & \cellcolor[HTML]{FF8080}$\begin{bmatrix}\ 0.0 & 0.0 & -0.5 & 0.0\ \end{bmatrix}^T$ & \cellcolor[HTML]{FF8080}$\begin{bmatrix}\ 0.0 & 0.0  & -0.51\ \end{bmatrix}^T$                         \\ \hline \end{tabular}  \end{center}
\end{table}

\begin{table}[p]
\caption{Hexacopter: Analysis of double-fault cases for the
PPNNPN configuration.\label{tab:Heksa_dvostruki_PPNNPN} }

\begin{center}  \begin{tabular}{|c|c|c|} \hline Fault    & $\boldsymbol{e_p}=\begin{bmatrix} 0 & 0 & 0 & 0 \end{bmatrix}^T$                               & $\boldsymbol{e_p}=\begin{bmatrix} 0 & 0 & 0  \end{bmatrix}^T$ \\ \hline         &                                                              &                                                                                      \\ \hline $M_{12}$ & \cellcolor[HTML]{FF8080}$\begin{bmatrix}\ 0.0 & -0.43 & 0.27 & 0.18\ \end{bmatrix}^T$ & \cellcolor[HTML]{FF8080}$\begin{bmatrix}\ 0.0 & -0.44  & 0.25\ \end{bmatrix}^T$ \\ \hline $M_{13}$ & \cellcolor[HTML]{67FD9A}$\begin{bmatrix}\ 0.0 & 0.0 & 0.0 & 0.0\ \end{bmatrix}^T$ & \cellcolor[HTML]{67FD9A}$\begin{bmatrix}\ 0.0 & 0.0  & 0.0\ \end{bmatrix}^T$                         \\ \hline $M_{14}$ & \cellcolor[HTML]{67FD9A}$\begin{bmatrix}\ 0.0 & 0.0 & 0.0 & 0.0\ \end{bmatrix}^T$ & \cellcolor[HTML]{67FD9A}$\begin{bmatrix}\ 0.0 & 0.0  & 0.0\ \end{bmatrix}^T$                         \\ \hline $M_{15}$ & \cellcolor[HTML]{FF8080}$\begin{bmatrix}\ 0.0 & 0.0 & -0.01 & 0.07\ \end{bmatrix}^T$ & \cellcolor[HTML]{67FD9A}$\begin{bmatrix}\ 0.0 & 0.0  & 0.0\ \end{bmatrix}^T$                         \\ \hline $M_{16}$ & \cellcolor[HTML]{FF8080}$\begin{bmatrix}\ 0.0 & -0.44 & -0.25 & -0.15\ \end{bmatrix}^T$ & \cellcolor[HTML]{FF8080}$\begin{bmatrix}\ 0.0 & -0.44  & -0.25\ \end{bmatrix}^T$                         \\ \hline $M_{23}$ & \cellcolor[HTML]{FF8080}$\begin{bmatrix}\ 0.0 & 0.0 & 0.51 & 0.0\ \end{bmatrix}^T$ & \cellcolor[HTML]{FF8080}$\begin{bmatrix}\ 0.0 & 0.0  & 0.5\ \end{bmatrix}^T$                         \\ \hline $M_{24}$ & \cellcolor[HTML]{67FD9A}$\begin{bmatrix}\ 0.0 & 0.0 & 0.0 & 0.0\ \end{bmatrix}^T$ & \cellcolor[HTML]{67FD9A}$\begin{bmatrix}\ 0.0 & 0.0  & 0.0\ \end{bmatrix}^T$                         \\ \hline $M_{25}$ & \cellcolor[HTML]{FF8080}$\begin{bmatrix}\ 0.0 & 0.0 & 0.0 & 0.04\ \end{bmatrix}^T$ & \cellcolor[HTML]{67FD9A}$\begin{bmatrix}\ 0.0 & 0.0  & 0.0\ \end{bmatrix}^T$                         \\ \hline $M_{26}$ & \cellcolor[HTML]{FF8080}$\begin{bmatrix}\ 0.0 & 0.0 & 0.0 & -0.05\ \end{bmatrix}^T$ & \cellcolor[HTML]{67FD9A}$\begin{bmatrix}\ 0.0 & 0.0  & 0.0\ \end{bmatrix}^T$                         \\ \hline $M_{34}$ & \cellcolor[HTML]{FF8080}$\begin{bmatrix}\ 0.0 & 0.43 & 0.27 & -0.02\ \end{bmatrix}^T$ & \cellcolor[HTML]{FF8080}$\begin{bmatrix}\ 0.0 & 0.44  & 0.25\ \end{bmatrix}^T$                         \\ \hline $M_{35}$ & \cellcolor[HTML]{FF8080}$\begin{bmatrix}\ 0.0 & 0.0 & 0.0 & 0.05\ \end{bmatrix}^T$ & \cellcolor[HTML]{67FD9A}$\begin{bmatrix}\ 0.0 & 0.0  & 0.0\ \end{bmatrix}^T$                         \\ \hline $M_{36}$ & \cellcolor[HTML]{FF8080}$\begin{bmatrix}\ 0.0 & 0.0 & 0.0 & -0.05\ \end{bmatrix}^T$ & \cellcolor[HTML]{67FD9A}$\begin{bmatrix}\ 0.0 & 0.0  & 0.0\ \end{bmatrix}^T$                         \\ \hline $M_{45}$ & \cellcolor[HTML]{FF8080}$\begin{bmatrix}\ 0.0 & 0.44 & -0.25 & 0.14\ \end{bmatrix}^T$ & \cellcolor[HTML]{FF8080}$\begin{bmatrix}\ 0.0 & 0.44  & -0.25\ \end{bmatrix}^T$                         \\ \hline $M_{46}$ & \cellcolor[HTML]{FF8080}$\begin{bmatrix}\ 0.0 & 0.0 & 0.0 & -0.08\ \end{bmatrix}^T$ & \cellcolor[HTML]{67FD9A}$\begin{bmatrix}\ 0.0 & 0.0  & 0.0\ \end{bmatrix}^T$                         \\ \hline $M_{56}$ & \cellcolor[HTML]{FF8080}$\begin{bmatrix}\ 0.0 & -0.51 & 0.0 & 0.0\ \end{bmatrix}^T$ & \cellcolor[HTML]{FF8080}$\begin{bmatrix}\ 0.0 & 0.0  & -0.51\ \end{bmatrix}^T$                         \\ \hline \end{tabular} \end{center}
\end{table}

\subsection{Fault-dependent controllability analysis for an octocopter system}
\color{black}
In this subsection we address two different octocopter configuration structures, the PNPNPNPN and the PPNNPPNN. The octocopters are designed based on the following parameters \cite{osmic_SMC} $m_o=1.8$ $[kg]$, $l=0.211$ $[m]$, $I_{xx}=I_{yy}=0.0429$ $[kg m^2]$, $I_{zz}=0.0748$ $[kg m^2]$, $I_{zzm}=2 \cdot 10^{-5}$ $[kg m^2]$, $b=8.5485 \cdot 10^{-6}$   $[\frac{N s^2}{rad^2}]$, $d=1.3678 \cdot 10^{-7}$ $[\frac{N m s^2}{rad^2}]$, $\omega_{max}=874$ $[rad/s]$. 

The analysis of single-fault cases for both considered octocopter configurations is presented in Table \ref{tab:Oktokopter_jednostruki}, while for double-fault cases in Tables  \ref{tab:Oktokopter_dvostruki_PNPNPNPN}
and \ref{tab:Oktokopter_dvostruki_PPNNPPNN}. It is evident that both octocopter configurations are fully insensitive with respect to all single failures in terms of their potential to continue the mission execution. 
From the results presented in Tables \ref{tab:Oktokopter_dvostruki_PNPNPNPN}  related to the PNPNPNPN configuration, $28\%$ of total double-fault cases lead to the loss of controllability (red color in the left column), while there are no unstable cases without possibility for a safe landing. This means that this octopcopter configuration will have a potential to continue the mission in $72\%$ of cases (green color in the left column) and to be safe in $100\%$ (green color in the right column). For the hexacopter PPNNPPNN configuration we have $14\%$ of controllability loss, no unstable cases, $86\%$ potential to continue mission and $100\%$ safety (see Table \ref{tab:Oktokopter_dvostruki_PPNNPPNN}). The obtained conclusions are the same as those obtained in \cite{FTC_ETH_Thomas_Schneider}. 
\begin{table}[p]
\caption{Octocopter: Analysis of single-fault cases for the
PNPNPNPN and PPNNPPNN configurations.\label{tab:Oktokopter_jednostruki} }

\begin{center}{\footnotesize \begin{tabular}{|c|c|c|} \hline Fault    & $\boldsymbol{e_p}$ for the PNPNPNPN                                                        & $e_p$ for the PPNNPPNN   \\ \hline         &                                                              &                                                                                      \\ \hline $M_1$ & \cellcolor[HTML]{67FD9A}$\begin{bmatrix}\ 0.0 & 0.0 & 0.0 & 0.0\ \end{bmatrix}^T$ & \cellcolor[HTML]{67FD9A}$\begin{bmatrix}\ 0.0 & 0.0 & 0.0 & 0.0\ \end{bmatrix}^T$ \\ \hline $M_2$ & \cellcolor[HTML]{67FD9A}$\begin{bmatrix}\ 0.0 & 0.0 & 0.0 & 0.0\ \end{bmatrix}^T$ & \cellcolor[HTML]{67FD9A}$\begin{bmatrix}\ 0.0 & 0.0 & 0.0 & 0.0\ \end{bmatrix}^T$                         \\ \hline $M_3$ & \cellcolor[HTML]{67FD9A}$\begin{bmatrix}\ 0.0 & 0.0 & 0.0 & 0.0\ \end{bmatrix}^T$ & \cellcolor[HTML]{67FD9A}$\begin{bmatrix}\ 0.0 & 0.0 & 0.0 & 0.0\ \end{bmatrix}^T$                         \\ \hline $M_4$ & \cellcolor[HTML]{67FD9A}$\begin{bmatrix}\ 0.0 & 0.0 & 0.0 & 0.0\ \end{bmatrix}^T$ & \cellcolor[HTML]{67FD9A}$\begin{bmatrix}\ 0.0 & 0.0 & 0.0 & 0.0\ \end{bmatrix}^T$                         \\ \hline $M_5$ & \cellcolor[HTML]{67FD9A}$\begin{bmatrix}\ 0.0 & 0.0 & 0.0 & 0.0\ \end{bmatrix}^T$ & \cellcolor[HTML]{67FD9A}$\begin{bmatrix}\ 0.0 & 0.0 & 0.0 & 0.0\ \end{bmatrix}^T$                         \\ \hline $M_6$ & \cellcolor[HTML]{67FD9A}$\begin{bmatrix}\ 0.0 & 0.0 & 0.0 & 0.0\ \end{bmatrix}^T$ & \cellcolor[HTML]{67FD9A}$\begin{bmatrix}\ 0.0 & 0.0 & 0.0 & 0.0\ \end{bmatrix}^T$                         \\ \hline $M_7$ & \cellcolor[HTML]{67FD9A}$\begin{bmatrix}\ 0.0 & 0.0 & 0.0 & 0.0\ \end{bmatrix}^T$ & \cellcolor[HTML]{67FD9A}$\begin{bmatrix}\ 0.0 & 0.0 & 0.0 & 0.0\ \end{bmatrix}^T$                         \\ \hline $M_8$ & \cellcolor[HTML]{67FD9A}$\begin{bmatrix}\ 0.0 & 0.0 & 0.0 & 0.0\ \end{bmatrix}^T$ & \cellcolor[HTML]{67FD9A}$\begin{bmatrix}\ 0.0 & 0.0 & 0.0 & 0.0\ \end{bmatrix}^T$                         \\ \hline \end{tabular}} \end{center}
\end{table}

\begin{table}[p]
\caption{Octocopter: Analysis of double-fault cases for the PNPNPNPN   configuration.\label{tab:Oktokopter_dvostruki_PNPNPNPN} }

\begin{center} {\footnotesize \begin{tabular}{|c|c|c|} \hline Fault    & $\boldsymbol{e_p}=\begin{bmatrix} 0 & 0 & 0 & 0 \end{bmatrix}^T$                               & $\boldsymbol{e_p}=\begin{bmatrix} 0 & 0 & 0  \end{bmatrix}^T$ \\ \hline         &                                                              &                                                                                      \\ \hline $M_{12}$ & \cellcolor[HTML]{67FD9A}$\begin{bmatrix}\ 0.0 & 0.0 & 0.0 & 0.0\ \end{bmatrix}^T$ & \cellcolor[HTML]{67FD9A}$\begin{bmatrix}\ 0.0 & 0.0  & 0.0\ \end{bmatrix}^T$ \\ \hline $M_{13}$ & \cellcolor[HTML]{FF8080}$\begin{bmatrix}\ 0.0 & 0.0 & 0.0 & 0.01\ \end{bmatrix}^T$ & \cellcolor[HTML]{67FD9A}$\begin{bmatrix}\ 0.0 & 0.0  & 0.0\ \end{bmatrix}^T$                         \\ \hline $M_{14}$ & \cellcolor[HTML]{67FD9A}$\begin{bmatrix}\ 0.0 & 0.0 & 0.0 & 0.0\ \end{bmatrix}^T$ & \cellcolor[HTML]{67FD9A}$\begin{bmatrix}\ 0.0 & 0.0  & 0.0\ \end{bmatrix}^T$                         \\ \hline $M_{15}$ & \cellcolor[HTML]{67FD9A}$\begin{bmatrix}\ 0.0 & 0.0 & 0.0 & 0.0\ \end{bmatrix}^T$ & \cellcolor[HTML]{67FD9A}$\begin{bmatrix}\ 0.0 & 0.0  & 0.0\ \end{bmatrix}^T$                         \\ \hline $M_{16}$ & \cellcolor[HTML]{67FD9A}$\begin{bmatrix}\ 0.0 & 0.0 & 0.0 & 0.0\ \end{bmatrix}^T$ & \cellcolor[HTML]{67FD9A}$\begin{bmatrix}\ 0.0 & 0.0  & 0.0\ \end{bmatrix}^T$                         \\ \hline $M_{17}$ & \cellcolor[HTML]{FF8080}$\begin{bmatrix}\ 0.0 & 0.0 & 0.0 & 0.01\ \end{bmatrix}^T$ & \cellcolor[HTML]{67FD9A}$\begin{bmatrix}\ 0.0 & 0.0  & 0.0\ \end{bmatrix}^T$                         \\ \hline $M_{18}$ & \cellcolor[HTML]{67FD9A}$\begin{bmatrix}\ 0.0 & 0.0 & 0.0 & 0.0\ \end{bmatrix}^T$ & \cellcolor[HTML]{67FD9A}$\begin{bmatrix}\ 0.0 & 0.0  & 0.0\ \end{bmatrix}^T$                         \\ \hline $M_{23}$ & \cellcolor[HTML]{67FD9A}$\begin{bmatrix}\ 0.0 & 0.0 & 0.0 & 0.0\ \end{bmatrix}^T$ & \cellcolor[HTML]{67FD9A}$\begin{bmatrix}\ 0.0 & 0.0  & 0.0\ \end{bmatrix}^T$                         \\ \hline $M_{24}$ & \cellcolor[HTML]{FF8080}$\begin{bmatrix}\ 0.0 & 0.0 & 0.0 & -0.01\ \end{bmatrix}^T$ & \cellcolor[HTML]{67FD9A}$\begin{bmatrix}\ 0.0 & 0.0  & 0.0\ \end{bmatrix}^T$                         \\ \hline $M_{25}$ & \cellcolor[HTML]{67FD9A}$\begin{bmatrix}\ 0.0 & 0.0 & 0.0 & 0.0\ \end{bmatrix}^T$ & \cellcolor[HTML]{67FD9A}$\begin{bmatrix}\ 0.0 & 0.0  & 0.0\ \end{bmatrix}^T$                         \\ \hline $M_{26}$ & \cellcolor[HTML]{67FD9A}$\begin{bmatrix}\ 0.0 & 0.0 & 0.0 & 0.0\ \end{bmatrix}^T$ & \cellcolor[HTML]{67FD9A}$\begin{bmatrix}\ 0.0 & 0.0  & 0.0\ \end{bmatrix}^T$                         \\ \hline $M_{27}$ & \cellcolor[HTML]{67FD9A}$\begin{bmatrix}\ 0.0 & 0.0 & 0.0 & 0.0\ \end{bmatrix}^T$ & \cellcolor[HTML]{67FD9A}$\begin{bmatrix}\ 0.0 & 0.0  & 0.0\ \end{bmatrix}^T$                         \\ \hline $M_{28}$ & \cellcolor[HTML]{FF8080}$\begin{bmatrix}\ 0.0 & 0.0 & 0.0 & 0.01\ \end{bmatrix}^T$ & \cellcolor[HTML]{67FD9A}$\begin{bmatrix}\ 0.0 & 0.0  & 0.0\ \end{bmatrix}^T$                         \\ \hline $M_{34}$ & \cellcolor[HTML]{67FD9A}$\begin{bmatrix}\ 0.0 & 0.0 & 0.0 & 0.0\ \end{bmatrix}^T$ & \cellcolor[HTML]{67FD9A}$\begin{bmatrix}\ 0.0 & 0.0  & 0.0\ \end{bmatrix}^T$                         \\ \hline $M_{35}$ & \cellcolor[HTML]{FF8080}$\begin{bmatrix}\ 0.0 & 0.0 & 0.0 & 0.01\ \end{bmatrix}^T$ & \cellcolor[HTML]{67FD9A}$\begin{bmatrix}\ 0.0 & 0.0  & 0.0\ \end{bmatrix}^T$                         \\ \hline $M_{36}$ & \cellcolor[HTML]{67FD9A}$\begin{bmatrix}\ 0.0 & 0.0 & 0.0 & 0.0\ \end{bmatrix}^T$ & \cellcolor[HTML]{67FD9A}$\begin{bmatrix}\ 0.0 & 0.0  & 0.0\ \end{bmatrix}^T$                         \\ \hline $M_{37}$ & \cellcolor[HTML]{67FD9A}$\begin{bmatrix}\ 0.0 & 0.0 & 0.0 & 0.0\ \end{bmatrix}^T$ & \cellcolor[HTML]{67FD9A}$\begin{bmatrix}\ 0.0 & 0.0  & 0.0\ \end{bmatrix}^T$                         \\ \hline $M_{38}$ & \cellcolor[HTML]{67FD9A}$\begin{bmatrix}\ 0.0 & 0.0 & 0.0 & 0.0\ \end{bmatrix}^T$ & \cellcolor[HTML]{67FD9A}$\begin{bmatrix}\ 0.0 & 0.0  & 0.0\ \end{bmatrix}^T$                         \\ \hline $M_{45}$ & \cellcolor[HTML]{67FD9A}$\begin{bmatrix}\ 0.0 & 0.0 & 0.0 & 0.0\ \end{bmatrix}^T$ & \cellcolor[HTML]{67FD9A}$\begin{bmatrix}\ 0.0 & 0.0  & 0.0\ \end{bmatrix}^T$                         \\ \hline $M_{46}$ & \cellcolor[HTML]{FF8080}$\begin{bmatrix}\ 0.0 & 0.0 & 0.0 & -0.01\ \end{bmatrix}^T$ & \cellcolor[HTML]{67FD9A}$\begin{bmatrix}\ 0.0 & 0.0  & 0.0\ \end{bmatrix}^T$                         \\ \hline $M_{47}$ & \cellcolor[HTML]{67FD9A}$\begin{bmatrix}\ 0.0 & 0.0 & 0.0 & 0.0\ \end{bmatrix}^T$ & \cellcolor[HTML]{67FD9A}$\begin{bmatrix}\ 0.0 & 0.0  & 0.0\ \end{bmatrix}^T$                         \\ \hline $M_{48}$ & \cellcolor[HTML]{67FD9A}$\begin{bmatrix}\ 0.0 & 0.0 & 0.0 & 0.0\ \end{bmatrix}^T$ & \cellcolor[HTML]{67FD9A}$\begin{bmatrix}\ 0.0 & 0.0  & 0.0\ \end{bmatrix}^T$                         \\ \hline $M_{56}$ & \cellcolor[HTML]{67FD9A}$\begin{bmatrix}\ 0.0 & 0.0 & 0.0 & 0.0\ \end{bmatrix}^T$ & \cellcolor[HTML]{67FD9A}$\begin{bmatrix}\ 0.0 & 0.0  & 0.0\ \end{bmatrix}^T$                         \\ \hline $M_{57}$ & \cellcolor[HTML]{FF8080}$\begin{bmatrix}\ 0.0 & 0.0 & 0.0 & 0.01\ \end{bmatrix}^T$ & \cellcolor[HTML]{67FD9A}$\begin{bmatrix}\ 0.0 & 0.0  & 0.0\ \end{bmatrix}^T$                         \\ \hline $M_{58}$ & \cellcolor[HTML]{67FD9A}$\begin{bmatrix}\ 0.0 & 0.0 & 0.0 & 0.0\ \end{bmatrix}^T$ & \cellcolor[HTML]{67FD9A}$\begin{bmatrix}\ 0.0 & 0.0  & 0.0\ \end{bmatrix}^T$                         \\ \hline $M_{67}$ & \cellcolor[HTML]{67FD9A}$\begin{bmatrix}\ 0.0 & 0.0 & 0.0 & 0.0\ \end{bmatrix}^T$ & \cellcolor[HTML]{67FD9A}$\begin{bmatrix}\ 0.0 & 0.0  & 0.0\ \end{bmatrix}^T$                         \\ \hline $M_{68}$ & \cellcolor[HTML]{FF8080}$\begin{bmatrix}\ 0.0 & 0.0 & 0.0 & -0.01\ \end{bmatrix}^T$ & \cellcolor[HTML]{67FD9A}$\begin{bmatrix}\ 0.0 & 0.0  & 0.0\ \end{bmatrix}^T$                         \\ \hline $M_{78}$ & \cellcolor[HTML]{67FD9A}$\begin{bmatrix}\ 0.0 & 0.0 & 0.0 & 0.0\ \end{bmatrix}^T$ & \cellcolor[HTML]{67FD9A}$\begin{bmatrix}\ 0.0 & 0.0  & 0.0\ \end{bmatrix}^T$                         \\ \hline \end{tabular} } \end{center}
\end{table}

\begin{table}[p]
\caption{Octocopter: Analysis of double-fault cases for the PPNNPPNN
configuration.\label{tab:Oktokopter_dvostruki_PPNNPPNN} }

\begin{center} {\footnotesize \begin{tabular}{|c|c|c|} \hline Fault    & $\boldsymbol{e_p}=\begin{bmatrix} 0 & 0 & 0 & 0 \end{bmatrix}^T$                               & $\boldsymbol{e_p}=\begin{bmatrix} 0 & 0 & 0  \end{bmatrix}^T$ \\ \hline         &                                                              &                                                                                      \\ \hline $M_{12}$ & \cellcolor[HTML]{FF8080}$\begin{bmatrix}\ 0.0 & 0.0 & 0.0 & 0.13\ \end{bmatrix}^T$ & \cellcolor[HTML]{67FD9A}$\begin{bmatrix}\ 0.0 & 0.0  & 0.0\ \end{bmatrix}^T$ \\ \hline $M_{13}$ & \cellcolor[HTML]{67FD9A}$\begin{bmatrix}\ 0.0 & 0.0 & 0.0 & 0.0\ \end{bmatrix}^T$ & \cellcolor[HTML]{67FD9A}$\begin{bmatrix}\ 0.0 & 0.0  & 0.0\ \end{bmatrix}^T$                         \\ \hline $M_{14}$ & \cellcolor[HTML]{67FD9A}$\begin{bmatrix}\ 0.0 & 0.0 & 0.0 & 0.0\ \end{bmatrix}^T$ & \cellcolor[HTML]{67FD9A}$\begin{bmatrix}\ 0.0 & 0.0  & 0.0\ \end{bmatrix}^T$                         \\ \hline $M_{15}$ & \cellcolor[HTML]{67FD9A}$\begin{bmatrix}\ 0.0 & 0.0 & 0.0 & 0.0\ \end{bmatrix}^T$ & \cellcolor[HTML]{67FD9A}$\begin{bmatrix}\ 0.0 & 0.0  & 0.0\ \end{bmatrix}^T$                         \\ \hline $M_{16}$ & \cellcolor[HTML]{67FD9A}$\begin{bmatrix}\ 0.0 & 0.0 & 0.0 & 0.0\ \end{bmatrix}^T$ & \cellcolor[HTML]{67FD9A}$\begin{bmatrix}\ 0.0 & 0.0  & 0.0\ \end{bmatrix}^T$                         \\ \hline $M_{17}$ & \cellcolor[HTML]{67FD9A}$\begin{bmatrix}\ 0.0 & 0.0 & 0.0 & 0.0\ \end{bmatrix}^T$ & \cellcolor[HTML]{67FD9A}$\begin{bmatrix}\ 0.0 & 0.0  & 0.0\ \end{bmatrix}^T$                         \\ \hline $M_{18}$ & \cellcolor[HTML]{67FD9A}$\begin{bmatrix}\ 0.0 & 0.0 & 0.0 & 0.0\ \end{bmatrix}^T$ & \cellcolor[HTML]{67FD9A}$\begin{bmatrix}\ 0.0 & 0.0  & 0.0\ \end{bmatrix}^T$                         \\ \hline $M_{23}$ & \cellcolor[HTML]{67FD9A}$\begin{bmatrix}\ 0.0 & 0.0 & 0.0 & 0.0\ \end{bmatrix}^T$ & \cellcolor[HTML]{67FD9A}$\begin{bmatrix}\ 0.0 & 0.0  & 0.0\ \end{bmatrix}^T$                         \\ \hline $M_{24}$ & \cellcolor[HTML]{67FD9A}$\begin{bmatrix}\ 0.0 & 0.0 & 0.0 & 0.0\ \end{bmatrix}^T$ & \cellcolor[HTML]{67FD9A}$\begin{bmatrix}\ 0.0 & 0.0  & 0.0\ \end{bmatrix}^T$                         \\ \hline $M_{25}$ & \cellcolor[HTML]{67FD9A}$\begin{bmatrix}\ 0.0 & 0.0 & 0.0 & 0.0\ \end{bmatrix}^T$ & \cellcolor[HTML]{67FD9A}$\begin{bmatrix}\ 0.0 & 0.0  & 0.0\ \end{bmatrix}^T$                         \\ \hline $M_{26}$ & \cellcolor[HTML]{67FD9A}$\begin{bmatrix}\ 0.0 & 0.0 & 0.0 & 0.0\ \end{bmatrix}^T$ & \cellcolor[HTML]{67FD9A}$\begin{bmatrix}\ 0.0 & 0.0  & 0.0\ \end{bmatrix}^T$                         \\ \hline $M_{27}$ & \cellcolor[HTML]{67FD9A}$\begin{bmatrix}\ 0.0 & 0.0 & 0.0 & 0.0\ \end{bmatrix}^T$ & \cellcolor[HTML]{67FD9A}$\begin{bmatrix}\ 0.0 & 0.0  & 0.0\ \end{bmatrix}^T$                         \\ \hline $M_{28}$ & \cellcolor[HTML]{67FD9A}$\begin{bmatrix}\ 0.0 & 0.0 & 0.0 & 0.0\ \end{bmatrix}^T$ & \cellcolor[HTML]{67FD9A}$\begin{bmatrix}\ 0.0 & 0.0  & 0.0\ \end{bmatrix}^T$                         \\ \hline $M_{34}$ & \cellcolor[HTML]{FF8080}$\begin{bmatrix}\ 0.0 & 0.01 & 0.02 & -0.13\ \end{bmatrix}^T$ & \cellcolor[HTML]{67FD9A}$\begin{bmatrix}\ 0.0 & 0.0  & 0.0\ \end{bmatrix}^T$                         \\ \hline $M_{35}$ & \cellcolor[HTML]{67FD9A}$\begin{bmatrix}\ 0.0 & 0.0 & 0.0 & 0.0\ \end{bmatrix}^T$ & \cellcolor[HTML]{67FD9A}$\begin{bmatrix}\ 0.0 & 0.0  & 0.0\ \end{bmatrix}^T$                         \\ \hline $M_{36}$ & \cellcolor[HTML]{67FD9A}$\begin{bmatrix}\ 0.0 & 0.0 & 0.0 & 0.0\ \end{bmatrix}^T$ & \cellcolor[HTML]{67FD9A}$\begin{bmatrix}\ 0.0 & 0.0  & 0.0\ \end{bmatrix}^T$                         \\ \hline $M_{37}$ & \cellcolor[HTML]{67FD9A}$\begin{bmatrix}\ 0.0 & 0.0 & 0.0 & 0.0\ \end{bmatrix}^T$ & \cellcolor[HTML]{67FD9A}$\begin{bmatrix}\ 0.0 & 0.0  & 0.0\ \end{bmatrix}^T$                         \\ \hline $M_{38}$ & \cellcolor[HTML]{67FD9A}$\begin{bmatrix}\ 0.0 & 0.0 & 0.0 & 0.0\ \end{bmatrix}^T$ & \cellcolor[HTML]{67FD9A}$\begin{bmatrix}\ 0.0 & 0.0  & 0.0\ \end{bmatrix}^T$                         \\ \hline $M_{45}$ & \cellcolor[HTML]{67FD9A}$\begin{bmatrix}\ 0.0 & 0.0 & 0.0 & 0.0\ \end{bmatrix}^T$ & \cellcolor[HTML]{67FD9A}$\begin{bmatrix}\ 0.0 & 0.0  & 0.0\ \end{bmatrix}^T$                         \\ \hline $M_{46}$ & \cellcolor[HTML]{67FD9A}$\begin{bmatrix}\ 0.0 & 0.0 & 0.0 & 0.0\ \end{bmatrix}^T$ & \cellcolor[HTML]{67FD9A}$\begin{bmatrix}\ 0.0 & 0.0  & 0.0\ \end{bmatrix}^T$                         \\ \hline $M_{47}$ & \cellcolor[HTML]{67FD9A}$\begin{bmatrix}\ 0.0 & 0.0 & 0.0 & 0.0\ \end{bmatrix}^T$ & \cellcolor[HTML]{67FD9A}$\begin{bmatrix}\ 0.0 & 0.0  & 0.0\ \end{bmatrix}^T$                         \\ \hline $M_{48}$ & \cellcolor[HTML]{67FD9A}$\begin{bmatrix}\ 0.0 & 0.0 & 0.0 & 0.0\ \end{bmatrix}^T$ & \cellcolor[HTML]{67FD9A}$\begin{bmatrix}\ 0.0 & 0.0  & 0.0\ \end{bmatrix}^T$                         \\ \hline $M_{56}$ & \cellcolor[HTML]{FF8080}$\begin{bmatrix}\ 0.0 & 0.02 & -0.01 & 0.13\ \end{bmatrix}^T$ & \cellcolor[HTML]{67FD9A}$\begin{bmatrix}\ 0.0 & 0.0  & 0.0\ \end{bmatrix}^T$                         \\ \hline $M_{57}$ & \cellcolor[HTML]{67FD9A}$\begin{bmatrix}\ 0.0 & 0.0 & 0.0 & 0.0\ \end{bmatrix}^T$ & \cellcolor[HTML]{67FD9A}$\begin{bmatrix}\ 0.0 & 0.0  & 0.0\ \end{bmatrix}^T$                         \\ \hline $M_{58}$ & \cellcolor[HTML]{67FD9A}$\begin{bmatrix}\ 0.0 & 0.0 & 0.0 & 0.0\ \end{bmatrix}^T$ & \cellcolor[HTML]{67FD9A}$\begin{bmatrix}\ 0.0 & 0.0  & 0.0\ \end{bmatrix}^T$                         \\ \hline $M_{67}$ & \cellcolor[HTML]{67FD9A}$\begin{bmatrix}\ 0.0 & 0.0 & 0.0 & 0.0\ \end{bmatrix}^T$ & \cellcolor[HTML]{67FD9A}$\begin{bmatrix}\ 0.0 & 0.0  & 0.0\ \end{bmatrix}^T$                         \\ \hline $M_{68}$ & \cellcolor[HTML]{67FD9A}$\begin{bmatrix}\ 0.0 & 0.0 & 0.0 & 0.0\ \end{bmatrix}^T$ & \cellcolor[HTML]{67FD9A}$\begin{bmatrix}\ 0.0 & 0.0  & 0.0\ \end{bmatrix}^T$                         \\ \hline $M_{78}$ & \cellcolor[HTML]{FF8080}$\begin{bmatrix}\ 0.0 & -0.01 & -0.03 & -0.13\ \end{bmatrix}^T$ & \cellcolor[HTML]{67FD9A}$\begin{bmatrix}\ 0.0 & 0.0  & 0.0\ \end{bmatrix}^T$                         \\ \hline \end{tabular}
}
\end{center}
\end{table}

%\section{Conclusion}

%The analysis, provides in this chapter, shows that a careful selection of the octocopter configuration may additionally influence the octocopter overall maneuverability and keep it ready to execute the mission under variety of fault states. For instance, when the probability of a double fault is high, we can increase mission reliability by choosing the PPNNPPNN configuration.

%Triple or quadruple faults can be also analyzed in the same way. However, the probability of such occurrence is much lower than for single or double faults, so these types of faults are not considered in this book. As already indicated, in order to fully exploit the presented maneuvarability analysis for each system regardless of the number of motors and their rotational directions, one should find the way how to include the obtained admissible set during the planning stage as well. The next section demonstrates an idea how to construct such a planner.

%\include{chapters/Chapter5/ch5}
%\include{chapters/Chapter6/ch6}
\chapter{Risk-sensitive motion planning}

 \color{black}

This section presents an idea how to construct a motion planner for an octocopter system based on the admissible set of thrust force and torques obtained through fault-dependent maneuverability analysis presented in Section 4. We present a risk-sensitive (or risk-aware) motion-planning algorithm capable of taking into account risks during the planning stage by means of mission-related fault-tolerant analysis. We showed in \cite{Osmic_automatika} that the approach was less conservative in terms of selected performance measures than a conservative risk planner (or risk-averse) that assumed that the considered fault would certainly occur during the mission execution. On the other hand, the risk-sensitive motion planner is also readier for accepting failures during the mission execution than the risk-Ignorant approach (or risk-prone) that assumes no failure will occur. In this section, we describe this approach and present the obtained results.

\section{Presentation of the admissible set of thrust force and torques with a set of inequality constraints}

As shown in Chapter 4, the admissible set for thrust force and torques can be determined depending on the number of DC motors used, the orientation configuration and the states of DC motors (with or without faults). This admissible set has a convex polytope-like form in four-dimensional space.

Each of the outer sides of the polytope can be represented by its related hyper-plane based on the four points that form that side, that is:
\begin{equation}
\boldsymbol{aT}+\boldsymbol{b\tau_{x}}+\boldsymbol{c\tau_{y}}+\boldsymbol{d\tau_{z}}\leqslant\boldsymbol{e},\label{hiper_ravan}
\end{equation}
where $\boldsymbol{a}$, $\boldsymbol{b}$, $\boldsymbol{c}$, $\boldsymbol{d}$ and $\boldsymbol{e}$ are the slope coefficients of the individual axes. 

Since the polytope-like admissible set is composed of a large number of such hyper-planes, it can be represented as a set of inequalities that fully describes the admissible set of thrust force and torques for each specific MAV design. For instance, for a healthy PNPNPNPN octocopter system, it turned out that the related polytope can be described by 617 inequalities. In case any of DC motor is in a fault state, the number of inequalities decreases, while in an extreme case when all DC motors are in failure modes, the admissible set is reduced to a single point at origin, that is $u_{ref}=[\begin{array}{cccc}
T & \tau_{x} & \tau_{y} & \tau_{z}\end{array}]=[\begin{array}{cccc}
0 & 0 & 0 & 0\end{array}]$, in which the system is fully uncontrollable.

The obtained inequalities can be further used in motion planning to generate a feasible trajectory that depends on the initial admissible set (i.e., the resulting polytope) of thrust force and torques. In the next section, we describe a risk-sensitive planner (RSP) based on a careful selection of some of the inequalities that describe the admissible set (only a few of them), where the selection process depends on the required mission.

\section{Selected optimization framework for motion planning}

In Chapter 3 it is shown that for the reference trajectory position tracking, the functions representing the coordinates $x$ and $y$  must be at least four times differentiable, while the functions representing the heights $z$ and orientations $\psi$ are at least twice differentiable. The references for $\phi$ and $\theta$ orientation coordinates are obtained as an output of controlling the $x$ and $y$  position coordinates. Accordingly, the height $z$ and orientation coordinates $\psi$ behave as double integrators, that is
\begin{equation}
\ddot{q},\label{double_int}
\end{equation}
and the $ x $ and $ y $ position coordinates can be approximated by a quadruple integrator: 
\begin{equation}
\ddddot{q},\label{quad_int}
\end{equation}
where $q=[x \ y \ z \ \psi]$, $\dot{q}$ and $\ddot{q}$ represent the generalized coordinates, velocity and acceleration, respectively.

The minization of acceleration  \eqref{double_int} and snap \eqref{quad_int} directly yields the minimization of the generalized forces acting on the system. This further results in the minimization of energy consumption while taking into account the constraints imposed on the trajectory. This consequently means that the battery consumption during the mission will be minimal. Detailed description of motion planning based on minimal acceleration and snap can be found in \cite{mellinger_snap}. 

For this reason, motion planning problem can be described as a fixed finite-time optimization problem given as
\begin{equation}
\label{optim_okvir}
\begin{tabular}{ccl}
               & $\underset{0\leqslant t \leqslant T}{minimize}(\Vert \boldsymbol{\ddot{q}}\Vert^{2})$                                                                                                                                                                                              &  \\
$subject$ $to$ &                                                                                                                                                                                                                                                 &  \\
               & $q_{min}\leqslant {q}\leqslant q_{max}$                                                                                                                                                                                                         &  \\
               & $\dot{q}_{min}\leqslant \dot{q}\leqslant \dot{q}_{max}$                                                                                                                                                                                         &  \\
               & $\ddot{q}_{min}\leqslant \ddot{q}\leqslant \ddot{q}_{max}$,                                                                                                                                                                                        \end{tabular}
\end{equation}
where the fixed finite-time represents the mission execution time $T$. The waypoints, as part of the given mission, through which the octocopter is supposed to pass should also be included in the optimization framework as desired constraints. One way to include these constraints is to impose hard constraints into the optimization. To ensure that the planner can be risk-averse, if necessary, we need to allow the motion planner to be capable of generating trajectories that may deviate from the waypoints which are designed for the selected mission. The deviation from the waypoints can be also used as a performance measure for the given mission. To do so, we include these constraints into the objective function by penalizing large deviations from the given waypoints as

\[
\ensuremath{\underset{0\leqslant t\leqslant T}{minimum}(\Vert\boldsymbol{\ddot{q}}\Vert^{2}+{\displaystyle \underset{i}{\sum}}\alpha_{i}(\Vert q-q_{i}\Vert^{2}))}
\]

\: \: \: \: \: \: \: \: \: \: \: \: \: \:  $subject$ $to$

\begin{equation}
\begin{array}{c}
\ensuremath{q_{min}\leqslant{q}\leqslant q_{max}}\\
\dot{q}_{min}\leqslant\dot{q}\leqslant\dot{q}_{max}\\
\ddot{q}_{min}\leqslant\ddot{q}\leqslant\ddot{q}_{max},
\end{array}\label{optim_okvir_prosireni}
\end{equation}
where the weights $0\leq\alpha_{i}\leq1$ are used to describe how important it is to pass through the waypoints $q_{i}$ during the mission execution. For the purpose of this presentation, we set $\alpha_{i}=1$ for all $i$. 

Since ({\ref{eq:lin_visine}}), ({\ref{eq:din_-model_orijentacije}}) and 
({\ref{eq:lin_koordinate}}) provide the relations between $T$, $\tau_x$, $\tau_y$ and $\tau_z$ and $q$, it is now possible to include the inequalities which describe the fault-dependent admissible set into the optimization framework. This gives the final form of the optimization framework used for the RSP motion planner: 
\begin{equation}
\label{optim_konacna}
\begin{tabular}{ccl}
               & $\underset{0\leqslant t \leqslant T}{minimize}(\Vert \boldsymbol{\ddot{q}}\Vert^{2}+{\displaystyle \underset{i}{\sum}}\alpha_{i}(\Vert q-q_{i}\Vert^{2}))$                                                                                                                                                                                              &  \\
$subject$ $to$ &                                                                                                                                                                                                                                                 &  \\
               & $q_{min}\leqslant {q}\leqslant q_{max}$                                                                                                                                                                                                         &  \\
               & $\dot{q}_{min}\leqslant \dot{q}\leqslant \dot{q}_{max}$                                                                                                                                                                                         &  \\
               & $\underset{i}{aT+b\boldsymbol{\tau_{x}}+c\boldsymbol{\tau_{y}}+d\boldsymbol{\tau_{z}}\leqslant e}.$\end{tabular}
\end{equation}

In this way, the control admissible set obtained through the maneuverability analysis presented in Chapter 4 can be included in the optimization framework as an additional set of contraints for each $i$, where $i$ indicates the number of waypoints. It should be noted that the optimization requires $\ddot{q}$ and $\ddddot{q}$ which can be approximated by the first-order approximation as
\begin{equation}
\dot{q}=\frac{q(i)-q(i-1)}{\varDelta t},\label{eq:prvi izvod}
\end{equation}
where $\varDelta t=T/N$, $T$ the mission execution time, while $N$ is the number of additional points included between waypoints. The role of these additional points is to increase a path resolution in order to provide a smooth reference path between waypoints. 

In case the environment contains obstacles preventing the octocopter from crossing some given waypoints, it is also possible to impose an additional constraint to secure collision-free paths. Namely, it is necessary to define a safety distance $R_{safe}$ to permit the system to stay distant from obstacles during the mission execution. The complete optimization framework that can be used for the RSP is given in the form

\[
\ensuremath{\underset{0\leqslant t\leqslant T}{minimum}(\Vert\boldsymbol{\ddot{q}}\Vert^{2}+{\displaystyle \underset{i}{\sum}}\alpha_{i}(\Vert q-q_{i}\Vert^{2})),}
\]

\: \: \: \: \: \: \: \: \: \: \:   $subject$ $to$

\begin{equation}
\begin{array}{c}
\ensuremath{q_{min}\leqslant{q}\leqslant q_{max}}\\
\text{\ensuremath{R_{safe}\leqslant\left|q-q_{ob}\right|}}\\
\ensuremath{\dot{q}_{min}\leqslant\dot{q}\leqslant\dot{q}_{max}}\\
\ensuremath{\underset{i}{aT+b\boldsymbol{\tau_{x}}+c\boldsymbol{\tau_{y}}+d\boldsymbol{\tau_{z}}\leqslant e.}}
\end{array}\label{eq:optim_s_preprekama}
\end{equation}
where $q_{ob}$ is the obstacle coordinate position.

In order to take into account any possible failure during the motion planning stage, one can include all constraints related to the admissible set of that failure. We call such a planner a risk-conservative planner (RCP). However, the RCP planner would be quite conservative, so in the following subsection we describe how to select some of the inequalities by carefully examining the given mission to form the RSP planner. It should be noted that the decision related to which failures and their related admissible sets to include should follow from the failure mode and effects analysis (FMEA) \cite{FTC_ETH_Thomas_Schneider}, \cite{osmic_FMEA}.

\section{Risk-sensitive motion planner based on mission-related fault-tolerant analysis}

In addition to the RCP planner, the RSP planner proposed in \cite{Osmic_automatika} takes into account only relevant inequalities from the total number included in the control addmissible set. The classical planner which does not consider any possible failure is referred to as Risk-Ignorant Planer (RIP). 

In this section we describe how to take possible failures into account during the motion planning stage by means of their related admissible sets. By doing so, we aim to include the associated risk into the planner in order to increase reliability of the mission execution in terms of satisfactory performance. This will be done at the cost of a much smaller performance deterioration than in case when the RCP planner is used for which all constraints related to risk-dependent admissible sets are included. As expected, the planner will require a bit more time to complete the mission than the RIP planner. 

When a fault occurs the octocopter system may be in a position and orientation such that the control allocation is capable to produce desired thrust force and torques for the remaining DC motors without any effect on the mission performance. On the contrary, the octocopter may be in such position and orientation to significantly deteriorate the performance. The idea behind the proposed RSP motion planner is to carefully select a certain number of fault-dependent inequalities to retain the overall performance of a healthy system as much as possible. By doing so, the planner aims to minimally reduce the vehicle maneuverability (much less than in case of the RCP motion planner) in order to decrease probability of being in those states in which the vehicle might significantly deteriorate the performance when any of selected fault occurs. 

The overall steps for mission-related fault-tolerant analysis and for designing the proposed RSP motion planner can be summarized as follows:

\begin{enumerate}

\item Select the failure modes of interest based on the FMEA analysis (e.g., single motor failures).\\  
\item Determine the minimum mission execution time for the RIP motion planner to achieve a feasible solution (from those that secure passing through waypoints).\\
\item For all selected failure modes, determine the minimum mission execution times of the RCP motion planner to achieve feasible solutions.\\
\item Set the maximum time of all minimum times obtained in step 3 to be the mission time in order to ensure that the RCP planner provides feasible solutions for each failure modes.\\
\item Find all inequalities associated to the fault dependent admissible sets for each selected failure mode from step 1, which are not satisfied during the mission execution based on the RIP motion planer and the mission time obtained in step 4.\\
\item Form the final optimization framework for the RSP motion planner by including all constraints found in step 5 and determine the minimum mission execution time for the RSP motion planner to achieve a feasible solution. This optimization framework represents the proposed RSP motion planner.\\ 

\end{enumerate}

The presented design steps can be explained as follows. First, we perform the FMEA analysis in order to find the most critical failure mode that will be taken into account during the planning stage (step 1). Second, we determine the minimum execution time for the RIP planner (step 2). The minimum mission execution time represents a time for which the optimization framework still gives a feasible solution, that is, the solution which ensures passing through the waypoints. Since the proposed design steps can be conducted off-line, that is before the mission execution, this minimum value can be easily found by incrementally decreasing the time and checking whether the related solution is feasible or not. Then, for all selected failure modes from step 1, we determine the minimum execution times (step 3) obtained with the RCP planner. In step 4, we select the worst-case (maximum time) from step 3 to be the mission time in order to ensure that all planners provide feasible solutions. This is also important for a fair comparison of all planners by means of the performance measured by deviation from the waypoints. Otherwise, some of the planners would be infeasible. In step 5, we first find the admissible sets for each failure mode and determine their related inequality sets. Then, we test the RIP motion planner, given the mission time from step 4, in order to find only those inequalities which are not possible to satisfy for the considered mission. To do so, we check the thrust force and the torques obtained by the RIP motion planner against the related admissible sets for each failure mode. In step 6, we form the final optimization framework by including a constraint set consisted of the inequalities extracted from step 5 and the admissible set of the healthy octocopter system. 

\section{Simulation results}

\begin{figure}
\centering{}\includegraphics[scale=0.5]{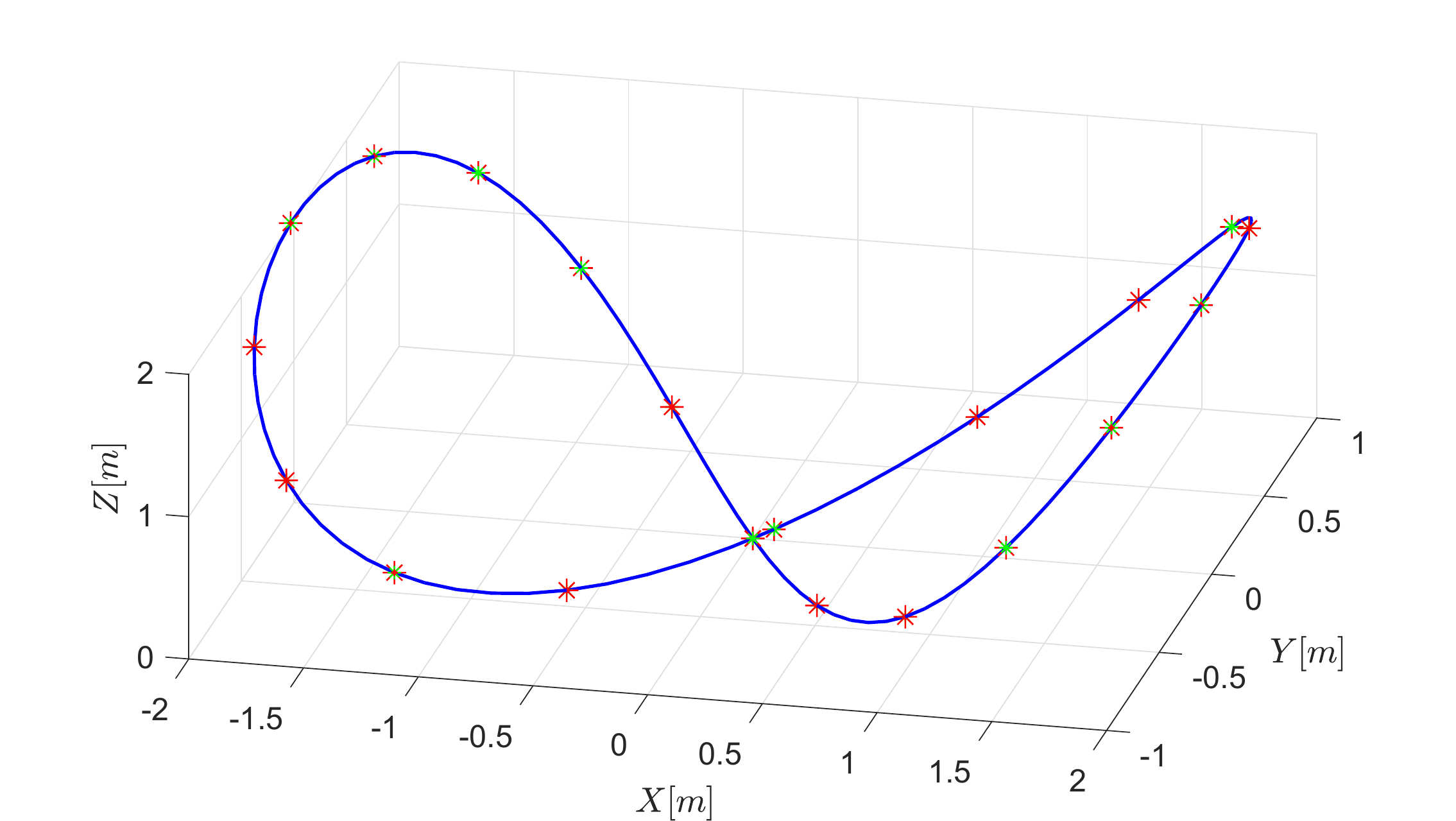}
\caption{{Uniform sampling of the Viviani curve with 21 points.\label{fig:Vivijani_uzorkovanje}}}
\end{figure}

The mission is defined in the form of Vivian curve as in preceding sections, in which 21 points have been generated uniformly along the curve to define the waypoints (Figure \ref{fig:Vivijani_uzorkovanje}). For the optimization framework (\ref{eq:optim_s_preprekama}), we generate additional 10 points between each successive waypoints, which gives the total number of $N=210$ points.

To test the quality of generated trajectories, we use two types of error, the first is related to the position $e_{R}$ and the second to the orientation $e_{\Psi}$, as:

\begin{equation}
e_{R}=\underset{i}{\sum}\sqrt{(x_{i}-x_{ref_{i}})^{2}+(y_{i}-y_{ref_{i}})^{2}+(z_{i}-z_{ref_{i}})^{2}}\label{Grska_poz}
\end{equation}

\begin{equation}
e_{\Psi}=\underset{i}{\sum}\sqrt{(\psi_{i}-\psi_{ref_{i}})^{2}}.\label{Grska_orj}
\end{equation}

\begin{figure}
\centering{}\includegraphics[scale=0.5]{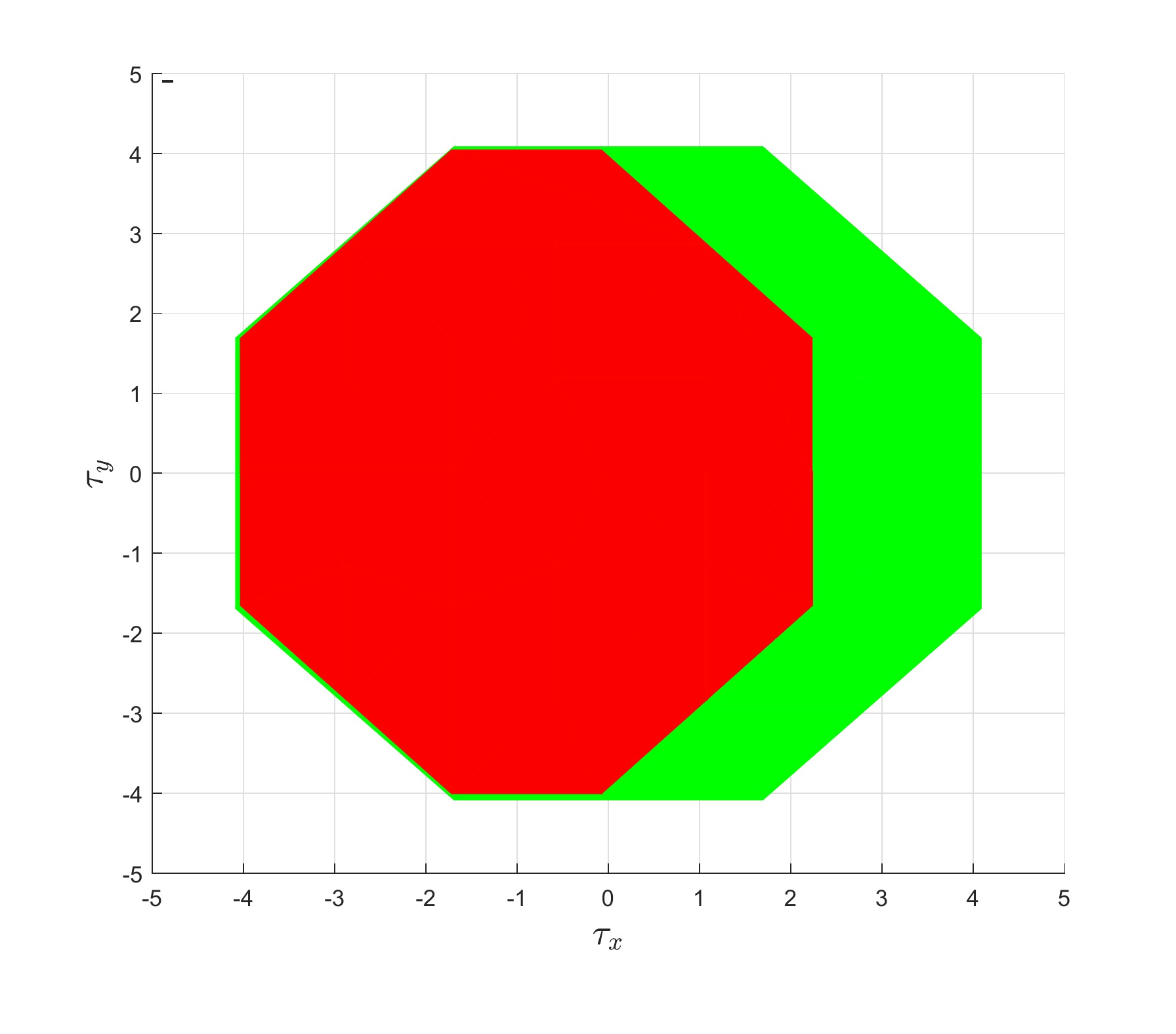}
\caption{{Control input domain (projection onto plane $T = mg$) in case without any faults (green), and in case of M1 in fault state (red).
\label{fig:Istovremeni_jednostruki}}}
\end{figure}

\textbf{Case 1:} First, we consider only a single failure occurred in the DC motor $M_1$ in order to take it into account in the motion planning stage (step 1 of the $RSP$ planning algorithm). Fig. \ref{fig:Istovremeni_jednostruki} shows the control admissible set obtained with the maneuverability analysis based on this failure. The obtained set is obviously reduced with respect to the admissible set for the system without failure consideration (the healthy octocopter system).

In steps 2 and 3 we determine the minimum times to get a feasible solution for the $RIP$ motion planner, which is $16$ $[s]$, and for the RCP motion planner, which is $20$ $[s]$ (see Table \ref{tab:Healty}).

In accordance to step 4, we then choose the mission time to be 20 [s] for the next step. In accordance with step 5, for the thrust forces and torques obtained by the RIP planner, we find all inequalities from the fault-dependent admissible set which are not satisfied during the mission execution. For the considered example, in the case of the RIP planner, the obtained thrust force and torques violate only two inequalities (out of 440 that describe the admissible set of thrust force and torques) in 111 cases related to 99 discrete positions along the mission curve (out of possible 210). By including only these two additional constraints into the final optimization (step 6), we get the RSP planner. One can observe that the RSP planner violates only one inequality constraint at only one position (see Table \ref{tab:Samo_jednostruki}).

\color{black}
\begin{table}\caption{{Performance comparison between the RIP and RCP approaches (Case 1, steps 
$1$, $2$ and $3$).\label{tab:Healty}}}

\centering{}%
\begin{tabular}{|l|c|c|c|c|c|}
\hline 

MAV
 & ${T}$ $[s]$ & $e_{R}[m]$  & $e_{R}/N[m]$  & $e_{\Psi}[rad]$  & $e_{\Psi}$/N {[}rad{]} \tabularnewline
\hline 
$RIP$  & 16  & 0,0412  & 0,0019  & 4e-12  & 4e-13 \tabularnewline
\hline 
$RCP$  & 20  & 0,0566  & 0,0027  & 1e-18  & 5e-21 \tabularnewline
\hline 
\end{tabular}
\end{table}

\begin{figure}
\centering{}\includegraphics[scale=0.5]{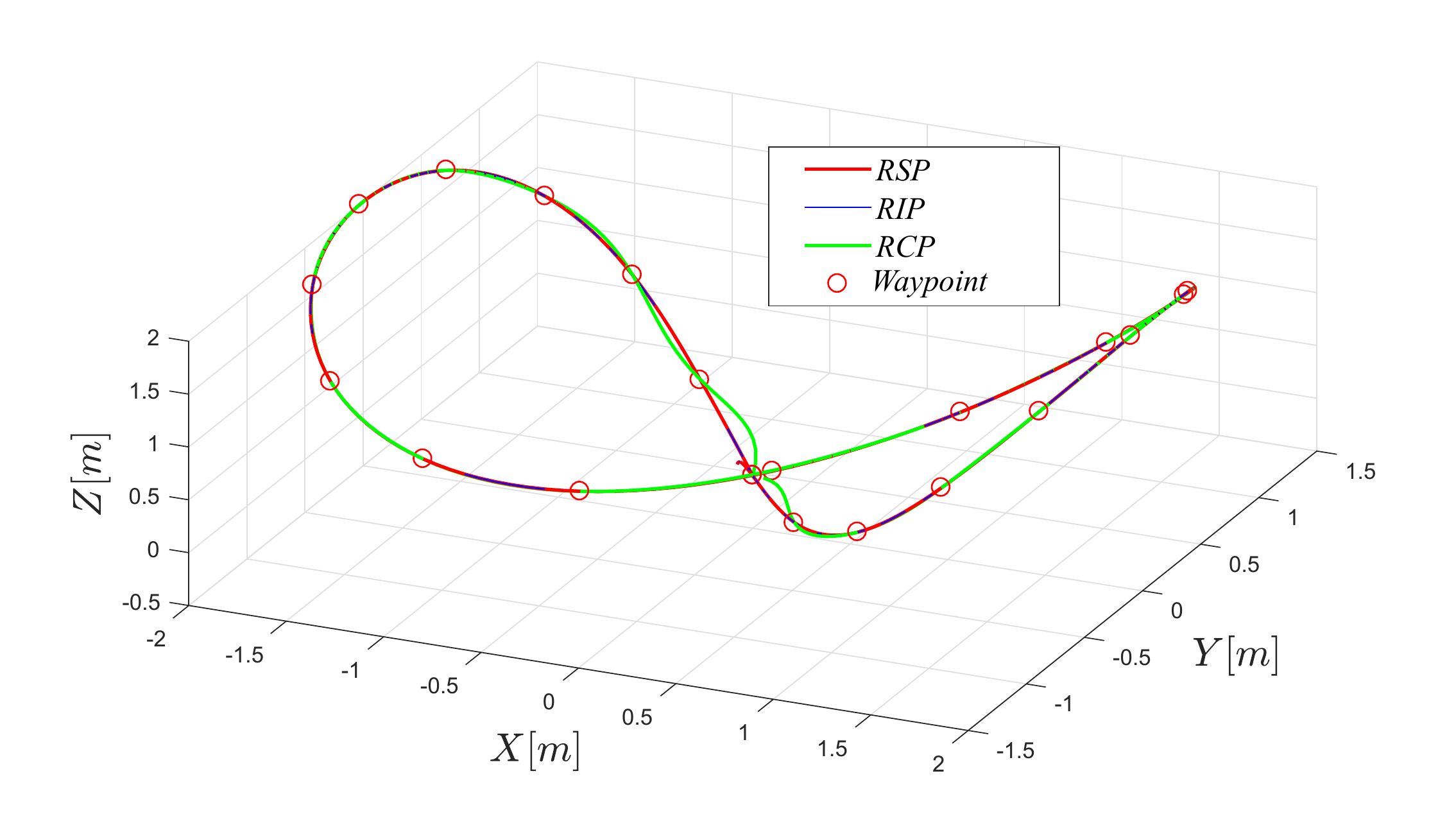}
\caption{{Paths generated using RIP, RCP, and RSP motion planners. The traversal time is 20 seconds.. \label{fig:Scenario_1}}}
\end{figure}

In order to fairly compare the results, the RIP planner is also executed for the mission time $T=20$ $[s]$. The paths generated by all three planners are shown in Fig. \ref{fig:Scenario_1}. One can observe that the RSP and RIP planners generated almost the same path, while the RCP has slight deviations with respect to the mission waypoints. However, the detailed comparison of all three planners (RIP, RCP and RSP) for the maximum time obtained ($T=20$, step 4), is given in Table \ref{tab:Samo_jednostruki}. It can be seen that the RIP and RSP planners obtained similar performances, better than the RCP approach. On the other hand, the RIP planner violated the constraints in 51 positions, which is less than in case when the mission execution time was $T=16$ $[s]$ when there were 111 such positions. Unlike the RIP, the RSP violated only one inequality constraint at only one position (see Table  \ref{tab:Samo_jednostruki}).

\begin{figure}
\centering{}\includegraphics[scale=0.55]{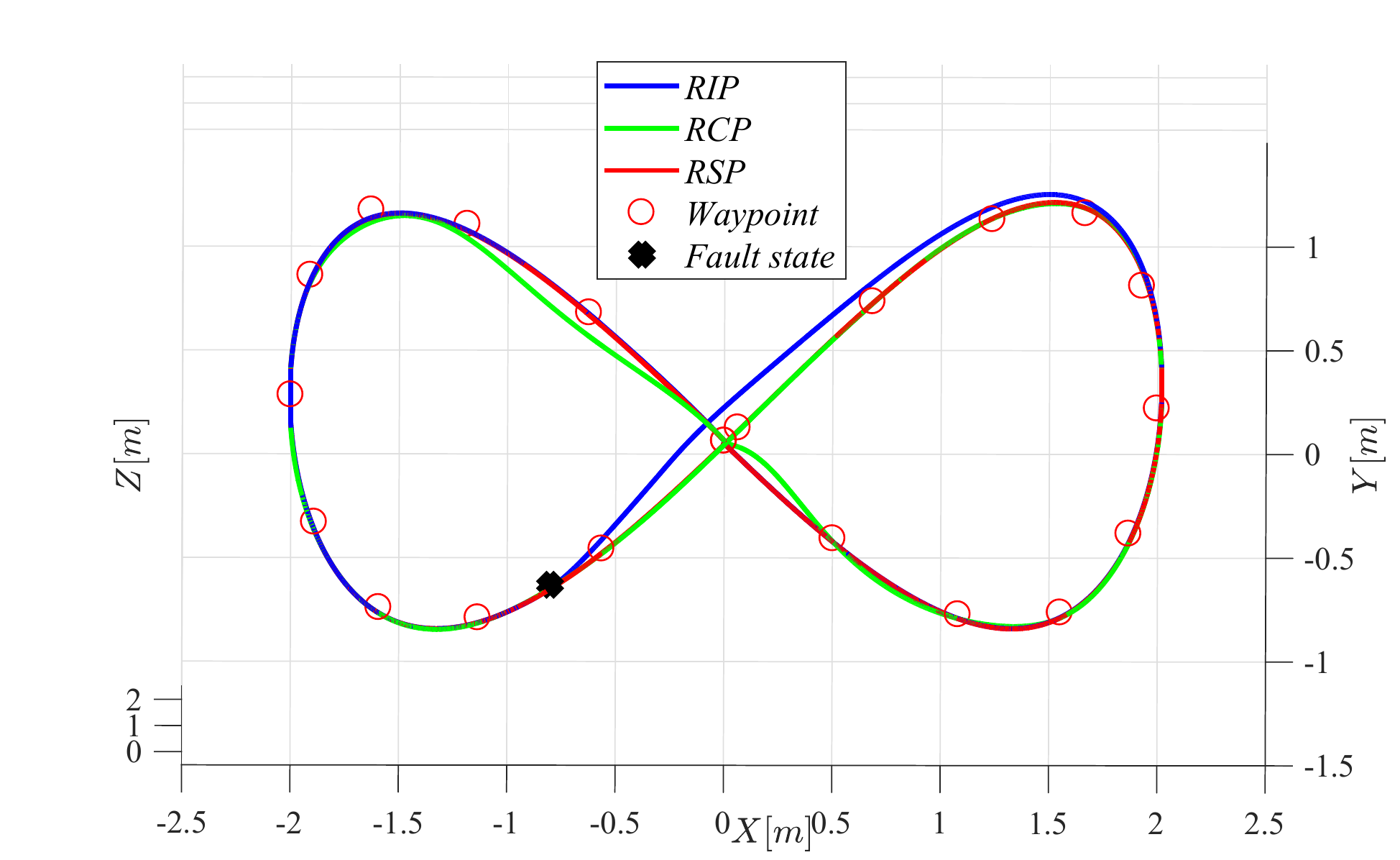}
\caption{{Tracking of trajectories obtained using RIP, RCP, and RSP planners. The nominal traversal time is 20s, with the fault state at M1 occurring at t = 8s. 
\label{fig:Scenario_1_simulacija}}}
\end{figure}

In order to justify the use of the RSP instead of the RIP, we illustrate one example with $T=20$ $[s]$ and a failure occurred in the motor $M_1$ at $t=8$ $[s]$. The resulting paths are shown in Fig. \ref{fig:Scenario_1_simulacija}. One can observe that the RSP planner (red) has the smallest total deviation from the waypoints. The RCP (green) has the worst deviation before the failure occurs, which is expected since during this period the planner produces the most conservative paths. The RIP (blue) generates the path without deviations before that critical event, but the significant deviation appears after that moment.

To further compare the planners, we additionally introduce three types of errors. 
$e_{Rp}[m]$ and $e_{\Psi p[rad]}$ represent the first error type indicating position and orientation errors with respect to the mission waypoints. The second type of errors are $e_{RRef}[m]$ and $e_{\Psi Ref}[rad]$ which indicate the position and orientation errors with respect to the no-failure path, while the third type of errors, $e_{RRIP}[m]$ and $e_{\Psi RIP}[rad]$, are the position and orientation errors with respect to the no-failure path obtained with the RIP planner which represents an ideal path in case without failures. From Table \ref{tab:Poredjenje_planera_jednostruki}, one can see that the RSP has the best overall performance.

\begin{table}
\caption{{Performance comparison for RIP, RCP and RSP motion planners with traversal time T=20s (Case 1, steps 5 and 6).
.\label{tab:Samo_jednostruki}}}

\centering{}%
\begin{tabular}{|l|c|c|c|c|}
\hline 
Performance & ${T}$ $[s]$ & $e_{R}[m]$  & $e_{\Psi}[rad]$  & $violated$ ${}constraints{}$ \tabularnewline
\hline 
$RIP$ & 20  & 0,038  & 5e-19  & 51 \tabularnewline
\hline 
$RCP$  & 20  & 0,057  & 1e-18  & 0 \tabularnewline
\hline 
$RSP$   & 20  & 0,038  & 1,25e-18  & 1 \tabularnewline
\hline 
\end{tabular}
\end{table}

\begin{figure}
\centering{}\includegraphics[scale=0.45]{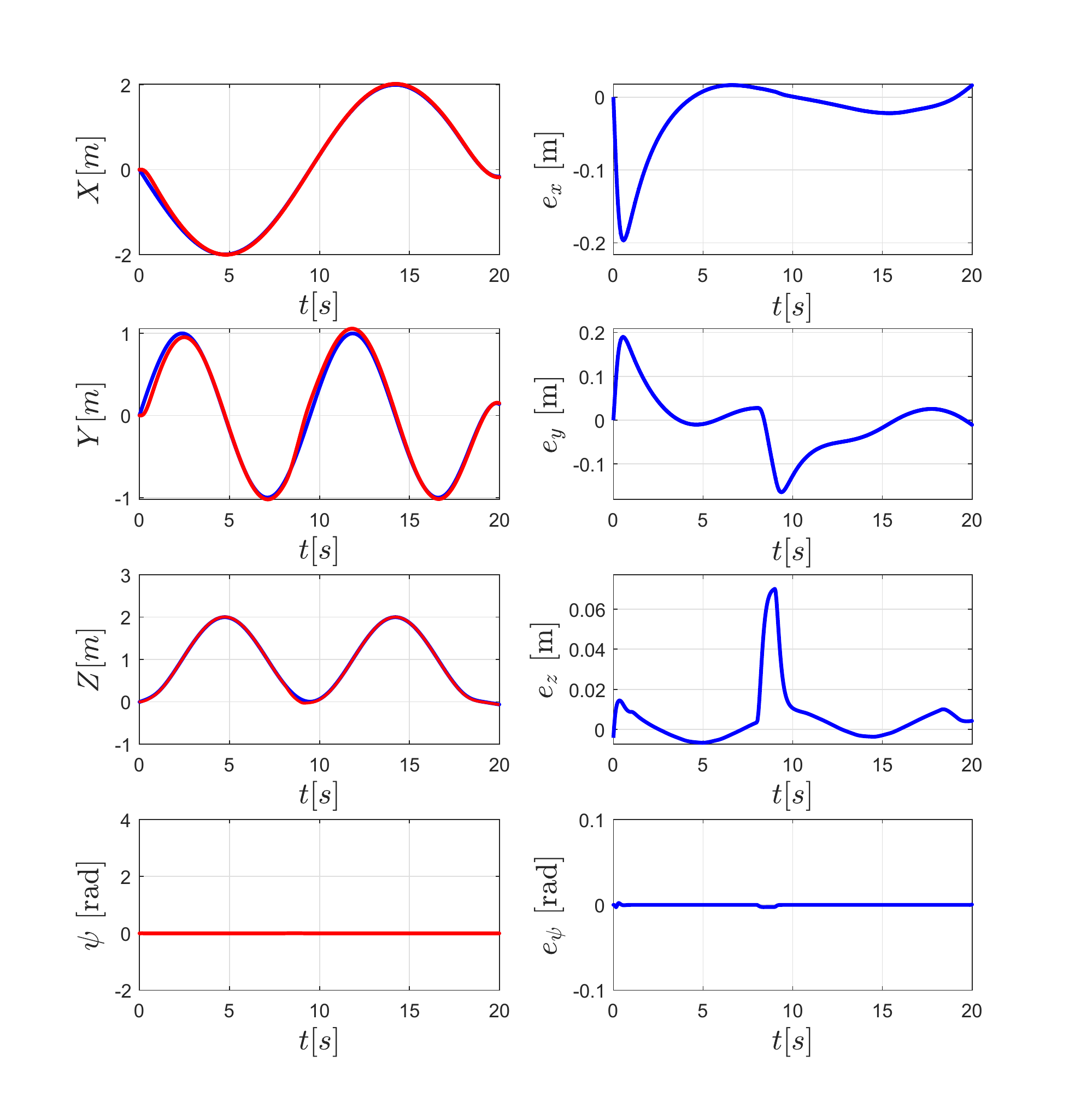}
\caption{{Tracking of Cartesian coordinates $x$, $y,$ $z$ orientation $\psi$
and respective tracking errors  $e_{x},$ $e_{y}$, $e_{z}$ i $e_{\psi}$ in case of trajectory generated by RIP planner. The traversal time is 20s, with the fault at M1 occurring at t=8s.  
\label{fig:RIP_jednostruka}}}
\end{figure}

\begin{figure}
\centering{}\includegraphics[scale=0.5]{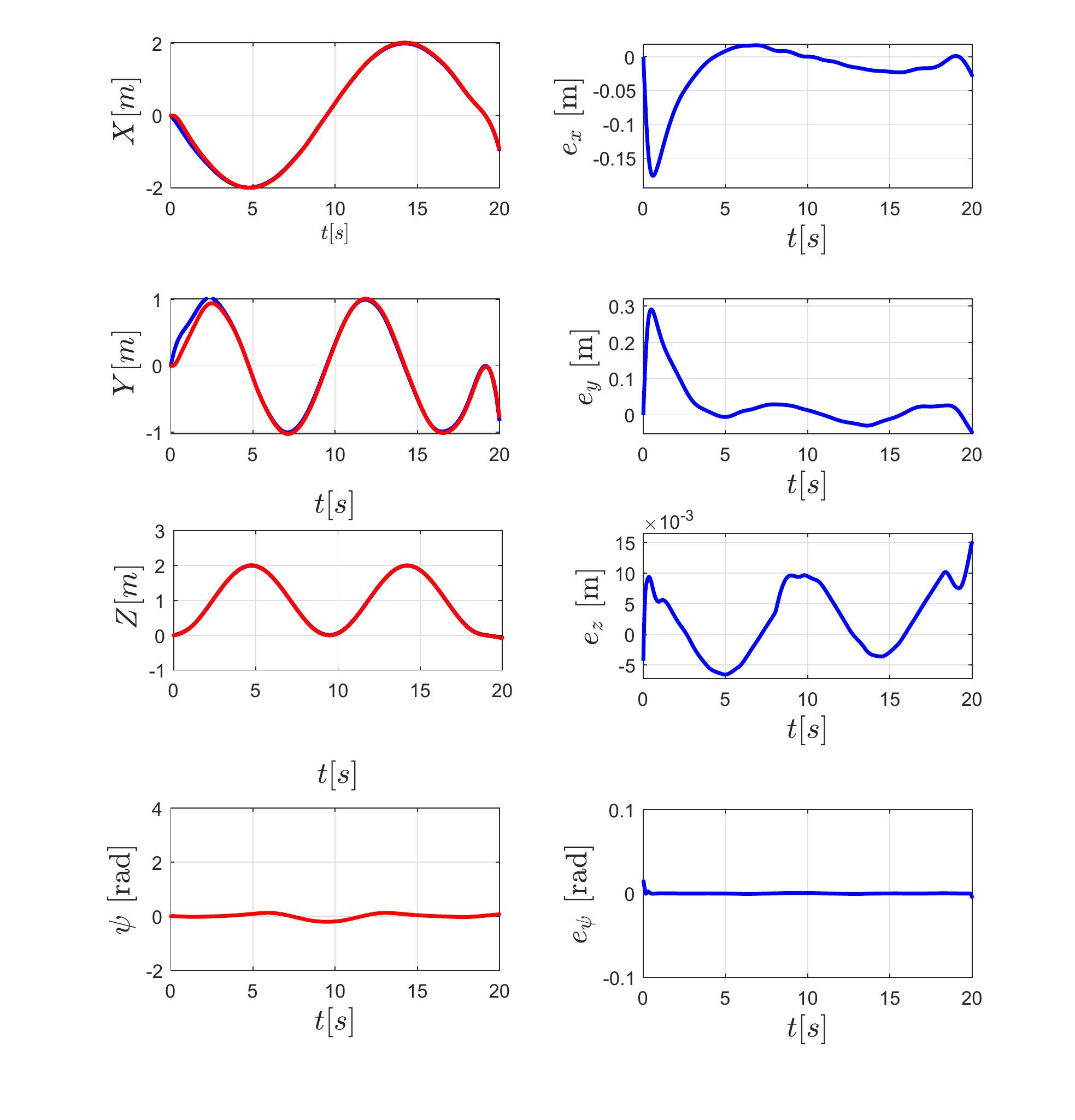}
\caption{{Tracking of Cartesian coordinates $x$, $y,$ $z$ orientation $\psi$
and respective tracking errors $e_{x},$ $e_{y}$, $e_{z}$ and $e_{\psi}$  in case of trajectory generated by RCP planner. The traversal time is 20s, with the fault at M1 occurring at t=8s.  
\label{fig:RCP_jednostruka}}}
\end{figure}

\begin{figure}
\centering{}\includegraphics[scale=0.5]{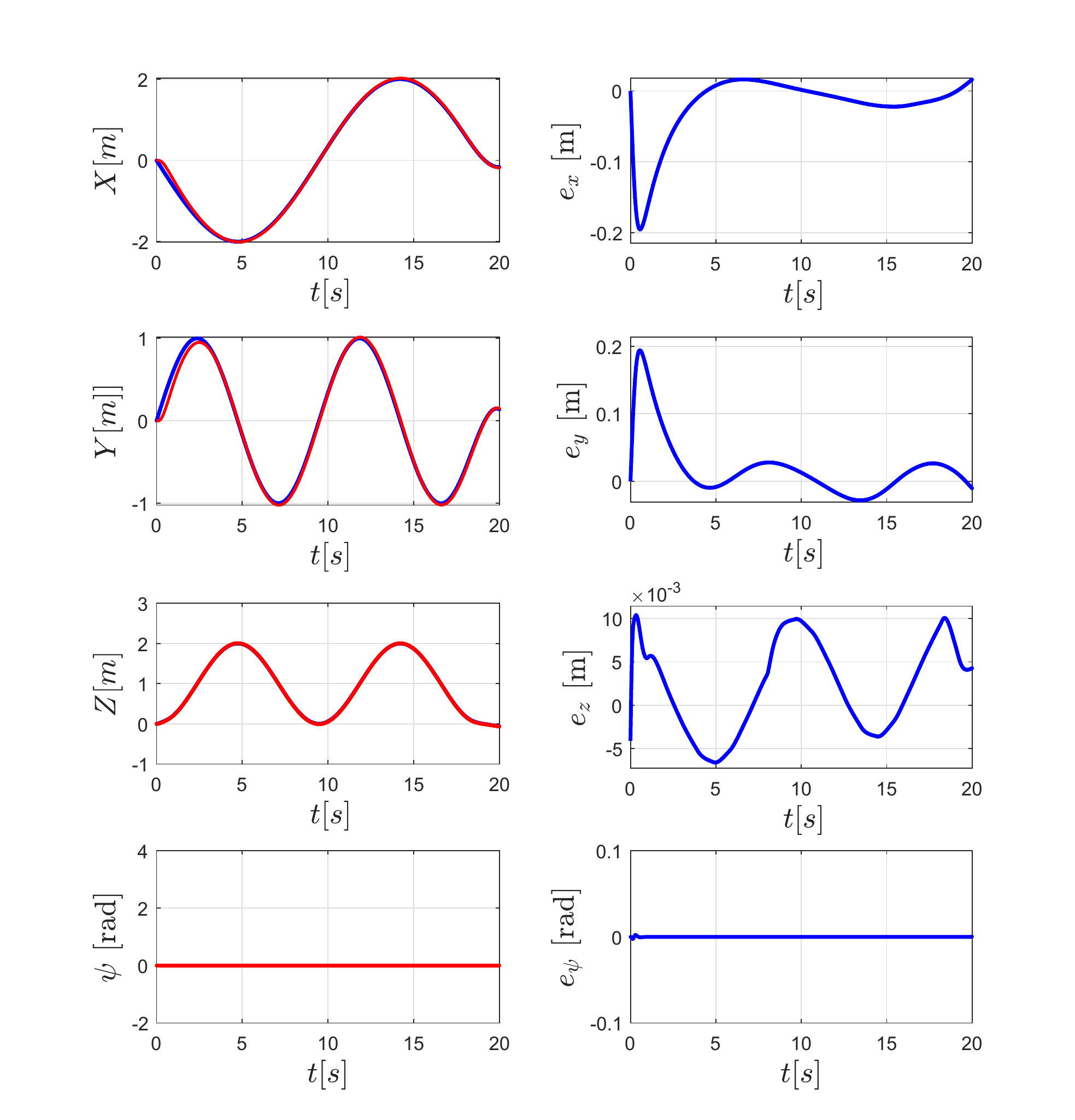}
\caption{{Tracking of Cartesian coordinates  $x$, $y,$ $z$ and orientation $\psi$
and respective tracking errors $e_{x},$ $e_{y}$, $e_{z}$ and $e_{\psi}$ in case of trajectory generated by $RSP$ planner. The traversal time is 20s, with the fault at M1 occurring at t=8s.  
\label{fig:RSP_jednostruka}}}
\end{figure}

Figs. \ref{fig:RIP_jednostruka},
\ref{fig:RCP_jednostruka} i \ref{fig:RSP_jednostruka} show the tracking results and related errors for each coordinate during the mission execution. Fig. \ref{fig:RIP_jednostruka} clearly indicates the previous conclusion that the RIP planner experiences the problems immediately after the failure occurs ($t=8s$). Namely, the octocopter needs several seconds to decrease the error with respect to $y$ coordinate after the failure occurs. Fig. \ref{fig:RCP_jednostruka} shows the results obtained by the RCP planner. As previously stressed, the octocopter experiences the worst deviations before the failure occurs due to the conservative maneuvers used during the no-failure stage. As expected, the RSP planner (see Fig. \ref{fig:RSP_jednostruka}) navigates the octocopter system to perceive the minimum deviation from the given waypoints during the whole mission duration. 

\begin{table}
\caption{{Performance comparison for RIP, RCP and RSP motion planners with traversal time T=20s (Case 1, steps 5 and 6).\label{tab:Poredjenje_planera_jednostruki}}}
\centering{}%
\begin{tabular}{|l|c|c|c|c|c|c|}
\hline 
Perf. & $e_{Rp[m]}$  & $e_{\Psi p[rad]}$  & $e_{RRef}[m]$  & $e_{\Psi Ref}[rad]$  & $e_{RRIP}[m]$  & $e_{\Psi RIP}[rad]$ \tabularnewline
\hline 
$RIP$ & 
1.75 & 0.251 & 1.29 & -2.7e-3 & 1.29 & -2.7e-3
\tabularnewline
\hline 
$RCP$ & 
1.55 & 2.56e-3 & 1.08 & 1.8e-3 & 1.41 & -1.58e-2
\tabularnewline
\hline 
$RSP$  & 1.44 & -1.22e-2 & 0.86 & -0.7e-5 & 1.01 & 2.9e-2
\tabularnewline
\hline 
\end{tabular}
\end{table}

\begin{figure}
\centering{}\includegraphics[width=7cm,height=7cm]{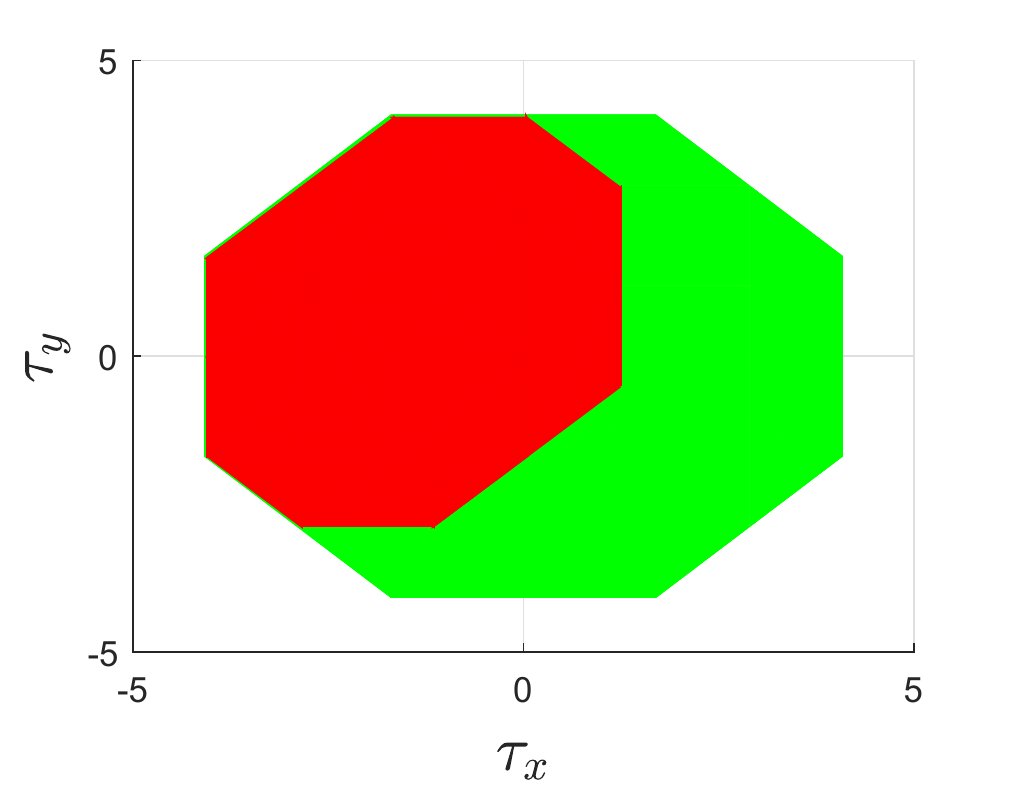}
\caption{{Control input domain (projection onto plane T = mg) in case without any faults (green), and in case of simultaneous faults at M1 and M6 (red).
\label{fig:Istovremeni_dvostruki}}}
\end{figure}

\textbf{Case 2:} Consider now a double-fault case related to the motors $M_1$ and $M_6$. Fig. \ref{fig:Istovremeni_dvostruki} shows the control admissible set for this case (red), which is reduced with respect to the admissible set obtained without considering any fault state (green). 

As expected, we can  see from Table \ref{tab:Dvostruki_planiranje}, that the RIP planner obtains the best performance. The RSP and RCP planner obtain similar performance, except that the RCP planner needs a bit more time ($T=26s$ vs. $T=18s$) to complete the mission. This is due to a more restrictive set of inequality constraints included in the optimization for the RCP planner.  However, in the case when the execution time is set to the maximum time (see Table \ref{tab:Dvostruki_planiranje}) obtained from these planners ($T=26s$), the RIP planner violates fault-dependent inequality constraints 80 times at 73 positions, while the RSP motion planner only 21 times at 21 positions. This indicates that the RSP motion planner is readier than the RIP planner in case this double fault occurs, while it needs a bit more time than the RIP planner to complete the mission.

\textbf{\medskip{}
}

\begin{table}
\caption{{Performance comparison for RIP, RCP and RSP motion planners in case of double fault, where their minimum respective traversal times are considered  (Case 2, steps 2, 3, and 5).\label{tab:Dvostruki_planiranje}}}

\centering{}%
\begin{tabular}{|l|c|c|c|c|}
\hline 
Performance & ${T}$ $[s]$ & $e_{R}[m]$  & $e_{\Psi}[rad]$  & $violated$ $constraints$ \tabularnewline
\hline 
$RIP$  & 16 & 0,0412 & 4e-19  & 80\tabularnewline
\hline 
$RCP M16$  & 26  & 0,284 & 2e-17  & 0 \tabularnewline
\hline 
$RSP$  & 18 & 0,234 & 1,2e-9 & 21\tabularnewline
\hline 
\end{tabular}
\end{table}

\begin{figure}
\centering{}\includegraphics[scale=0.55]{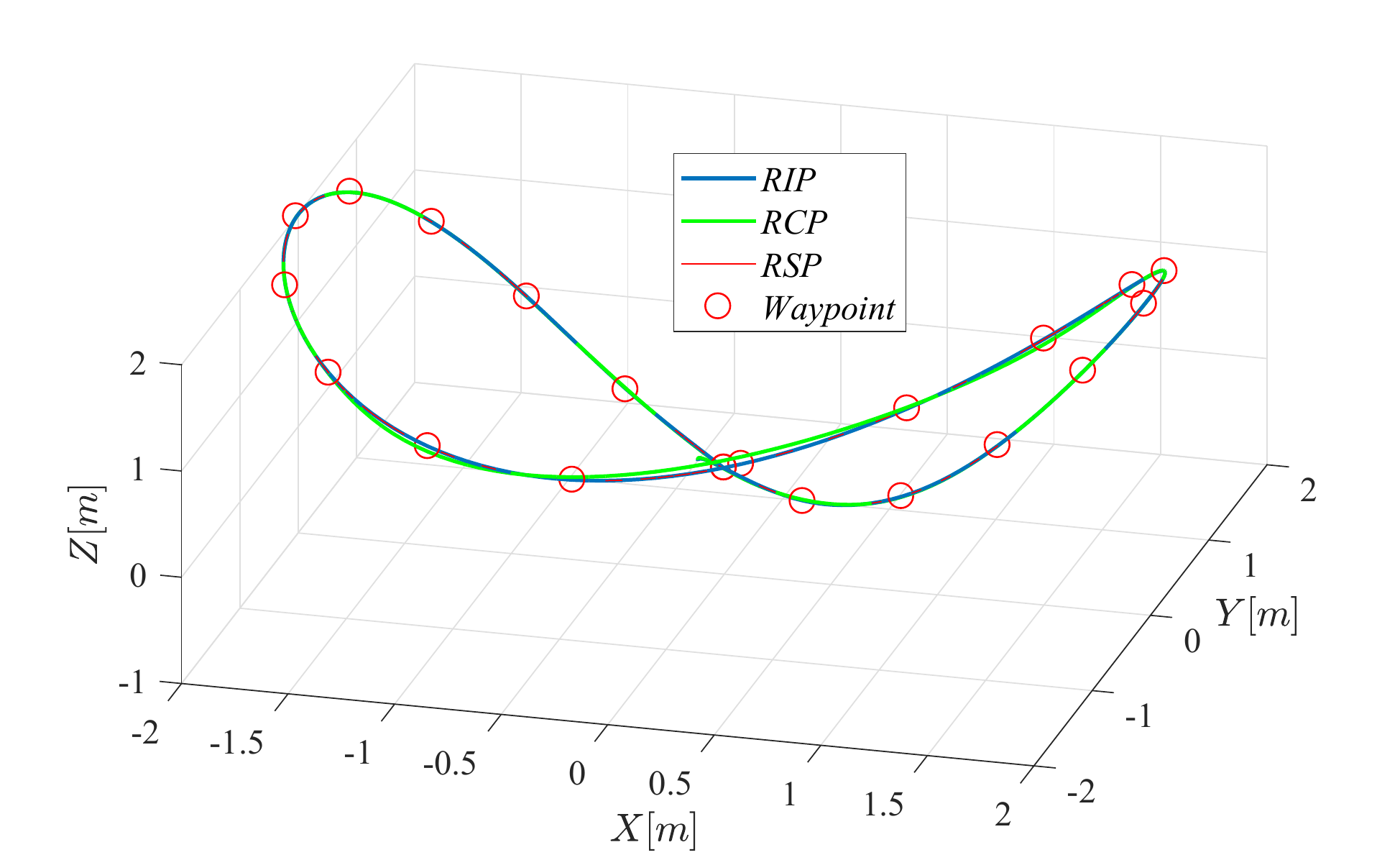}
\caption{{Paths generated using RIP (without fault states), RCP (faults at M1 and M6), and RSP (faults at M1 and M6 -- selected inequalities) motion planners. The traversal time is 26 seconds. 
\label{fig:Scenario_2}}}
\end{figure}

The planners are compared for the mission execution time $T=26$ $[s]$. The paths generated by three considered planners are shown in Fig. \ref{fig:Scenario_2}, from which one can see that the smallest and largest deviations from the waypoints are obtained with the RIP and RCP planners, respectively.

\begin{figure}
\centering{}\includegraphics[scale=0.5]{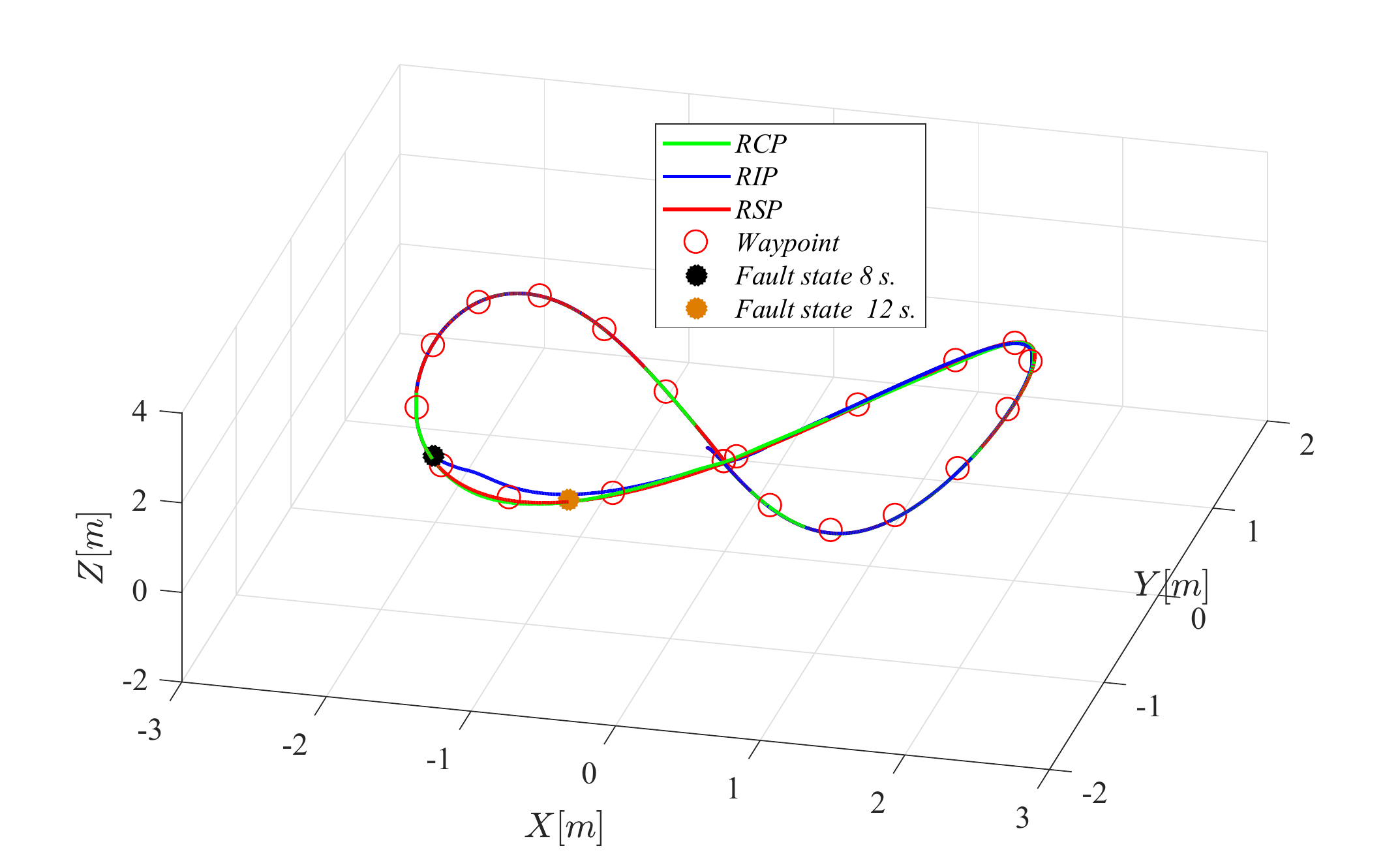}
\caption{Tracking of trajectories obtained using RIP, RCP, and RSP planners. The nominal traversal time is 26s, with the fault states at M1 and M6 occurring at t=8s and t=12s respectively.}
 
\label{fig:Scenario_2_simulacija}
\end{figure}

In order to fully compare the planners, we simulate the case when the motors $M_1$ and $M_6$ fail at $t=8s$ and $t=12s$, respectively. The obtained paths are shown in Fig. \ref{fig:Scenario_2_simulacija}, while the overall performance is summarized in Table \ref{tab:Poredjenje_planera_dvostruko}. It can be seen that the RSP planner obtains the best performance in comparison to the worst performance obtained by the RIP planner. The results are expected since the RIP planner does not take any information about motor failures into account during the planning stage. 

\begin{table}
\caption{{Performance comparison for RIP, RCP and RSP motion planners in case of simultaneous fault at M1 and M6 (Case 2, steps 5 and 6). Traversal time is T=26s.
\label{tab:Poredjenje_planera_dvostruko}}}

\centering{}
\small
\begin{tabular}[!ht]{|l|c|c|c|c|c|c|}
\hline 
Perf. & $e_{Rp}[m]$  & $e_{\Psi p}[rad]$  & $e_{RRef}[m]$  & $e_{\Psi Ref}[rad]$  & $e_{RRIP}[m]$  & $e_{\Psi RIP}[rad]$ \tabularnewline
\hline 
$RIP$  & 1.32 & 5.6e-3 & 1.04 & -3.8e-3 & 1.08 & -3.9e-3\tabularnewline
\hline 
$RCP$  & 1.15 & 2.79e-3 & 0.72 & -7.9e-4 & 0.85 & -8.21e-4\tabularnewline
\hline 
$RSP$   & 1.01 & 2.38e-2 & 0.65 & -2.9e-5 & 0.67 & -3e-5\tabularnewline
\hline 
\end{tabular}
\end{table}

\begin{figure}
\centering{}\includegraphics[scale=0.5]{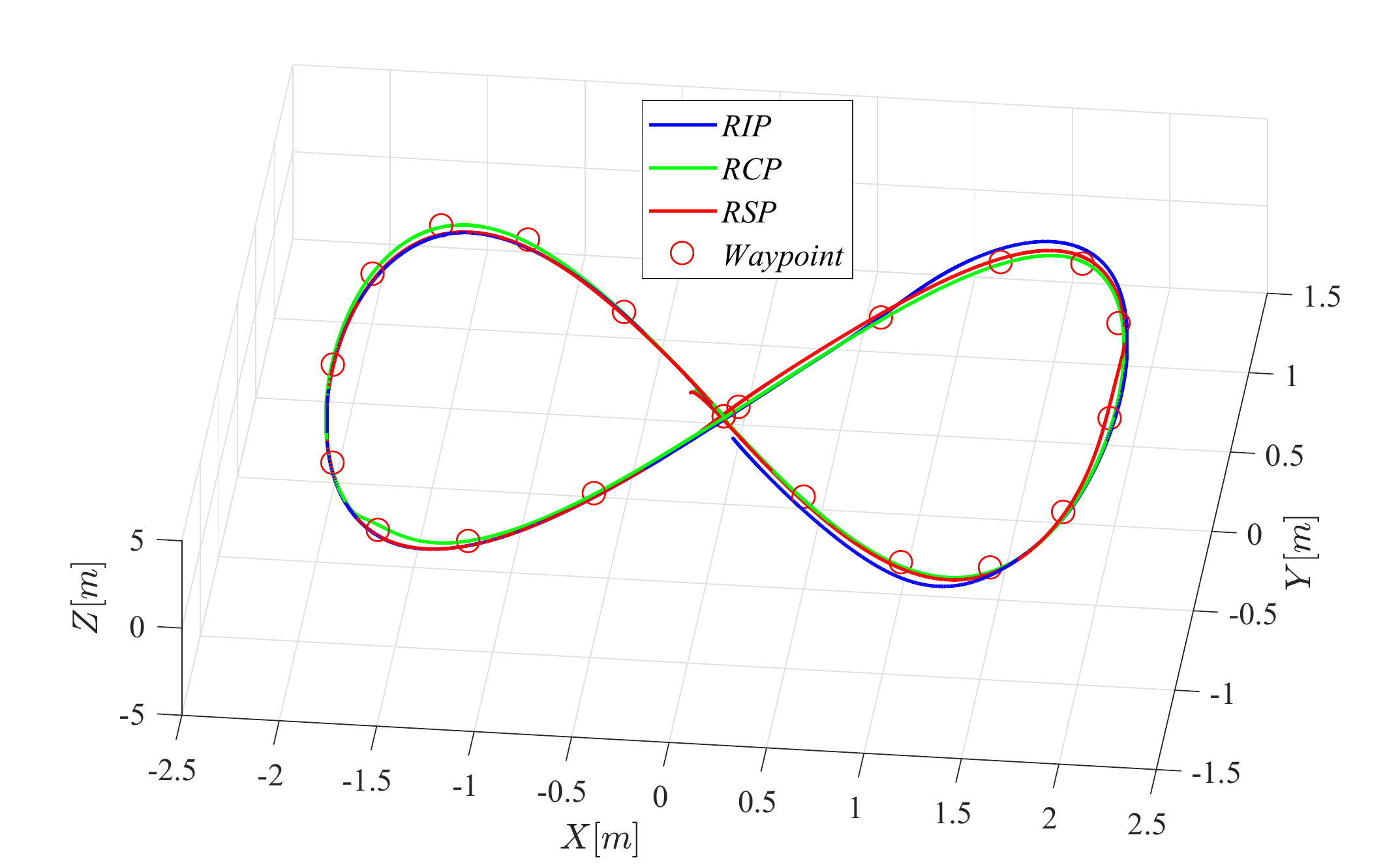}
\caption{{Tracking of trajectories obtained using RIP, RCP, and RSP planners. The traversal times are 16s, 26s and 18s respectively. The fault states at M1 and M6 occur at t=8s and t=12s respectively.
\label{fig:Scenario_2_simulacija_raz_vremena}}}
\end{figure}

It is also interesting to test the results such that the mission execution times for each planner are the same as in case when the considered fault does not occur, that is $T_{RIP}=16s$, $T_{RCP}=26s$ and $T_{RSP}=16s$. The paths obtained by these three planners are shown in Fig. \ref{fig:Scenario_2_simulacija_raz_vremena}, while the performance is summarized in Table \ref{tab:Poredjenje_planera_dvostruko_raz_vremena}. One can derive the same conclusion and see that the RIP planner generates the worst deviation from the mission waypoints, while the RCP and RSP have the similar performance. However, the execution time of the RSP is much smaller than in case of the RCP planner. This means that the RSP generates the paths with good perfromance with mission execution times close to the one obtained with the RIP planner.  

\begin{table}
\caption{{Performance comparison for RIP, RCP and RSP motion planners in case of simultaneous fault at M1 and M6 (Case 2, steps 5 and 6). Traversal times are T=16s, 26s, and 18s respectively.
\label{tab:Poredjenje_planera_dvostruko_raz_vremena}}}

\centering{}%
\begin{tabular}{|l|c|c|c|c|}
\hline 
Performance & $e_{Rp}[m]$  & $e_{\Psi p}[rad]$  & $e_{RRef}[m]$  & $e_{\Psi Ref}[rad]$ \tabularnewline
\hline 
$RIP$  & 1.74 & -7.65e-3 & 1.43 & -7.8e-3\tabularnewline
\hline 
$RCP$  & 1.26 & -7.65e-3 & 0.71 & -1.3e-3\tabularnewline
\hline 
$RSP$   & 1.32 & -7.6e-3 & 1.02 & -2.8e-5\tabularnewline
\hline 
\end{tabular}
\end{table}

The similar analysis can be further conducted for different multiple-faults. However, every additional fault would substantially reduce the control admissible set which would further deteriorate the results of the RCP planner. 

\textbf{\medskip{}
}

\textbf{Case 3:} In this case we address all eight possible single faults for an octocopter system. Table \ref{tab:Planer_svi_jednostruki} shows the results obtained by the RIP, RSP as well as different variants of the RCP planner related to different single-faults, that is the RCP-$M_i$, $i = \overline{1..8}$. The RCP-$M_i$ planner takes into account only the admissible set related to the fault of the motor $M_i$, meaning that the planner is conservatively prepared only for that fault. It should be noted that the RCP planner is not able to take all 8 single faults simultaneosly into account during the planning stage since the final admissible set would be an empty set. However, the RSP planner is capable to address all $M_i$ single faults simultaneously. As previously explained, this is possible since the RSP planner takes only a few inequality constraints for the admissible sets related to all $M_i$ faults, in order to form the optimization framework. For this reason, the RSP is capable to provide a feasible solution, unlike the RCP planner. 

One can observe from Table \ref{tab:Planer_svi_jednostruki} that the RSP planner needs more time ($T=28s$) to complete the mission in a satisfactory manner. This is due to the fact that it is the only planner that takes all 8 single-faults into account. However, for the mission execution time set to the maximum $T=28s$, one can see from Table \ref{tab:Narusene_nejednakosti} \cite{Osmic_automatika} that the number of violated inequalities related to all single-fault admissible sets ($M_i$) was significantly smaller for the RSP motion planner with respect to other planners. As expected, the RIP planner violates the largest number of those constraints which makes it unprepared for any single-fault occurrence during the mission executions.  
It is worth mentioning that the constraints of the admissible set related to the motor $M_3$ are violated in a huge number except by the RCP-$M_3$ planner that takes into account those constraints in the planning stage. This is probably due to the selected mission which requires such maneuvers sensitive to those constraints. An additional interesting observation regarding the admissible sets related to the motors $M_2$ and $M_5$ is that all planners have managed to satisfy all related constraints during the whole mission. Finally, as expected, all RCP-$M_i$ planners satisfy all constraints related to their own admissible sets $M_i$. 

\begin{table}
\caption{{Performance comparison between the RIP, RCP and RSP planners (Case 3. step $1$, $2$ and $3$).\label{tab:Planer_svi_jednostruki}}}

\centering{}%
\begin{tabular}{|l|c|c|c|c|c|}
\hline 
Performance  & ${T}$ $[s]$ & $e_{R}[m]$  & $e_{R}/N[m]$  & $e_{\Psi}[rad]$  & $e_{\Psi}/N[rad]$ \tabularnewline
\hline 
$RIP$  & 16  & 0,0412  & 0,0019  & 4e-12  & 4e-13 \tabularnewline
\hline 
$RCP$ ($M_1$)   & 20  & 0,057  & 0,0027  & 1e-18  & 5e-21 \tabularnewline
\hline 
$RCP$ ($M_2$)   & 20  & 0,038  & 0,018  & 4e-18  & 2e-19 \tabularnewline
\hline 
$RCP$ ($M_3$)   & 20  & 0,038  & 0,018  & 7e-12  & 3,4e-13 \tabularnewline
\hline 
$RCP$ ($M_4$)   & 28  & 0,5  & 0,0238  & 1,4e-8  & 6,9e-10 \tabularnewline
\hline 
$RCP$ ($M_5$)   & 24  & 0,26  & 0,012  & 0,0012  & 5,9e-4 \tabularnewline
\hline 
$RCP$ ($M_6$)   & 24  & 0,43  & 0,021  & 9e-12  & 4,4e-13 \tabularnewline
\hline 
$RCP$ ($M_7$)   & 20  & 0,073  & 0,0035  & 1,5e-10  & 7,4e-12 \tabularnewline
\hline 
$RCP$ ($M_8$)   & 20  & 0,038  & 0,018  & 7e-12  & 3,4e-13 \tabularnewline
\hline 
$RSP$   & 28  & 0,54  & 0,025  & 7e-6  & 3,5e-7 \tabularnewline
\hline 
\end{tabular}
\end{table}

\begin{table}
\caption{{The number of unsatisfied inequalities for RIP, RCP and RSP planners for each possible single fault state $(M_i)$, with respect to feasible control inputs (Case 3, steps 4,5, and 6).\label{tab:Narusene_nejednakosti}}}
\centering{}%
\begin{tabular}{|l|c|c|c|c|c|c|c|c|}
\hline 
MAV  & $M_1$  & $M_2$  & $M_3$  & $M_4$  & $M_5$  & $M_6$  & $M_7$  & $M_8$ \tabularnewline
\hline 
$RIP$  & 52  & 0  & 108  & 206  & 0  & 108  & 24  & 136 \tabularnewline
\hline 
$RCP$ ($M_1$)   & 0  & 0  & 107  & 179  & 0  & 89  & 20  & 116 \tabularnewline
\hline 
$RCP$ ($M_2$)   & 52  & 0  & 110  & 205  & 0  & 108  & 24  & 136 \tabularnewline
\hline 
$RCP$ ($M_3$)   & 53  & 0  & 0  & 185  & 0  & 87  & 23  & 137 \tabularnewline
\hline 
$RCP$ ($M_4$)   & 52  & 0  & 99  & 0  & 0  & 99  & 24  & 136 \tabularnewline
\hline 
$RCP$ ($M_5$)   & 52  & 0  & 101  & 197  & 0  & 99  & 24  & 136 \tabularnewline
\hline 
$RCP$ ($M_6$)   & 52  & 0  & 101  & 197  & 0  & 0  & 24  & 136 \tabularnewline
\hline 
$RCP$ ($M_7$)   & 52  & 0  & 101  & 191  & 0  & 95  & 0  & 134 \tabularnewline
\hline 
$RCP$ ($M_8$)   & 52  & 0  & 108  & 197  & 0  & 108  & 24  & 0 \tabularnewline
\hline 
$RSP$   & 0  & 0  & 118  & 15  & 0  & 5  & 0  & 0 \tabularnewline
\hline 
\end{tabular}
\end{table}

As in Case 1 and Case 2, the results are compared using the mission execution time obtained with step 3 of the RSP planning algorithm which is now set to $T=28${ $[s]$}. We have conducted 30 single-fault simulations for each planner, where a single fault has been randomly generated at a random time moment. The statistical results obtained for all previously considered errors are summarized in Table \ref{tab:Metrika_simulacija}.

\begin{table}
\caption{{Feasibility comparison for trajectories obtained using RIP, RCP and RSP planners, for each possible single fault state $(M_i)$. The time instant of fault occurrence is random.\label{tab:Metrika_simulacija}}}
\centering{}%
\begin{tabular}{|l|c|c|c|}
\hline 

 & $RIP$   & $RCP$   & $RSP$ \tabularnewline
\hline 
Mean value ($e_{Rp}[m]$ ) & 1,21 & 1,3 & 1,105\tabularnewline
\hline 
Standard deviation ($e_{Rp}[m]$) & 7,44e-2 & 1,86e-1 & 3,98e-2\tabularnewline
\hline 
Mean value ($e_{\Psi p}[rad]$ ) & -2,28e-4 & 3,26e-2 & 4,42e-3\tabularnewline
\hline 
Standard deviationa ($e_{\Psi p}[rad]$ ) & 1,63e-3 & 4,44e-2 & 8,93e-4
\tabularnewline
\hline 
Mean value ($e_{RRef}[m]$) & 0,738 & 0,743 & 0,608\tabularnewline
\hline 
Standard deviation ($e_{RRef}[m]$) & 0,058 & 0,145 & 0,0506\tabularnewline
\hline 
Mean value ($e_{\Psi Ref}[rad]$)  & 8,133e-5 & -3,74e-3 & -9,93e-5\tabularnewline
\hline 
Standard deviation ($e_{\Psi Ref}[rad]$) & 1,22e-3 & 5,43e-3 & 6,48e-4\tabularnewline
\hline 
Mean value($e_{RRIP}[m]$)  & 0,738 & 0,937 & 0,635\tabularnewline
\hline 
Standard deviationa ($e_{RRIP}[m]$)  & 0,058 & 0,257 & 0,056\tabularnewline
\hline 
Mean value($e_{\Psi RIP}[rad]$ ) & 8,133e-5 & -1,04e-2 & -2,82e-3\tabularnewline
\hline 
Standard deviation ($e_{\Psi RIP}[rad]$ ) & 1,22e-3 & 1,69e-2 & 6,48e-4\tabularnewline
\hline 
\end{tabular}
\end{table}
As expected, the RCP which was customized only to one particular single-fault, generates the worst statistical results, since the simulated single-failures could be linked to any motor. Moreover, two simulations based on the RCP planner were terminated since the octocopter became uncontrollable immediately after the fault occurred. One can also observe that the best statistical results have been obtained by the RSP planner, since it was inherently constructed to be prepared for any possible single fault.

\hspace{0.5cm}

\begin{figure}
\centering{}\includegraphics[scale=0.45]{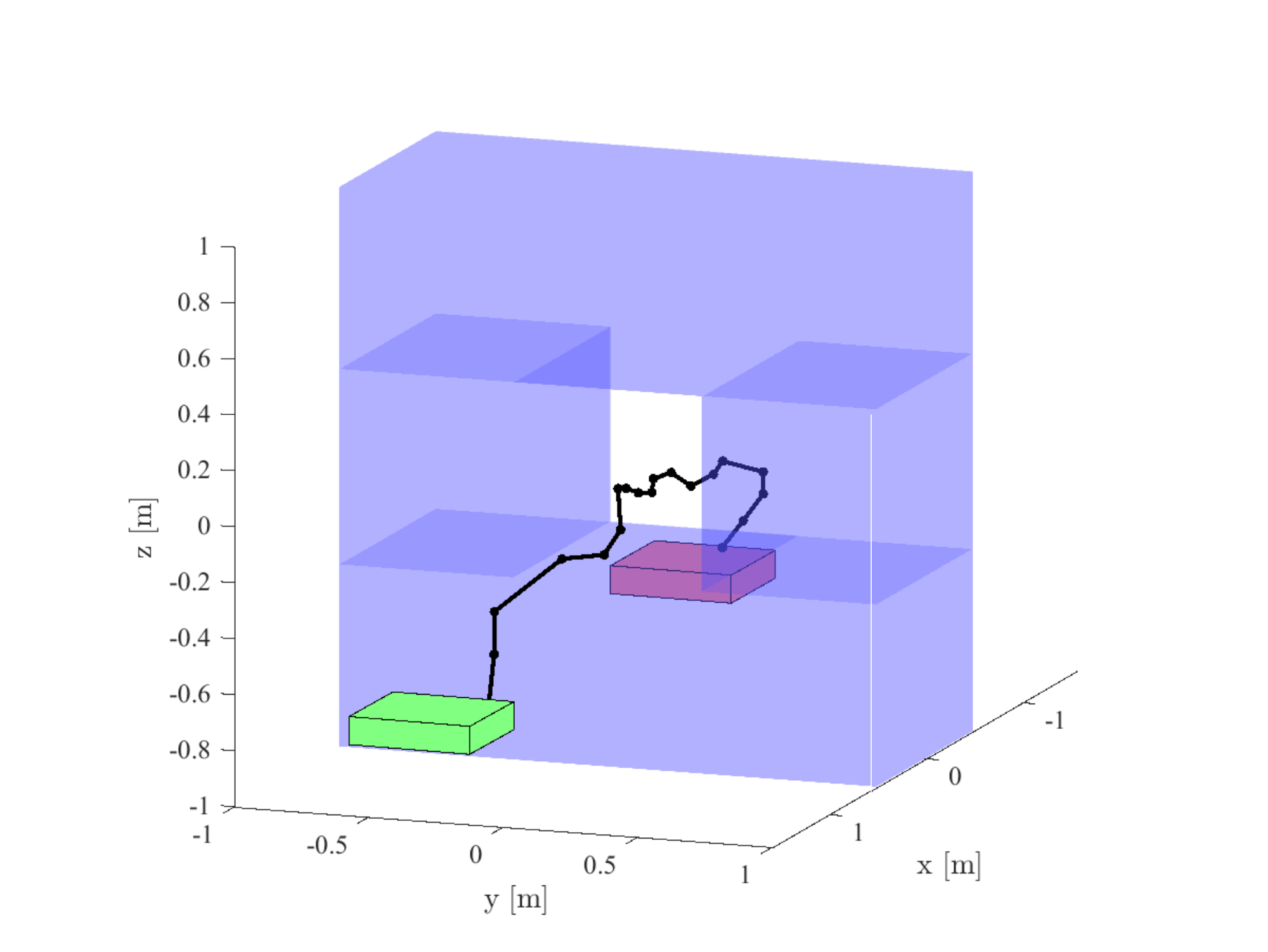}
\caption{{Octocopter motion through a narrow corridor (dimensions 1m x 1m x 1m). RRT algorithm is used for generating waypoints..
\label{fig:Koridor}}}
\end{figure}

\textbf{Case 4:} In the following set of scenarios, the octocopter has to pass through a narrow corridor, which is 1m long, with the square cross-section of dimensions 1m x 1m. The dimensions of the craft are 0,5m x 0,5m x 0,2m. For computing the collision free path, the RRT planning algorithm is used, where the octocopter is considered as a free-flying rigid body. RRT has generated 20 intermediate waypoints, which serve as milestones for the octocopter motion. Figure \ref{fig:Koridor} depicts the corridor, the initial configuration (red), the final configuration (green) and the waypoints. This setup is used to test several scenarios. The first one (Case 4.1), the planning is performed using RIP, RCP and RSP planners. In a simulation scenario, we assume, no fault state occurs. 
Following the steps of the RSP planner, the mission duration is set to 12s. For such setting, all three planners generate paths with the octocopter passing the narrow corridor without colliding with the walls. Figure (\color{blue} fali slika \color{black}) shows the diagram of minimum distance between the octocopter and the corridor walls. 

\begin{figure}
\centering{}\includegraphics[scale=0.45]{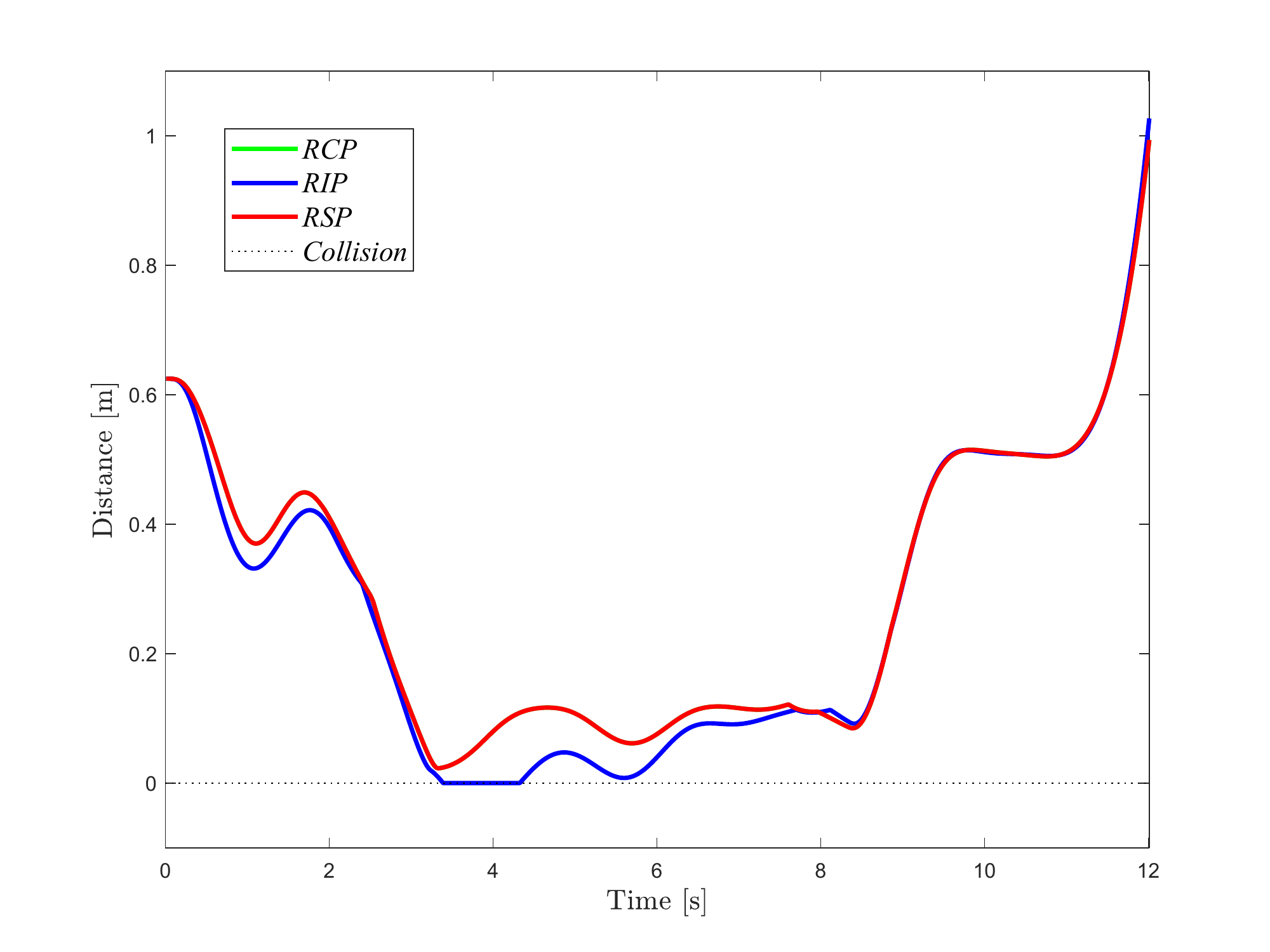}
\caption{{Octocopter motion through the corridor with a single fault state at $M_1$ at $t=3$s. RCP anticipates the fault state at $M_1$.
\label{fig:Scenario_1_koridor}}}
\end{figure}

In Case 4.2, RCP planner anticipates the fault state at motor $M_1$. In simulation, the actual fault occurs at motor $M_1$ at $t=3$s. The minimum distance diagram for all three planners is given in Figure \ref{fig:Scenario_1_koridor}. It can be noted that RCP and RSP planners enable collision-free motion, whereas the motion generated by the RIP planner causes collision of octocopter with the corridor wall. The fault is generated at the time when the octocopter enters the corridor and likely represents the most sensitive region within the whole mission. This can also be noted from the motion along the path generated by RCP and RSP planners, which are in this case identical. Though collision-free, the considered path comes rather close to the wall. When the fault occurs at the motor $M_1$ ($t=3$s), the maneuver performed by octocopter, according to the path planned by RIP, clearly does not compensate for the fault state. A significant deviation from the reference path occurs and the octocopter collides with the wall. On the other hand, RCP planner accounts for the fault state at $M_1$. Therefore, when the specific fault occurs, it does not cause the deviation from the nominal path computed by RCP. A similar behavior is observed when the octocopter follows the path obtained by the RSP planner.

\begin{figure}
\centering{}\includegraphics[scale=0.45]{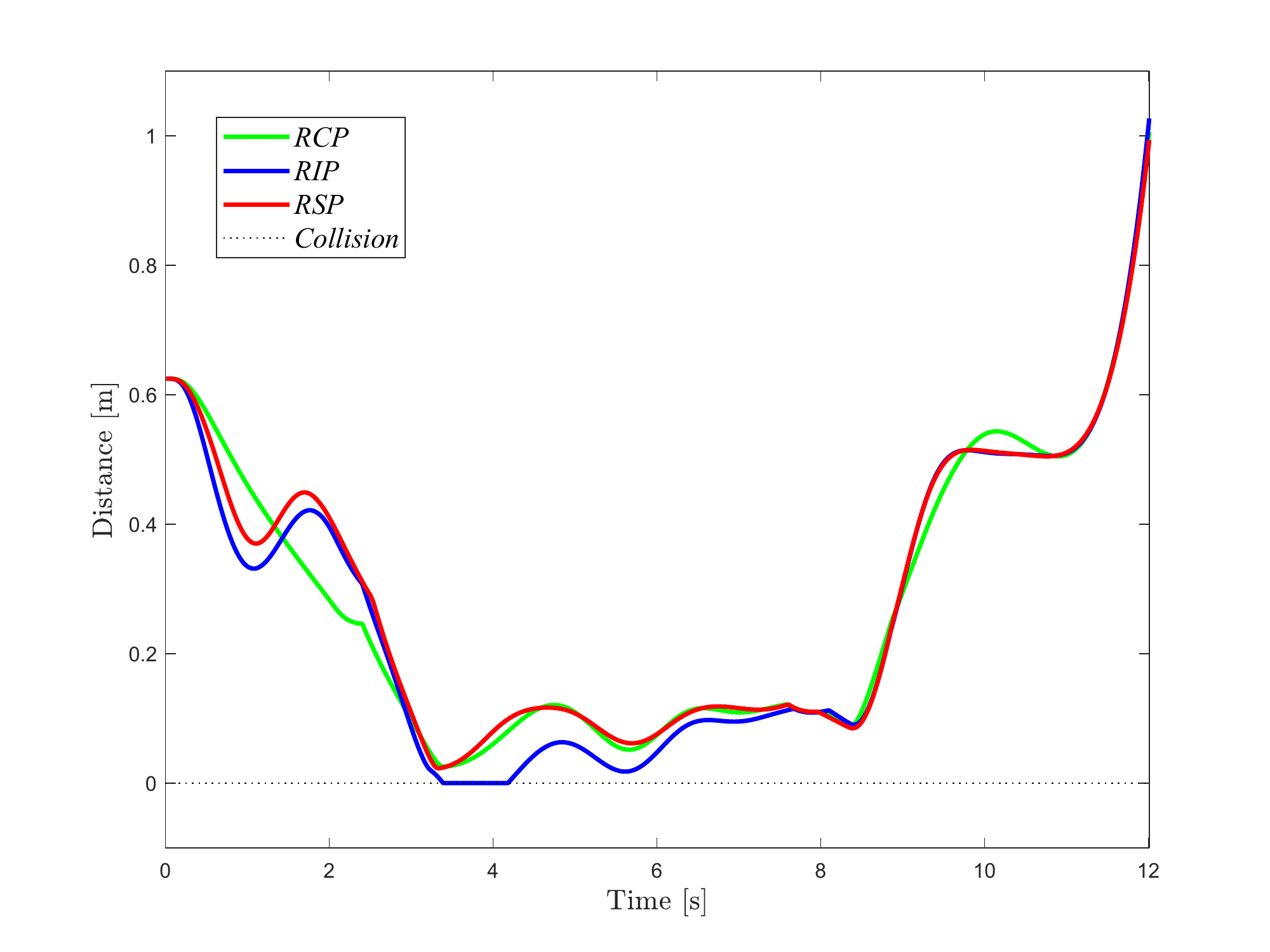}
\caption{{Octocopter motion through the corridor with a single fault state at $M_1$ at $t=3$s. RCP anticipates the fault state at $M_8$.
\label{fig:Scenario_2_koridor}}}
\end{figure}

In scenario 4.3, the fault state occurs again at $t=3$s at motor $M_1$, however, RCP planner anticipates the fault state at motor $M_8$. Similarly, as in the previous case, octocopter collides with the wall in case of following the path generated by the RIP planner. On the other hand, the motions generated by RCP and RSP planners are collision-free (see Figure \ref{fig:Scenario_2_koridor}).  The reason collision does not occur for the RCP-planned motion may be the fact that the control region implied by the possible fault state at $M_8$ prevents aggressive maneuvers near the entrance of the corridors. Thus, the octocopter may adapt to the actual fault at $M_1$, though the fault at this motor has not been anticipated.  

\begin{figure}
\centering{}\includegraphics[scale=0.45]{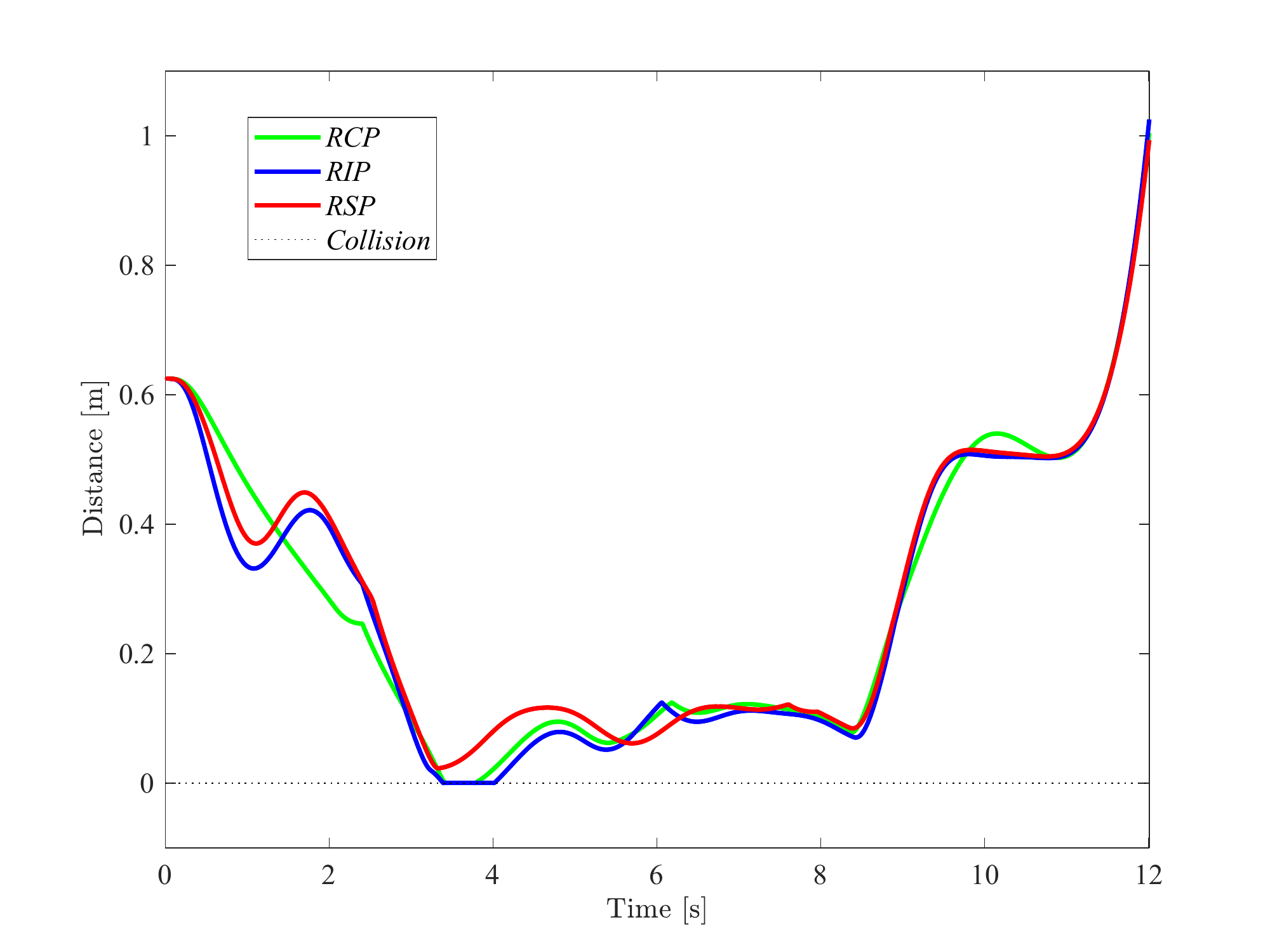}
\caption{{Octocopter motion through the corridor with a double fault states at $M_1$ and $M_4$ at $t=3$s and $t=5$s respectively. RCP anticipates the fault state at $M_8$.
\label{fig:Scenario_3_koridor}}}
\end{figure}

In case 4.4, a double fault occurs, at motors $M_1$ and $M_4$, at $t=3$s and $t=5$s respectively. Only RSP planner generates the collision-free motion, while the paths computed by RIP and RCP are not feasible. The minimum distance diagram is shown in Figure \ref{fig:Scenario_3_koridor}.

\begin{figure}
\centering{}\includegraphics[scale=0.45]{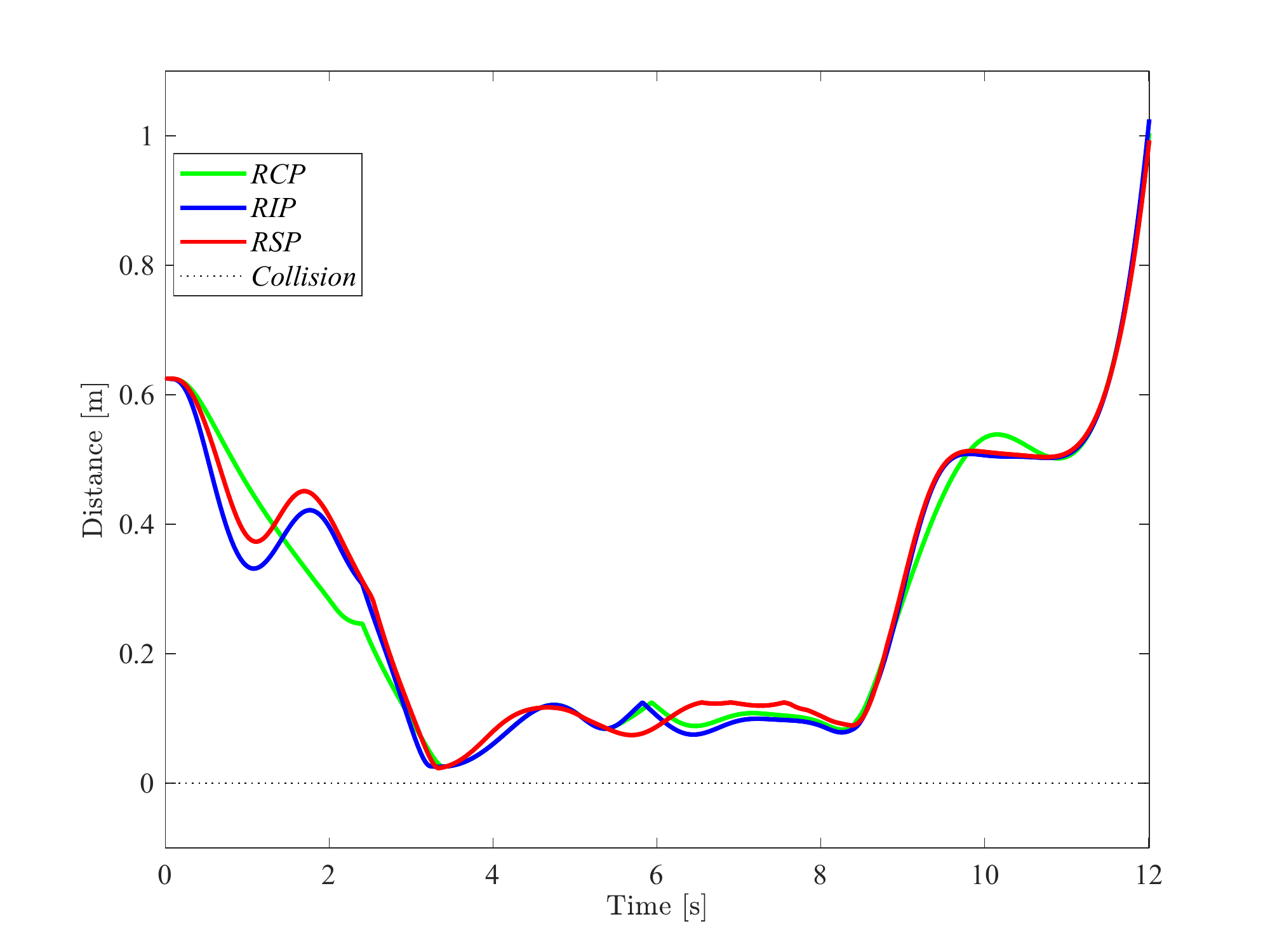}
\caption{{Octocopter motion through the corridor with a double fault states at $M_1$ and $M_4$ at $t=5$s and $t=8$s respectively. RCP anticipates the fault state at $M_8$.
\label{fig:Scenario_4_koridor}}}
\end{figure}

The case 4.5.  assumes fault states at motors $M_1$ and $M_4$, at $t=5$s and $t=8$s respectively. This time, all three planners generate collision-free motions. This is likely due to the fact that at $t=5$s, the octocopter is already inside the corridor, with ample clearance margins. Hence, even when the fault occurs, there is sufficient room for adjustment without causing collisions. The minimum distance diagram is shown in Figure \ref{fig:Scenario_4_koridor}.

Based on presented scenarios for motion planning through a narrow corridor, it can be seen that the RSP planner shows the best performance in cases of single and double fault states. For cases 4.1 and 4.5, all three planners show reliable performance. However, since we cannot predict the time of fault occurrence with certainty, or the maneuver performed by the craft at the time, the RSP-generated motion turns out to be most reliable. Clearly, this approach encourages those octocopter maneuvers that are the most beneficial to recovering the path following.

\chapter{Conclusion}

This book includes all necessary ingredients to design control and motion planning algorithms for an autonomous octocopter system. 
Chapter 2 describes motion principles of an octocopter design and includes the octocopter kinematics and dynamics equations as well as motor dynamic models. It also provides a general octocopter state-space model upon which it is possible to construct a variety of control design algorithms. 

Chapter 3 provides a fault-tolerant control, control allocation and a simple PD controller that controls the octocopter system to track the given reference position and orientation. In order to include information about potentially active fault states of DC motors into a control design, it is necessary to implement an algorithm that can include those information into the actuation matrix and adjust the control so that the octocopter system is still optimally controlled during such circumstances. The presented algorithm which is capable of identifying and isolating the active failures is based on the recursive least-squares algorithm. The information of the fault state identification is then fed to the control allocation algorithm which adapts the distribution of control signals only to the remaining active motors in order to achieve a feasible control whenever is possible. It is shown that the octocopter system based on such a control architecture is capable of achieving good performance in case all eight DC motors of the octocopter system are fully available. However, in case of a single motor failure, the control may be feasible but the tracking performance could be significantly deteriorated.  

Chapter 4 introduces a fault-dependant controllability analysis which thoroughly examines the potential of an octocopter system to continue the mission execution under variety of possible fault states. The analysis considers single-fault, double faults, and the effects those faults may have on the system behaviour, depending on the distribution of the motor rotational directions. The analysis shows that a careful selection of an octocopter configuration structure may additionally influence the overall maneuverability of the system and it can make the octocopter more fault-tolerant to variety of fault states. For instance, in case the probability of a double motor fault state is high, it is possible to select an octocopter based on the PPNNPPNN configuration structure to increase reliability of the system. Other multiple faults, including triple and quadruple faults, can be analyzed under the same framework as well. However, the occurrence probabilities of such faults are much lower, so these faults are not considered in the book.

Chapter 5 provides a full algorithm to construct a risk-sensitive motion planner (RSP). The RSP is a risk-aware planner which is capable of including the relevant information about mission-dependant constraints which are related to potential fault states and imposed to the maneuverability of an octocopter system into the planning stage. In this way, the reference trajectories which are designed to follow a given sequence of mission waypoints are more appropriate in case a motor failure occurs during the mission execution than in case when this information are ignored as in case of a risk-insensitive motion planner (RIP). The obtained results of the RSP are compared to the results obtained with the RIP and the risk-conservative planner (RCP). Unlike the RSP planner, the RCP takes full information about fault states regardless of their relevance to the selected mission. The results related to any single- and double motor faults show that the RSP planner outperforms the RIP and RCP planners in case any of those faults occur during the mission execution, while it preserves good performance of the RIP and safety of the RCP approach.

\bibliographystyle{plain}
\bibliography{bibtex_example}

\begin{thebibliography}{10}

\bibitem{basson_pseudo}
Lionel Basson.
\newblock {\em Control allocation as part of a fault-tolerant control
  architecture for {{UAV} }s}.
\newblock PhD thesis, Stellenbosch: University of Stellenbosch, 2011.

\bibitem{Berbra}
C~Berbra, S~Lesecq, and JJ~Martinez.
\newblock A multi-observer switching strategy for fault-tolerant control of a
  quadrotor helicopter.
\newblock In {\em Control and Automation, 2008 16th Mediterranean Conference
  on}, pages 1094--1099. IEEE, 2008.

\bibitem{Istorija_UAV_blom2010}
John~David Blom.
\newblock {\em Unmanned Aerial Systems: A Historical Perspective}, volume~45.
\newblock Combat Studies Institute Press, 2010.

\bibitem{Samir_B.}
Samir Bouabdallah.
\newblock {\em Design and control of Quad rotors with Application to Autonomous
  flying.}
\newblock PhD thesis, PhDM thesis, Ecole Polytechnique Federale De Lausanne,
  Laboratoire de systemes autonomes 1, Section De microtechnique, 2007.

\bibitem{bouabdallah2004design}
Samir Bouabdallah, Pierpaolo Murrieri, and Roland Siegwart.
\newblock Design and control of an indoor micro quadrotor.
\newblock In {\em IEEE International Conference on Robotics and Automation,
  2004. Proceedings. ICRA'04. 2004}, volume~5, pages 4393--4398. IEEE, 2004.

\bibitem{bouabdallah2005backstepping}
Samir Bouabdallah and Roland Siegwart.
\newblock Backstepping and sliding-mode techniques applied to an indoor micro
  quadrotor.
\newblock In {\em Proceedings of the 2005 IEEE international conference on
  robotics and automation}, pages 2247--2252. IEEE, 2005.

\bibitem{Control_Allocation_1_casavola2010fault}
Alessandro Casavola and Emanuele Garone.
\newblock Fault-tolerant adaptive control allocation schemes for overactuated
  systems.
\newblock {\em International journal of robust and nonlinear control},
  20(17):1958--1980, 2010.

\bibitem{Inspekcija_hrane_chang2013improvement}
Hsiang-Kuan Chang, Yih-Chi Tan, Jihn-Sung Lai, Tsung-Yi Pan, Tzu-Ming Liu, and
  Ching-Pin Tung.
\newblock Improvement of a drainage system for flood management with assessment
  of the potential effects of climate change.
\newblock {\em Hydrological Sciences Journal}, 58(8):1581--1597, 2013.

\bibitem{Termalna_inspekcija_dios2006automatic}
Martinez-De Dios, A~Ollero, et~al.
\newblock Automatic detection of windows thermal heat losses in buildings using
  {UAV} s.
\newblock In {\em Automation Congress, 2006. WAC'06. World}, pages 1--6. IEEE,
  2006.

\bibitem{NAdgledanje_2_doherty2007uav}
Patrick Doherty and Piotr Rudol.
\newblock A uav search and rescue scenario with human body detection and
  geolocalization.
\newblock In {\em AI 2007: Advances in Artificial Intelligence}, pages 1--13.
  Springer, 2007.

\bibitem{Rekontigurabilni_FTC_drozeski2005fault}
Graham~R Drozeski, Bhaskar Saha, and George~J Vachtsevanos.
\newblock A fault detection and reconfigurable control architecture for
  unmanned aerial vehicles.
\newblock In {\em Aerospace Conference, 2005 IEEE}, pages 1--9. IEEE, 2005.

\bibitem{Korejci_controllability_II}
Guang-Xun Du, Quan Quan, and Kai-Yuan Cai.
\newblock Controllability analysis and degraded control for a class of
  hexacopters subject to rotor failures.
\newblock {\em Journal of Intelligent \& Robotic Systems}, 78(1):143--157,
  2015.

\bibitem{Korejci_controllability_I}
Guangxun Du, Quan Quan, Binxian Yang, and K~Cai.
\newblock Controllability analysis for a class of multirotors subject to rotor
  failure/wear.
\newblock {\em Comput. Res. Repository (CoRR)}, 2014.

\bibitem{Fotografiranje_UAV_flener2013seamless}
Claude Flener, Matti Vaaja, Anttoni Jaakkola, Anssi Krooks, Harri Kaartinen,
  Antero Kukko, Elina Kasvi, Hannu Hyypp{\"a}, Juha Hyypp{\"a}, and Petteri
  Alho.
\newblock Seamless mapping of river channels at high resolution using mobile
  lidar and {UAV} -photography.
\newblock {\em Remote Sensing}, 5(12):6382--6407, 2013.

\bibitem{8}
Claude Flener, Matti Vaaja, Anttoni Jaakkola, Anssi Krooks, Harri Kaartinen,
  Antero Kukko, Elina Kasvi, Hannu Hyypp{\"a}, Juha Hyypp{\"a}, and Petteri
  Alho.
\newblock Seamless mapping of river channels at high resolution using mobile
  lidar and {UAV} -photography.
\newblock {\em Remote Sensing}, 5(12):6382--6407, 2013.

\bibitem{Mapiranje_fraundorfer2012vision}
Friedrich Fraundorfer, Lionel Heng, Dominik Honegger, Gim~Hee Lee, Lorenz
  Meier, Petri Tanskanen, and Marc Pollefeys.
\newblock Vision-based autonomous mapping and exploration using a quadrotor
  mav.
\newblock In {\em Intelligent Robots and Systems (IROS), 2012 IEEE/RSJ
  International Conference on}, pages 4557--4564. IEEE, 2012.

\bibitem{Freddi}
Alessandro Freddi, Sauro Longhi, and Andrea Monteri{\`u}.
\newblock A diagnostic thau observer for a class of unmanned vehicles.
\newblock {\em Journal of Intelligent \& Robotic Systems}, 67(1):61--73, 2012.

\bibitem{Nadgledanje_freed2004human}
Michael Freed, Robert Harris, and M~Shafto.
\newblock Human-interaction challenges in {UAV} -based autonomous surveillance.
\newblock In {\em Proceedings of the 2004 Spring Symposium on Interactions
  Between Humans and Autonomous Systems Over Extended Operations}, 2004.

\bibitem{Euler_garcia2006modelling}
Pedro~Castillo Garcia, Rogelio Lozano, and Alejandro~Enrique Dzul.
\newblock {\em Modelling and control of mini-flying machines}.
\newblock Springer Science \& Business Media, 2006.

\bibitem{NAdgledanje_granica_girard2004border}
Anouck~R Girard, Adam~S Howell, and J~Karl Hedrick.
\newblock Border patrol and surveillance missions using multiple unmanned air
  vehicles.
\newblock In {\em Decision and Control, 2004. CDC. 43rd IEEE Conference on},
  volume~1, pages 620--625. IEEE, 2004.

\bibitem{Backsteping_2_glad2000flight}
Torkel Glad and Ola H{\"a}rkeg{\aa}rd.
\newblock Flight control design using backstepping.
\newblock {\em Linkoping University Electronic press}, 2000.

\bibitem{Inspekcija_EE_Mreze_golightly2005visual}
Ian Golightly and Dewi Jones.
\newblock Visual control of an unmanned aerial vehicle for power line
  inspection.
\newblock In {\em Advanced Robotics, 2005. ICAR'05. Proceedings., 12th
  International Conference on}, pages 288--295. IEEE, 2005.

\bibitem{Morse}
Michael~A Goodrich, Bryan~S Morse, Damon Gerhardt, Joseph~L Cooper, Morgan
  Quigley, Julie~A Adams, and Curtis Humphrey.
\newblock Supporting wilderness search and rescue using a camera-equipped mini
  {UAV}.
\newblock {\em Journal of Field Robotics}, 25(1-2):89--110, 2008.

\bibitem{backstepping_1_harkegard2000backstepping}
Ola Harkegard and S~Torkel Glad.
\newblock A backstepping design for flight path angle control.
\newblock In {\em Decision and Control, 2000. Proceedings of the 39th IEEE
  Conference on}, volume~4, pages 3570--3575. IEEE, 2000.

\bibitem{vrste_UAV}
Mostafa Hassanalian and Abdessattar Abdelkefi.
\newblock Classifications, applications, and design challenges of drones: A
  review.
\newblock {\em Progress in Aerospace Sciences}, 91:99--131, 2017.

\bibitem{Klasicna_mehanika}
John~Safko Herbert~Goldstein, Charles~Poole.
\newblock {\em Classical Machanics}.
\newblock Addison Wesley, 2014.

\bibitem{hwang2009survey}
Inseok Hwang, Sungwan Kim, Youdan Kim, and Chze~Eng Seah.
\newblock A survey of fault detection, isolation, and reconfiguration methods.
\newblock {\em IEEE transactions on control systems technology},
  18(3):636--653, 2009.

\bibitem{isermann_ident}
Rolf Isermann and Marco M{\"u}nchhof.
\newblock {\em Identification of dynamic systems: an introduction with
  applications}.
\newblock Springer Science \& Business Media, 2010.

\bibitem{Izadi}
Hojjat~A Izadi, Youmin Zhang, and Brandon~W Gordon.
\newblock Fault tolerant model predictive control of quad-rotor helicopters
  with actuator fault estimation.
\newblock {\em IFAC Proceedings Volumes}, 44(1):6343--6348, 2011.

\bibitem{Adaptivno_robusni_FTC_jin2012robust}
Xiaozheng Jin, Guanghong Yang, and Li~Peng.
\newblock Robust adaptive tracking control of distributed delay systems with
  actuator and communication failures.
\newblock {\em Asian Journal of Control}, 14(5):1282--1298, 2012.

\bibitem{johansen2013control}
Tor~A Johansen and Thor~I Fossen.
\newblock Control allocation—a survey.
\newblock {\em Automatica}, 49(5):1087--1103, 2013.

\bibitem{MPC_FTC_1_khan2011fault}
Rudaba Khan, Paul Williams, Robin Hill, Cees Bil, et~al.
\newblock Fault tolerant flight control system design for {UAV} 's using
  nonlinear model predictive contro.
\newblock {\em Australian Control Conference, 10-11 November 2011, Melbourne,
  Australia}, 2011.

\bibitem{Edukacijske_platforme}
Tom{\'a}{\v{s}} Krajn{\'\i}k, Vojt{\v{e}}ch Von{\'a}sek, Daniel Fi{\v{s}}er,
  and Jan Faigl.
\newblock Ar-drone as a platform for robotic research and education.
\newblock In {\em Research and Education in Robotics-EUROBOT 2011}, pages
  172--186. Springer, 2011.

\bibitem{Osmic_AIM}
Muhamed Kuric, Bakir Lacevic, Nedim Osmic, and Adnan Tahirovic.
\newblock Rls-based fault-tolerant tracking control of multirotor aerial
  vehicles.
\newblock In {\em 2017 IEEE International Conference on Advanced Intelligent
  Mechatronics (AIM)}, pages 1148--1153. IEEE, 2017.

\bibitem{osmic_modelica}
Muhamed Kuric, Nedim Osmic, and Adnan Tahirovic.
\newblock Multirotor aerial vehicle modeling in modelica.
\newblock In {\em Proceedings of12th International Modelica Conference, Prague,
  Chezh Republic, May 15-17, 2017}, pages 373--380. Linkoping University
  Electronic Press, 2017.

\bibitem{Kumar_micro_UAV}
Alex Kushleyev, Daniel Mellinger, Caitlin Powers, and Vijay Kumar.
\newblock Towards a swarm of agile micro quadrotors.
\newblock {\em Autonomous Robots}, 35(4):287--300, 2013.

\bibitem{LQG_FTC_lemos2013actuator}
Joao~M Lemos, In{\^e}s Sampaio, Manuel Rijo, and Lu{\i}s~M Rato.
\newblock Actuator fault tolerant lqg control of a water delivery canal.
\newblock In {\em Control and Fault-Tolerant Systems (SysTol), 2013 Conference
  on}, pages 432--437. IEEE, 2013.

\bibitem{Inspekcija_EE_MREZA2_li2008knowledge}
Zhengrong Li, Yuee Liu, Ross Hayward, Jinglan Zhang, and Jinhai Cai.
\newblock Knowledge-based power line detection for {UAV} surveillance and
  inspection systems.
\newblock In {\em Image and Vision Computing New Zealand, 2008. IVCNZ 2008.
  23rd International Conference}, pages 1--6. IEEE, 2008.

\bibitem{Kumar_konstrukcije}
Quentin Lindsey, Daniel Mellinger, and Vijay Kumar.
\newblock Construction of cubic structures with quadrotor teams.
\newblock {\em Proc. Robotics: Science \& Systems VII}, 2011.

\bibitem{lorincz2021novel}
Josip Lorincz, Adnan Tahirovi{\'c}, and Biljana~Risteska Stojkoska.
\newblock A novel real-time unmanned aerial vehicles-based disaster management
  framework.
\newblock In {\em 2021 29th Telecommunications Forum (TELFOR)}, pages 1--4.
  IEEE, 2021.

\bibitem{Lunze_II}
Jan Lunze.
\newblock From fault diagnosis to reconfigurable control: A unified concept.
\newblock In {\em 2016 3rd Conference on Control and Fault-Tolerant Systems
  (SysTol)}, pages 413--421. IEEE, 2016.

\bibitem{flaying_arena_Muler}
Sergei Lupashin, Markus Hehn, Mark~W Mueller, Angela~P Schoellig, Michael
  Sherback, and Raffaello D’Andrea.
\newblock A platform for aerial robotics research and demonstration: The flying
  machine arena.
\newblock {\em Mechatronics}, 24(1):41--54, 2014.

\bibitem{Euler_madani2006control}
Tarek Madani and Abdelaziz Benallegue.
\newblock Control of a quadrotor mini-helicopter via full state backstepping
  technique.
\newblock In {\em Decision and Control, 2006 45th IEEE Conference on}, pages
  1515--1520. IEEE, 2006.

\bibitem{mahony2012multirotor}
Robert Mahony, Vijay Kumar, and Peter Corke.
\newblock Multirotor aerial vehicles: Modeling, estimation, and control of
  quadrotor.
\newblock {\em IEEE robotics \& automation magazine}, 19(3):20--32, 2012.

\bibitem{Katastrofe_maza2011experimental}
Iv{\'a}n Maza, Fernando Caballero, Jes{\'u}s Capit{\'a}n, JR~Mart{\'\i}nez-de
  Dios, and An{\'\i}bal Ollero.
\newblock Experimental results in multi-{UAV} coordination for disaster
  management and civil security applications.
\newblock {\em Journal of intelligent \& robotic systems}, 61(1-4):563--585,
  2011.

\bibitem{mehmood_analiza}
Hamza Mehmood, Takuma Nakamura, and Eric~N Johnson.
\newblock A maneuverability analysis of a novel hexarotor {UAV} concept.
\newblock In {\em 2016 International Conference on Unmanned Aircraft Systems
  (ICUAS)}, pages 437--446. IEEE, 2016.

\bibitem{mellinger_snap}
Daniel Mellinger and Vijay Kumar.
\newblock Minimum snap trajectory generation and control for quadrotors.
\newblock In {\em 2011 IEEE International Conference on Robotics and
  Automation}, pages 2520--2525. IEEE, 2011.

\bibitem{mellinger2012trajectory}
Daniel Mellinger, Nathan Michael, and Vijay Kumar.
\newblock Trajectory generation and control for precise aggressive maneuvers
  with quadrotors.
\newblock {\em The International Journal of Robotics Research}, 31(5):664--674,
  2012.

\bibitem{Fuzzy_FTC_1681912}
L.F. Mendonca, S.M. Vieira, J.M.C. Sousa, and J.M.G. Sa~da Costa.
\newblock Fault accommodation using fuzzy predictive control.
\newblock In {\em Fuzzy Systems, 2006 IEEE International Conference on}, pages
  1535--1542, 2006.

\bibitem{Inspekcija_mostova_metni2007uav}
Najib Metni and Tarek Hamel.
\newblock A {UAV} for bridge inspection: Visual servoing control law with
  orientation limits.
\newblock {\em Automation in construction}, 17(1):3--10, 2007.

\bibitem{michael2010grasp}
Nathan Michael, Daniel Mellinger, Quentin Lindsey, and Vijay Kumar.
\newblock The grasp multiple micro-{UAV} testbed.
\newblock {\em IEEE Robotics \& Automation Magazine}, 17(3):56--65, 2010.

\bibitem{franchi_heksa}
Giulia Michieletto, Markus Ryll, and Antonio Franchi.
\newblock Control of statically hoverable multi-rotor aerial vehicles and
  application to rotor-failure robustness for hexarotors.
\newblock In {\em 2017 IEEE International Conference on Robotics and Automation
  (ICRA)}, pages 2747--2752. IEEE, 2017.

\bibitem{Paralelni_thorem_morin2008introduction}
David Morin.
\newblock {\em Introduction to classical mechanics: with problems and
  solutions}.
\newblock Cambridge University Press, 2008.

\bibitem{mueller_I}
Mark~W Mueller and Raffaello D'Andrea.
\newblock Stability and control of a quadrocopter despite the complete loss of
  one, two, or three propellers.
\newblock In {\em 2014 IEEE international conference on robotics and automation
  (ICRA)}, pages 45--52. IEEE, 2014.

\bibitem{mueller2014stability}
Mark~W Mueller and Raffaello D'Andrea.
\newblock Stability and control of a quadrocopter despite the complete loss of
  one, two, or three propellers.
\newblock In {\em 2014 IEEE international conference on robotics and automation
  (ICRA)}, pages 45--52. IEEE, 2014.

\bibitem{mueller_II}
Mark~W Mueller and Raffaello D’Andrea.
\newblock Relaxed hover solutions for multicopters: Application to algorithmic
  redundancy and novel vehicles.
\newblock {\em The International Journal of Robotics Research}, 35(8):873--889,
  2016.

\bibitem{muller2011_tenis}
Mark M{\"u}ller, Sergei Lupashin, and Raffaello D'Andrea.
\newblock Quadrocopter ball juggling.
\newblock In {\em 2011 IEEE/RSJ international conference on Intelligent Robots
  and Systems}, pages 5113--5120. IEEE, 2011.

\bibitem{Control_Allocation_2_oppenheimer2006control}
Michael~W Oppenheimer, David~B Doman, and Michael~A Bolender.
\newblock Control allocation for over-actuated systems.
\newblock In {\em Control and Automation, 2006. MED'06. 14th Mediterranean
  Conference on}, pages 1--6. IEEE, 2006.

\bibitem{osmic_SMC}
Nedim Osmi{\'c}, Muhamed Kuri{\'c}, and Ivan Petrovi{\'c}.
\newblock Detailed octorotor modeling and pd control.
\newblock In {\em 2016 IEEE International Conference on Systems, Man, and
  Cybernetics (SMC)}, pages 002182--002189. IEEE, 2016.

\bibitem{osmic_FMEA}
Nedim Osmic, Anel Tahirbegovic, Adnan Tahirovic, and Stjepan Bogdan.
\newblock Failure mode and effects analysis for large scale multirotor unmanned
  aerial vehicle controlled by moving mass system.
\newblock In {\em 2018 IEEE International Systems Engineering Symposium
  (ISSE)}, pages 1--8. IEEE, 2018.

\bibitem{Osmic_automatika}
Nedim Osmic, Adnan Tahirovic, and Ivan Petrovic.
\newblock Risk-sensitive motion planning for mavs based on mission-related
  fault-tolerant analysis.
\newblock {\em Automatika}, 61(2):295--311, 2020.

\bibitem{Nadzor_saobracaja_puri2005survey}
Anuj Puri.
\newblock A survey of unmanned aerial vehicles ({UAV}) for traffic
  surveillance.
\newblock {\em Department of computer science and engineering, University of
  South Florida}, 2005.

\bibitem{Lagrange_raffo2010integral}
Guilherme~V Raffo, Manuel~G Ortega, and Francisco~R Rubio.
\newblock An integral predictive/nonlinear hinf control structure for a
  quadrotor helicopter.
\newblock {\em Automatica}, 46(1):29--39, 2010.

\bibitem{Konfiguracije_rinaldi2014pid}
Filippo Rinaldi, A~Gargioli, and Fulvia Quagliotti.
\newblock Pid and lq regulation of a multirotor attitude: Mathematical
  modelling, simulations and experimental results.
\newblock {\em Journal of Intelligent and Robotic Systems}, 73(1-4):33--50,
  2014.

\bibitem{Muler_cooperative}
Robin Ritz, Mark~W M{\"u}ller, Markus Hehn, and Raffaello D'Andrea.
\newblock Cooperative quadrocopter ball throwing and catching.
\newblock In {\em 2012 IEEE/RSJ International Conference on Intelligent Robots
  and Systems}, pages 4972--4978. IEEE, 2012.

\bibitem{coaxial_octo}
Majd Saied, Hassan Shraim, Clovis Francis, Isabelle Fantoni, and Benjamin
  Lussier.
\newblock Controllability analysis and motors failures symmetry in a coaxial
  octorotor.
\newblock In {\em 2015 Third International Conference on Technological Advances
  in Electrical, Electronics and Computer Engineering (TAEECE)}, pages
  245--250. IEEE, 2015.

\bibitem{FTC_ETH_Thomas_Schneider}
Thomas Schneider.
\newblock {\em Fault-tolerant Multirotor Systems}.
\newblock MSC thesis, ETH Zurich, Swiss Federal Institute of Technology Zurich,
  2011.

\bibitem{Sharifi}
Farid Sharifi, Mostafa Mirzaei, Brandon~W Gordon, and Youmin Zhang.
\newblock Fault tolerant control of a quadrotor {UAV} using sliding mode
  control.
\newblock In {\em Control and Fault-Tolerant Systems (SysTol), 2010 Conference
  on}, pages 239--244. IEEE, 2010.

\bibitem{shen2011autonomous}
Shaojie Shen, Nathan Michael, and Vijay Kumar.
\newblock Autonomous multi-floor indoor navigation with a computationally
  constrained mav.
\newblock In {\em 2011 IEEE International Conference on Robotics and
  Automation}, pages 20--25. IEEE, 2011.

\bibitem{shen2014multi}
Shaojie Shen, Yash Mulgaonkar, Nathan Michael, and Vijay Kumar.
\newblock Multi-sensor fusion for robust autonomous flight in indoor and
  outdoor environments with a rotorcraft mav.
\newblock In {\em 2014 IEEE International Conference on Robotics and Automation
  (ICRA)}, pages 4974--4981. IEEE, 2014.

\bibitem{shi_Analiza}
Dongjie Shi, Binxian Yang, and Quan Quan.
\newblock Reliability analysis of multicopter configurations based on
  controllability theory.
\newblock In {\em 2016 35th Chinese Control Conference (CCC)}, pages
  6740--6745. IEEE, 2016.

\bibitem{Sicilijano_Robotika}
Villani~L. Siciliano~B., Sciavicco~L. and Oriolo G.
\newblock {\em Robotics: Modelling, Planning and Control}.
\newblock Springer, 2009.

\bibitem{Cross_produkt_Gilbert_Strang}
Gilbert Strang.
\newblock {\em Linear Algebra and Its Applications}.
\newblock Thomson Learning Academic Resource Center, 2006.

\bibitem{tahirovic2013discussion}
Adnan Tahirovic.
\newblock Discussion on:" control and navigation in manoeuvres of formations of
  unmanned mobile vehicles".
\newblock {\em European Journal of Control}, 19(2):172, 2013.

\bibitem{tahirovic2013convergent}
Adnan Tahirovic and Alessandro Astolfi.
\newblock A convergent solution to the multi-vehicle coverage problem.
\newblock In {\em 2013 American Control Conference}, pages 4635--4641. IEEE,
  2013.

\bibitem{tahirovic2016receding}
Adnan Tahirovic, Mehmed Brkic, Aldin Bostan, and Benjamin Seferagic.
\newblock A receding horizon scheme for constrained multi-vehicle coverage
  problems.
\newblock In {\em 2016 IEEE International Conference on Systems, Man, and
  Cybernetics (SMC)}, pages 004652--004656. IEEE, 2016.

\bibitem{Tomic_istr_platforme}
Teodor Tomic, Korbinian Schmid, Philipp Lutz, Andreas Domel, Michael Kassecker,
  Elmar Mair, Iris~Lynne Grixa, Felix Ruess, Michael Suppa, and Darius
  Burschka.
\newblock Toward a fully autonomous {UAV} : Research platform for indoor and
  outdoor urban search and rescue.
\newblock {\em IEEE robotics \& automation magazine}, 19(3):46--56, 2012.

\bibitem{Lagrange_MARTINEZ_TORRES}
Cesar~Martinez Torres.
\newblock {\em Fault Tolerant Controln by Flatneess approach.}
\newblock Phd thesis, San Nicolas De Los Garza, Nuevo Leon, 2013.

\bibitem{Lunze_I}
Daniel Vey and Jan Lunze.
\newblock Structural reconfigurability analysis of multirotor {{UAV} }s after
  actuator failures.
\newblock In {\em 2015 54th IEEE Conference on Decision and Control (CDC)},
  pages 5097--5104. IEEE, 2015.

\bibitem{Sliding_mode_FTC_wang2012sliding}
Tao Wang, Wenfang Xie, and Youmin Zhang.
\newblock Sliding mode fault tolerant control dealing with modeling
  uncertainties and actuator faults.
\newblock {\em ISA transactions}, 51(3):386--392, 2012.

\bibitem{Adaptivni_FTC_xu2013adaptive}
Qing Xu, Hao Yang, Bin Jiang, Donghua Zhou, and Youmin Zhang.
\newblock Adaptive fault-tolerant control design for {UAV} s formation flight
  under actuator faults.
\newblock In {\em Unmanned Aircraft Systems (ICUAS), 2013 International
  Conference on}, pages 1097--1105. IEEE, 2013.

\bibitem{Zhang_agrikultura}
Chunhua Zhang and John~M Kovacs.
\newblock The application of small unmanned aerial systems for precision
  agriculture: a review.
\newblock {\em Precision agriculture}, 13(6):693--712, 2012.

\bibitem{zhang2008bibliographical}
Youmin Zhang and Jin Jiang.
\newblock Bibliographical review on reconfigurable fault-tolerant control
  systems.
\newblock {\em Annual reviews in control}, 32(2):229--252, 2008.

\end{thebibliography}

\printindex

\end{document}